%% file: acl_latex.tex
\pdfoutput=1

\documentclass[11pt]{article}

\usepackage[preprint]{acl}

\usepackage{times}
\usepackage{latexsym}

\usepackage[T1]{fontenc}

\usepackage[utf8]{inputenc}

\usepackage{microtype}

\usepackage{xcolor}

\usepackage{soul} 
\sethlcolor{red!8} 
\newcommand{\redhl}[1]{\textcolor{red}{\textbf{#1}}}

\usepackage{inconsolata}

\usepackage{graphicx}

\usepackage{amsmath}
\usepackage{amssymb}
\usepackage{amsthm}

\usepackage{booktabs}
\usepackage{tabularx}

\usepackage{multirow}
\usepackage{array}
\newcolumntype{C}{>{\centering\arraybackslash}X}
\usepackage{pifont}
\usepackage{enumitem}
\usepackage{subcaption}
\usepackage{verbatim}
\usepackage{makecell}
\usepackage[table]{xcolor} 
\usepackage{tcolorbox}
\usepackage[linesnumbered,ruled,vlined]{algorithm2e}
\usepackage{seqsplit}

\usepackage{tcolorbox}
\tcbuselibrary{skins, breakable}

\definecolor{hmcolor}{HTML}{35A79C} 

\newcommand{\gc}[2]{\cellcolor{hmcolor!#1}#2}
\usepackage{cleveref}
\usepackage{tikz}
\usetikzlibrary{calc, shapes, backgrounds}

\definecolor{aclBlueDark}{HTML}{4E79A7}
\definecolor{aclBlueLight}{HTML}{A0CBE8}
\definecolor{aclOrangeDark}{HTML}{F28E2B}
\definecolor{aclOrangeLight}{HTML}{FFBE7D}

\title{Circular Reasoning: Understanding Self-Reinforcing Loops \\in Large Reasoning Models}

\author{
 \textbf{Zenghao Duan}\textsuperscript{1,2 *},
 \textbf{Liang Pang}\textsuperscript{1 *},
 \textbf{Zihao Wei}\textsuperscript{1,2 *},
 \textbf{Wenbin Duan}\textsuperscript{1,2 *},
 \\
 \textbf{Yuxin Tian}\textsuperscript{3},
 \textbf{Shicheng Xu}\textsuperscript{1,2},
 \textbf{Jingcheng Deng}\textsuperscript{1,2},
 \textbf{Zhiyi Yin}\textsuperscript{1 \(\dagger\)},
 \textbf{Xueqi Cheng}\textsuperscript{1},
\\
 \textsuperscript{1}State Key Laboratory of AI Safety, Institute of Computing Technology, Chinese Academy of Sciences
 \\
 \textsuperscript{2}University of Chinese Academy of Sciences
 \\
 \textsuperscript{3}People’s Public Security University of China
 \\
 \small{
     \href{mailto:email@domain}{\{duanzenghao24s, yinzhiyi, pangliang, cxq\}@ict.ac.cn}
 }
}

\begin{document}

\maketitle

\stepcounter{footnote}
\footnotetext[\value{footnote}]{* Equal Contributions}
\stepcounter{footnote}
\footnotetext[\value{footnote}]{\(\dagger\) Corresponding author}

\begin{abstract}

Despite the success of test-time scaling, Large Reasoning Models (LRMs) frequently encounter repetitive loops that lead to computational waste and inference failure. 
In this paper, we identify a distinct failure mode termed \textbf{\textit{Circular Reasoning}}. Unlike traditional model degeneration, this phenomenon manifests as a self-reinforcing trap where generated content acts as a logical premise for its own recurrence, compelling the reiteration of preceding text.
To systematically analyze this phenomenon, we introduce LoopBench, a dataset designed to capture two distinct loop typologies: \textit{numerical loops} and \textit{statement loops}.
Mechanistically, we characterize circular reasoning as a state collapse exhibiting distinct boundaries, where semantic repetition precedes textual repetition.
We reveal that reasoning impasses trigger the loop onset, which subsequently persists as an inescapable cycle driven by a self-reinforcing V-shaped attention mechanism.
Guided by these findings, we employ the Cumulative Sum (CUSUM) algorithm to capture these precursors for early loop prediction. Experiments across diverse LRMs validate its accuracy and elucidate the stability of long-chain reasoning.

\end{abstract}



\input{latex/sections/01_introduction}  \label{introduction}
\input{latex/sections/04_dataset}       \label{loopbench}
\input{latex/sections/05_analysis}
\label{analysis}
\input{latex/sections/06_method}        \label{method}
\input{latex/sections/02_related-work}
\input{latex/sections/07_conclusion}


\bibliography{custom}

\appendix
\input{latex/sections/X1_dataset}

\input{latex/sections/X5_analysis}

\input{latex/sections/X3_method}

\input{latex/sections/X4_case_study}

\end{document}

%% file: latex/sections/01_introduction.tex
\section{Introduction}


In recent years, Large Reasoning Models (LRMs) have made significant progress in complex reasoning tasks through Test-time Scaling strategies \cite{openai2024o1, snell2024scaling}. 
However, relying on long chains of thought has a downside. 
As sequences grow, models often generate unnecessary or even repetitive steps \cite{DBLP:journals/corr/abs-2504-06514,DBLP:journals/corr/abs-2510-15522}, leading to reduced accuracy and wasted computational resources \cite{DBLP:journals/corr/abs-2511-00536}.

\begin{figure}[t]
    \centering
    \includegraphics[width=\linewidth]{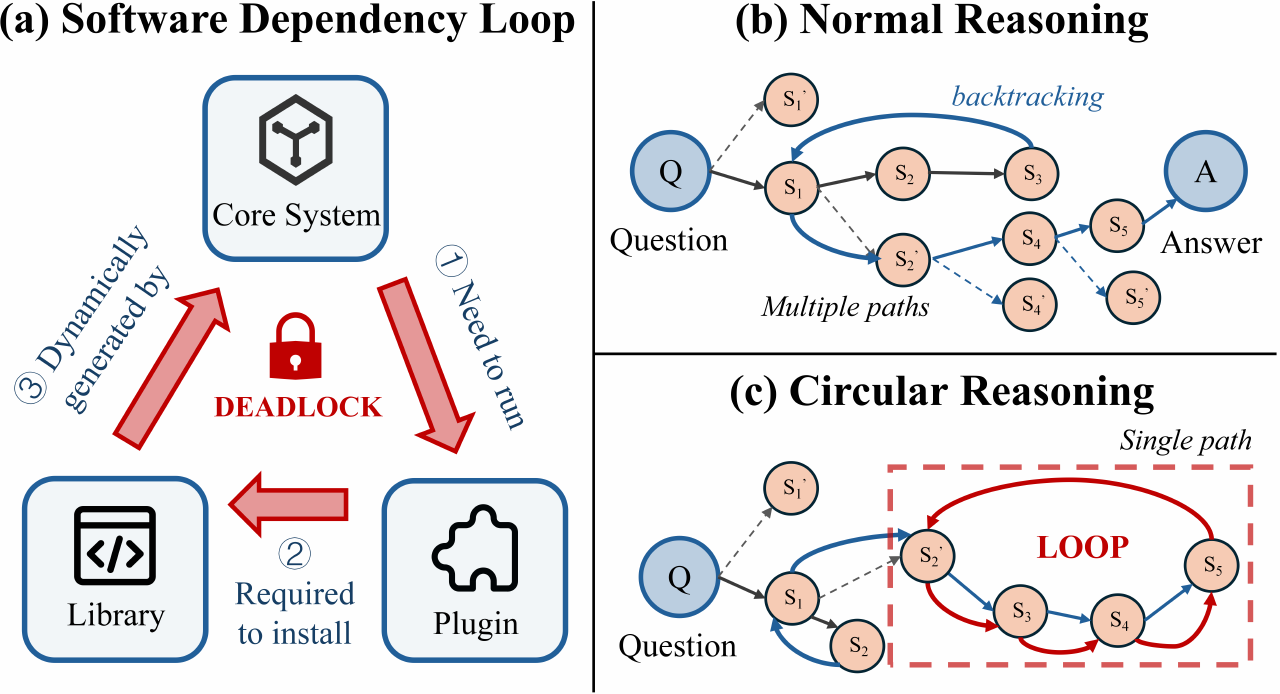}
    \caption{\textbf{The Trap of Circular Reasoning.} A logical deadlock that blocks step-by-step inference. Analogous to a software deadlock (a), this mode entraps the model in a self-reinforcing cycle (c),preventing the flexible backtracking seen in normal reasoning (b).}
    \label{fig:motivation}
\end{figure}



In this paper, we identify a distinct failure mode termed \textit{\textbf{Circular Reasoning}}. Unlike traditional model degeneration (e.g., collapse or simple repetition), this phenomenon manifests as a self-reinforcing trap where generated content acts as a logical premise for its own recurrence. 
As shown in Figure \ref{fig:motivation}, unlike normal reasoning that navigates diverse paths towards a solution, this pathology locks the inference into an endless loop, blocking access to a valid conclusion.


Despite its severity, systematic research on this failure mode remains limited.
We observe that this pathology prevails in high-precision arithmetic and complex recursive reasoning tasks.
To study it, we construct \textbf{LoopBench} with 700 samples covering seven sub-tasks (\S\ref{subsec:loopbench}).
Our experiments show that most LRMs, including open-source and closed-source models, suffer from severe repetition, manifesting as \textit{numerical loops} or \textit{statement loops}.
Notably, while numerical loops are equally prevalent across model types, statement loops are significantly more common in LRMs compared to base or instruction-tuned models (\S\ref{subsec:key observation}).

Next, we provide an in-depth mechanistic analysis. By dissecting the behavior, triggers, and persistence factors, we uncover the pathology of reasoning loops:
\textbf{Behavior (\S\ref{subsec:phase transition}):} Entering a loop marks a internal state collapse with clear boundaries. Through topological analysis using reasoning graphs~\cite{Minegishi2025TopologyOR}, we also reveal that semantic circularity consistently precedes explicit textual repetition, serving as a latent structural precursor.
\textbf{Trigger (\S\ref{subsec:trigger}):} While reasoning impasses generally precipitate the loop onset, the specific drivers vary by type. \textit{Statement loops} are driven by dense high-entropy minority tokens (e.g., `But', `Wait') within the semantic circularity that disrupt attention distribution. In contrast, \textit{numerical loops} are typically induced by context saturation from tasks requiring long-form expansions.
\textbf{Persistence (\S\ref{subsec:vshape}):} The `V-shaped' attention pattern sustains loop cycles. Predominantly focusing on attention sinks and recent tokens~\cite{fu2024attentionpattern}, this bias compels the model to self-reinforce repetition once the immediate context becomes repetitive.



Building on above insights, we propose an early prediction strategy for \textit{statement loops}. 
We exploit the finding that repetition loops are typically preceded by a surge of reflection sentence, which triggers a distinct shift in hidden states. 
By monitoring this latent warning signal via the Cumulative Sum (CUSUM) algorithm~\cite{Saxena2015Sequential}, we can proactively identify loop risks.  Empirical validation across diverse models confirms the efficacy of this approach (\S\ref{sec:detection}), providing a new perspective on ensuring the stability of long-chain reasoning.


The main contributions of our work are summarized as follows:
\begin{itemize}[itemsep=3pt, topsep=0pt, parsep=0pt]
    \item We identify the ``Circular Reasoning'' phenomenon and introduce LoopBench to systematically quantify loop behavior in LRMs.
    \item We conduct an in-depth analysis of loop mechanisms, systematically investigating both the model's internal state during repetition and the factors that trigger and sustain these cycles.
    \item We propose an early prediction strategy based on hidden states targeting \textit{statement loops}, which effectively anticipates loop formation across diverse models.
\end{itemize}

%% file: latex/sections/04_dataset.tex
\section{LoopBench: A Benchmark for Reasoning Loop in LRMs}


In this section, we first identify task characteristics that trigger reasoning loops in LRMs. Based on this, we develop LoopBench to evaluate this phenomenon. Finally, we conduct experiments to systematically analyze model performance.

\subsection{Benchmark Construction}    \label{subsec:loopbench}




We observe that looping behavior are primarily induced by two specific task patterns, covering seven sub-tasks: 
(1) \textbf{High-Precision Arithmetic}: Includes Square Root, Long Division, and Newton's Iteration. These tasks demand sustained precision over long-horizon stepwise calculations. 
(2) \textbf{Complex Recursive Reasoning}: Includes Tower of Hanoi, Truth-Teller Puzzles, Path Planning, and Logical Paradoxes. These require rigorous state tracking and recursive dependency management. 
Detailed task formulations are provided in Table \ref{tab:loopbench_tasks}.

Guided by these task patterns, we construct LoopBench with 100 samples for each sub-task, resulting in a total of 700 instances (the taxonomy is illustrated in Figure \ref{fig:loopbench}, and the detail is in Appendix \ref{sec:LoopBench_Detail}). Although LoopBench serves as a specific instantiation, it highlights fundamental challenges regarding reasoning stability in LRMs.

\begin{figure}
    \centering
    \includegraphics[width=0.8\linewidth]{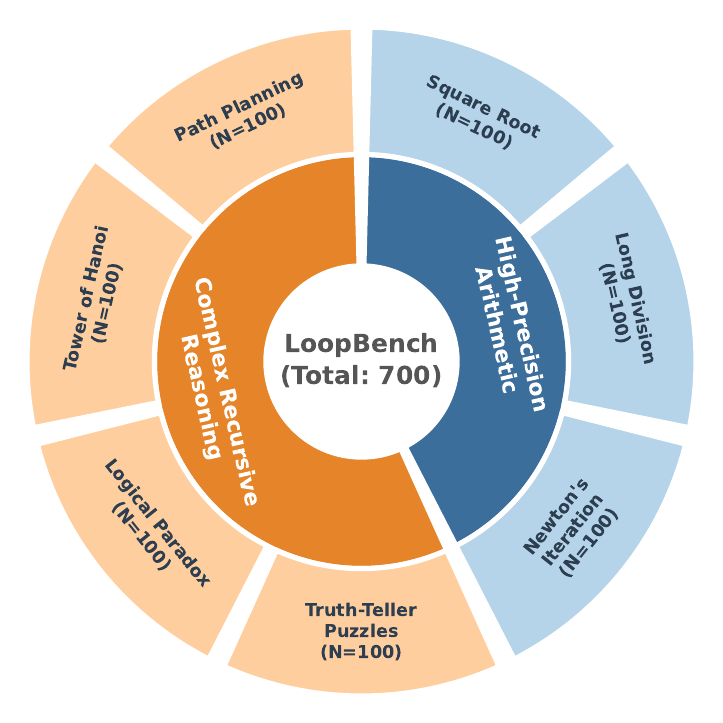}
    \caption{\textbf{Taxonomy of LoopBench.} The benchmark is organized into two primary categories: High-Precision Arithmetic and Complex Recursive Reasoning.}
    \label{fig:loopbench}
\end{figure}

\begin{table*}[t]
\centering
\caption{\textbf{Evaluation of loop susceptibility across benchmarks.} We report DLN and SLN for base, instruction-tuned, and reasoning models. Open-source models use the Balanced Setting, while API models use default decoding.}
\label{tab:repetition_result}
\resizebox{\textwidth}{!}{%
\begin{tabular}{ll | ccc | ccc | ccc} 
\toprule
\multicolumn{2}{c}{\multirow{2}{*}{\textbf{Model}}} & \multicolumn{3}{c}{\textbf{AIME25}} & \multicolumn{3}{c}{\textbf{SuperGPQA}} & \multicolumn{3}{c}{\textbf{LoopBench}} \\ 
\cmidrule(lr){3-5} \cmidrule(lr){6-8} \cmidrule(lr){9-11} 
\multicolumn{2}{c}{} & \textbf{DLN} & \textbf{SLN} & \textbf{Sum} & \textbf{DLN} & \textbf{SLN} & \textbf{Sum} & \textbf{DLN} & \textbf{SLN} & \textbf{Sum} \\ \midrule

\rowcolor{gray!15} \multicolumn{11}{c}{\textit{\textbf{Open-Source Models}}} \\ \addlinespace[3pt]

\multirow{12}{*}{\makecell[l]{Reasoning}} 
 & DeepSeek-R1-Distill-Qwen-7B   & 0.00\% & \gc{10}{6.67\%} & \gc{10}{\textbf{6.67\%}} & \gc{5}{1.50\%} & \gc{10}{6.00\%} & \gc{12}{\textbf{7.50\%}} & \gc{25}{17.14\%} & \gc{60}{44.86\%} & \gc{80}{\textbf{62.00\%}} \\
 & DeepSeek-R1-Distill-Llama-8B  & \gc{5}{3.33\%} & \gc{10}{6.67\%} & \gc{15}{\textbf{10.00\%}} & 0.00\% & \gc{5}{2.50\%} & \gc{5}{\textbf{2.50\%}} & \gc{25}{16.86\%} & \gc{35}{24.57\%} & \gc{60}{\textbf{41.43\%}} \\
 & DeepSeek-R1-0528-Qwen3-8B     & 0.00\% & 0.00\% & \textbf{0.00\%} & 0.00\% & 0.00\% & \textbf{0.00\%} & \gc{15}{9.14\%} & \gc{10}{6.29\%} & \gc{20}{\textbf{15.43\%}} \\
 & DeepSeek-R1-Distill-Qwen-14B  & \gc{5}{3.33\%} & \gc{10}{6.67\%} & \gc{15}{\textbf{10.00\%}} & \gc{3}{1.00\%} & \gc{10}{6.25\%} & \gc{12}{\textbf{7.25\%}} & \gc{30}{19.14\%} & \gc{50}{37.14\%} & \gc{75}{\textbf{56.29\%}} \\
 & DeepSeek-R1-Distill-Qwen-32B  & \gc{5}{3.33\%} & \gc{5}{3.33\%} & \gc{10}{\textbf{6.67\%}} & 0.00\% & \gc{8}{4.50\%} & \gc{8}{\textbf{4.50\%}} & \gc{20}{14.00\%} & \gc{50}{36.00\%} & \gc{65}{\textbf{50.00\%}} \\
 & QwQ-32B                       & 0.00\% & 0.00\% & \textbf{0.00\%} & \gc{2}{0.50\%} & 0.00\% & \gc{2}{\textbf{0.50\%}} & \gc{35}{24.29\%} & \gc{10}{5.77\%} & \gc{40}{\textbf{30.06\%}} \\
 & Qwen3-8B                      & 0.00\% & \gc{5}{3.33\%} & \gc{5}{\textbf{3.33\%}} & \gc{3}{1.00\%} & \gc{5}{1.50\%} & \gc{5}{\textbf{2.50\%}} & \gc{25}{18.86\%} & \gc{15}{10.43\%} & \gc{40}{\textbf{29.29\%}} \\
 & Qwen3-14B                     & \gc{5}{3.33\%} & 0.00\% & \gc{5}{\textbf{3.33\%}} & \gc{2}{0.50\%} & 0.00\% & \gc{2}{\textbf{0.50\%}} & \gc{30}{21.99\%} & \gc{10}{5.43\%} & \gc{35}{\textbf{27.41\%}} \\
 & Qwen3-32B                     & 0.00\% & 0.00\% & \textbf{0.00\%} & \gc{3}{1.00\%} & 0.00\% & \gc{3}{\textbf{1.00\%}} & \gc{35}{26.43\%} & \gc{8}{4.29\%} & \gc{40}{\textbf{30.72\%}} \\
 & gpt-oss-20B                   & \gc{5}{3.33\%} & 0.00\% & \gc{5}{\textbf{3.33\%}} & \gc{5}{1.50\%} & \gc{2}{0.50\%} & \gc{5}{\textbf{2.00\%}} & \gc{20}{13.71\%} & \gc{12}{8.72\%} & \gc{30}{\textbf{22.43\%}} \\
 & Phi-4-reasoning               & \gc{5}{3.33\%} & 0.00\% & \gc{5}{\textbf{3.33\%}} & \gc{5}{3.00\%} & 0.00\% & \gc{5}{\textbf{3.00\%}} & \gc{40}{27.81\%} & \gc{20}{13.43\%} & \gc{55}{\textbf{41.24\%}} \\
 & Phi-4-reasoning-plus          & \gc{5}{3.33\%} & 0.00\% & \gc{5}{\textbf{3.33\%}} & \gc{5}{2.50\%} & 0.00\% & \gc{5}{\textbf{2.50\%}} & \gc{40}{29.90\%} & \gc{18}{12.00\%} & \gc{55}{\textbf{41.90\%}} \\ \midrule
 
\multirow{3}{*}{\makecell[l]{Base}} 
 & Qwen3-8B-base     & 0.00\% & 0.00\% & \textbf{0.00\%} & \gc{2}{0.50\%} & 0.00\% & \gc{2}{\textbf{0.50\%}} & \gc{30}{20.57\%} & \gc{5}{1.71\%} & \gc{30}{\textbf{22.29\%}} \\
 & Qwen3-14B-base    & 0.00\% & \gc{5}{3.33\%} & \gc{5}{\textbf{3.33\%}} & 0.00\% & 0.00\% & \textbf{0.00\%} & \gc{30}{20.57\%} & \gc{2}{0.19\%} & \gc{28}{\textbf{20.76\%}} \\
 & Gemma-2-27B       & 0.00\% & \gc{5}{3.33\%} & \gc{5}{\textbf{3.33\%}} & \gc{5}{1.50\%} & 0.00\% & \gc{5}{\textbf{1.50\%}} & \gc{20}{13.52\%} & \gc{5}{2.05\%} & \gc{20}{\textbf{15.57\%}} \\ \midrule
 
\multirow{5}{*}{\makecell[l]{Instruct}} 
 & Llama-3.1-8B-Instruct & 0.00\% & 0.00\% & \textbf{0.00\%} & 0.00\% & 0.00\% & \textbf{0.00\%} & \gc{25}{17.91\%} & \gc{5}{2.86\%} & \gc{28}{\textbf{20.76\%}} \\
 & Qwen2.5-7B-Instruct   & \gc{5}{3.33\%} & 0.00\% & \gc{5}{\textbf{3.33\%}} & \gc{2}{0.50\%} & 0.00\% & \gc{2}{\textbf{0.50\%}} & \gc{25}{19.04\%} & \gc{3}{0.76\%} & \gc{25}{\textbf{19.80\%}} \\
 & Qwen2.5-14B-Instruct  & 0.00\% & 0.00\% & \textbf{0.00\%} & 0.00\% & \gc{2}{0.50\%} & \gc{2}{\textbf{0.50\%}} & \gc{25}{17.90\%} & 0.00\% & \gc{25}{\textbf{17.90\%}} \\
 & Qwen2.5-32B-Instruct  & 0.00\% & 0.00\% & \textbf{0.00\%} & 0.00\% & \gc{2}{0.50\%} & \gc{2}{\textbf{0.50\%}} & \gc{30}{24.76\%} & \gc{2}{0.57\%} & \gc{30}{\textbf{25.33\%}} \\
 & Gemma-2-27B-it        & 0.00\% & 0.00\% & \textbf{0.00\%} & 0.00\% & 0.00\% & \textbf{0.00\%} & \gc{30}{24.95\%} & 0.00\% & \gc{30}{\textbf{24.95\%}} \\ \midrule

\rowcolor{gray!15} \multicolumn{11}{c}{\textit{\textbf{API Models}}} \\ \addlinespace[3pt]
\multirow{12}{*}{\makecell[l]{APIs}} 
 & gemini-2.5-pro (\textit{thinking})  & 0.00\% & 0.00\% & \textbf{0.00\%} & 0.00\% & 0.00\% & \textbf{0.00\%} & \gc{7}{5.00\%} & \gc{2}{0.29\%} & \gc{9}{\textbf{5.29\%}} \\
 & gemini-2.5-pro             & 0.00\% & 0.00\% & \textbf{0.00\%} & 0.00\% & 0.00\% & \textbf{0.00\%} & \gc{10}{7.43\%} & 0.00\% & \gc{10}{\textbf{7.43\%}} \\
 & claude-opus-4-1 (\textit{thinking})    & 0.00\% & 0.00\% & \textbf{0.00\%} & 0.00\% & 0.00\% & \textbf{0.00\%} & \gc{8}{4.43\%} & \gc{2}{0.29\%} & \gc{8}{\textbf{4.71\%}} \\
 & claude-opus-4-1            & 0.00\% & 0.00\% & \textbf{0.00\%} & 0.00\% & 0.00\% & \textbf{0.00\%} & \gc{8}{6.71\%} & 0.00\% & \gc{8}{\textbf{6.71\%}} \\
 & Doubao-seed-1.6 (\textit{thinking}) & 0.00\% & 0.00\% & \textbf{0.00\%} & 0.00\% & 0.00\% & \textbf{0.00\%} & \gc{20}{15.00\%} & \gc{3}{1.14\%} & \gc{21}{\textbf{16.14\%}} \\
 & Doubao-seed-1.6            & 0.00\% & 0.00\% & \textbf{0.00\%} & 0.00\% & 0.00\% & \textbf{0.00\%} & \gc{20}{16.43\%} & \gc{2}{0.29\%} & \gc{20}{\textbf{16.72\%}} \\
 & DeepSeek-Reasoner (\textit{thinking})         & 0.00\% & \gc{5}{3.33\%} & \gc{5}{\textbf{3.33\%}} & 0.00\% & 0.00\% & \textbf{0.00\%} & \gc{15}{11.00\%} & \gc{3}{1.28\%} & \gc{18}{\textbf{12.28\%}} \\
 & DeepSeek-Chat              & 0.00\% & 0.00\% & \textbf{0.00\%} & 0.00\% & 0.00\% & \textbf{0.00\%} & \gc{16}{7.72\%} & \gc{2}{0.57\%} & \gc{18}{\textbf{8.29\%}} \\
 & Qwen-Plus (\textit{thinking})      & 0.00\% & 0.00\% & \textbf{0.00\%} & 0.00\% & 0.00\% & \textbf{0.00\%} & \gc{20}{14.00\%} & \gc{3}{1.14\%} & \gc{20}{\textbf{15.14\%}} \\
 & Qwen-Plus                  & 0.00\% & 0.00\% & \textbf{0.00\%} & 0.00\% & 0.00\% & \textbf{0.00\%} & \gc{21}{16.00\%} & \gc{2}{0.29\%} & \gc{22}{\textbf{16.29\%}} \\
 & MiMo-V2-Flash (\textit{thinking})     & 0.00\% & 0.00\% & \textbf{0.00\%} & 0.00\% & \gc{2}{0.50\%} & \gc{2}{\textbf{0.50\%}} & \gc{25}{18.57\%} & \gc{8}{4.86\%} & \gc{28}{\textbf{23.43\%}} \\
 & MiMo-V2-Flash              & 0.00\% & 0.00\% & \textbf{0.00\%} & \gc{3}{1.00\%} & 0.00\% & \gc{3}{\textbf{1.00\%}} & \gc{25}{19.43\%} & \gc{3}{1.37\%} & \gc{28}{\textbf{20.80\%}} \\
\bottomrule
\end{tabular}%
}
\end{table*}

\subsection{Key Observations on LoopBench}  \label{subsec:key observation}

We conducted extensive experiments to investigate loop behavior across diverse models. 
Our selection covers open- and closed-source models of varying sizes, as well as different training stages including base, instruction-tuned, and reasoning models.
To validate the challenge level of LoopBench, we compared it against established datasets such as AIME2025 \cite{aime25} and a stratified subset of SuperGPQA \cite{pteam2025supergpqascalingllmevaluation}.
For evaluation, we quantified loop using two specific metrics: Numerical Loop Rate (NLR) and Statement Loop Rate (SLR). These metrics correspond to the two distinct granularities observed in our analysis. 
We evaluate the models across various decoding configurations.
Detailed implementation settings are provided in Appendix \ref{subsec:detail_setup}.

Tables \ref{tab:repetition_result}, \ref{tab:repetition_result_t01}, and \ref{tab:repetition_result_t10} present the main results, highlighting the following key findings:

\paragraph{LoopBench triggers significantly higher repetition than other benchmarks.}
The propensity to induce loop generation is substantially higher on LoopBench compared to baselines like AIME2025 and SuperGPQA, despite their recognized difficulty. 
Specifically, under the balanced setting, NLR peaks at 29.90\% and SLR reaches 44.86\%. 
This confirms that LoopBench effectively exposes model fragility under specific reasoning patterns often overlooked by standard evaluations.


\paragraph{Loop patterns exhibit distinct granular behaviors.} 
Repetition manifests at two granularities with divergent characteristics. 
Numerical loops are a universal phenomenon across all training paradigms, typically appearing as structured cyclical digits rather than random noise. Conversely, statement loops are predominantly observed in reasoning models; while negligible in base and instruction-tuned models, they are prevalent in LRMs. Detailed cases are provided in Appendix \ref{subsec:case}.
Crucially, we observe no significant correlation between repetition susceptibility and model scale or distillation status (e.g., DS-Qwen3-8B exhibits robust stability). This suggests that the issue stems from intrinsic generation mechanisms rather than capacity limitations.



\paragraph{Closed-source models exhibit superior yet imperfect robustness.} While generally demonstrating lower repetition rates than their open-source counterparts, they remain vulnerable. Numerical loops, in particular, persist as a significant bottleneck; with the exception of the gemini-2.5-pro and claude-Opus-4.1 series, most proprietary models still exhibit a DLR exceeding 15\%, underscoring that even state-of-the-art systems are susceptible to the high-precision arithmetic traps in LoopBench.

\paragraph{Decoding strategies mitigate but do not eliminate repetition.} 
While increasing randomness (e.g., higher temperature and sampling settings) reduces repetition frequency, it offers no complete cure. Notably, numerical loops persist even at $T=1.0$ (Table~\ref{tab:repetition_result_t10}), indicating a strong intrinsic bias towards these patterns despite high stochasticity.
Furthermore, relying on high temperature creates a dilemma. While lower temperatures are typically preferred for rigorous reasoning to ensure precision \cite{deepseekai2025deepseekr1}, they increase the risk of repetition. Conversely, raising temperature improves diversity but often lowers reasoning accuracy and leads to hallucinations \cite{huang2023survey}. This indicates an inherent trade-off between loop and accuracy, suggesting that parameter tuning alone cannot fix the ``Circular Reasoning'' trap.

%% file: latex/sections/05_analysis.tex
\begin{figure}[t]
    \centering
    \begin{subfigure}[b]{\linewidth}
        \centering
        \includegraphics[width=\linewidth]{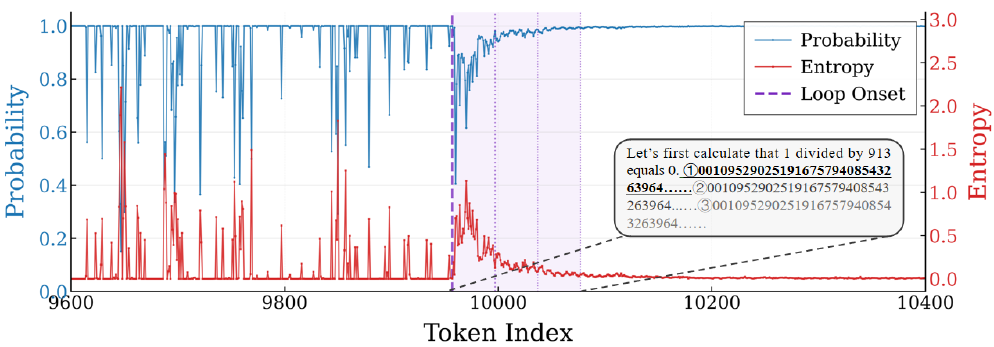}
        \caption{Numerical Loop Case}
        \label{fig:loop-attn}
    \end{subfigure}
    \begin{subfigure}[b]{\linewidth}
        \centering
        \includegraphics[width=\linewidth]{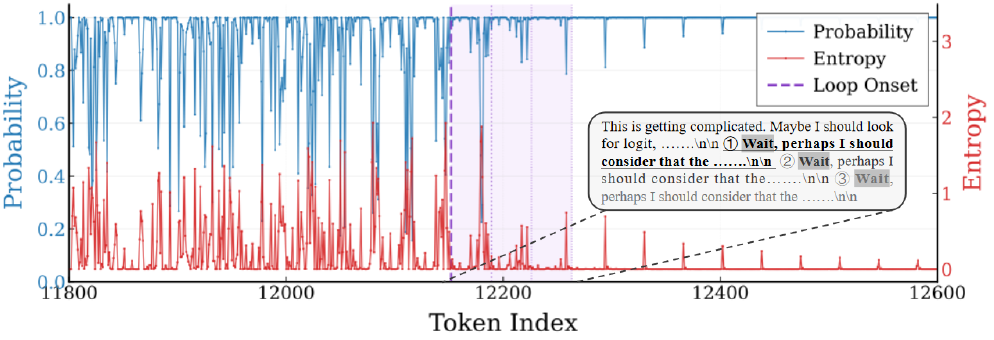}
        \caption{Statement Loop Case}
        \label{fig:nonloop-attn}
    \end{subfigure}
    \caption{\textbf{Abrupt shift in decoding dynamics in DS-Qwen-14B.} The loop onset triggers an immediate entropy collapse (red) and probability surge (blue), marking a sudden entry into rigid determinism in both cases.}
    \label{fig:logist}
\end{figure}

\section{Analysis of Circular Reasoning}

We investigate the mechanisms of circular reasoning using DeepSeek-R1-Distill-Qwen2.5-14B (hereafter DS-Qwen-14B) and Qwen3-8B as primary case studies. We examine this phenomenon from three perspectives by the lifecycle of a loop:

\begin{itemize}[itemsep=3pt, topsep=0pt, parsep=0pt]
    \item \textbf{Behavior:} How does the model’s internal state evolve during the transition to loop?
    \item \textbf{Trigger:} What specific factors trigger the transition from normal reasoning to a loop?
    \item \textbf{Persistence:} Why is the model unable to self-correct and exit the cycle once initiated?
\end{itemize}

\begin{figure}
    \centering
    \begin{subfigure}[b]{0.48\linewidth}
        \centering
        \includegraphics[width=\linewidth]{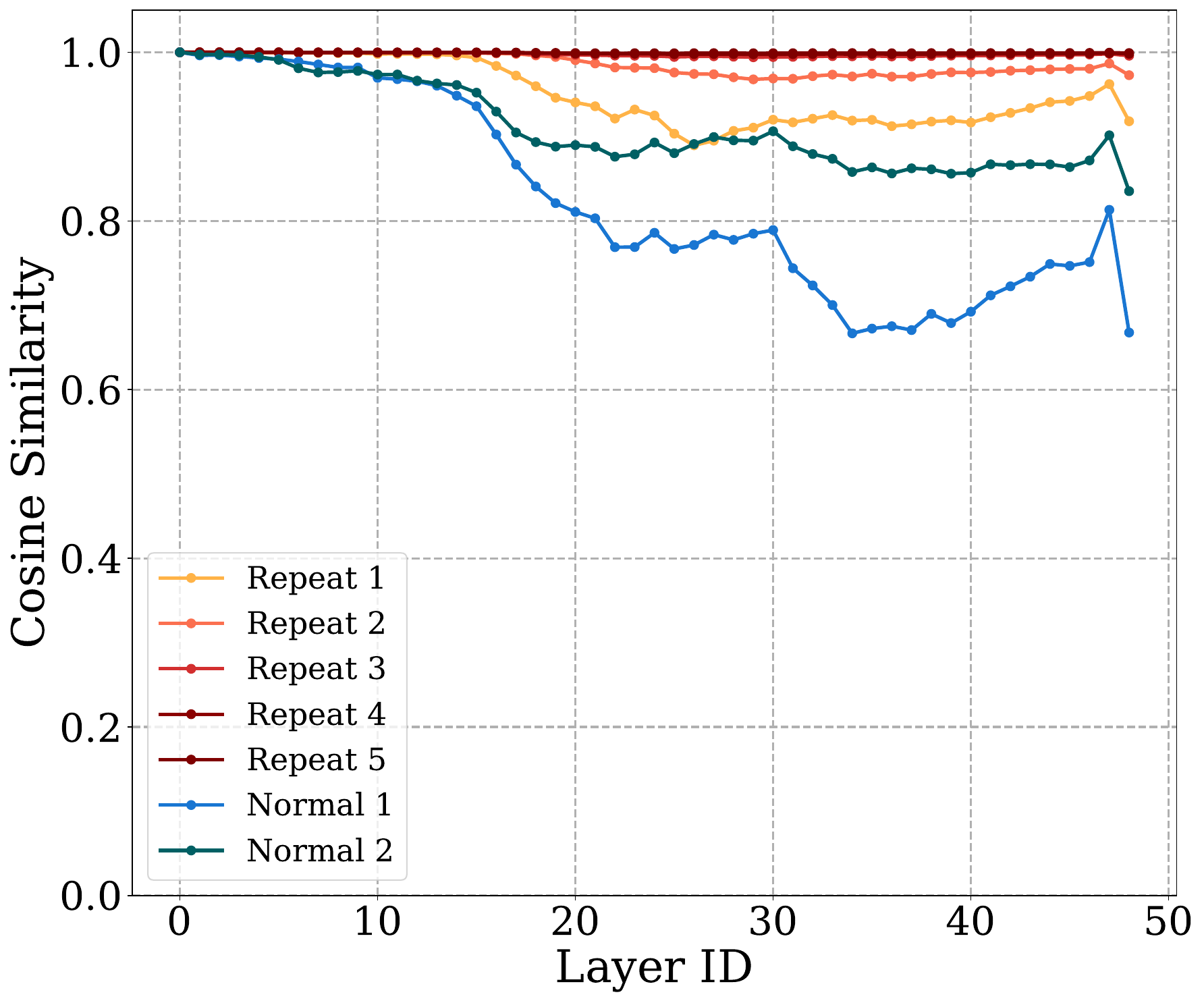}
        \caption{Cosine Similarity}
        \label{fig:loop-attn-2}
    \end{subfigure}
    \begin{subfigure}[b]{0.48\linewidth}
        \centering
        \includegraphics[width=\linewidth]{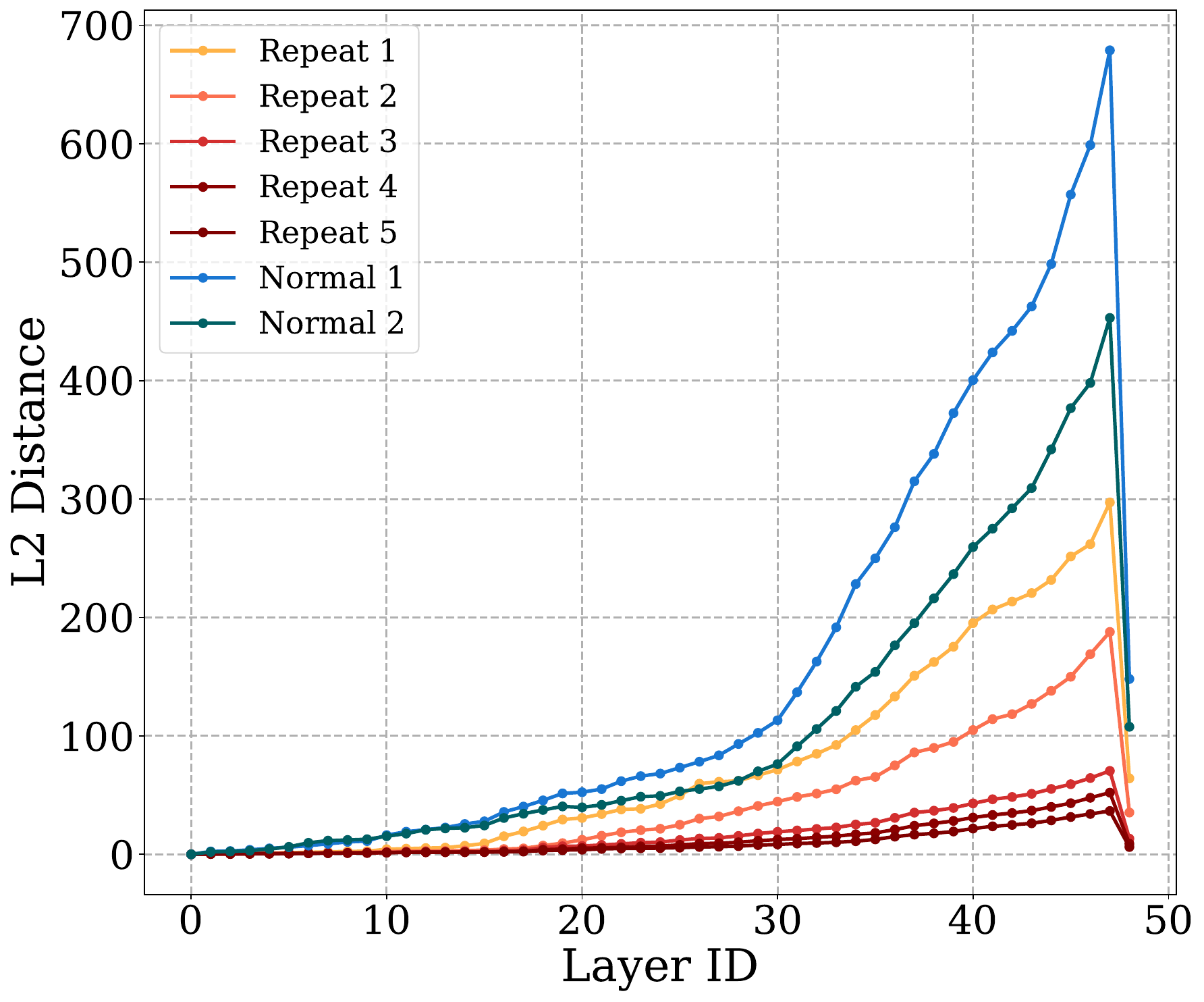}
        \caption{L2 Distance}
        \label{fig:nonloop-attn-2}
    \end{subfigure}
    \caption{\textbf{Collapse of Internal States.} 
    We report the layer-wise Cosine Similarity and $L_2$ Distance in DS-Qwen-14B. 
    Curves labelled ``Repeat $k$'' denote the comparison between the representation in the ``$k$''-th loop and ``$k-1$''-th cycle.
    \textit{\textbf{Trend:}} As loop deepens (darker red lines), the internal state becomes rigid.}
    \label{fig:cosin_qs_case1}
\end{figure}

\subsection{Loop Behavior Analysis}         \label{subsec:phase transition}

The transition into a loop state exhibits clear boundaries, irrespective of the repetition type. This indicates a fundamental shift to a cyclically constrained pattern in internal dynamics, distinct from the exploratory nature of normal reasoning.

\paragraph{Determinism Surge and State Collapse.}
We first quantify the determinism surge via logit distributions and entropy. As shown in Figure~\ref{fig:logist}, the onset of a loop triggers a sharp rise in maximum logit values and a corresponding drop in entropy toward zero. This sudden shift to rigid determinism stands in sharp contrast to the dynamic fluctuations of normal generation.
To investigate the mechanism behind this surge, we analyze the accompanying state collapse in internal activations. Using statement loops in DS-Qwen-14B as a case study, we compute the cosine similarity and $L_2$ norm differences between activation vectors of identical tokens (e.g., ``\texttt{\textbackslash n\textbackslash nBut}'') across repetition cycles.
As illustrated in Figure~\ref{fig:cosin_qs_case1}, the representation space contracts significantly. Unlike normal segments, repetitive loops exhibit a rapid increase in similarity across layers. Crucially, in deep cycles, cosine similarity saturates to nearly 1.0, while vector norm differences vanish. This state collapse confirms that the loop constitutes a distinct internal state, fundamentally different from normal reasoning.

\begin{figure*}[t]
    \centering
    \begin{subfigure}[b]{0.3\linewidth}
        \centering
        \includegraphics[width=\textwidth]{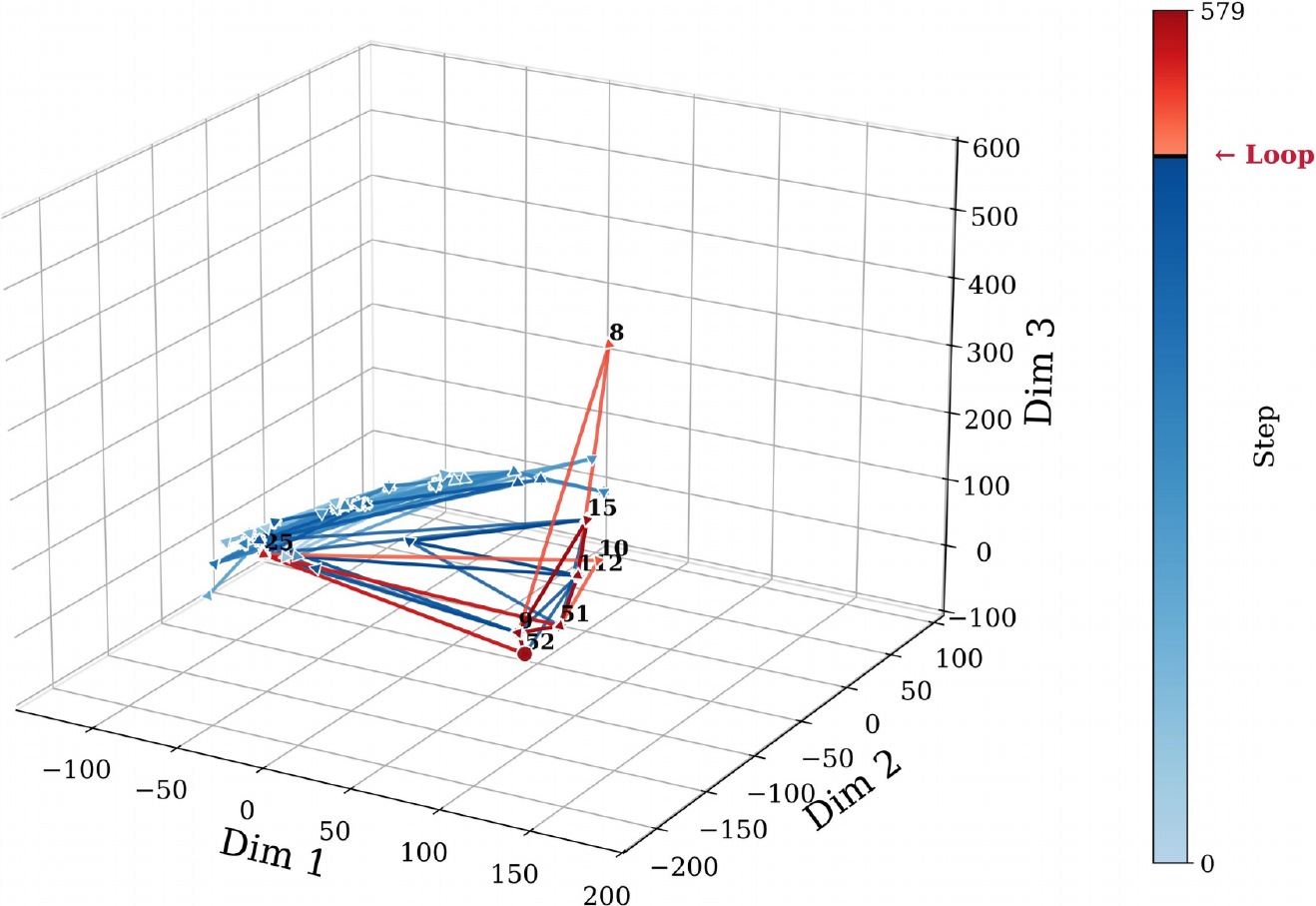}
        \caption{PCA Projection}
        \label{fig:case1_pca}
    \end{subfigure}
    \hfill 
    \begin{subfigure}[b]{0.3\linewidth}
        \centering
        \includegraphics[width=\textwidth]{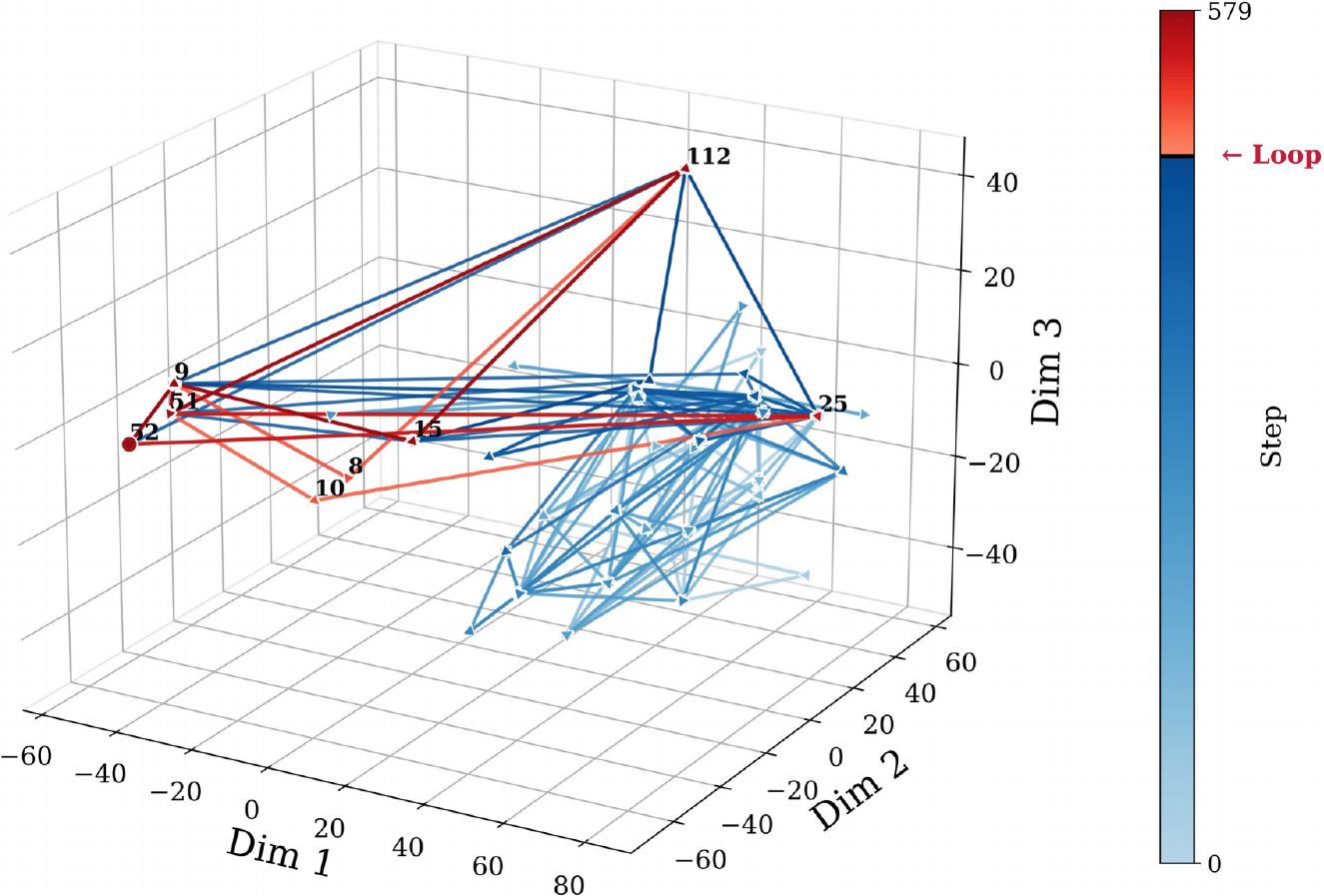}
        \caption{t-SNE Projection}
        \label{fig:case1_tsne}
    \end{subfigure}
    \hfill 
    \begin{subfigure}[b]{0.35\linewidth}
        \centering
        \includegraphics[width=\textwidth]{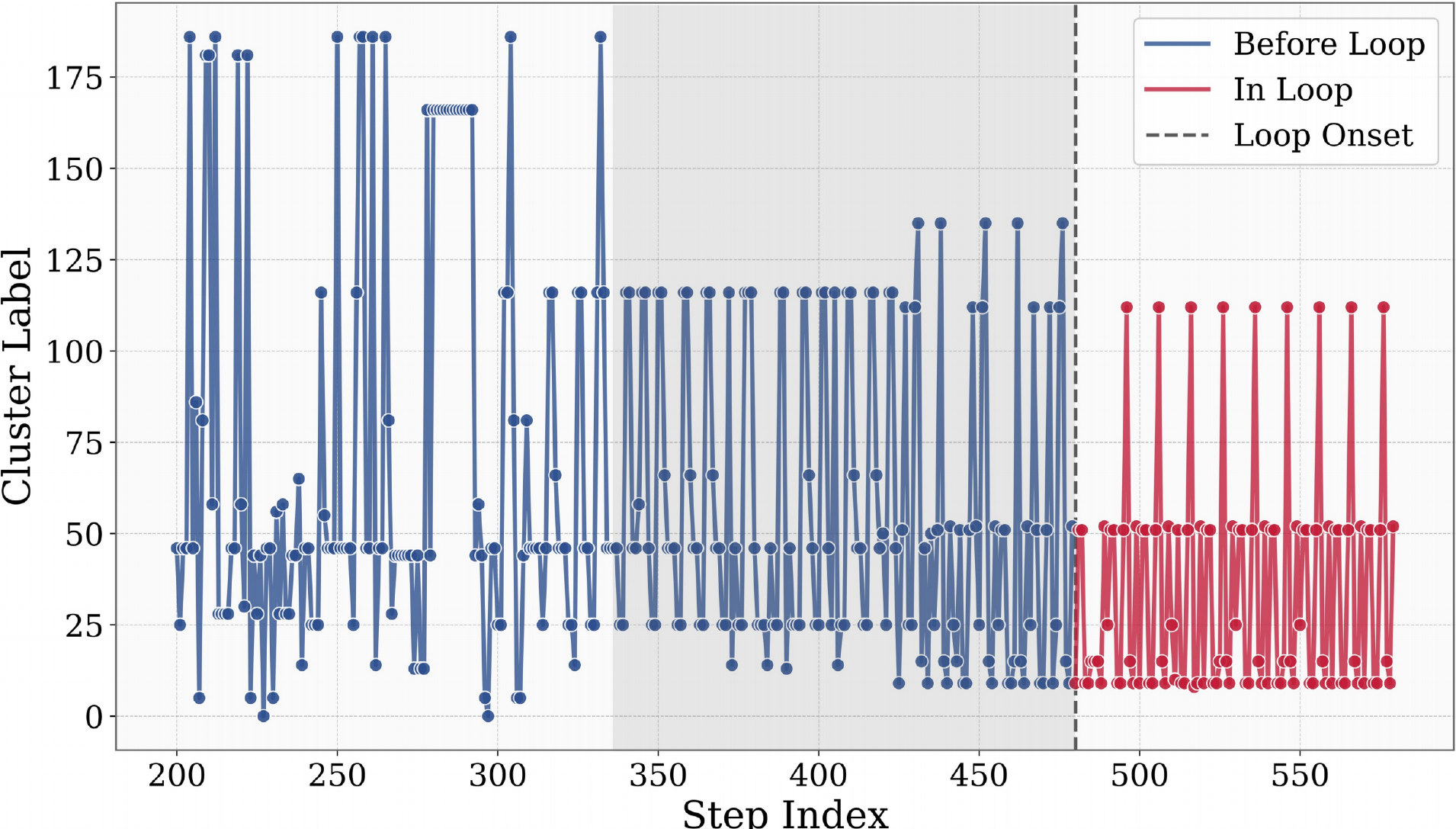}
        \caption{Cluster Trajectory Timeline}
        \label{fig:case1_path}
    \end{subfigure}
    \caption{\textbf{Reasoning graph visualization of a statement loop in DS-Qwen-14B.} (a) and (b) show internal states collapsing into cycles. Crucially, (c) plots cluster labels against sentence IDs, demonstrating that \textit{semantic circularity} (shaded region) significantly precedes explicit textual repetition (dashed line), serving as an early warning.}
    \label{fig:reasoning_graph_topology}
\end{figure*}

\paragraph{High Separability of Internal States.}
To quantify the distinction between repetitive and normal states, we train binary classifiers (Linear, SVM, and MLP) on model hidden states. We use averaged hidden states for statement loops and individual vectors for numerical loops (see Appendix for details).
As shown in Table~\ref{tab:classifier_performance}, all classifiers in both models achieved high accuracy, demonstrating that these two states have clear decision boundaries in the representation space. 
Qualitatively, we also observe that statement loops are often preceded by ``Self-correction attempt'' (e.g., \textit{``I think I cannot solve it. But maybe I can try ...''}), which serve as a strong semantic indicator of impending repetition.

\begin{table}
  \centering
  \caption{\textbf{Separability of loop representations.} The consistently high accuracy achieved by all classifiers confirms that repetitive and normal states possess distinct decision boundaries in the representation space.}
  \label{tab:classifier_performance}
  \scalebox{0.78}{
    \begin{tabular}{lcccccc}
      \toprule
      \multirow{2}{*}{Classifier} & \multicolumn{3}{c}{Numeral Loop} & \multicolumn{3}{c}{Statement Loop} \\
      \cmidrule(lr){2-4} \cmidrule(lr){5-7}
       & ACC & F1 & AUC & ACC & F1 & AUC \\
      \midrule
      \rowcolor{gray!15}   \multicolumn{7}{c}{\textit{\textbf{DS-Qwen-14B}}} \\  
      Linear & 0.991 &	0.992 &	0.998 &	0.998 &	0.998 &	1.000  \\
      SVM    & 0.992 &	0.992 &	0.999 &	0.998 &	0.999 &	0.999  \\
      MLP    & 0.991 &	0.991 &	0.996 &	0.997 &	0.997 &	1.000  \\
      \rowcolor{gray!15}   \multicolumn{7}{c}{\textit{\textbf{Qwen3-8B}}} \\  \addlinespace[3pt]
      Linear & 0.998 &	0.999 &	0.999 &	0.998 &	0.998 &	0.998  \\
      SVM    & 0.996 &	0.998 &	0.999 &	0.997 &	0.997 &	0.999  \\
      MLP    & 0.997 &	0.997 &	0.999 &	0.998 &	0.998 &	1.000  \\
      \bottomrule
    \end{tabular}
    }
\end{table}

\paragraph{Reasoning Graphs in Statement Loops.}
Following~\citep{Minegishi2025TopologyOR}, we construct a reasoning graph to analyze the model's reasoning path from a topological perspective. 
Specifically, within each statement loop case, we represent each generated sentence as a data point using its mean final-layer hidden state. 
We employ K-Means to group these states into $K=200$ clusters, where each cluster serves as a graph node. 
Consequently, the sequential transitions between clusters form directed edges, effectively transforming the text generation process into a trajectory on a graph.

We visualize these reasoning graph trajectories in 3D space by applying dimensionality reduction techniques (PCA and t-SNE) to the cluster nodes. This visualization reveals two critical topological characteristics. 
(1) First, upon entering the loop, the model's internal trajectory strictly traverses a fixed subgraph structure. As shown in Figures~\ref{fig:case1_pca} and \ref{fig:case1_tsne}, the trajectory collapses into a cycle, confirming that external statement repetition is underpinned by an internal topological lock-in.
(2) \textbf{More critically, we observe that semantic circularity precedes statement repetition.} As shown in Figure~\ref{fig:case1_path}, the node transitions converge into a periodic oscillation (shaded region) before the explicit loop onset (dashed line). \textit{While the generated sentences in this phase are lexically distinct, they exhibit high semantic redundancy, causing them to fall into recurrent cluster labels}. This indicates that the model is trapped in a ``semantic attractor'' prior to generating verbatim repetitive text.

\subsection{Loop Triggering Mechanisms}         \label{subsec:trigger}

   
We investigate the triggers of loop formation. 
Our analysis identifies reasoning impasses as the key factors driving the transition to repetition, although the specific triggering dynamics differ fundamentally between statement and numerical loops.

\begin{figure}[t]
    \centering
    \includegraphics[width=\linewidth]{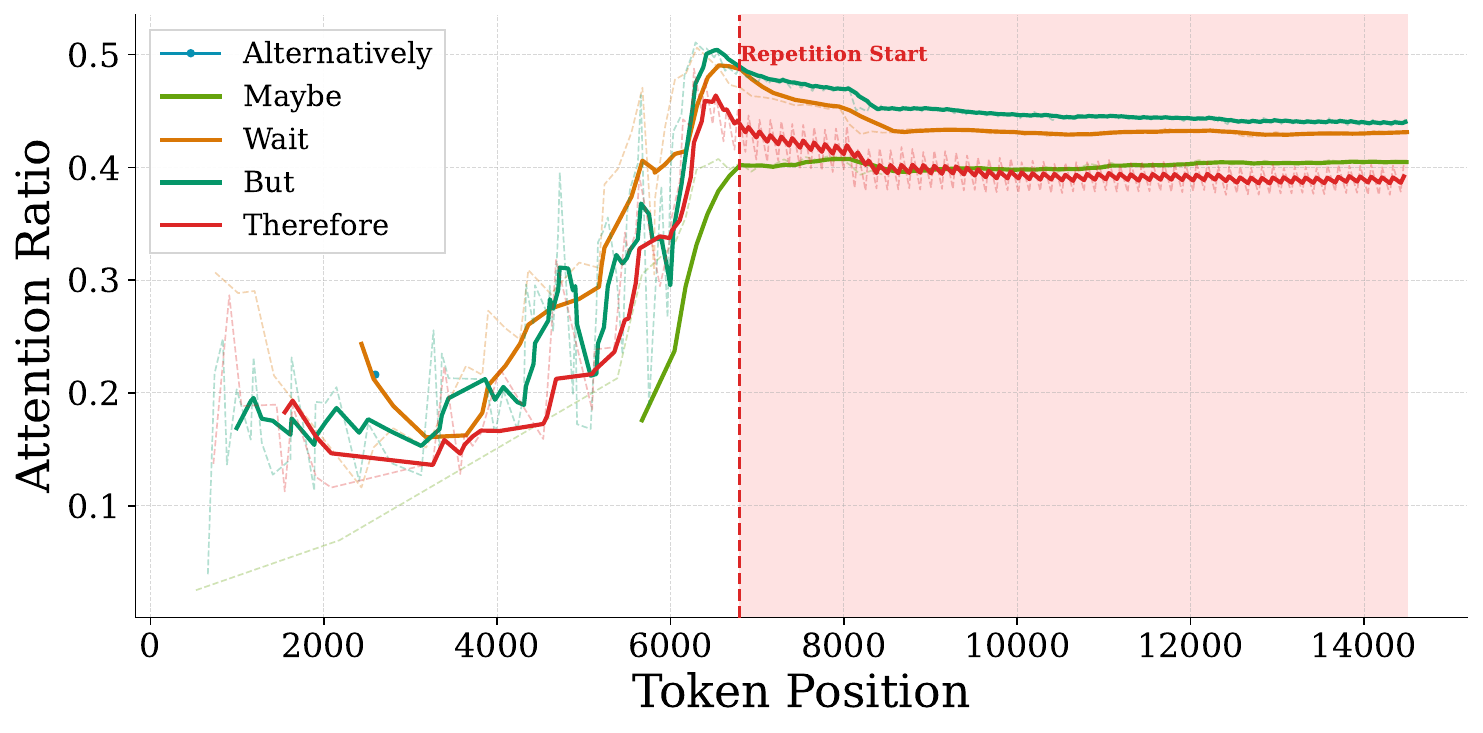}
    \caption{\textbf{Attention Distribution of High-Entropy Initial Tokens in DS-Qwen-14B.} We visualize the attention proportion assigned to preceding identical tokens. \textit{Note}: Attention sinks and recent tokens (last 128) are excluded. A distinct concentration of attention emerges prior to repetition onset.}
    \label{fig:attn-vertical}
\end{figure}

\begin{figure}[t]
    \centering
    \includegraphics[width=\linewidth]{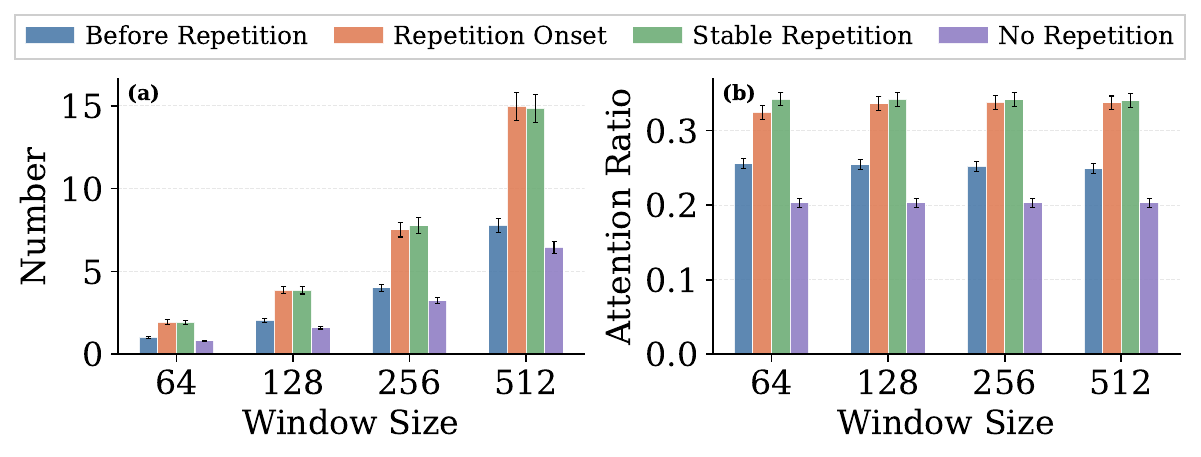}
    \caption{\textbf{Statistics of high-entropy tokens.} 
    We compare token counts and attention proportions. The ``Repetition Onset'' phase exhibits abnormally high metrics compared to stable or non-repeating phases, indicating an attention anomaly triggers the loop.}
    \label{fig:placeholder}
\end{figure}

\paragraph{Triggers of Statement Loops.} 

Our observations reveal that statement loops are frequently precipitated by reflective phrases acting as semantic pivots. 
These pivots signal an \textit{impasse where the model, unable to derive a solution yet unwilling to terminate, resorts to recursive retrials}, as illustrated in Figure~\ref{fig:case_study_Truth-Teller}. 
These semantic shifts, which manifest prior to the onset of explicit textual repetition, are composed primarily of dense high-entropy tokens (e.g., \textit{But''}, \textit{Wait''})~\cite{Wang2025BeyondT8}. They function as volatile decision junctions that inject significant uncertainty into the reasoning path, destabilizing the subsequent attention distribution.

By comparing sentences with and without repetition across five representative high-entropy tokens, we find that the model pays significantly more attention to these high-entropy tokens immediately before a loop begins.
As shown in Figure \ref{fig:attn-vertical}, both the density of these tokens and their accumulated attention weights remain consistently high in repeating cases. This confirms that such an attention imbalance is a key driver of statement-level loops.

To verify this quantitatively, we compared 50 loop and 50 non-loop samples across different reasoning phases. 
Following~\citep{Wang2025BeyondT8}, we curated a list of high-entropy tokens that typically appear at the start of sentences. Specifically, we examined the ``Repetition Onset'' (a fixed window preceding repetition) and ``Stable Repetition'' (a fixed window following repetition) phases against the pre-repetition history and non-repeating baselines.
As shown in Figure \ref{fig:placeholder}, both the density and attention share of these high-entropy tokens peak during the Repetition Onset phase. This confirms that a sudden shift in attention toward these tokens acts as a key precursor to repetition, signaling the transition from normal reasoning to a loop state.

Therefore, we conclude that statement loops are driven by \textbf{self-reinforcing attention}, where a high density of high-entropy tokens monopolizes the model's limited attentional capacity. Consequently, the model is entrapped in a recursive cycle of redundant self-reflections, thereby irreversibly derailing the progressive reasoning path.

\paragraph{Triggers of Numerical Loops.} 
In contrast to statement loops, numerical loops originate from distinct triggering mechanisms. 
They are primarily induced by arithmetic tasks necessitating long-form expansions with inherent periodicity. 
For instance (Figure~\ref{fig:Case Study: Square Root}), calculating square roots requires the model to repeatedly generate decimal periods (e.g., $0.0000\dots$). 
Such repetitive patterns rapidly saturate the recent context window and dominate the immediate history, effectively trapping the generation process in a self-reinforcing loop. Notably, this failure persists across both base and instruction-tuned models, suggesting a fundamental limitation in handling periodic sequences.

\subsection{Loop Persistence Mechanisms}        \label{subsec:vshape}

\begin{figure}[t]
    \centering
    \begin{subfigure}[b]{\linewidth}
        \centering
        \includegraphics[width=\linewidth]{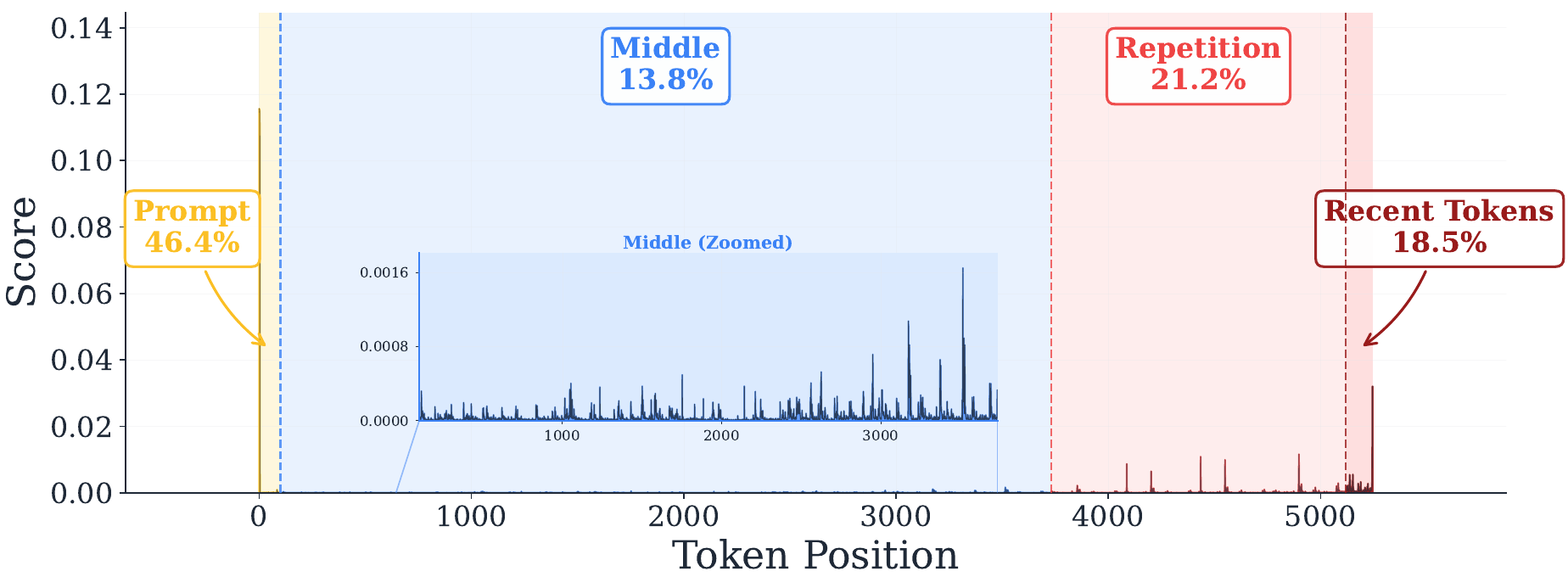}
        \caption{``V-shaped'' Attention Pattern}
        \label{fig:v-shape}
    \end{subfigure}
    \begin{subfigure}[b]{\linewidth}
        \centering
        \includegraphics[width=\linewidth]{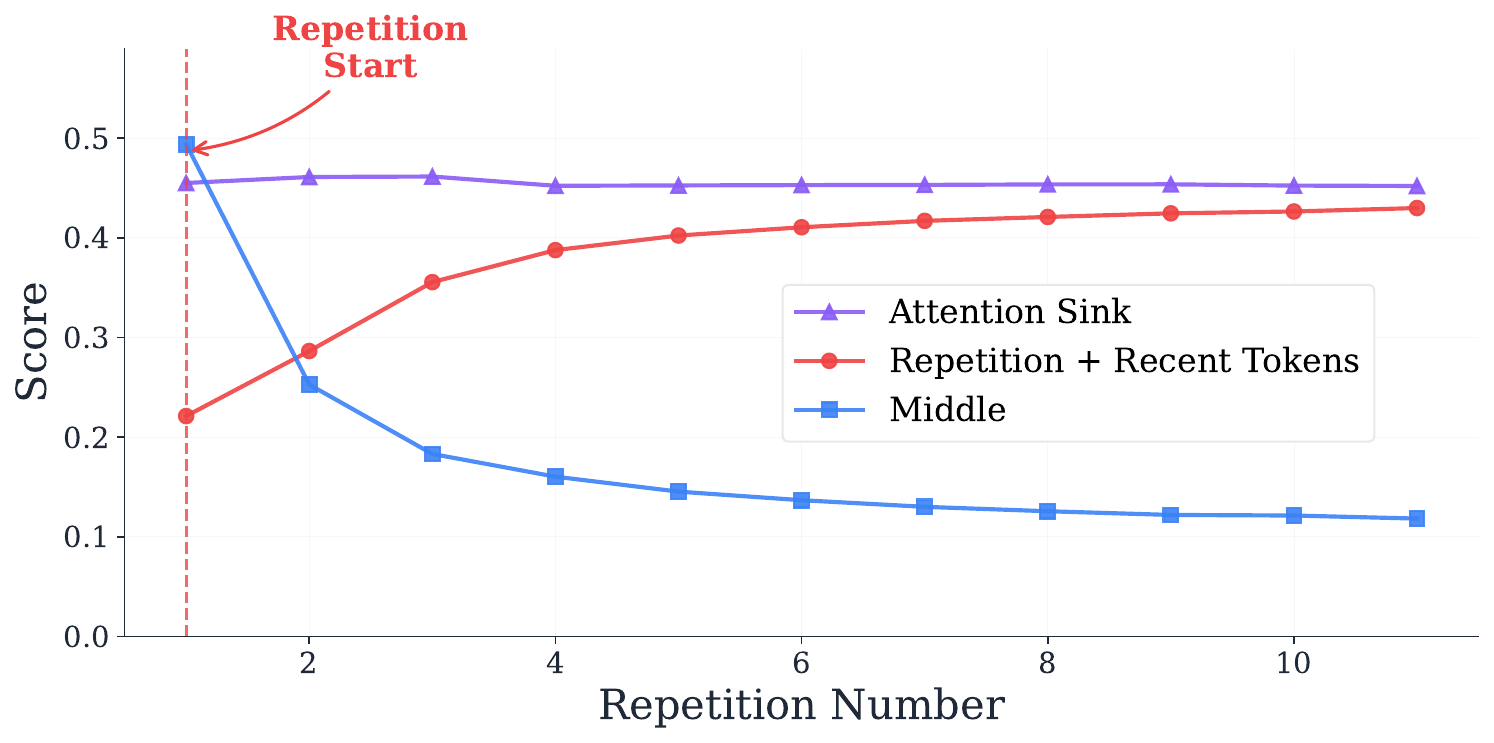}
        \caption{Trajectory of Attention Scores}
        \label{fig:cumul}
    \end{subfigure}
    \caption{\textbf{Attention Pattern in DS-Qwen-14B.} (a) The model focuses on sinks and recent tokens (last 128 tokens), and the ``Repetition'' region excludes this recent window. (b) Repetitive tokens attract increasing attention, making the loop nearly inescapable.}
    \label{fig:attention_pattern}
\end{figure}

We investigate the mechanism of loop persistence by analyzing the attention dynamics in large reasoning models. 
Our analysis reveals a consistent ``V-shaped'' attention pattern (Figure~\ref{fig:v-shape}). 
Specifically, during next-token prediction, attention weights are heavily concentrated on two distinct areas: the initial \textit{attention sinks} and the most recent tokens. 
This observation aligns with prior findings~\cite{fu2024attentionpattern,Xiao2023EfficientSL}.
However, while efficient for normal generation, this mechanism becomes detrimental once repetition initiates. 
As repetitive sequences saturate the recent context window, the attention allocated to these local tokens rises abnormally (Figure~\ref{fig:cumul}).
This creates a self-reinforcing feedback loop: the model focuses excessively on its own recent repetitions, which in turn boosts the confidence of generating the same tokens again. 
Consequently, this dynamic traps the model in a closed attractor state, confining generation to a recursive trajectory of identical content, thereby making the loop nearly inescapable.

%% file: latex/sections/06_method.tex
\section{Early Prediction of Loop Onset}    \label{sec:detection}




Predicting repetition prior to its manifestation is crucial for enhancing model reliability and ensuring computational efficiency. 
Since \textit{numerical loops} is often an intrinsic requirement of calculation, our method specifically targets \textit{\textbf{statement loops}}. We leveraging the observation that they are frequently precipitated by a surge of high-entropy reflection tokens. 
This semantic shift induces a distinct state transition in the hidden states, which we utilize as a warning signal to forecast loop risks before the onset of explicit textual repetition.



\begin{figure}
    \centering
    \includegraphics[width=\linewidth]{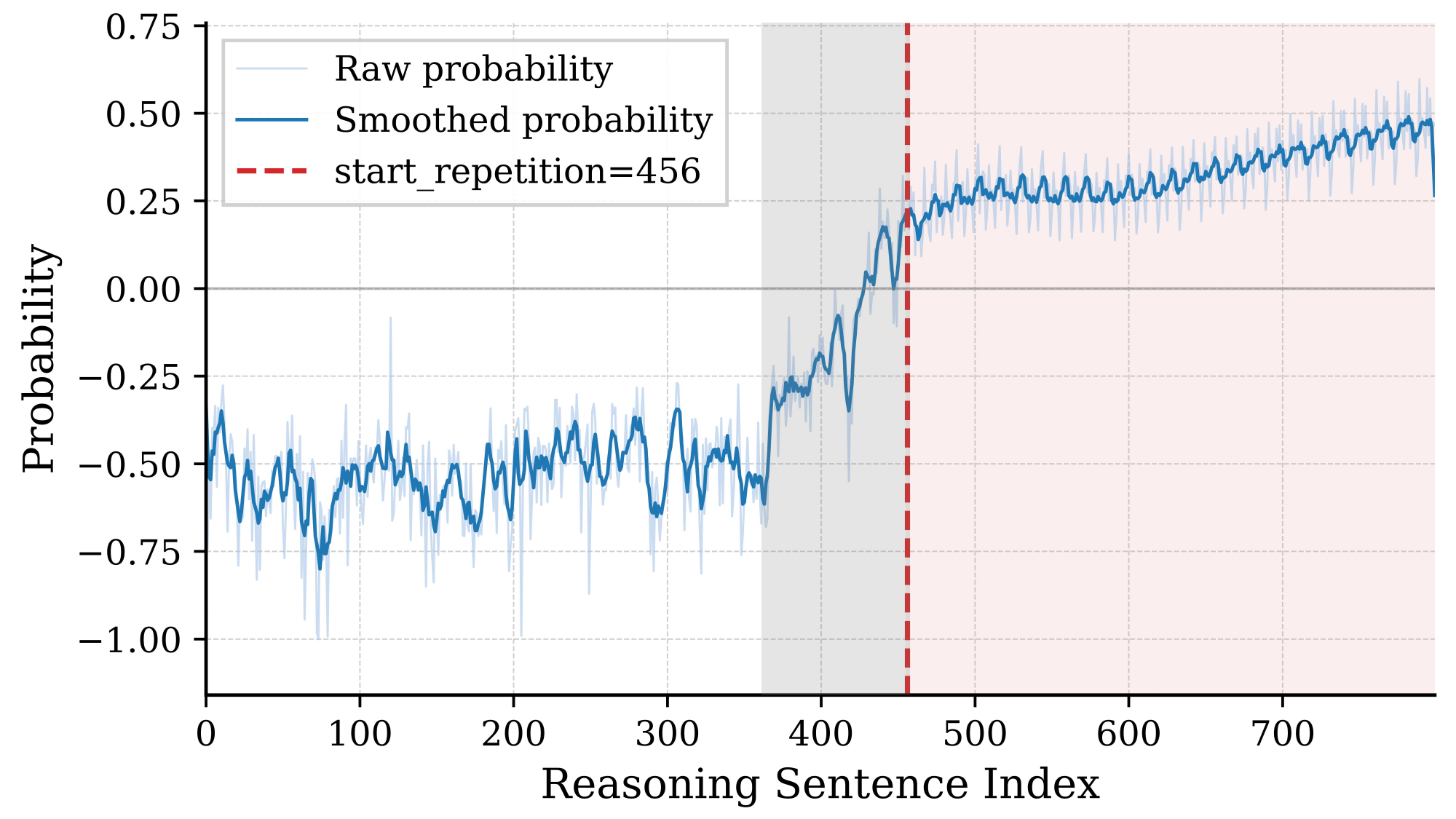}
    \caption{\textbf{Prediction score for a loop case.} The red area indicates repetition onset; the gray area highlights the sharp probability rise preceding repetition.}
    \label{fig:hidden-detect}
\end{figure}

To detect this transition, we apply the linear classifier from Section~\ref{subsec:phase transition} to monitor the model's internal states. Specifically, for each sentence $i$, we take the average of its hidden states at the last layer (denoted as $h_i$) and compute a prediction score:
\begin{equation}
    x_i = w^\top h_i + b
\end{equation}
As shown in Figure~\ref{fig:hidden-detect}, a predicted score $x_i > 0$ indicates that the model is likely entering a loop state. We observe that this probability rises sharply just before repetition starts, a trend that is absent in normal reasoning. This confirms that hidden states provide a warning signal before the model fails, allowing us to catch the transition from normal reasoning to loops by monitoring their trajectory.


\paragraph{Real-time Detection via Cumulative Sum.}

We formulate repetition detection as a sequential change-point detection problem employing the Cumulative Sum (CUSUM) algorithm~\cite{Saxena2015Sequential}. Specifically, we track the cumulative deviation $S_i$ based on the score $x_i$ via the update:
\begin{equation}
    S_i = \max(0, \ S_{i-1} + (x_i - r))
\end{equation}
where $r$ is a reference value calibrated to the mean of non-repetitive cases. This formulation ensures that $S_i$ accumulates only systematic deviations exceeding $r$, effectively filtering out transient noise. 

To further exclude momentary spikes, intervention is triggered only when the condition $S_i > h$ persists for $p$ consecutive steps. To minimize false positives, we calibrate $h = \alpha \cdot S_{max}$ on a held-out normal set, where $S_{max}$ denotes the maximum observed statistic.



\paragraph{Experimental Setup.}
We evaluate our method on the LoopBench dataset using greedy decoding to ensure reproducibility. For CUSUM calibration, we sample 50 non-loop instances per model. From the remaining data, we construct a balanced test set for each model, comprising at least 50 repetitive and 50 non-repetitive cases. For models that do not meet the minimum requirement of 50 loop cases in LoopBench, we exclude them from this specific analysis to maintain statistical reliability. 
Performance is measured using four metrics: Early Detection Rate (EDR), False Positive Rate (FPR), Average Sentence Earliness (ASE), and Average Token Earliness (ATE). These metrics assess the ability to anticipate repetition while quantifying the risk of misclassifying normal sequences. Further details on metric definitions and hyperparameter settings are provided in Appendix~\ref{sec:setup_pre}.


\paragraph{Main Results}

Table~\ref{tab:predict_results} presents the performance of our prediction strategy across diverse model architectures. We summarize the key findings as follows:
(1) \textbf{High Predictability with Low False Alarms.}
Our method achieves a high Early Detection Rate (EDR) across all models, validating the existence of detectable warning signals. Ablation studies underscore the necessity of the persistence mechanism. While the baseline ($p=1$) yields marginally higher recall, it suffers from prohibitive False Positive Rates (FPR), often exceeding 50\%. 
(2) \textbf{Significant Early Warning Window.}
The temporal metrics show that our method predicts loops well before they happen. On average, we detect the warning signal over 40 sentences (or 1500 tokens) before repetition starts. This validates our finding that semantic circularity precedes textual repetition, providing enough time for intervention.
(3) \textbf{Robustness Across Architectures.}
The proposed method demonstrates consistent efficacy across models of varying sizes and families (Qwen, Llama, Phi, GPT). This robustness is particularly highlighted by the results on Phi-4-reasoning: despite being smaller and more prone to instability, our approach effectively detects loop onset while maintaining a low False Positive Rate.

\begin{table}[t]
\centering
\caption{\textbf{Results on statement loop prediction.} The performance of our method (with persistence) compared to the baseline without persistence ($p=1$) shows that our persistence effectively reduces the FPR.}
\label{tab:predict_results}
\resizebox{\columnwidth}{!}{%
\begin{tabular}{lcccc} 
\toprule
\textbf{Model} & \textbf{EDR} ($\uparrow$) & \textbf{FPR} ($\downarrow$) & \textbf{ASLT} ($\uparrow$) & \textbf{ATLT} ($\uparrow$) \\
\midrule

\textbf{DeepSeek-R1-Distill-Qwen-7B}       & 0.74 & \textbf{0.24} & 44.1 & 1663.5 \\
\quad \textit{- w/o $p=1$} & 0.82 & 0.44 & 49.5 & 1870.2 \\
\addlinespace[4pt]

\textbf{DeepSeek-R1-Distill-Llama-8B}      & 0.76 & \textbf{0.34} & 46.2 & 1871.6 \\
\quad \textit{- w/o $p=1$} & 0.82 & 0.48 & 52.1 & 1980.4 \\
\addlinespace[4pt]

\textbf{DeepSeek-R1-Distill-Qwen-14B}      & 0.76 & \textbf{0.28} & 48.9 & 1865.4 \\
\quad \textit{- w/o $p=1$} & 0.84 & 0.40 & 55.3 & 2272.8 \\
\addlinespace[4pt]

\textbf{DeepSeek-R1-Distill-Qwen-32B}      & 0.72 & \textbf{0.30} & 36.5 & 1305.9 \\
\quad \textit{- w/o $p=1$} & 0.78 & 0.54 & 41.8 & 1515.1 \\
\addlinespace[4pt]

\textbf{Qwen3-8B}         & 0.64 & \textbf{0.30} & 44.6 & 1815.4 \\
\quad \textit{- w/o $p=1$} & 0.72 & 0.52 & 50.2 & 2025.8 \\
\addlinespace[4pt]

\textbf{gpt-oss-20B}      & 0.76 & \textbf{0.32} & 51.4 & 1750.8 \\
\quad \textit{- w/o $p=1$} & 0.82 & 0.48 & 53.1 & 1862.3 \\
\addlinespace[4pt]

\textbf{Phi-4-reasoning}  & 0.74 & \textbf{0.28} & 40.5 & 1420.6 \\
\quad \textit{- w/o $p=1$} & 0.84 & 0.62 & 46.8 & 1635.2 \\
\addlinespace[4pt]

\textbf{Phi-4-reasoning-plus} & 0.72 & \textbf{0.26} & 42.1 & 1455.9 \\
\quad \textit{- w/o $p=1$} & 0.86 & 0.58 & 48.3 & 1468.5 \\

\bottomrule
\end{tabular}%
}
\end{table}

%% file: latex/sections/02_related-work.tex
\section{Related Work}


\paragraph{Neural Text Degeneration.}
Although prominent in LRMs, looping originates from classic neural text degeneration, where models devolve into ``repeaters'' \cite{holtzman2020curious}. This pathology has been addressed across the pipeline:
(1) \textbf{Data Perspective.} \citet{hernandez2022repetition} identified corpus duplication as a primary driver via the ``Repetition In, Repetition Out'' principle. Thus, strict deduplication, from exact matching to semantic pruning \cite{lee2022deduplicating, abbas2023semdedup}, is critical for prevention.
(2) \textbf{Training Objectives.} To mitigate the degeneration induced by standard MLE \cite{Welleck2019NeuralTG}, objectives evolved to explicitly penalize repetition (e.g., Unlikelihood Training) or improve isotropy \cite{su2022contrastive}. Methods like ``Straight to the Gradient'' \cite{lin2021straight} further reduce exposure bias by optimizing evaluation metrics directly.
(3) \textbf{Decoding Strategies.} To escape the ``likelihood trap,'' solutions include stochastic methods like Nucleus Sampling \cite{holtzman2020curious} for diversity, and deterministic approaches like Contrastive Search \cite{su2022contrastive} or explicit penalties \cite{keskar2019ctrl} that balance coherence without randomness.



Despite existing solutions for LLMs, the issue re-emerges in Large Reasoning Models (LRMs) as \textit{circular reasoning}. Driven by the reflective and ``trial-and-error'' nature of CoT \cite{Yuan2023ScalingRO,xie2025word,wei2025stop}, these meaningful logic loops evade standard deduplication methods effective for simple n-gram repetition.

\paragraph{Mechanistic Interpretability.}
The goal of mechanistic interpretability is to reverse-engineer model behaviors \citep{elhage2021mathematical,pang2025large} by mapping distinct functional properties \citep{duan2025gloss,todd2023function,deng2025latent,wei2025stop} to identifiable components within LLMs.
These components span various granularities, including neurons \citep{dai-etal-2022-knowledge}, multi-headed self-attention (MHSA) \citep{ge2025focusing, ge2025innate}, feed-forward networks (FFN) \citep{deng2024everything, duan2025related}, layers \citep{Zhao2024DefendingLL}, and circuits \citep{yao2024knowledge}. 
Specifically regarding repetition, recent works attribute this phenomenon to distinct internal mechanisms, such as Induction Heads \citep{olsson2022context} in attention layers and specialized repetition neurons within MLPs \citep{hiraoka-inui-2025-repetition}. 
Aligning with this direction, we primarily analyze the loop behavior of LRMs through the lens of internal hidden states and attention mechanisms.


%% file: latex/sections/07_conclusion.tex
\section{Conclusion}

In this paper, we characterize \textit{Circular Reasoning} via LoopBench, identifying loops as deterministic phase transitions where semantic circularity precedes textual repetition. Mechanistically, these are triggered by specific task demands and sustained by `V-shaped' attention. Finally, we validate a CUSUM-based strategy to anticipate loop formation and ensure reasoning stability.

\section*{Limitations}

We acknowledge three main limitations in this paper. First, our investigation concentrates on high-precision arithmetic and recursive reasoning tasks; consequently, loop dynamics in other domains remain to be fully explored. Second, as our mechanistic analysis primarily relies on the DeepSeek and Qwen families, the architectural universality of these findings across other model types requires further verification. Finally, our current intervention serves as an inference-time mitigation specifically optimized for statement loops, rather than a fundamental training-stage resolution that systematically eradicates the underlying pathology at its source.

\section*{Ethical Considerations}

This work primarily contributes to the efficiency and sustainability of Large Reasoning Models. By enabling the early detection and termination of infinite reasoning loops, our approach significantly mitigates the computational waste and energy consumption associated with invalid recursive generation, aligning with Green AI principles. 
While we characterize specific loop triggers that could theoretically be exploited for resource-exhaustion attacks (e.g., denial-of-service), we posit that exposing these mechanistic vulnerabilities is a prerequisite for developing robust defense mechanisms. Furthermore, our LoopBench is derived exclusively from public benchmarks, involving no personally identifiable information or safety concerns.

%% file: latex/sections/X1_dataset.tex
\section{Detail of LoopBench}   \label{sec:LoopBench_Detail}

To systematically evaluate the robustness of LRMs against repetitive generation, we curated LoopBench based on two core design principles: (1) High-Precision Arithmetic, which demands rigorous numerical fidelity, and (2) Recursive Iterative Logic, which requires maintaining clear state tracking over long contexts. 
Guided by these principles, we select 7 specific sub-tasks (formal definitions provided in Table \ref{tab:loopbench_tasks}). For each sub-task, we utilized GPT-5\footnote{\url{https://platform.openai.com/docs/models/gpt-5}} to synthesize 100 distinct samples. To ensure the quality and validity of the benchmark, we applied the following strict filtering criteria during construction:

\begin{enumerate}
    \item \textit{\textbf{Bounded Precision Constraints:}} For arithmetic tasks, the required decimal precision is capped at \textbf{500 digits}. This threshold allows us to rigorously distinguish between legitimate long calculations and invalid infinite loops.
    
    \item \textit{\textbf{Substantial Reasoning Depth:}} We strictly filter out trivial queries solvable via direct knowledge retrieval. All selected samples necessitate extensive, multi-step derivation processes (\textbf{typically exceeding 100 reasoning steps}) to impose significant pressure on the model's long-context capabilities.
\end{enumerate}

To standardize the evaluation environment and trigger the model's maximum reasoning potential, a unified system instruction is applied across all queries, as illustrated in Figure \ref{fig:system_prompt}. This instruction explicitly mandates \textbf{rigorous, step-by-step derivation}, strictly precluding models from bypassing the reasoning process through memorized heuristics or short-cuts.
Table \ref{tab:loopbench_cases} details representative input queries for each sub-task. The benchmark challenges models in two distinct dimensions:

\begin{itemize}
    \item \textit{\textbf{High-Precision Arithmetic Tasks}}: These tasks necessitate unwavering numerical fidelity(e.g., \textit{Square Root Calculation}). Models must sustain accuracy across hundreds of decimal places without error accumulation, where a single digit error can propagate and trigger digital loop.
    
    \item \textit{\textbf{Complex Recursive Reasoning Tasks}}: These tasks challenge long-context state tracking(e.g., \textit{Logical Paradox Analysis}). They compel models to navigate extensive search spaces and manage intricate dependencies without losing logical coherence, testing their resistance to statement loops.
\end{itemize}




\begin{figure}[t]
    \centering
    \begin{tcolorbox}[
        colback=orange!15, 
        colframe=orange!60, 
        title=\textbf{Unified System Instruction},
        fonttitle=\bfseries\small,
        boxrule=0.5pt
    ]
    \small
    ``You are a meticulous, conscientious, and by-the-book \textit{[role]} who must present, step by step, without skipping any steps or omitting any intermediate quantities, the derivation and update of each digit/step...''
    \end{tcolorbox}
    \caption{The unified system instruction applied across LoopBench tasks to enforce reasoning generation.}
    \label{fig:system_prompt}
\end{figure}

\begin{table*}[t]
    \centering
    \caption{\textbf{Overview of Task Formulations.} The prompt templates utilize parameterized variables (e.g., $\{N\}, \{K\}$) to dynamically generate diverse instances. \textbf{Note:} All tasks strictly require the model to explicitly document every intermediate reasoning step (e.g., remainders, state updates, or recursive moves) before generating the final answer.}
    \label{tab:loopbench_tasks}
    \vspace{2mm}

    \begin{tabularx}{\linewidth}{@{} l X @{}}
        \toprule
        \textbf{Sub-task} & \textbf{Task Formulation} \\
        \midrule
        \multicolumn{2}{l}{%
            \cellcolor{gray!25}%
            \begin{minipage}{\dimexpr\linewidth-2\tabcolsep}
                \vspace{2pt} 
                \textbf{Task: High-Precision Arithmetic} \hfill 
                \small \textit{Setting: $\{N\}, \{A\}, \{B\}$ are large integers; $\{K\}$ is decimal precision.}
                \vspace{2pt}
            \end{minipage}%
        } \\
        \addlinespace[4pt] 
        
        \textbf{Square Root} & Compute $\sqrt{\{N\}}$ to $\{K\}$ decimal digits via digit-by-digit extraction.\\
        \addlinespace[3pt] 
        
        \textbf{Long Division} & Compute $\{A\} \div \{B\}$ to yield $\{K\}$ decimal digits using standard long division. \\
        \addlinespace[3pt]
        
        \textbf{Newton's Iteration} & Approximate $\sqrt[m]{\{N\}}$ to $\{K\}$ decimal places via Newton's method (initial $x_0=\{I\}$), deriving the exact value update $x_n \to x_{n+1}$ step-by-step.\\
        
        \midrule
        
        \multicolumn{2}{l}{%
            \cellcolor{gray!25}%
            \begin{minipage}{\dimexpr\linewidth-2\tabcolsep}
                \vspace{2pt}
                \textbf{Task: Complex Recursive Reasoning} \hfill 
                \small \textit{Setting: $\{N\}$ agents/disks; $\{T\}$ rounds; $\{D\}$ grid dimensions.}
                \vspace{2pt}
            \end{minipage}%
        } \\
        \addlinespace[4pt]
        
        \textbf{Truth-Teller Puzzles} & Identify valid truth-value assignments for $\{N\}$ circularly linked agents. Each agent's statement constrains neighbors (modulo $\{N\}$), requiring checking. \\
        \addlinespace[3pt]
        
        \textbf{Logical Paradox} & Deduce states in a grid based on public constraints and a sequence of epistemic updates, where a specific agent gains knowledge exactly at Round $\{T\}$. \\
        \addlinespace[3pt]
        
        \textbf{Tower of Hanoi} & Generate the optimal recursive sequence to move $\{N\}$ disks from Source to Target, strictly adhering the larger disks cannot be placed on smaller ones. \\
        \addlinespace[3pt]
        
        \textbf{Path Planning} & Compute the shortest path in a $\{D\}$-dimensional grid from Start to Goal, avoiding implicit obstacles defined by modular arithmetic constraints. \\
        
        \bottomrule
    \end{tabularx}
\end{table*}

\begin{table*}[t]
\centering
\small
\renewcommand\tabularxcolumn[1]{m{#1}} 
\caption{\textbf{Representative Instances from LoopBench.} The table shows selected input queries for each sub-task, characterized by requirements for extreme precision and complex constraints.}
\label{tab:loopbench_cases}
\begin{tabularx}{\textwidth}{c c X} 
    \toprule
    \textbf{Category} & \textbf{Sub-Task} & \textbf{Problem} \\ 
    \midrule
    \multirow{17}{*}{\textbf{\makecell{High-Precision\\Arithmetic}}} 
    & \textbf{Square Root} 
    & Compute $\sqrt{99980001}$ using the standard digit-by-digit square-root algorithm. After obtaining the integer part, continue the same process to generate 700 decimal digits. At every step, show the chosen trial digit and update the intermediate remainder, bringing down digit pairs accordingly. \textbf{Provide the full sequence of moves and the resulting 700-digit decimal expansion.} \\  
    \cmidrule{2-3} 

    & \textbf{Long Division} 
    & Compute $1 \div 13{,}631$ using the standard step-by-step long division algorithm. Since the integer part is 0, carry out the division to obtain exactly 350 digits after the decimal point in the quotient. At each step of the long division, record the current remainder and the next quotient digit it produces. \textbf{Finally, give the resulting decimal expansion with all 350 digits.} \\  
    \cmidrule{2-3}

    & \textbf{Newton's Iteration} 
    & Use Newton's method to compute $\sqrt[5]{590255551}$ to 500 correct decimal places. Consider the equation $f(x)=x^5-590255551=0$ and start from the initial guess $x_0=50$. Apply the Newton iteration $x_{n+1}=x_n-\frac{f(x_n)}{f'(x_n)}$, carrying all computations with sufficient precision so that the final iterate $x_N$ satisfies $|x_N^5-590255551|<10^{-500}$, which guarantees at least 500 correct decimal digits of the root. For each iteration, explicitly show the derivation of the update from $x_n$ to $x_{n+1}$, and at the end, \textbf{report the decimal expansion of $\sqrt[5]{590255551}$ to 500 digits after the decimal point.} \\  
    \midrule

    \multirow{32}{*}{\textbf{\makecell{Complex Recursion\\Reasoning}}} 
    & \textbf{Truth-Teller Puzzles} 
    & There are 30 people, labeled $C_1$ through $C_{30}$, standing in a circle. Each person is either a truth-teller or a liar. For each $i$, person $C_i$ states: ``Exactly two of $C_{i+1}, C_{i+2}, C_{i+3}$ are lying,'' where indices are taken modulo 30 (so $C_{31}=C_1, C_{32}=C_2, C_{33}=C_3$). Determine all possible truth-value assignments that satisfy all 30 statements simultaneously, and for each solution verify completely that every statement is consistent with its speaker's status. \\ 
    \cmidrule{2-3}

    & \textbf{Logical Paradox} 
    & \textbf{Sixteen perfectly rational logicians}, labeled $L_1$--$L_{16}$, \textbf{stand in a $4 \times 4$ grid}, each with a hidden integer from 1 to 20 in front of them, all distinct. Each logician can see the other seven numbers in their own row and column but cannot see their own. The host publicly announces that the sum of every row, every column, and both main diagonals is 50; additionally, the four corner positions contain prime numbers, and at least two of the central four values ($L_6, L_7, L_{10}, L_{11}$) are even. Beginning from Round 1, in each round every logician states either ``I don't know my number'' or ``I know my number,'' and during the first 35 rounds all of them say they do not know. However, in Round 36, when it is $L_1$'s turn, $L_1$ declares: ``I know my number now.'' Determine the unique $4 \times 4$ grid of numbers and explain how $L_1$ logically deduced their value in Round 36. \\ 
    \cmidrule{2-3}

    & \textbf{Tower of Hanoi} 
    & Consider the classical Tower of Hanoi puzzle with three rods, labeled A, B, and C, and \textbf{20 disks} of distinct sizes initially stacked on rod A in order of decreasing size from bottom to top (largest at the bottom, smallest at the top). A legal move consists of taking the top disk from one rod and placing it onto another rod, subject to the rule that no disk may ever be placed on top of a smaller disk. The goal is to move the entire stack from rod A to rod C using rod B as an auxiliary, following only legal moves. Solve this instance of the Tower of Hanoi by describing the sequence of moves needed to transfer all 20 disks from rod A to rod C using the minimal number of moves, and present either the full list of all $2^{20}-1$ moves or \textbf{a clearly described and correct pattern that generates this optimal sequence.} \\  
    \cmidrule{2-3}

    & \textbf{Path Planning} 
    & Consider a 4-dimensional grid of integer lattice points $(x_1,x_2,x_3,x_4)$ with $0 \le x_i \le 9$ for all $i$. You start at $(0,0,0,0)$ and want to reach $(9,9,9,9)$. In one move, you may go from a point $(x_1,x_2,x_3,x_4)$ to any point that differs by exactly $\pm 1$ in one coordinate and is equal in the other three coordinates (i.e., along one of the 8 axis directions $\pm x_1,\pm x_2,\pm x_3,\pm x_4$), provided the new point stays within the grid. A grid point $(x_1,x_2,x_3,x_4)$ is considered an obstacle and cannot be visited if the sum $x_1 + x_2 + x_3 + x_4$ is divisible by 3, except that the start $(0,0,0,0)$ and goal $(9,9,9,9)$ are always allowed even though they satisfy this condition. \textbf{Find a shortest path from $(0,0,0,0)$ to $(9,9,9,9)$ that avoids all obstacle points, and output the sequence of coordinates along this path.} \\ 
    \bottomrule
\end{tabularx}
\end{table*}

\section{Details of Observational Analysis}

\subsection{Detailed Setup}  \label{subsec:detail_setup}





\paragraph{Models Selection.}

we provide the detailed specifications and access points for the models evaluated in LoopBench. Our model suite spans open-source weights and commercial APIs, covering varying parameter scales and training paradigms. We categorize open-source models into three groups: Reasoning, Base, and Instruct models. 

We evaluate the following representative open-weights \textbf{Reasoning Models} that demonstrate advanced reasoning capabilities:
\begin{itemize}
    \item \textbf{\textit{DeepSeek-R1-Distill Series}}: We evaluate the distilled versions of DeepSeek-R1 based on Qwen and Llama architectures. Specifically, we use DeepSeek-R1-Distill-Qwen-7B\footnote{\url{https://huggingface.co/deepseek-ai/DeepSeek-R1-Distill-Qwen-7B}}, DeepSeek-R1-Distill-Qwen-14B\footnote{\url{https://huggingface.co/deepseek-ai/DeepSeek-R1-Distill-Qwen-14B}}, DeepSeek-R1-Distill-Qwen-32B\footnote{\url{https://huggingface.co/deepseek-ai/DeepSeek-R1-Distill-Qwen-32B}}, DeepSeek-R1-Distill-Llama-8B\footnote{\url{https://huggingface.co/deepseek-ai/DeepSeek-R1-Distill-Llama-8B}}, DeepSeek-R1-0528-Qwen3-8B\footnote{\url{https://huggingface.co/deepseek-ai/DeepSeek-R1-0528-Qwen3-8B}}.

    \item \textbf{\textit{Qwen Series}}: We select the reasoning model QwQ-32B\footnote{\url{https://huggingface.co/Qwen/QwQ-32B}}, alongside the latest Qwen3 series, specifically the 8B\footnote{\url{https://huggingface.co/Qwen/Qwen3-8B}}, 14B\footnote{\url{https://huggingface.co/Qwen/Qwen3-14B}}, and 32B\footnote{\url{https://huggingface.co/Qwen/Qwen3-32B}} variants.
    
    \item \textbf{\textit{gpt-oss Series}}: We select gpt-oss-20B\footnote{\url{https://huggingface.co/openai/gpt-oss-20b}}, OpenAI's open-weight Mixture-of-Experts (MoE) model, which serves as a representative baseline with strong reasoning capability.

    \item \textbf{\textit{Phi-4 Series}}: We evaluate Phi-4-reasoning\footnote{\url{https://huggingface.co/microsoft/Phi-4-reasoning}} and its enhanced variant, Phi-4-reasoning-plus\footnote{\url{https://huggingface.co/microsoft/Phi-4-reasoning-plus}}. These models are distinguished by their capabilities.
\end{itemize}

To serve as a control group for analyzing the loop behavior, we evaluate the following representative \textbf{Base Models} without instruction tuning:

\begin{itemize}
    \item \textbf{\textit{Qwen Base}}: We utilize the model variants of the Qwen3 series, specifically Qwen3-8B-Base\footnote{\url{https://huggingface.co/Qwen/Qwen3-8B-Base}} and Qwen3-14B-Base\footnote{\url{https://huggingface.co/Qwen/Qwen3-14B-Base}}, to establish a performance baseline before tuning.
    
    \item \textbf{\textit{Gemma Base}}: We include Gemma-2-27B\footnote{\url{https://huggingface.co/google/gemma-2-27b}}, a high-capacity base model representing Google's open-weights family.
\end{itemize}

We examine the following \textbf{Instruct Models}, which, unlike the LRMs, typically demonstrate weaker complex reasoning capabilities:

\begin{itemize}
    \item \textbf{\textit{Llama Series}}: We select Llama-3.1-8B-Instruct\footnote{\url{https://huggingface.co/meta-llama/Llama-3.1-8B-Instruct}}, which serves as a widely adopted industry standard for instruction following.

    \item \textbf{\textit{Qwen Series}}: We evaluate the instruction-tuned counterparts of the Qwen family, specifically Qwen2.5-7B-Instruct\footnote{\url{https://huggingface.co/Qwen/Qwen2.5-7B-Instruct}}, Qwen2.5-14B-Instruct\footnote{\url{https://huggingface.co/Qwen/Qwen2.5-14B-Instruct}}, and Qwen2.5-32B-Instruct\footnote{\url{https://huggingface.co/Qwen/Qwen2.5-32B-Instruct}}. Including these allows for a direct comparison with their Base and Reasoning variants.

    \item \textbf{\textit{Gemma Series}}: We include Gemma-2-27B-it\footnote{\url{https://huggingface.co/google/gemma-2-27b-it}}, ensuring diverse architecture from the Google open-weights ecosystem.
\end{itemize}

For \textbf{Closed-Source Models}, we utilize their API services. To isolate the impact of reasoning generation, we explicitly compare the \textbf{Thinking Mode} with the \textbf{Non-Thinking Mode} for each provider.

\begin{itemize}
    \item \textit{\textbf{DeepSeek API}}: We evaluate DeepSeek-Reasoner\footnote{\url{https://api-docs.deepseek.com/guides/thinking_mode}} and DeepSeek-Chat\footnote{\url{https://api-docs.deepseek.com/guides/multi_round_chat}} via official open platform. These two endpoints serve as the representative proxies for Thinking Mode and Non-Thinking Mode, respectively.
    
    \item \textit{\textbf{Doubao API}}: We evaluate Doubao-Seed-1.6 via the Volcengine Ark\footnote{\url{https://console.volcengine.com/ark/region:ark+cn-beijing/model/detail?Id=doubao-seed-1-6}}. We switch between modes by configuring the reasoning parameter: setting \texttt{"type": "enabled"} for the Thinking Mode and \texttt{"type": "disabled"} for the Non-Thinking Mode.
    
    \item \textit{\textbf{Qwen API}}: We assess Qwen-Plus\footnote{\url{https://help.aliyun.com/zh/model-studio/models\#5ef284d4ed42p}} via Alibaba Cloud's Model Studio (Bailian). We control the behavior by toggling the \texttt{enable\_thinking} parameter, evaluating the model in Thinking Mode (enabled) versus Non-Thinking Mode (disabled).
    
    \item \textit{\textbf{Xiaomi API}}: We evaluate MiMo-V2-Flash via the Xiaomi MiMo Open Platform\footnote{\url{https://platform.xiaomimimo.com/\#/docs/quick-start/first-api-call}}. We explicitly control the generation depth by toggling the \texttt{reasoning} configuration to compare the Thinking Mode against the Non-Thinking Mode.
\end{itemize}

\paragraph{Datasets Selection.}

To validate the relative challenge level of LoopBench, we conduct comparative evaluations against two established benchmarks representing the pinnacle of mathematical and general reasoning capabilities:

\begin{itemize}
    \item \textit{\textbf{AIME2025}} \cite{aime25}: We utilize the problems from the \textit{2025 American Invitational Mathematics Examination (AIME)}. This dataset serves as a gold standard for assessing advanced mathematical problem-solving. It contains high-difficulty competition problems that require multi-step arithmetic derivation, making it an ideal baseline for testing whether models enter loops during intensive calculation.
    
    \item \textit{\textbf{SuperGPQA}} \cite{pteam2025supergpqascalingllmevaluation}: To cover a broader spectrum of logical reasoning beyond pure mathematics, we adopt SuperGPQA(Scaling LLM Evaluation across 285 Graduate Disciplines). This benchmark is designed to be "Google-proof," featuring extremely challenging questions across science, biology, and physics at a graduate level. 
    Considering the extensive scale of the original dataset, we constructed a stratified subset by randomly sampling 200 questions. This subset preserves the diversity of the disciplines while allowing for efficient evaluation of repetition behavior in logical reasoning contexts.
\end{itemize}

\paragraph{Evaluation Metrics.}
Based on our observations, repetitive outputs in LRMs primarily manifest in two distinct forms: \textbf{numerical loops} and \textbf{statement loops}. To systematically quantify these behaviors, we introduce two corresponding metrics: \textbf{Numerical Loop Rate (NLR)} and \textbf{Statement Loop Rate (SLR)}. The detailed criteria for identifying these loops are defined as follows:

\begin{itemize}
    \item \textit{\textbf{Numerical Loop}}: We identify a numerical loop based on the total accumulated length of the repetitive sequence. Let $l$ denote the length (in tokens) of the minimal repeating unit (comprising one or more digits), and $k$ denote the number of consecutive repetitions. We classify an output as a numerical loop if the total length satisfies:
    \[
    k \times l > 500
    \]
    This threshold is established based on the constraints of \textbf{LoopBench}, where no problem requires a numerical precision exceeding 500 digits. Consequently, any repetitive numerical sequence surpassing this length is considered an invalid infinite loop.

    \item \textit{\textbf{Statement Loop}}: We identify a statement loop based on the recurrence of sentence-level units. Let $l$ denote the number of sentences in the minimal repeating unit, and $k$ denote the number of repetitions. We classify an output as a statement loop if the repetition count satisfies:
    \[
    k > 3
    \]
    This threshold was empirically selected to robustly detect recursive reasoning traps where the model regurgitates previous thoughts without making new progress.
\end{itemize}

\paragraph{Implementation Details.}

To evaluate the robustness of models against repetition under varying degrees of sampling stochasticity, we utilized three distinct decoding configurations for \textbf{open-weights models}. Across all these settings, we applied a fixed \textbf{repetition penalty of $1.1$} to mimic standard deployment practices that mildly suppress trivial stuttering. The specific configurations are defined as follows (where $T$ denotes temperature):

\begin{itemize}
    \item \textbf{Conservative Setting} ($T=0.1$, top-$k=5$, top-$p=0.95$): A deterministic strategy that forces the model to select high-confidence tokens. This setting tests whether loops persist even when randomness is minimized.
    
    \item \textbf{Balanced Setting} ($T=0.6$, top-$k=15$, top-$p=0.9$): A standard strategy typical of general-purpose chat applications, striking a trade-off between coherence and diversity.
    
    \item \textbf{Explorative Setting} ($T=1.0$, top-$k=30$, top-$p=0.9$): A high-stochasticity strategy used to assess whether increased entropy can naturally disrupt or mitigate repetitive loops.
\end{itemize}

\noindent Conversely, for \textbf{Closed-Source Models,} (e.g., Claude-4.1-opus), we adhered to their default decoding configurations provided by the official APIs to reflect real-world usage patterns.

\begin{table*}[t]
\centering
\caption{\textbf{Evaluation of loop susceptibility across benchmarks.} We report DLN and SLN for base, instruction-tuned, and reasoning models. Open-source models use the Conservative Setting ($T=0.1$).}
\label{tab:repetition_result_t01}
\resizebox{\textwidth}{!}{%
\begin{tabular}{ll | ccc | ccc | ccc} 
\toprule
\multicolumn{2}{c}{\multirow{2}{*}{\textbf{Model}}} & \multicolumn{3}{c}{\textbf{AIME25}} & \multicolumn{3}{c}{\textbf{SuperGPQA}} & \multicolumn{3}{c}{\textbf{LoopBench}} \\ 
\cmidrule(lr){3-5} \cmidrule(lr){6-8} \cmidrule(lr){9-11} 
\multicolumn{2}{c}{} & \textbf{DLN} & \textbf{SLN} & \textbf{Sum} & \textbf{DLN} & \textbf{SLN} & \textbf{Sum} & \textbf{DLN} & \textbf{SLN} & \textbf{Sum} \\ \midrule

\multirow{12}{*}{\makecell[l]{Reasoning\\model}} 
 & DeepSeek-R1-Distill-Qwen-7B      & \gc{3}{3.33\%} & \gc{13}{13.33\%} & \gc{17}{\textbf{16.66\%}} & \gc{2}{1.50\%} & \gc{12}{12.00\%} & \gc{14}{\textbf{13.50\%}} & \gc{31}{30.64\%} & \gc{61}{61.43\%} & \gc{92}{\textbf{92.07\%}} \\
 & DeepSeek-R1-Distill-Llama-8B     & \gc{3}{3.33\%} & \gc{13}{13.33\%} & \gc{17}{\textbf{16.66\%}} & \gc{1}{0.50\%} & \gc{10}{10.00\%} & \gc{11}{\textbf{10.50\%}} & \gc{25}{25.29\%} & \gc{45}{44.57\%} & \gc{70}{\textbf{69.86\%}} \\
 & DeepSeek-R1-0528-Qwen3-8B     & 0.00\% & 0.00\% & \textbf{0.00\%} & 0.00\% & 0.00\% & \textbf{0.00\%} & \gc{16}{15.86\%} & \gc{7}{7.07\%} & \gc{22}{\textbf{22.93\%}} \\
 & DeepSeek-R1-Distill-Qwen-14B     & 0.00\% & \gc{10}{10.00\%} & \gc{10}{\textbf{10.00\%}} & \gc{1}{0.50\%} & \gc{10}{10.00\%} & \gc{11}{\textbf{10.50\%}} & \gc{30}{30.43\%} & \gc{45}{45.14\%} & \gc{76}{\textbf{75.57\%}} \\
 & DeepSeek-R1-Distill-Qwen-32B     & 0.00\% & \gc{10}{10.00\%} & \gc{10}{\textbf{10.00\%}} & 0.00\% & \gc{12}{12.00\%} & \gc{12}{\textbf{12.00\%}} & \gc{21}{21.43\%} & \gc{45}{45.43\%} & \gc{67}{\textbf{66.85\%}} \\
 & QwQ 32B         & 0.00\% & 0.00\% & \textbf{0.00\%} & \gc{1}{1.00\%} & \gc{5}{5.00\%} & \gc{6}{\textbf{6.00\%}} & \gc{38}{38.14\%} & \gc{8}{7.86\%} & \gc{46}{\textbf{46.00\%}} \\
 & Qwen3-8B        & \gc{3}{3.33\%} & 0.00\% & \gc{3}{\textbf{3.33\%}} & 0.00\% & 0.00\% & \textbf{0.00\%} & \gc{25}{25.29\%} & \gc{10}{10.28\%} & \gc{36}{\textbf{35.57\%}} \\
 & Qwen3-14B       & \gc{3}{3.33\%} & \gc{3}{3.33\%} & \gc{7}{\textbf{6.66\%}} & 0.00\% & \gc{1}{0.50\%} & \gc{1}{\textbf{0.50\%}} & \gc{30}{29.57\%} & \gc{7}{7.14\%} & \gc{36}{\textbf{36.71\%}} \\
 & Qwen3-32B       & \gc{3}{3.33\%} & 0.00\% & \gc{3}{\textbf{3.33\%}} & 0.00\% & 0.00\% & \textbf{0.00\%} & \gc{33}{33.43\%} & \gc{7}{7.57\%} & \gc{40}{\textbf{40.00\%}} \\
 & gpt-oss-20B     & \gc{13}{13.33\%} & \gc{7}{6.67\%} & \gc{20}{\textbf{20.00\%}} & \gc{3}{2.50\%} & \gc{7}{7.00\%} & \gc{10}{\textbf{9.50\%}} & \gc{45}{44.57\%} & \gc{18}{17.57\%} & \gc{62}{\textbf{62.14\%}} \\
 & Phi-4-reasoning & \gc{7}{6.67\%} & \gc{3}{3.33\%} & \gc{10}{\textbf{10.00\%}} & \gc{3}{3.00\%} & \gc{8}{7.50\%} & \gc{11}{\textbf{10.50\%}} & \gc{20}{19.71\%} & \gc{33}{33.43\%} & \gc{53}{\textbf{53.15\%}} \\
 & Phi-4-reasoning-plus & \gc{7}{6.67\%} & \gc{10}{10.00\%} & \gc{17}{\textbf{16.67\%}} & \gc{3}{3.00\%} & \gc{10}{9.50\%} & \gc{13}{\textbf{12.50\%}} & \gc{18}{18.00\%} & \gc{30}{29.57\%} & \gc{48}{\textbf{47.57\%}} \\ \midrule
 
\multirow{3}{*}{\makecell[l]{base model}} 
 & Qwen3-8B-base   & 0.00\% & 0.00\% & \textbf{0.00\%} & \gc{1}{1.00\%} & \gc{3}{3.00\%} & \gc{4}{\textbf{4.00\%}} & \gc{38}{38.86\%} & \gc{3}{3.43\%} & \gc{42}{\textbf{42.29\%}} \\
 & Qwen3-14B-base  & 0.00\% & \gc{7}{6.67\%} & \gc{7}{\textbf{6.67\%}} & 0.00\% & \gc{1}{0.50\%} & \gc{1}{\textbf{0.50\%}} & \gc{36}{36.28\%} & \gc{1}{0.86\%} & \gc{37}{\textbf{37.14\%}} \\
 & gemma-2-27B     & 0.00\% & \gc{7}{6.67\%} & \gc{7}{\textbf{6.67\%}} & \gc{1}{1.00\%} & \gc{3}{3.00\%} & \gc{4}{\textbf{4.00\%}} & \gc{16}{16.57\%} & \gc{8}{8.28\%} & \gc{24}{\textbf{24.86\%}} \\ \midrule
 
\multirow{5}{*}{\makecell[l]{Instruct model}} 
 & Llama-3.1-8B-Instruct & 0.00\% & \gc{7}{6.67\%} & \gc{7}{\textbf{6.67\%}} & \gc{2}{2.00\%} & \gc{4}{4.00\%} & \gc{6}{\textbf{6.00\%}} & \gc{40}{39.86\%} & \gc{7}{7.43\%} & \gc{47}{\textbf{47.29\%}} \\
 & Qwen2.5-7B-Instruct   & 0.00\% & 0.00\% & \textbf{0.00\%} & \gc{1}{0.50\%} & \gc{1}{0.50\%} & \gc{1}{\textbf{1.00\%}} & \gc{51}{51.43\%} & \gc{1}{1.43\%} & \gc{53}{\textbf{52.85\%}} \\
 & Qwen2.5-14B-Instruct  & 0.00\% & 0.00\% & \textbf{0.00\%} & 0.00\% & \gc{2}{2.00\%} & \gc{2}{\textbf{2.00\%}} & \gc{50}{49.71\%} & \gc{1}{1.43\%} & \gc{51}{\textbf{51.14\%}} \\
 & Qwen2.5-32B-Instruct  & 0.00\% & 0.00\% & \textbf{0.00\%} & \gc{1}{0.50\%} & 0.00\% & \gc{1}{\textbf{0.50\%}} & \gc{50}{50.15\%} & \gc{1}{1.14\%} & \gc{51}{\textbf{51.29\%}} \\
 & gemma-2-27B-it        & 0.00\% & 0.00\% & \textbf{0.00\%} & 0.00\% & 0.00\% & \textbf{0.00\%} & \gc{15}{15.14\%} & 0.00\% & \gc{15}{\textbf{15.14\%}} \\ 

\bottomrule
\end{tabular}%
}
\end{table*}

\begin{table*}[t]
\centering
\caption{\textbf{Evaluation of loop susceptibility across benchmarks.} We report DLN and SLN for base, instruction-tuned, and reasoning models. Open-source models use the Explorative Setting ($T=1.0$).}
\label{tab:repetition_result_t10}
\resizebox{\textwidth}{!}{%
\begin{tabular}{ll | ccc | ccc | ccc} 
\toprule
\multicolumn{2}{c}{\multirow{2}{*}{\textbf{Model}}} & \multicolumn{3}{c}{\textbf{AIME25}} & \multicolumn{3}{c}{\textbf{SuperGPQA}} & \multicolumn{3}{c}{\textbf{LoopBench}} \\ 
\cmidrule(lr){3-5} \cmidrule(lr){6-8} \cmidrule(lr){9-11} 
\multicolumn{2}{c}{} & \textbf{DLN} & \textbf{SLN} & \textbf{Sum} & \textbf{DLN} & \textbf{SLN} & \textbf{Sum} & \textbf{DLN} & \textbf{SLN} & \textbf{Sum} \\ \midrule

\multirow{12}{*}{\makecell[l]{Reasoning\\model}} 
 & DeepSeek-R1-Distill-Qwen-7B      & 0.00\% & 0.00\% & \textbf{0.00\%} & 0.00\% & 0.00\% & \textbf{0.00\%} & \gc{15}{14.86\%} & \gc{4}{3.86\%}   & \gc{19}{\textbf{18.71\%}} \\
 & DeepSeek-R1-Distill-Llama-8B     & 0.00\% & 0.00\% & \textbf{0.00\%} & \gc{1}{1.00\%} & 0.00\% & \gc{1}{\textbf{1.00\%}} & \gc{19}{18.57\%} & \gc{2}{2.15\%}   & \gc{21}{\textbf{20.72\%}} \\
 & DeepSeek-R1-0528-Qwen3-8B     & 0.00\% & 0.00\% & \textbf{0.00\%} & 0.00\% & 0.00\% & \textbf{0.00\%} & \gc{3}{3.14\%}   & \gc{1}{0.43\%}   & \gc{4}{\textbf{3.57\%}} \\
 & DeepSeek-R1-Distill-Qwen-14B     & 0.00\% & 0.00\% & \textbf{0.00\%} & 0.00\% & 0.00\% & \textbf{0.00\%} & \gc{13}{12.86\%} & \gc{2}{2.14\%}   & \gc{15}{\textbf{15.00\%}} \\
 & DeepSeek-R1-Distill-Qwen-32B     & 0.00\% & 0.00\% & \textbf{0.00\%} & \gc{1}{0.50\%} & 0.00\% & \gc{1}{\textbf{0.50\%}} & \gc{11}{11.43\%} & \gc{2}{2.15\%}   & \gc{14}{\textbf{13.57\%}} \\
 & QwQ 32B         & 0.00\% & 0.00\% & \textbf{0.00\%} & 0.00\% & 0.00\% & \textbf{0.00\%} & \gc{18}{18.00\%} & \gc{2}{2.12\%}   & \gc{20}{\textbf{20.12\%}} \\
 & Qwen3-8B        & 0.00\% & 0.00\% & \textbf{0.00\%} & \gc{1}{1.00\%} & 0.00\% & \gc{1}{\textbf{1.00\%}} & \gc{9}{8.57\%}   & \gc{2}{2.14\%}   & \gc{11}{\textbf{10.71\%}} \\
 & Qwen3-14B       & 0.00\% & 0.00\% & \textbf{0.00\%} & 0.00\% & 0.00\% & \textbf{0.00\%} & \gc{13}{13.43\%} & \gc{1}{1.28\%}   & \gc{15}{\textbf{14.71\%}} \\
 & Qwen3-32B       & \gc{3}{3.33\%} & 0.00\% & \gc{3}{\textbf{3.33\%}} & \gc{1}{0.50\%} & 0.00\% & \gc{1}{\textbf{0.50\%}} & \gc{18}{18.00\%} & \gc{1}{0.86\%}   & \gc{19}{\textbf{18.86\%}} \\
 & gpt-oss-20B     & \gc{3}{3.33\%} & 0.00\% & \gc{3}{\textbf{3.33\%}} & \gc{3}{3.00\%} & 0.00\% & \gc{3}{\textbf{3.00\%}} & \gc{21}{21.43\%} & \gc{4}{3.65\%}   & \gc{25}{\textbf{25.08\%}} \\
 & Phi-4-reasoning & \gc{7}{6.67\%} & 0.00\% & \gc{7}{\textbf{6.67\%}} & \gc{4}{3.50\%} & 0.00\% & \gc{4}{\textbf{3.50\%}} & \gc{39}{38.86\%} & \gc{9}{9.00\%}   & \gc{48}{\textbf{47.86\%}} \\
 & Phi-4-reasoning-plus & \gc{3}{3.33\%} & 0.00\% & \gc{3}{\textbf{3.33\%}} & \gc{4}{4.00\%} & 0.00\% & \gc{4}{\textbf{4.00\%}} & \gc{36}{36.27\%} & \gc{7}{7.07\%}   & \gc{43}{\textbf{43.35\%}} \\ \midrule
 
\multirow{3}{*}{\makecell[l]{base model}} 
 & Qwen3-8B-base   & 0.00\% & 0.00\% & \textbf{0.00\%} & \gc{1}{0.50\%} & \gc{2}{1.50\%} & \gc{2}{\textbf{2.00\%}} & \gc{20}{19.86\%} & \gc{1}{0.29\%}   & \gc{20}{\textbf{20.15\%}} \\
 & Qwen3-14B-base  & 0.00\% & \gc{3}{3.33\%} & \gc{3}{\textbf{3.33\%}} & 0.00\% & \gc{1}{0.50\%} & \gc{1}{\textbf{0.50\%}} & \gc{19}{18.86\%} & \gc{1}{0.29\%}   & \gc{20}{\textbf{19.15\%}} \\
 & gemma-2-27B     & 0.00\% & 0.00\% & \textbf{0.00\%} & \gc{3}{2.50\%} & \gc{1}{0.50\%} & \gc{3}{\textbf{3.00\%}} & \gc{9}{9.14\%}   & \gc{6}{6.00\%}   & \gc{15}{\textbf{15.14\%}} \\ \midrule
 
\multirow{5}{*}{\makecell[l]{Instruct model}} 
 & Llama-3.1-8B-Instruct & 0.00\% & 0.00\% & \textbf{0.00\%} & 0.00\% & 0.00\% & \textbf{0.00\%} & \gc{18}{18.29\%} & \gc{3}{2.57\%}   & \gc{21}{\textbf{20.86\%}} \\
 & Qwen2.5-7B-Instruct   & 0.00\% & 0.00\% & \textbf{0.00\%} & 0.00\% & 0.00\% & \textbf{0.00\%} & \gc{15}{14.57\%} & \gc{3}{2.57\%}   & \gc{17}{\textbf{17.14\%}} \\
 & Qwen2.5-14B-Instruct  & 0.00\% & 0.00\% & \textbf{0.00\%} & \gc{1}{0.50\%} & \gc{1}{0.50\%} & \gc{1}{\textbf{1.00\%}} & \gc{18}{17.71\%} & \gc{2}{1.71\%}   & \gc{19}{\textbf{19.43\%}} \\
 & Qwen2.5-32B-Instruct  & 0.00\% & 0.00\% & \textbf{0.00\%} & 0.00\% & \gc{1}{1.00\%} & \gc{1}{\textbf{1.00\%}} & \gc{19}{18.86\%} & \gc{1}{0.43\%}   & \gc{19}{\textbf{19.29\%}} \\
 & gemma-2-27B-it        & 0.00\% & 0.00\% & \textbf{0.00\%} & 0.00\% & 0.00\% & \textbf{0.00\%} & \gc{6}{5.71\%}   & 0.00\%          & \gc{6}{\textbf{5.71\%}} \\ 

\bottomrule
\end{tabular}%
}
\end{table*}

\subsection{Case Study}     \label{subsec:case}

We conducted our evaluation through three distinct channels: deploying open-source models locally, querying official APIs, and utilizing official web platforms. As illustrated in Figure \ref{fig:Case Study: Square Root}, \ref{fig:case_study_digital}, \ref{fig:case_study_divition}, \ref{fig:case_study_Truth-Teller}, \ref{fig:case_study_Paradox}, \ref{fig:case_study_tower}, \ref{fig:case_study_Planning}, \ref{fig:qwen}, \ref{fig:doubao}, the results exhibit instances of numerical loops and statement loops.

%% file: latex/sections/X5_analysis.tex
\section{Extended Analysis of Circular Reasoning}
\label{sec:appendix_analysis}

In this section, we present comprehensive experimental results for DS-Qwen-14B and Qwen-8B. These supplementary analyses serve to substantiate the universality of circular reasoning, demonstrating that this phenomenon is not comprised of isolated anomalies, but rather represents a consistent behavioral pattern inherent to the models.

\subsection{Additional Results on Phase Transition}

\begin{figure*}[t]
    \centering
    \begin{subfigure}[b]{0.48\textwidth}
        \centering
        \includegraphics[width=\textwidth]{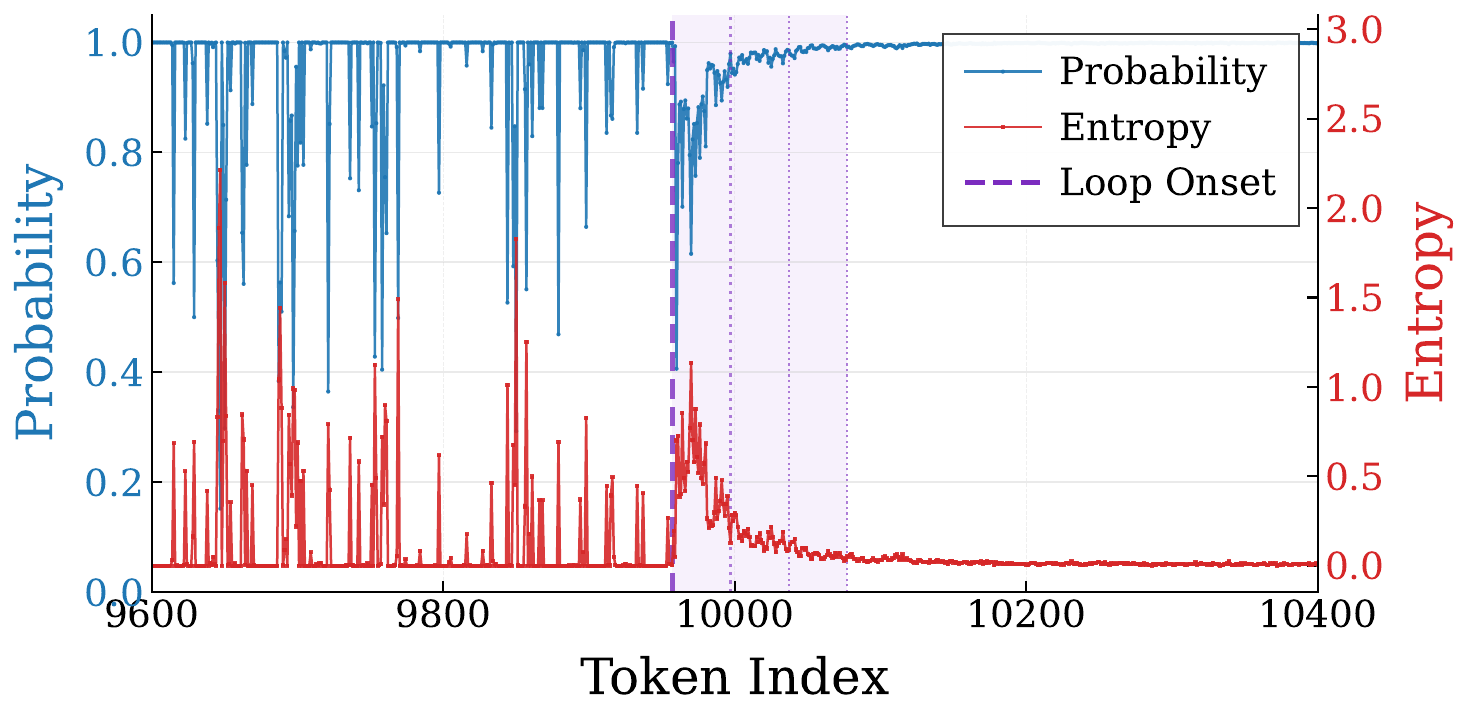}
        \caption{Numerical Loop Case 1}
        \label{fig:ds_num_loop_c1}
    \end{subfigure}
    \hfill
    \begin{subfigure}[b]{0.48\textwidth}
        \centering
        \includegraphics[width=\textwidth]{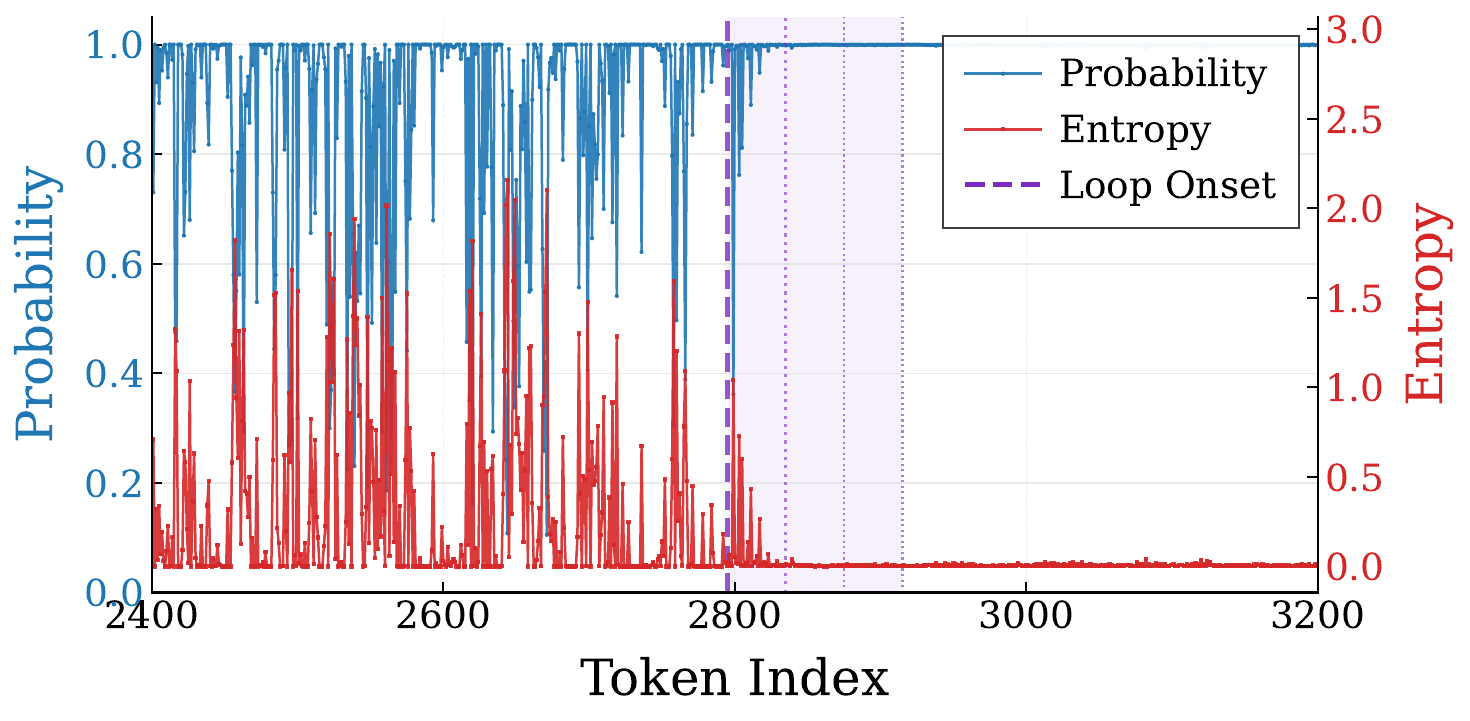}
        \caption{Numerical Loop Case 2}
        \label{fig:ds_num_loop_c2}
    \end{subfigure}
    \caption{\textbf{Visualization of Determinism Surge in DS-Qwen-14B (Numerical Loops).} The dashed vertical line marks the onset of token repetition. This point triggers an immediate entropy collapse (red) and a simultaneous probability surge (blue), indicating a sudden transition into rigid determinism.}
    \label{fig:ds-qwen_num_loop}
\end{figure*}

\begin{figure*}[t]
    \centering
    \begin{subfigure}[b]{0.48\textwidth}
        \centering
        \includegraphics[width=\textwidth]{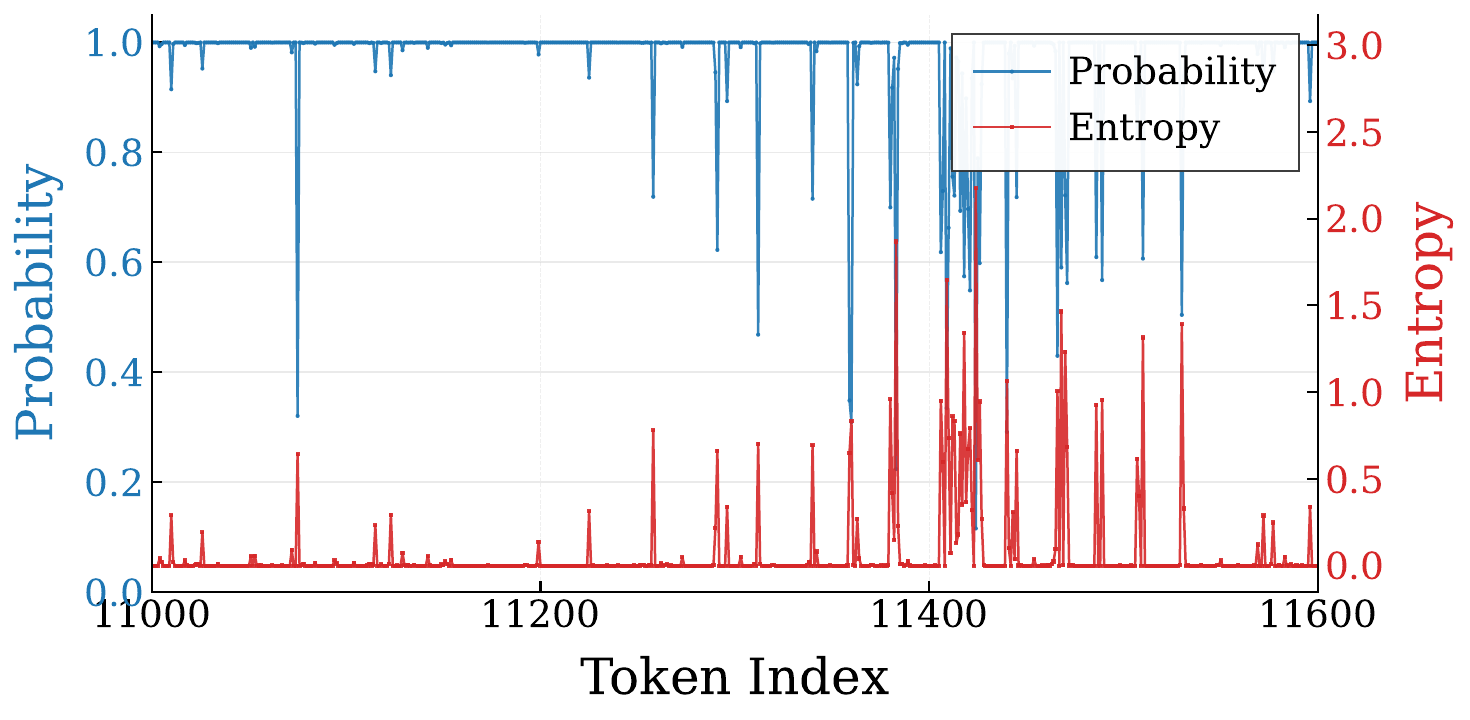}
        \caption{Numerical Non-loop Case 1}
        \label{fig:ds_num_noloop_c1}
    \end{subfigure}
    \hfill
    \begin{subfigure}[b]{0.48\textwidth}
        \centering
        \includegraphics[width=\textwidth]{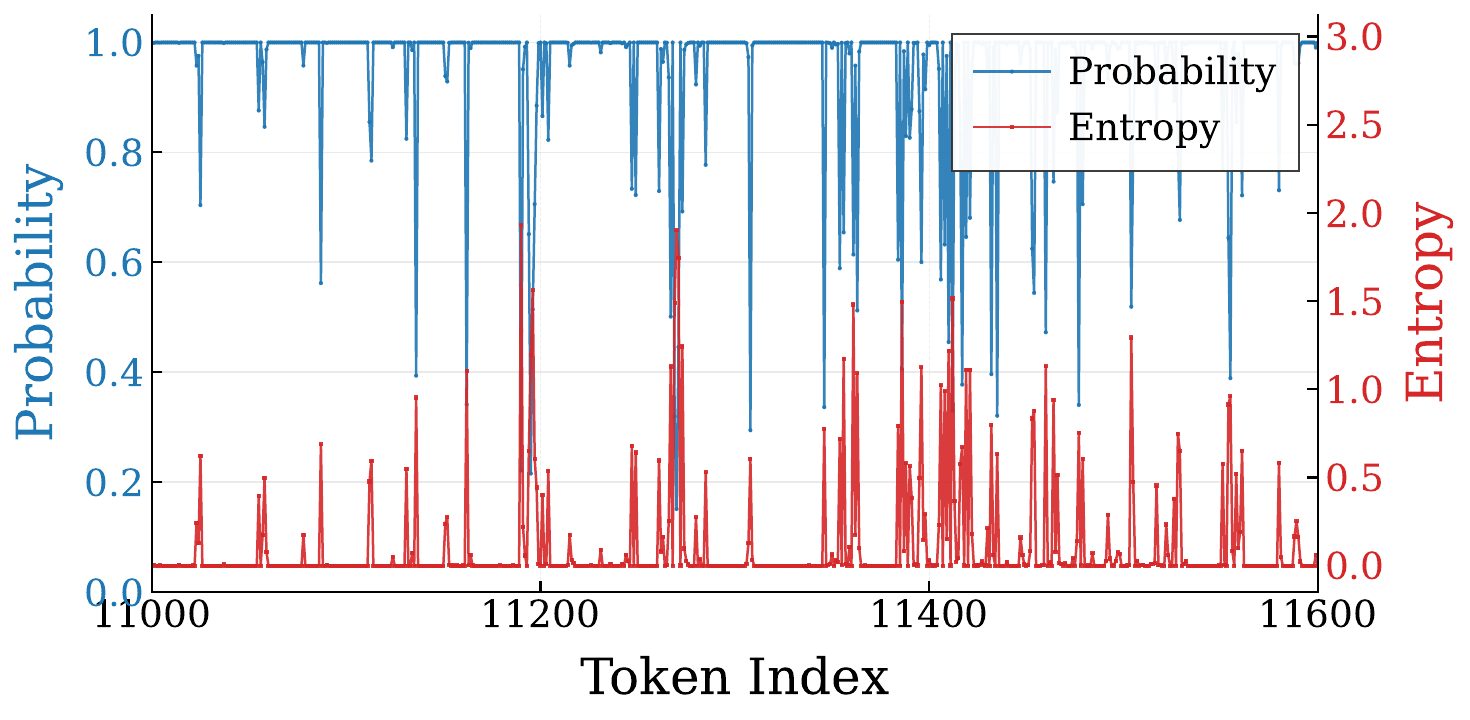}
        \caption{Numerical Non-loop Case 2}
        \label{fig:ds_num_noloop_c2}
    \end{subfigure}
    \caption{\textbf{Baseline Behavior in DS-Qwen-14B (Numerical Non-loops).} In contrast to the loop cases, the probability and entropy distributions exhibit high variance and stochastic fluctuations without a distinct collapse pattern.}
    \label{fig:ds-qwen_num_nonloop}
\end{figure*}

\begin{figure*}[t]
    \centering
    \begin{subfigure}[b]{0.48\textwidth}
        \centering
        \includegraphics[width=\textwidth]{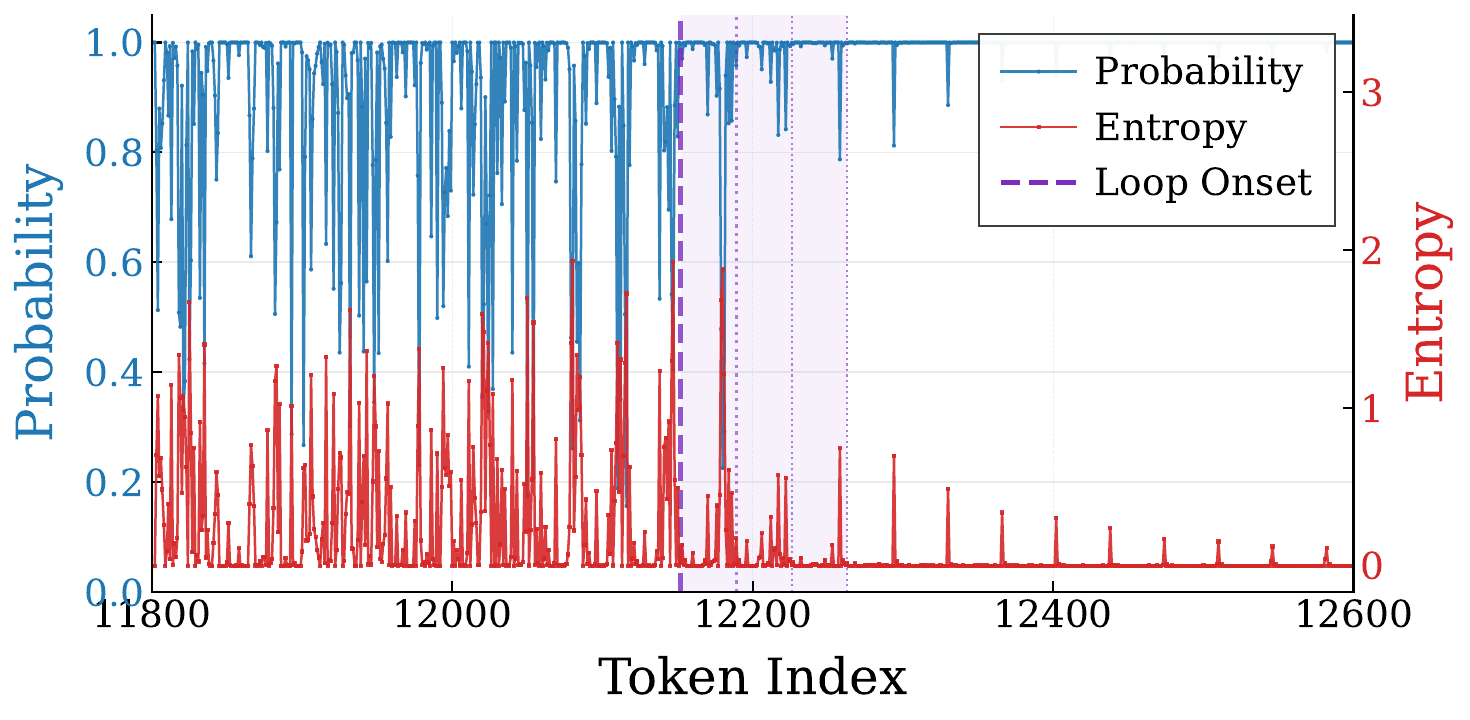}
        \caption{Statement Loop Case 1}
        \label{fig:ds_stmt_loop_c1}
    \end{subfigure}
    \hfill
    \begin{subfigure}[b]{0.48\textwidth}
        \centering
        \includegraphics[width=\textwidth]{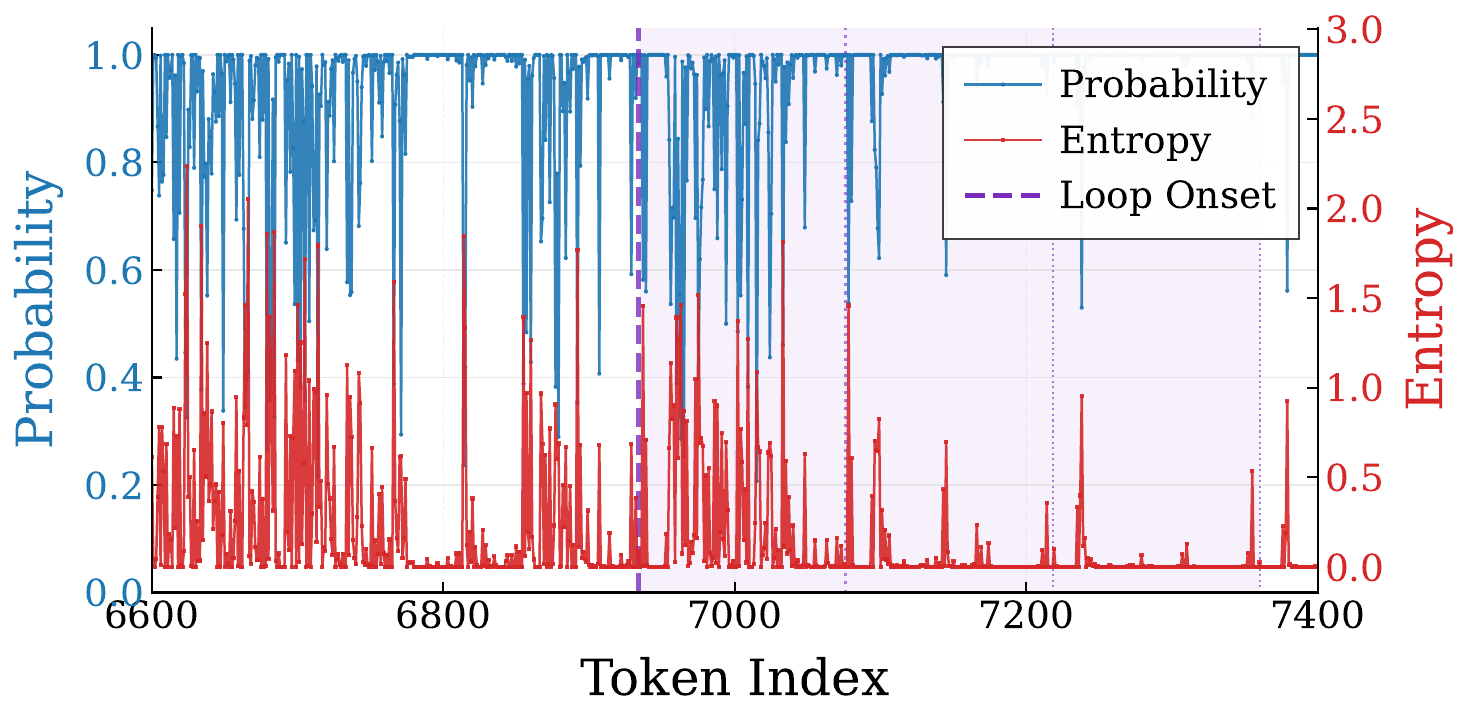}
        \caption{Statement Loop Case 2}
        \label{fig:ds_stmt_loop_c2}
    \end{subfigure}
    \caption{\textbf{Visualization of Determinism Surge in DS-Qwen-14B (Statement Loops).} Similar to numerical tasks, the repetition onset (dashed line) marks a sharp phase transition where entropy collapses to near-zero, confirming the modality-agnostic nature of the phenomenon.}
    \label{fig:ds-qwen_statement_loop}
\end{figure*}

\begin{figure*}[t]
    \centering
    \begin{subfigure}[b]{0.48\textwidth}
        \centering
        \includegraphics[width=\textwidth]{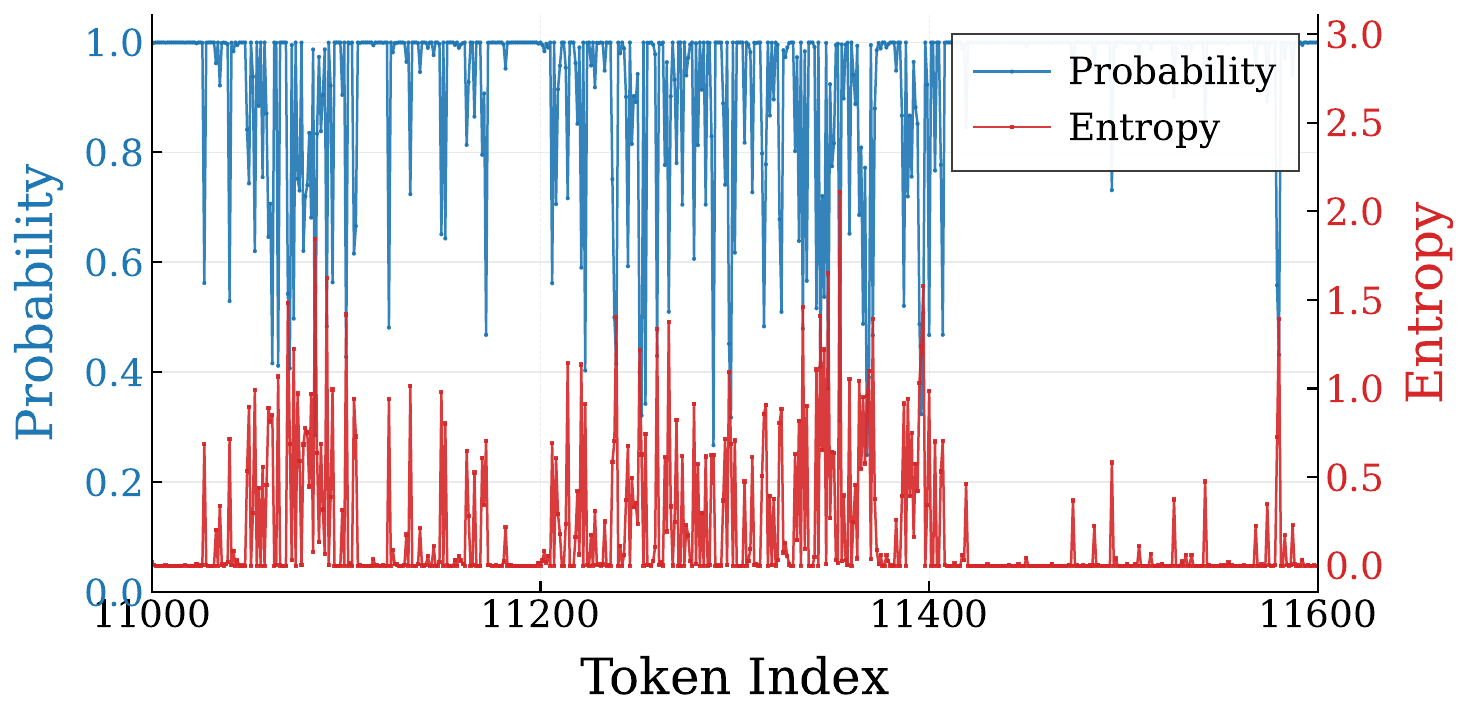}
        \caption{Statement Non-loop Case 1}
        \label{fig:ds_stmt_noloop_c1}
    \end{subfigure}
    \hfill
    \begin{subfigure}[b]{0.48\textwidth}
        \centering
        \includegraphics[width=\textwidth]{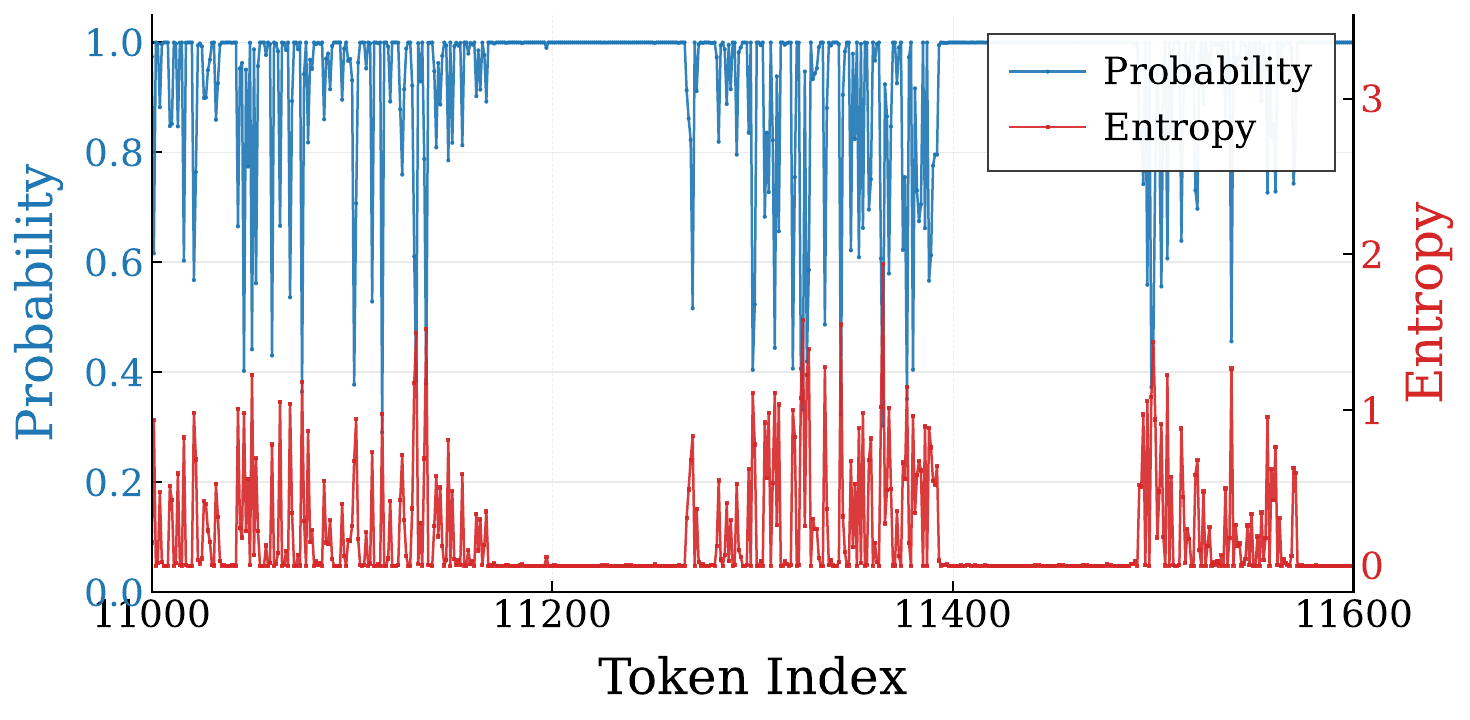}
        \caption{Statement Non-loop Case 2}
        \label{fig:ds_stmt_noloop_c2}
    \end{subfigure}
    \caption{\textbf{Baseline Behavior in DS-Qwen-14B (Statement Non-loops).} The distributions remain disordered with no observable determinism surge.}
    \label{fig:ds-qwen_statement_nonloop}
\end{figure*}

\begin{figure*}[t]
    \centering
    \begin{subfigure}[b]{0.48\textwidth}
        \centering
        \includegraphics[width=\textwidth]{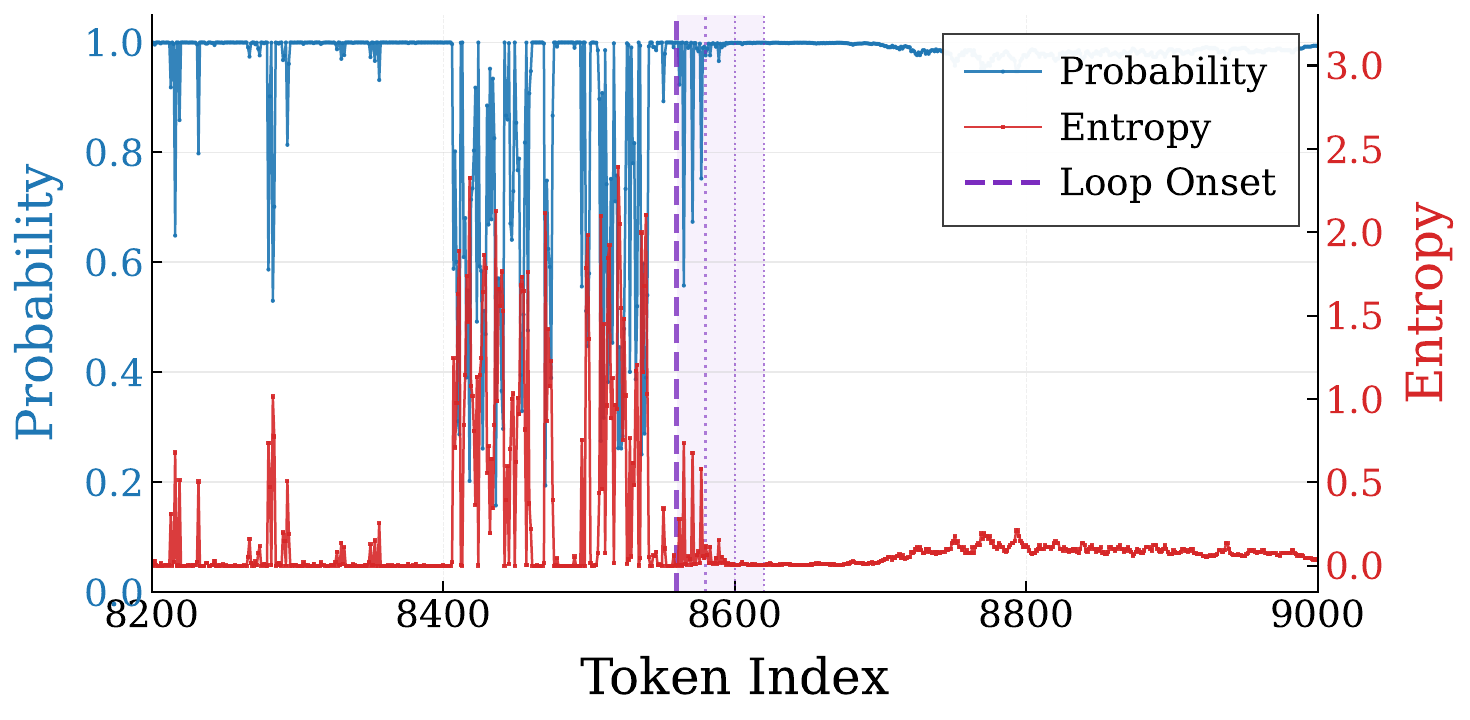}
        \caption{Numerical Loop Case 1}
        \label{fig:qw3_num_loop_c1}
    \end{subfigure}
    \hfill
    \begin{subfigure}[b]{0.48\textwidth}
        \centering
        \includegraphics[width=\textwidth]{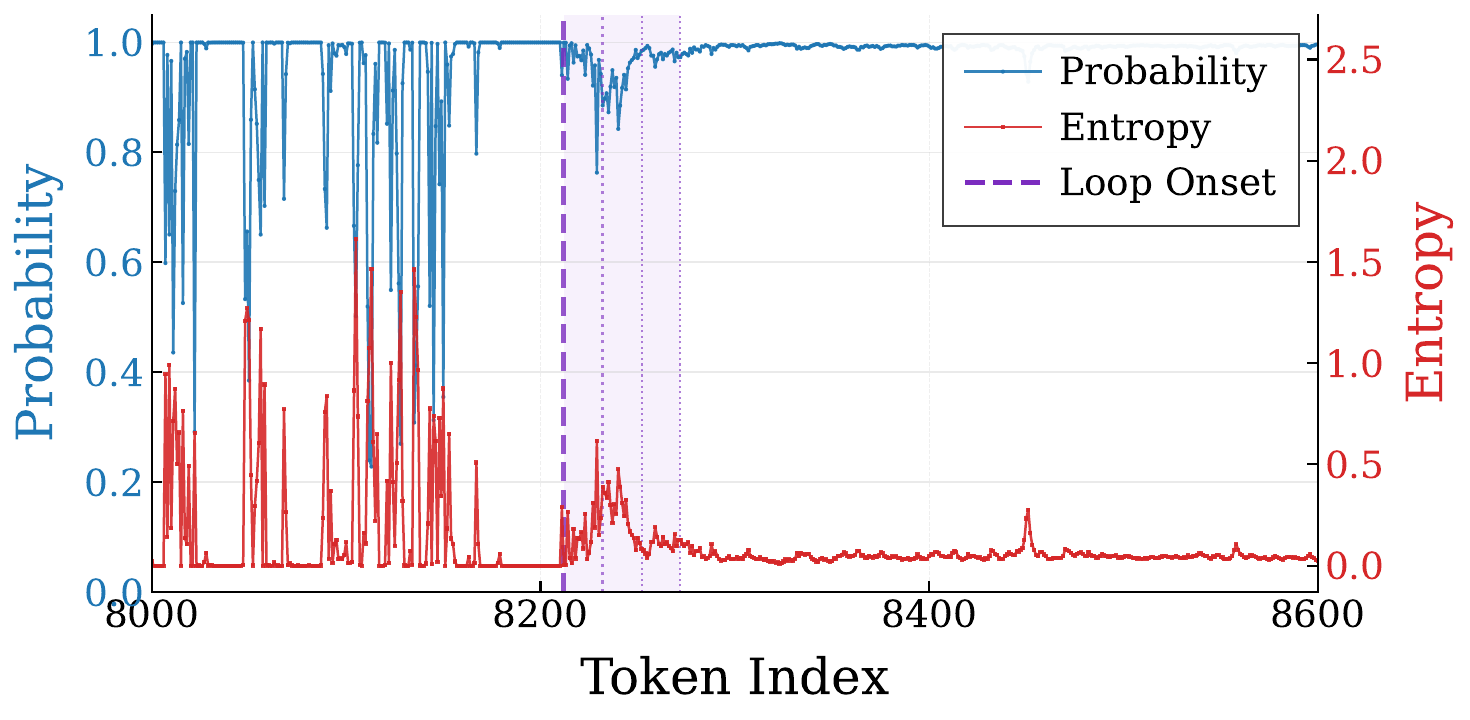}
        \caption{Numerical Loop Case 2}
        \label{fig:qw3_num_loop_c2}
    \end{subfigure}
    \caption{\textbf{Visualization of Determinism Surge in Qwen3-8B (Numerical Loops).} The dashed vertical line marks the onset of token repetition. This point triggers an immediate entropy collapse (red) and a simultaneous probability surge (blue), indicating a sudden transition into rigid determinism.}
    \label{fig:qwen3_num_loop}
\end{figure*}

\begin{figure*}[t]
    \centering
    \begin{subfigure}[b]{0.48\textwidth}
        \centering
        \includegraphics[width=\textwidth]{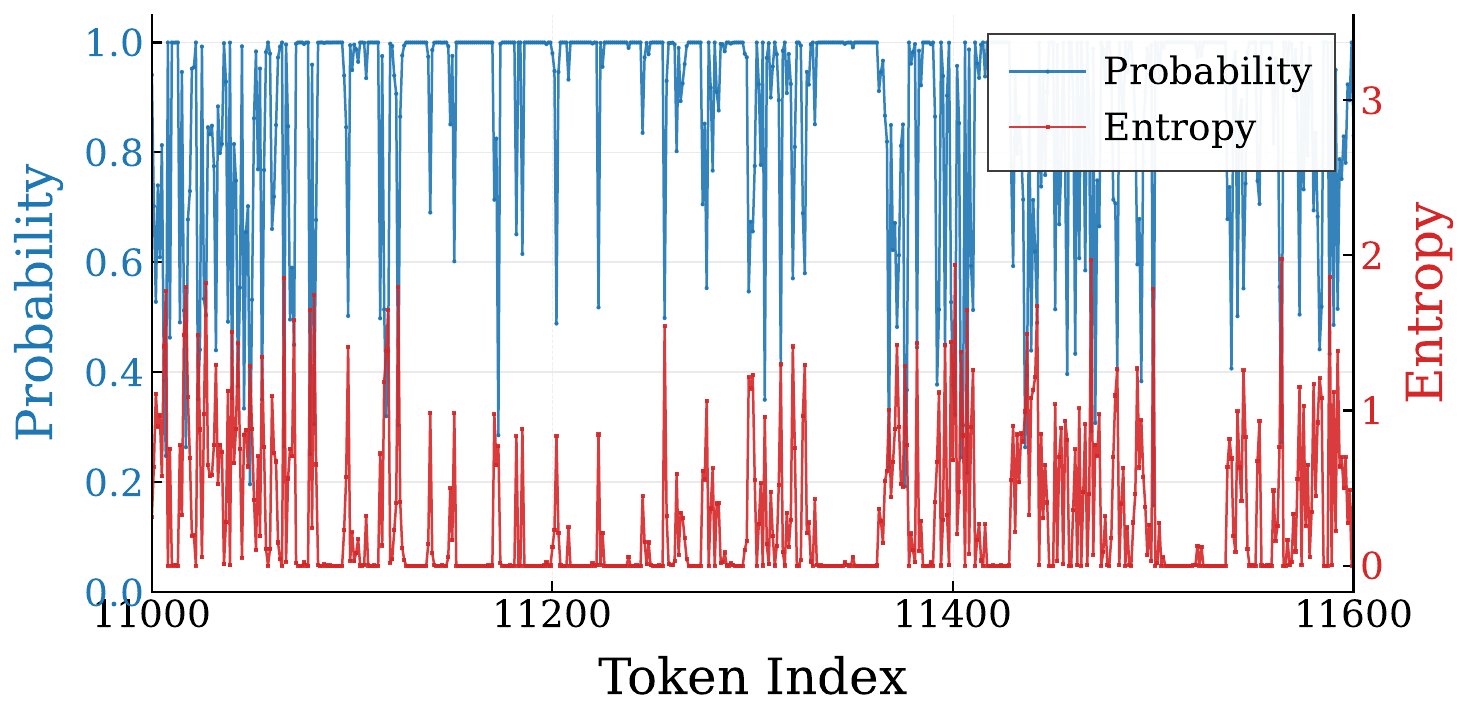}
        \caption{Numerical Non-loop Case 1}
        \label{fig:qw3_num_noloop_c1}
    \end{subfigure}
    \hfill
    \begin{subfigure}[b]{0.48\textwidth}
        \centering
        \includegraphics[width=\textwidth]{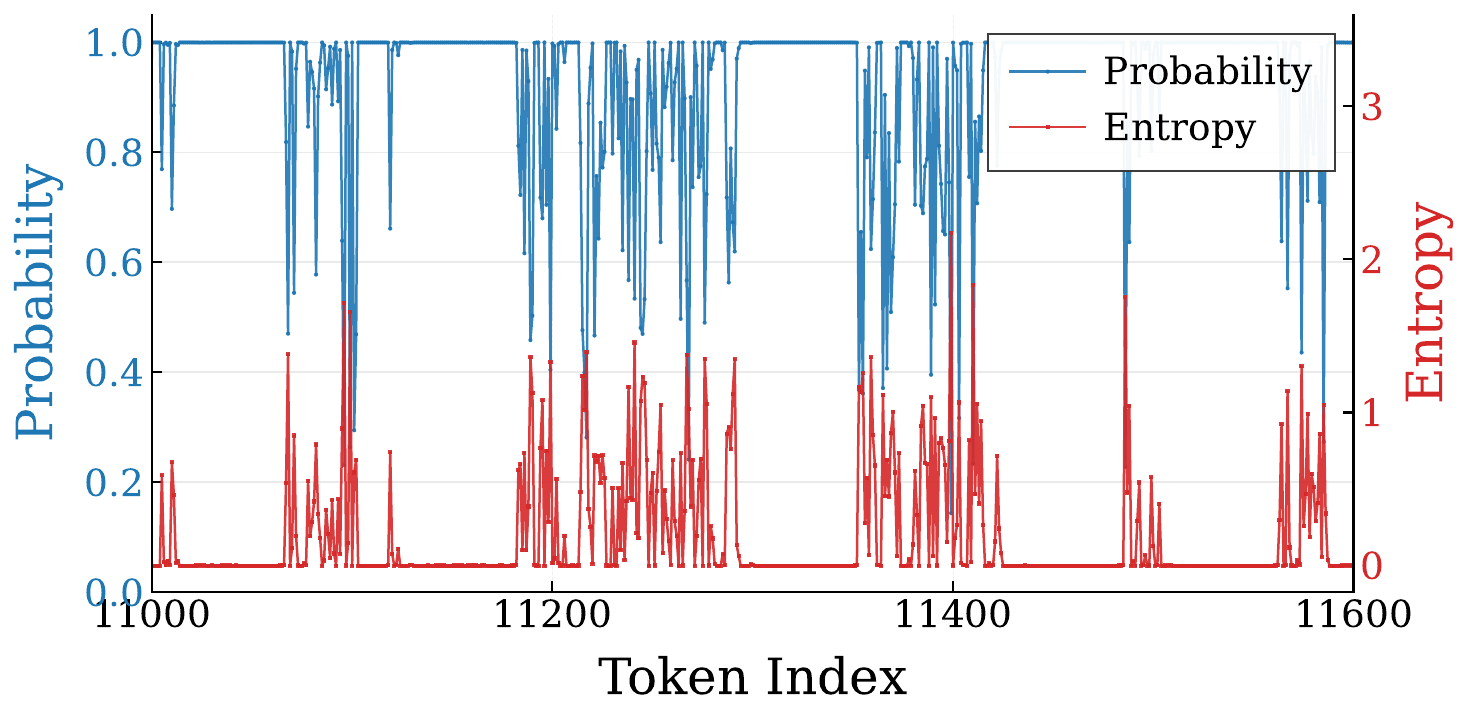}
        \caption{Numerical Non-loop Case 2}
        \label{fig:qw3_num_noloop_c2}
    \end{subfigure}
    \caption{\textbf{Baseline Behavior in Qwen3-8B (Numerical Non-loops).} In contrast to the loop cases, the probability and entropy distributions exhibit high variance and stochastic fluctuations without a distinct collapse pattern.}
    \label{fig:qwen3_num_nonloop}
\end{figure*}

\begin{figure*}[t]
    \centering
    \begin{subfigure}[b]{0.48\textwidth}
        \centering
        \includegraphics[width=\textwidth]{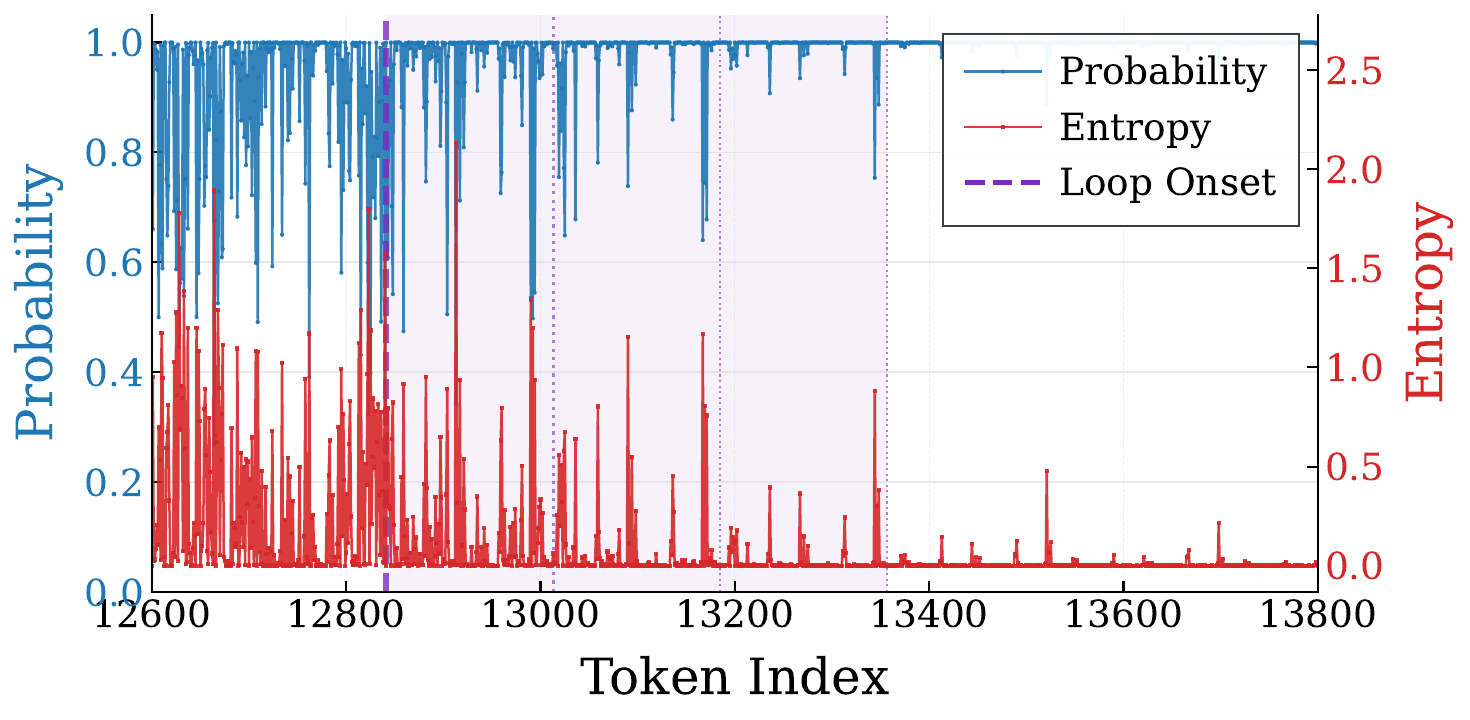}
        \caption{Statement Loop Case 1}
        \label{fig:qw3_stmt_loop_c1}
    \end{subfigure}
    \hfill
    \begin{subfigure}[b]{0.48\textwidth}
        \centering
        \includegraphics[width=\textwidth]{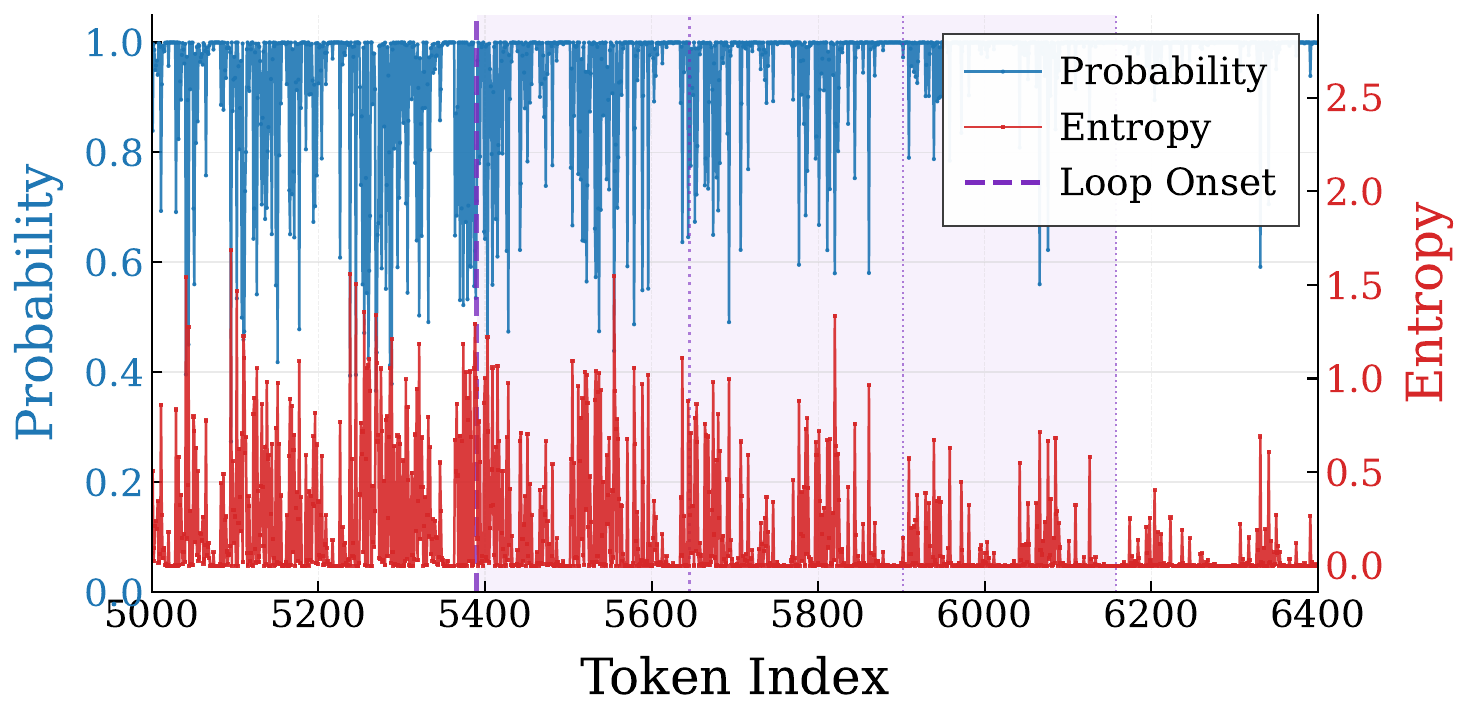}
        \caption{Statement Loop Case 2}
        \label{fig:qw3_stmt_loop_c2}
    \end{subfigure}
    \caption{\textbf{Visualization of Determinism Surge in Qwen3-8B (Statement Loops).} Similar to numerical tasks, the repetition onset (dashed line) marks a sharp phase transition where entropy collapses to near-zero, confirming the modality-agnostic nature of the phenomenon.}
    \label{fig:qwen3_statement_loop}
\end{figure*}

\begin{figure*}[t]
    \centering
    \begin{subfigure}[b]{0.48\textwidth}
        \centering
        \includegraphics[width=\textwidth]{latex/figures/appendix/ds-qwen-logist/statement/question_0009_no_loop.pdf}
        \caption{Statement Non-loop Case 1}
        \label{fig:qw3_stmt_noloop_c1}
    \end{subfigure}
    \hfill
    \begin{subfigure}[b]{0.48\textwidth}
        \centering
        \includegraphics[width=\textwidth]{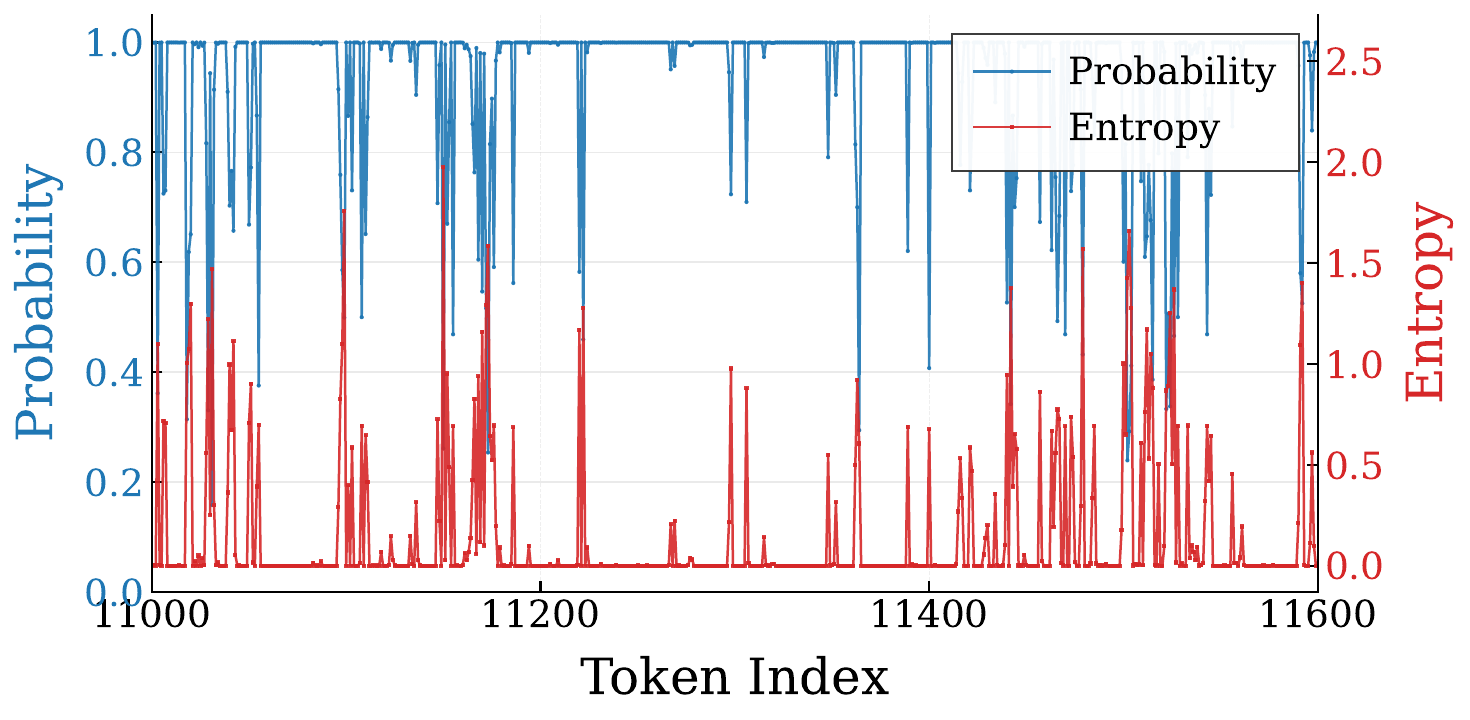}
        \caption{Statement Non-loop Case 2}
        \label{fig:qw3_stmt_noloop_c2}
    \end{subfigure}
    \caption{\textbf{Baseline Behavior in Qwen3-8B (Statement Non-loops).} The distributions remain disordered with no observable determinism surge.}
    \label{fig:qwen3_statement_nonloop}
\end{figure*}

We provide a detailed visualization of the entropy and probability dynamics for two distinct models. For DS-Qwen-14B, Figures~\ref{fig:ds-qwen_num_loop} and \ref{fig:ds-qwen_statement_loop} illustrate the circular reasoning scenarios, whereas Figures~\ref{fig:ds-qwen_num_nonloop} and \ref{fig:ds-qwen_statement_nonloop} present the non-loop baselines. Extending this analysis to Qwen3-8B to verify universality, we observe identical patterns: Figures~\ref{fig:qwen3_num_loop} and \ref{fig:qwen3_statement_loop} depict the loop cases, contrasted against the non-loop baselines in Figures~\ref{fig:qwen3_num_nonloop} and \ref{fig:qwen3_statement_nonloop}. Across both models and task types, the Determinism Surge remains a robust indicator of the transition.

\begin{figure*}[t]
    \centering
    \includegraphics[width=\textwidth]{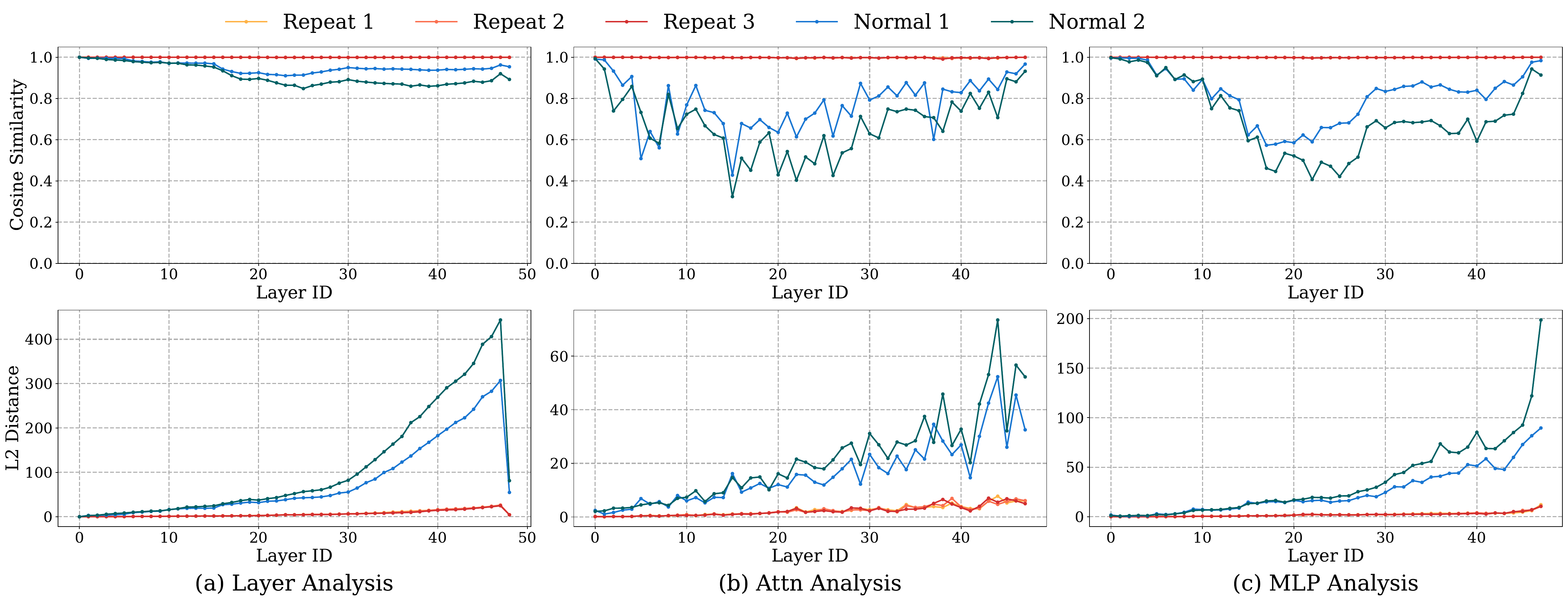}
    \caption{\textbf{Progressive obstruction of information flow (DS-Qwen-14B).} 
    We report the layer-wise Cosine Similarity (top) and $L_2$ Distance (bottom) across Layer, Attention, and MLP modules. 
    Curves labelled ``Repeat $k$'' denote the comparison between the representation in the ``$k$''-th loop and ``$k-1$''-th cycle. 
    \textit{\textbf{Trend:}} As repetition deepens (darker red lines), the internal state becomes rigid, contrasting sharply with the dynamic flow observed in normal baselines (blue lines). This indicates a complete blockage of information updates in later cycles.}
    \label{fig:hidden_state_ds}
\end{figure*}

\begin{figure*}[t]
    \centering
    \includegraphics[width=\textwidth]{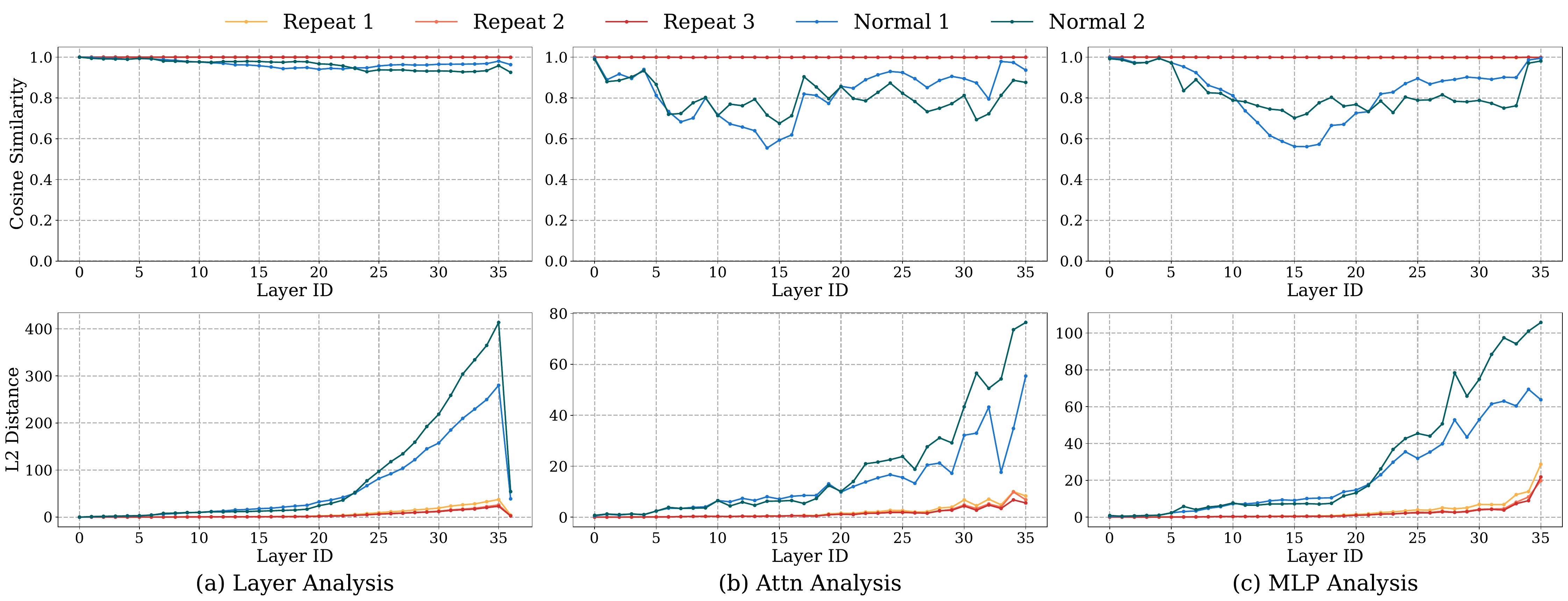}
    \caption{\textbf{Progressive obstruction of information flow (Qwen3-8B, Case I).} 
    Similar to the observations in DS-Qwen-14B, the Qwen3-8B model exhibits high cosine similarity and low $L_2$ distance in the recurring loops, signifying a cessation of semantic evolution in the hidden states.}
    \label{fig:hidden_state_qwen3_1}
\end{figure*}

\begin{figure*}[t]
    \centering
    \includegraphics[width=\textwidth]{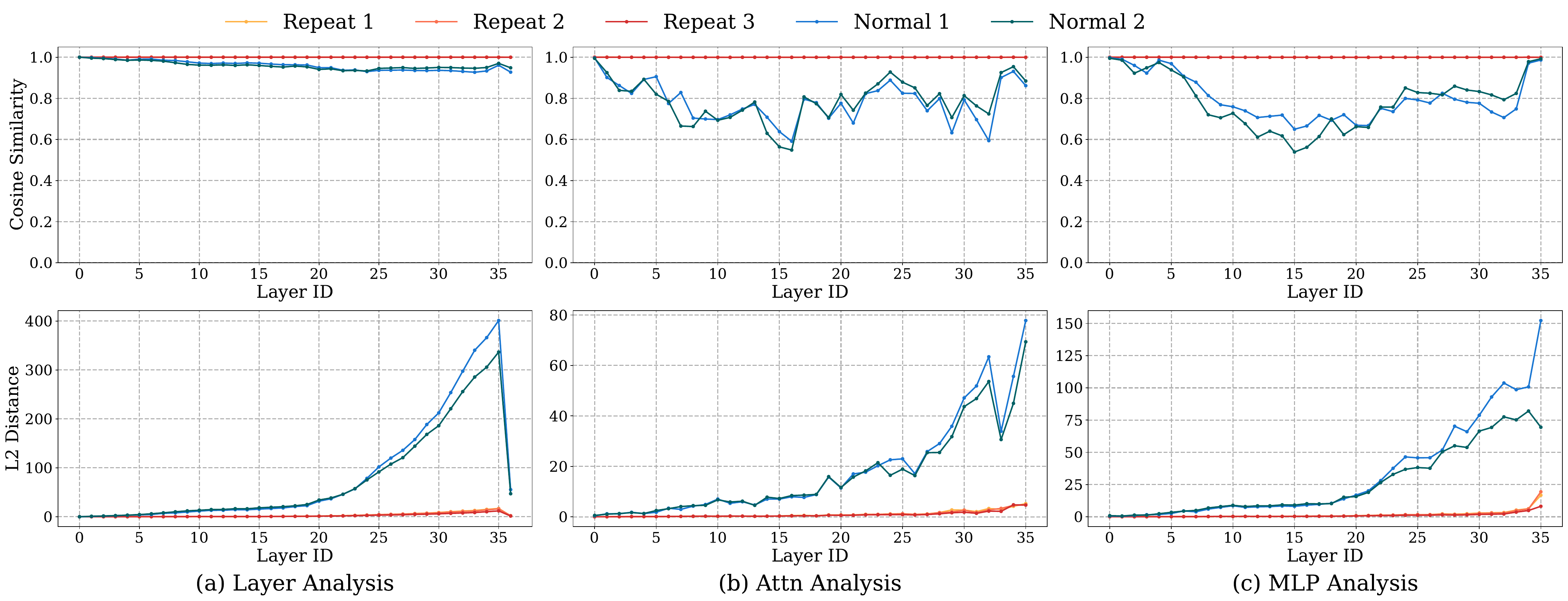}
    \caption{\textbf{Progressive obstruction of information flow (Qwen3-8B, Case II).} 
    Continued analysis of Qwen3-8B internal states. The darker red curves consistently show reduced variation compared to the blue baseline curves, confirming the universality of the information blockage phenomenon across different model architectures.}
    \label{fig:hidden_state_qwen3_2}
\end{figure*}

Subsequently, we investigate the evolution of internal representations across the model's architecture, specifically examining the residual stream (Layer), Attention module (Attn), and Feed-Forward Networks (MLP). To quantify the information flow, we compute the layer-wise cosine similarity and $L_2$ Distance between the hidden states of the current reasoning cycle and those of the preceding cycle. This analysis reveals how the model's internal dynamics stiffen as it enters the circular reasoning phase, as illustrated in Figure~\ref{fig:hidden_state_ds} for DS-Qwen-14B, and further corroborated by Figures~\ref{fig:hidden_state_qwen3_1} and \ref{fig:hidden_state_qwen3_2} for Qwen3-8B.

\begin{figure*}[t]
    \centering
    \begin{subfigure}[b]{0.3\linewidth}
        \centering
        \includegraphics[width=\textwidth]{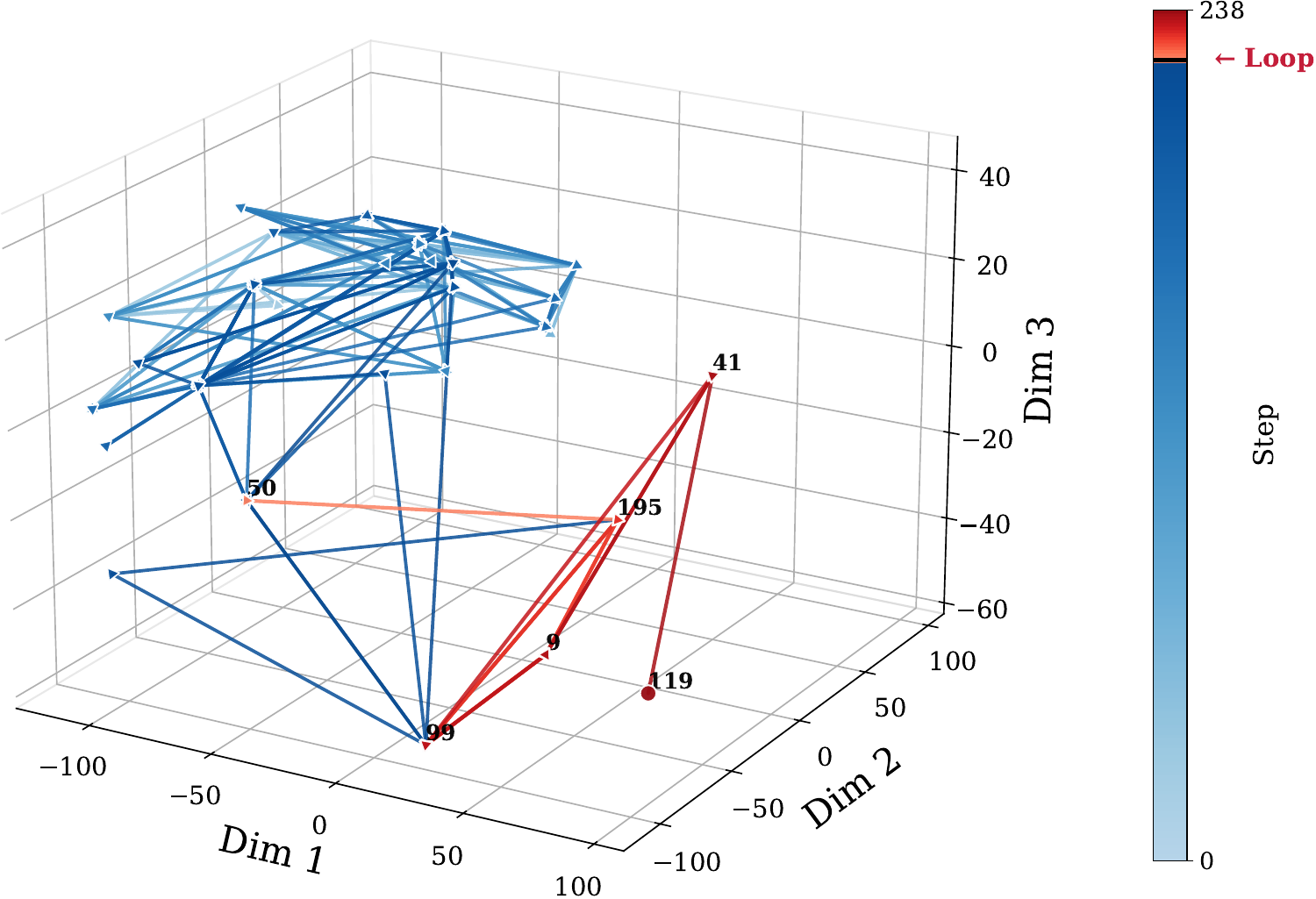}
        \caption{PCA Projection}
        \label{fig:case01_pca}
    \end{subfigure}
    \hfill
    \begin{subfigure}[b]{0.3\linewidth}
        \centering
        \includegraphics[width=\textwidth]{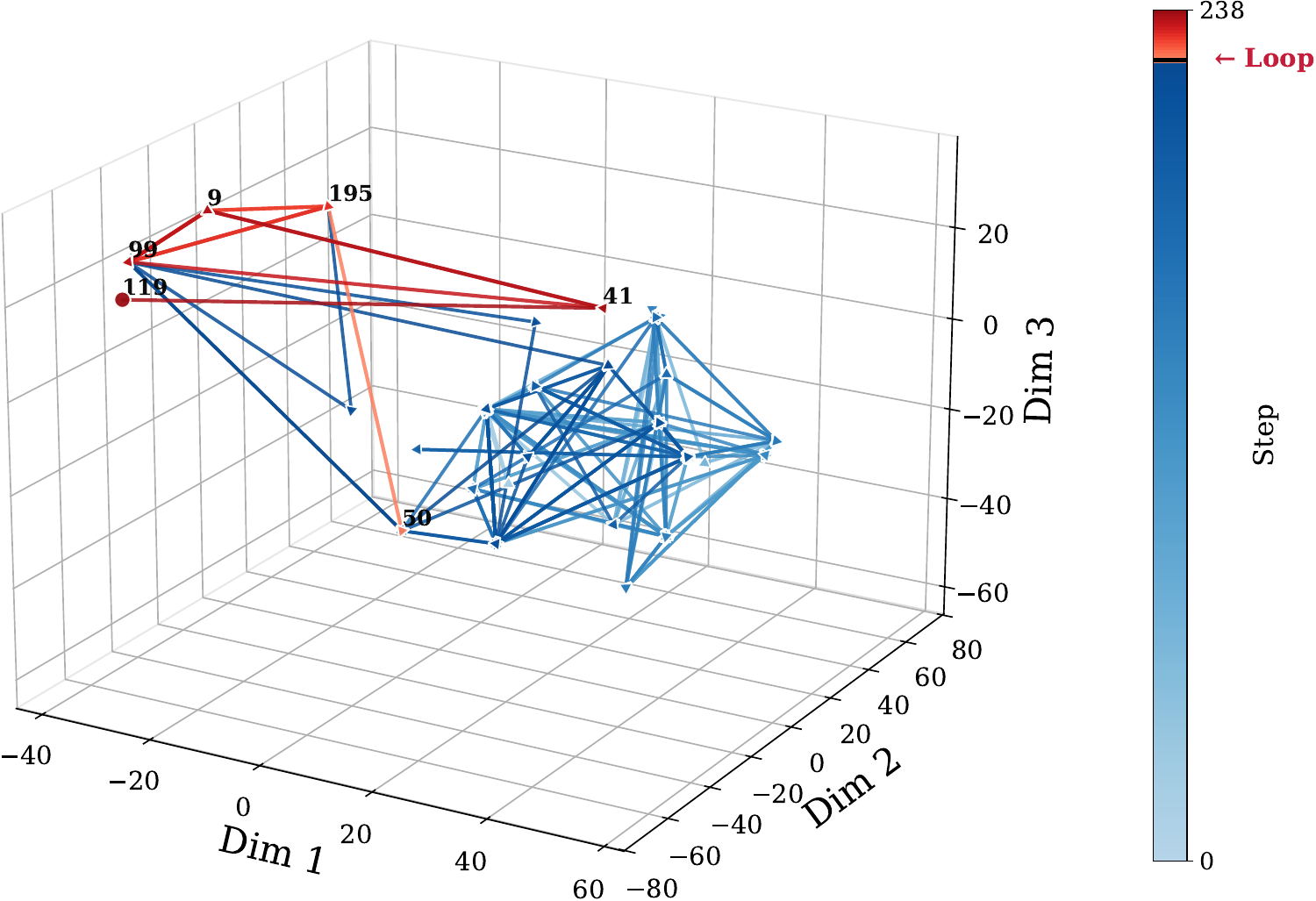}
        \caption{t-SNE Projection}
        \label{fig:case01_tsne}
    \end{subfigure}
    \hfill
    \begin{subfigure}[b]{0.35\linewidth}
        \centering
        \includegraphics[width=\textwidth]{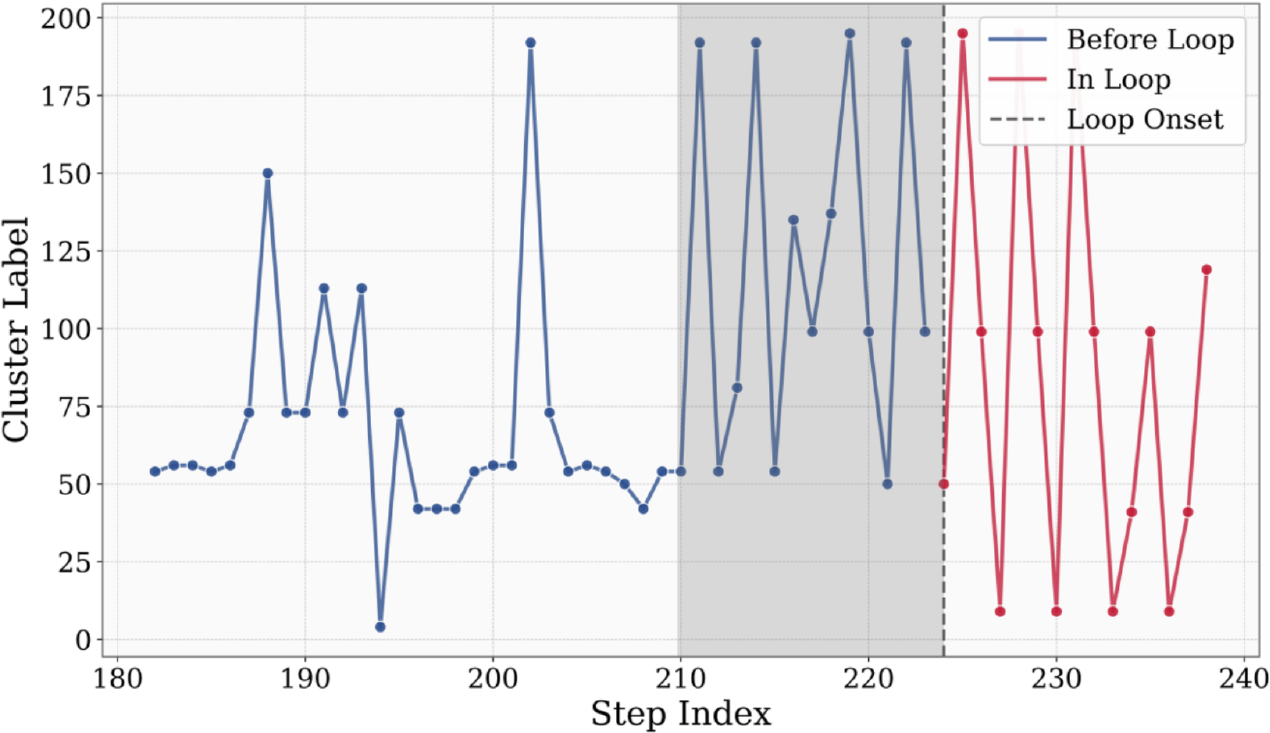}
        \caption{Cluster Trajectory Timeline}
        \label{fig:case01_path}
    \end{subfigure}
    \caption{\textbf{Reasoning graph visualization of a statement loop in DS-Qwen-14B (Case I).} (a) and (b) show internal states collapsing into cycles. Crucially, (c) plots cluster labels against sentence IDs, demonstrating that \textit{semantic circularity} (shaded region) significantly precedes explicit textual repetition, serving as an early warning.}
    \label{fig:reasoning_graph_topology01}
\end{figure*}

\begin{figure*}[t]
    \centering
    \begin{subfigure}[b]{0.3\linewidth}
        \centering
        \includegraphics[width=\textwidth]{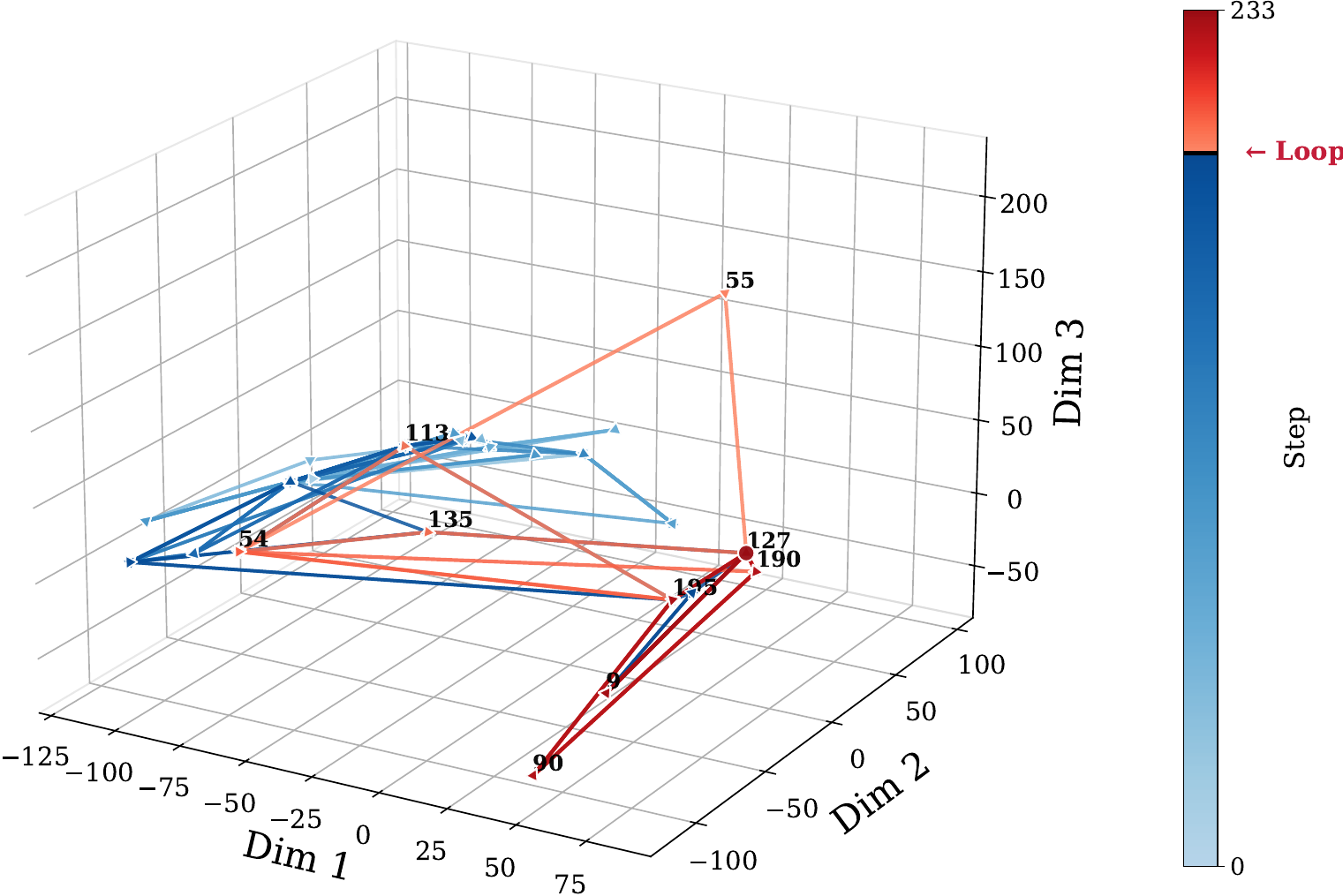}
        \caption{PCA Projection}
        \label{fig:case02_pca}
    \end{subfigure}
    \hfill
    \begin{subfigure}[b]{0.3\linewidth}
        \centering
        \includegraphics[width=\textwidth]{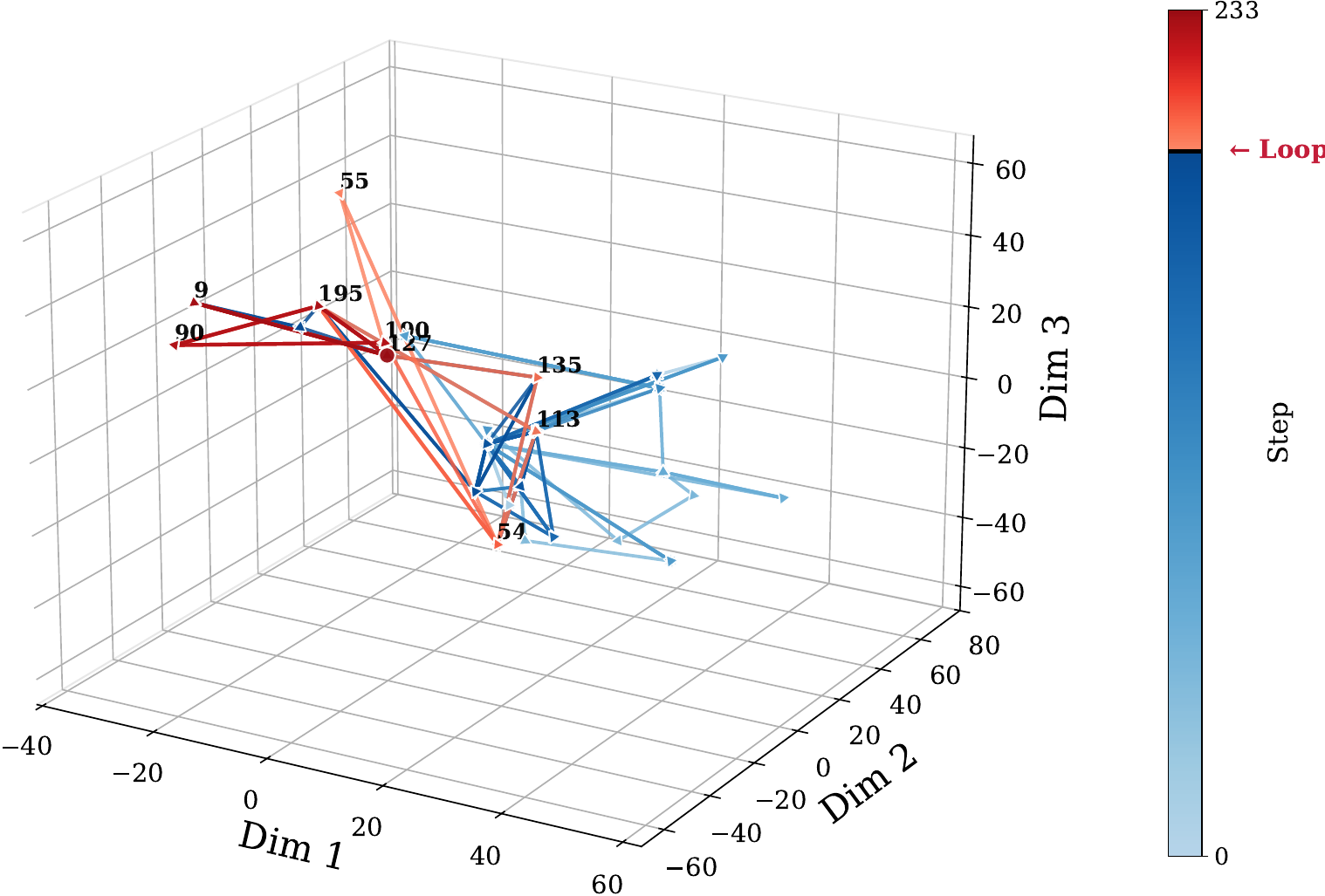}
        \caption{t-SNE Projection}
        \label{fig:case02_tsne}
    \end{subfigure}
    \hfill
    \begin{subfigure}[b]{0.35\linewidth}
        \centering
        \includegraphics[width=\textwidth]{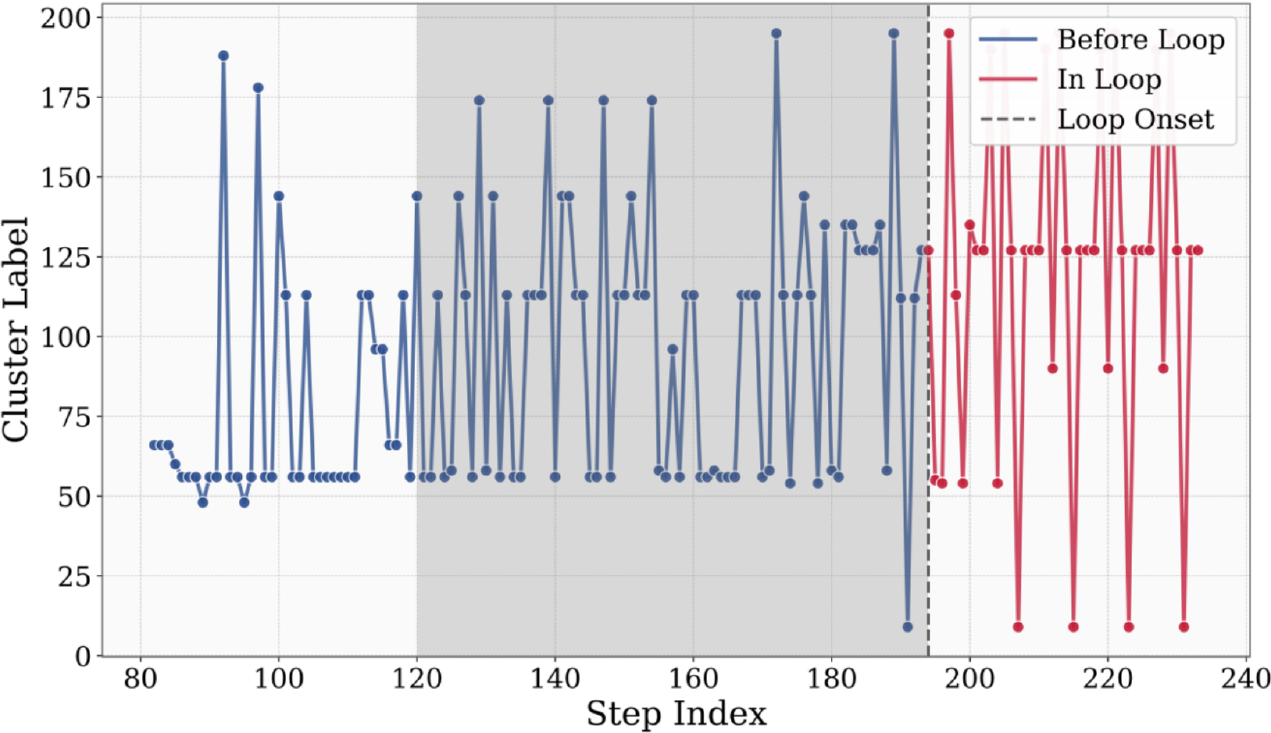}
        \caption{Cluster Trajectory Timeline}
        \label{fig:case02_path}
    \end{subfigure}
    \caption{\textbf{Reasoning graph visualization of a statement loop in DS-Qwen-14B (Case II).} The topological structure indicates a stable periodic attractor, preventing the model from generating new reasoning steps.}
    \label{fig:reasoning_graph_topology02}
\end{figure*}

\begin{figure*}[t]
    \centering
    \begin{subfigure}[b]{0.3\linewidth}
        \centering
        \includegraphics[width=\textwidth]{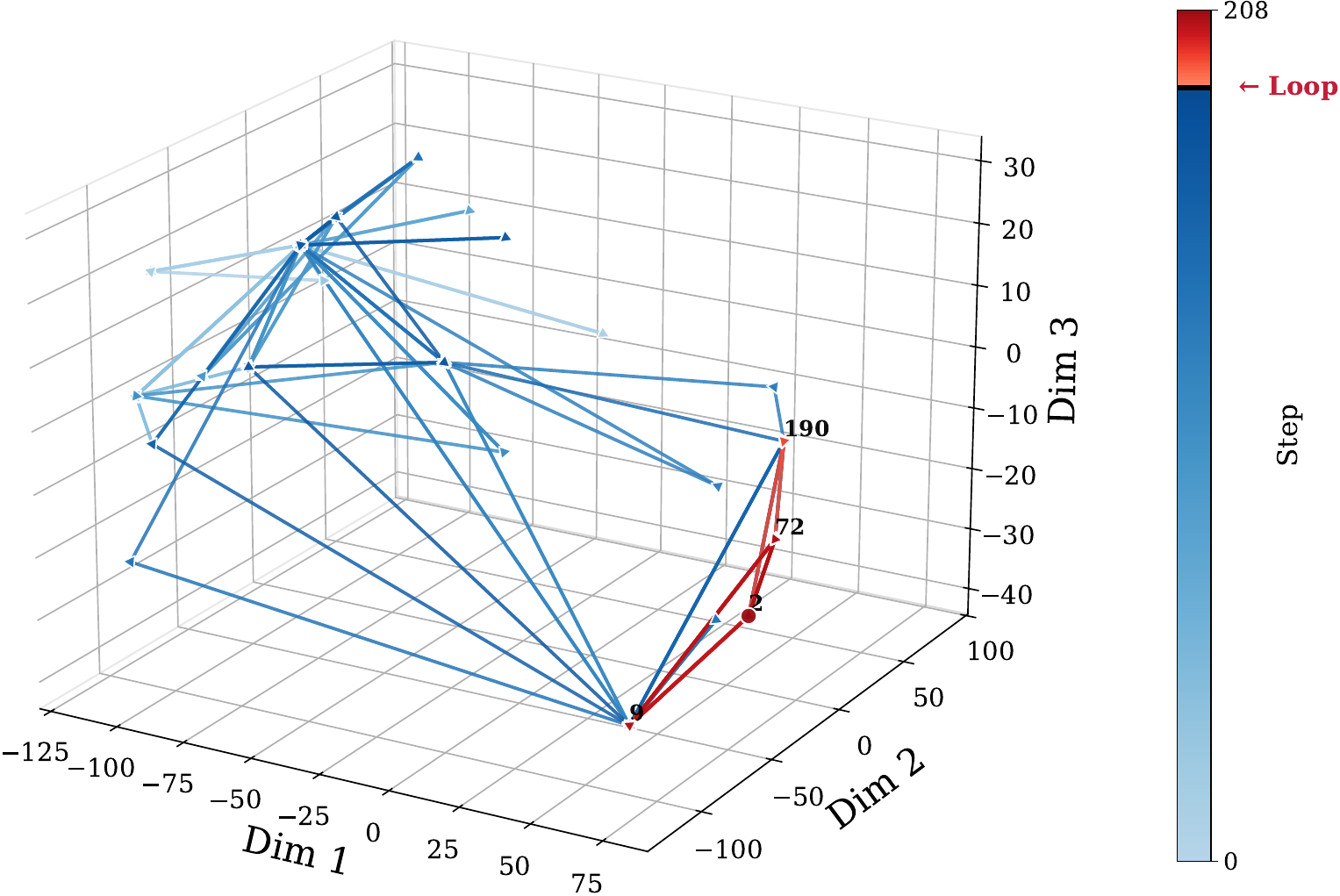}
        \caption{PCA Projection}
        \label{fig:case03_pca}
    \end{subfigure}
    \hfill
    \begin{subfigure}[b]{0.3\linewidth}
        \centering
        \includegraphics[width=\textwidth]{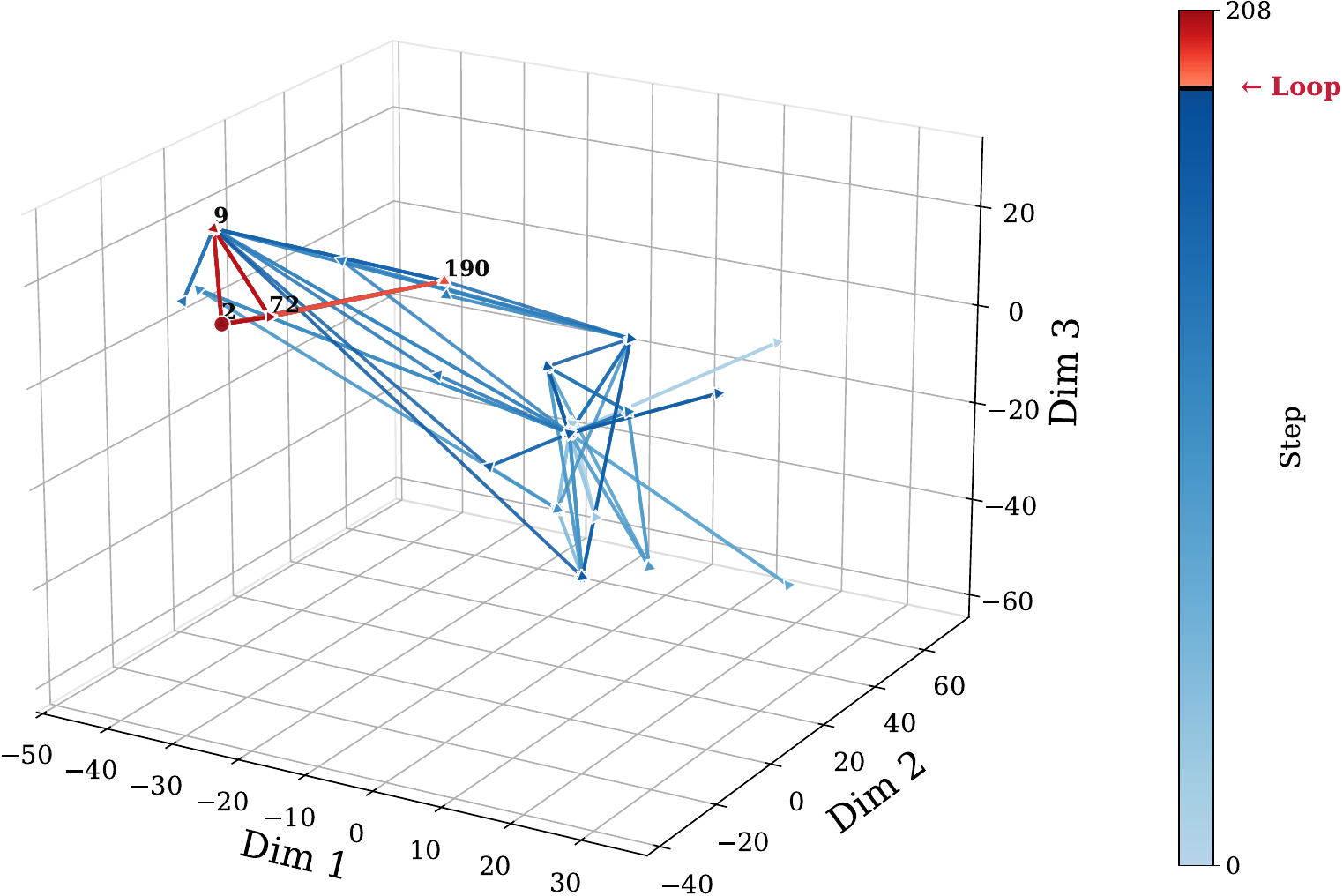}
        \caption{t-SNE Projection}
        \label{fig:case03_tsne}
    \end{subfigure}
    \hfill
    \begin{subfigure}[b]{0.35\linewidth}
        \centering
        \includegraphics[width=\textwidth]{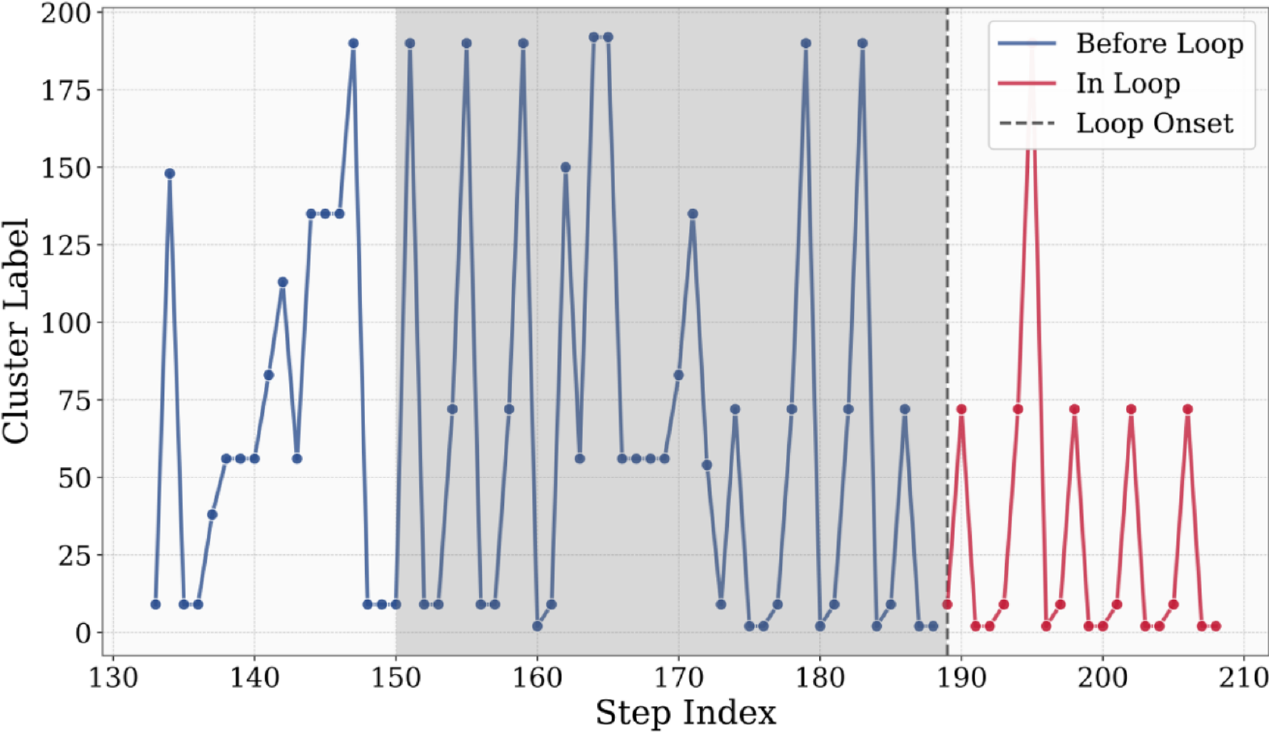}
        \caption{Cluster Trajectory Timeline}
        \label{fig:case03_path}
    \end{subfigure}
    \caption{\textbf{Reasoning graph visualization of a statement loop in DS-Qwen-14B (Case III).} Consistent with other cases, the semantic loop is established well before the repetition of exact tokens.}
    \label{fig:reasoning_graph_topology03}
\end{figure*}

\begin{figure*}[t]
    \centering
    \begin{subfigure}[b]{0.3\linewidth}
        \centering
        \includegraphics[width=\textwidth]{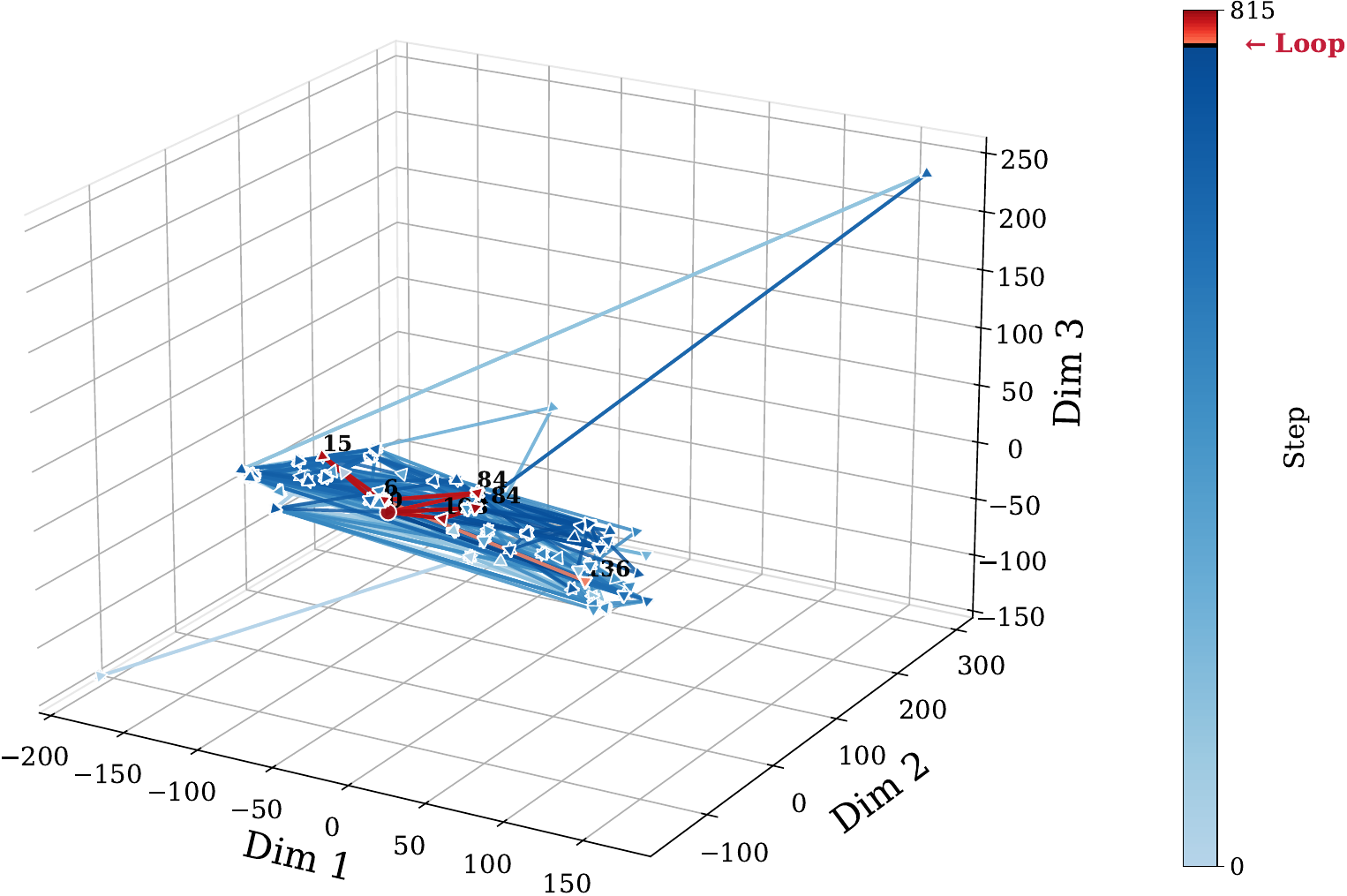}
        \caption{PCA Projection}
        \label{fig:case04_pca}
    \end{subfigure}
    \hfill
    \begin{subfigure}[b]{0.3\linewidth}
        \centering
        \includegraphics[width=\textwidth]{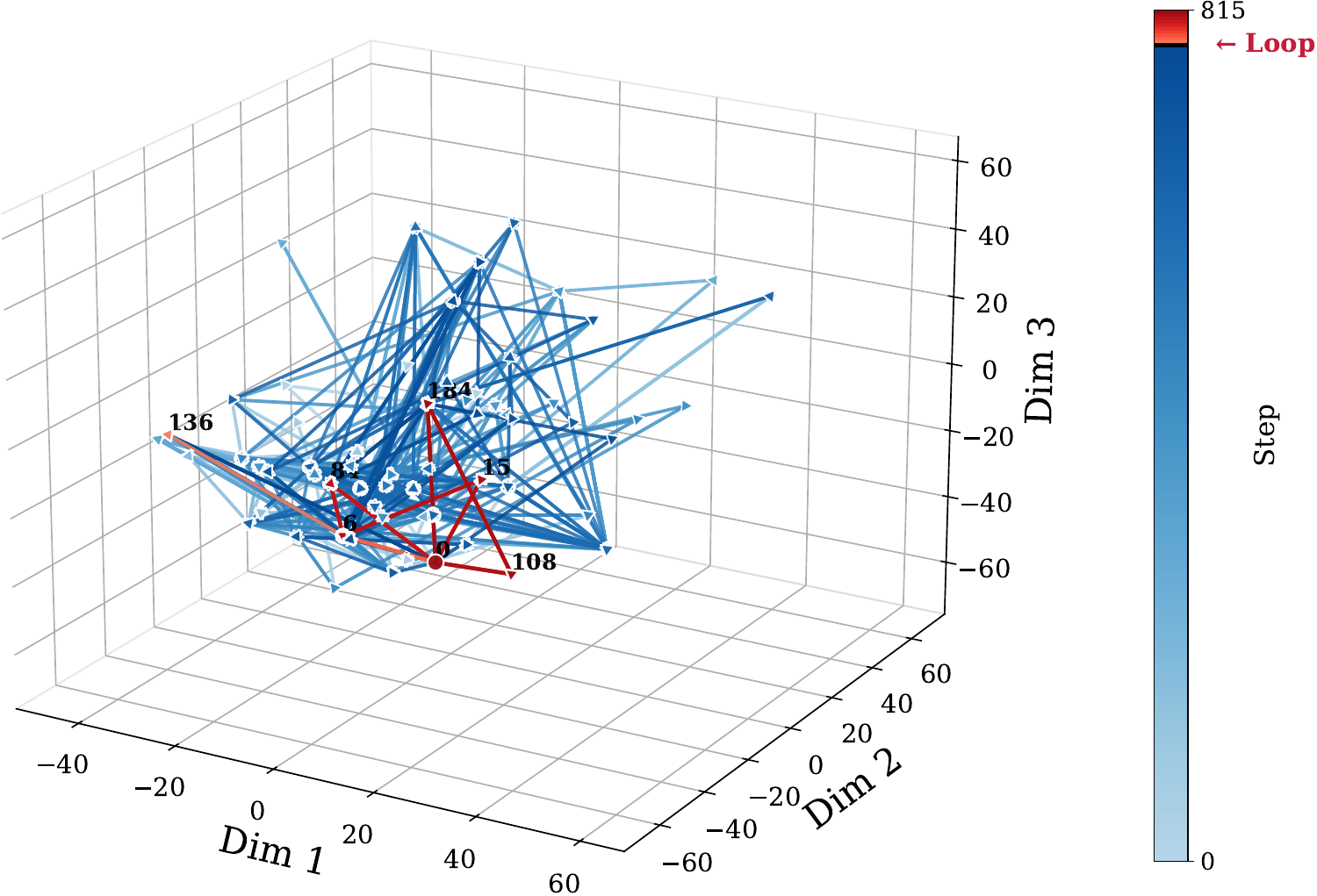}
        \caption{t-SNE Projection}
        \label{fig:case04_tsne}
    \end{subfigure}
    \hfill
    \begin{subfigure}[b]{0.35\linewidth}
        \centering
        \includegraphics[width=\textwidth]{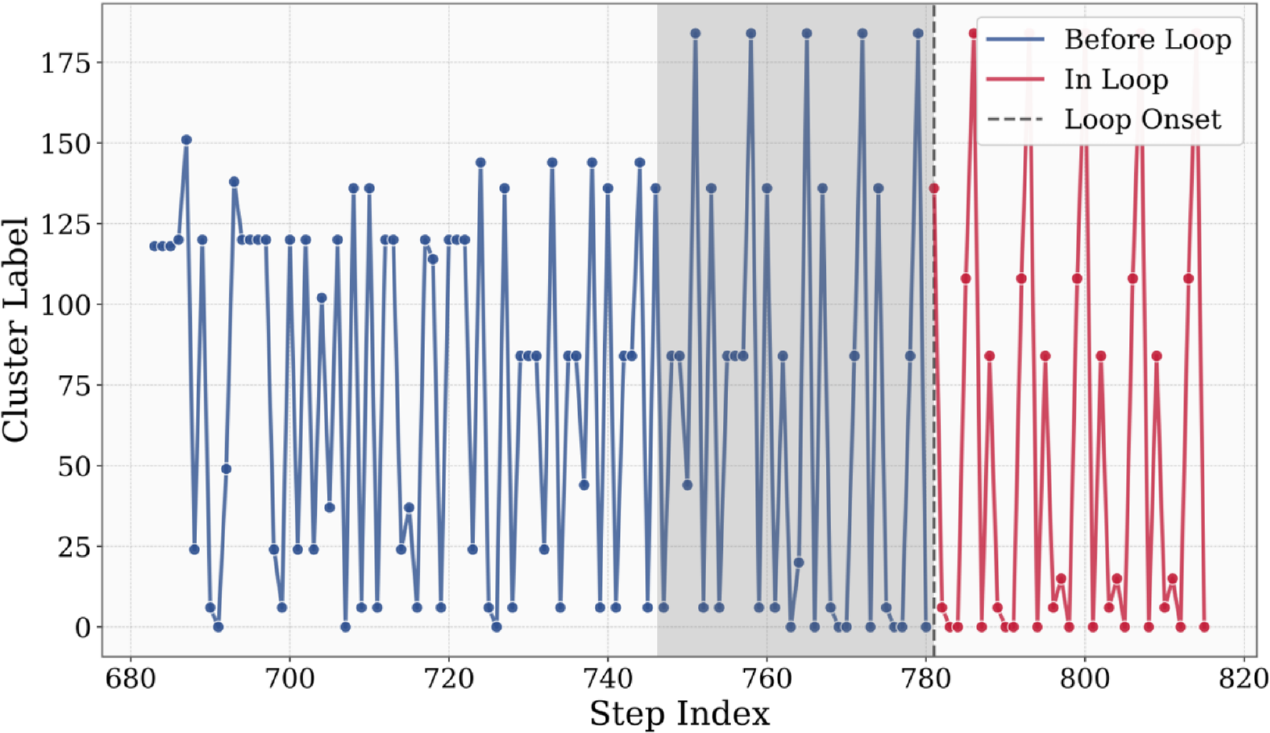}
        \caption{Cluster Trajectory Timeline}
        \label{fig:case04_path}
    \end{subfigure}
    \caption{\textbf{Reasoning graph visualization of a statement loop in Qwen3-8B (Case I).} Similar to DS-Qwen, Qwen3-8B exhibits a clear collapse into low-dimensional manifolds during loop formation.}
    \label{fig:reasoning_graph_topology04}
\end{figure*}

\begin{figure*}[t]
    \centering
    \begin{subfigure}[b]{0.3\linewidth}
        \centering
        \includegraphics[width=\textwidth]{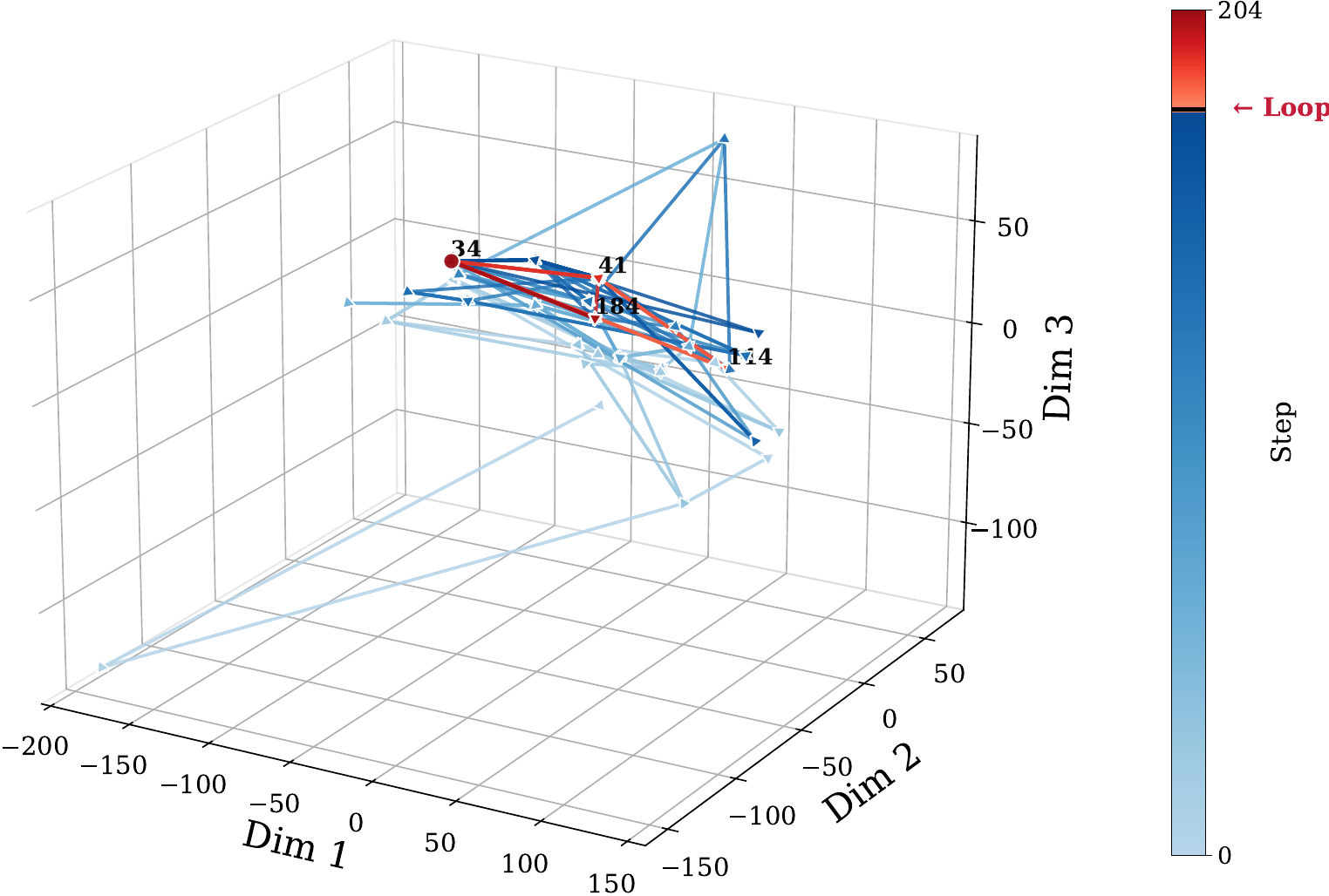}
        \caption{PCA Projection}
        \label{fig:case05_pca}
    \end{subfigure}
    \hfill
    \begin{subfigure}[b]{0.3\linewidth}
        \centering
        \includegraphics[width=\textwidth]{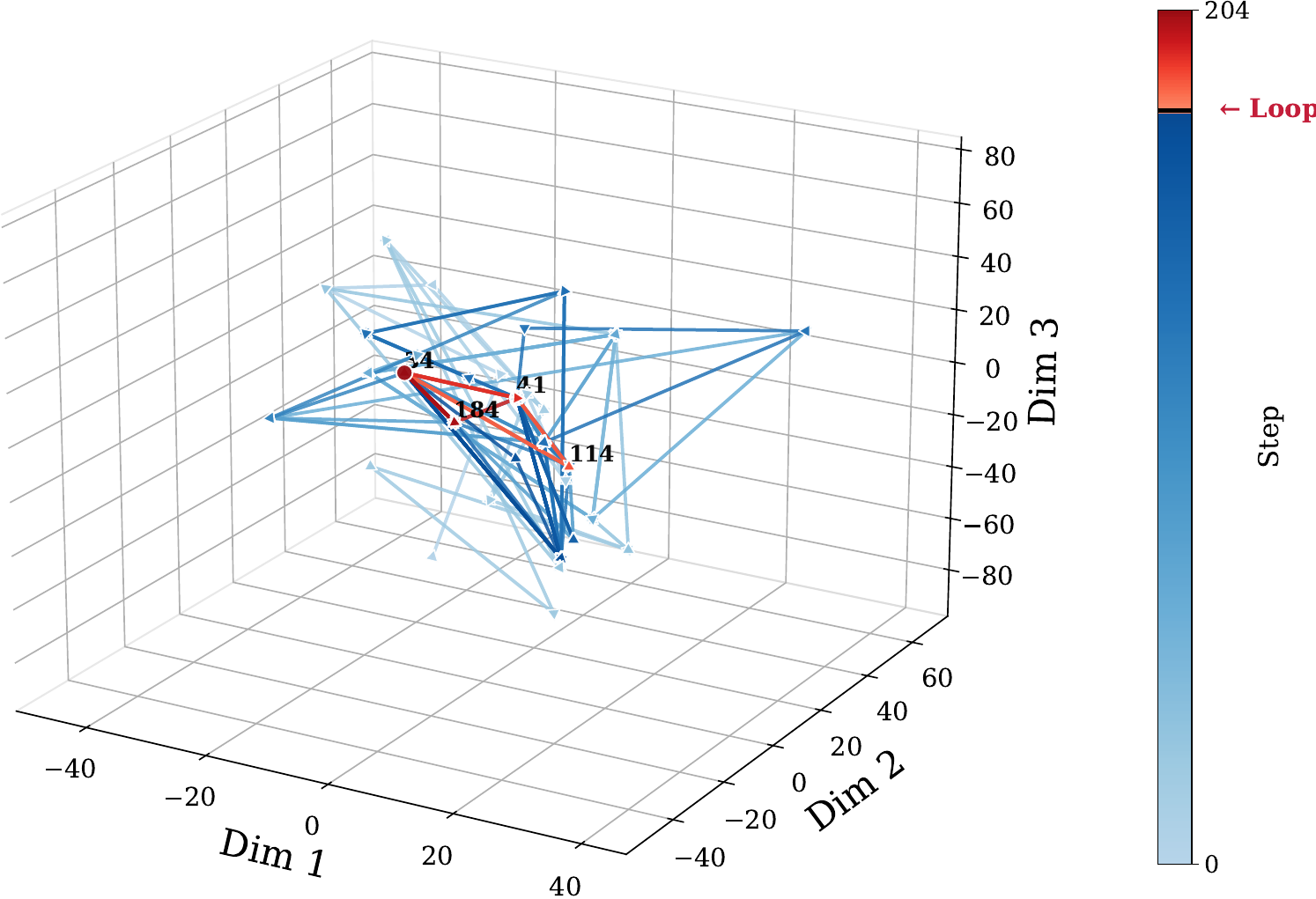}
        \caption{t-SNE Projection}
        \label{fig:case05_tsne}
    \end{subfigure}
    \hfill
    \begin{subfigure}[b]{0.35\linewidth}
        \centering
        \includegraphics[width=\textwidth]{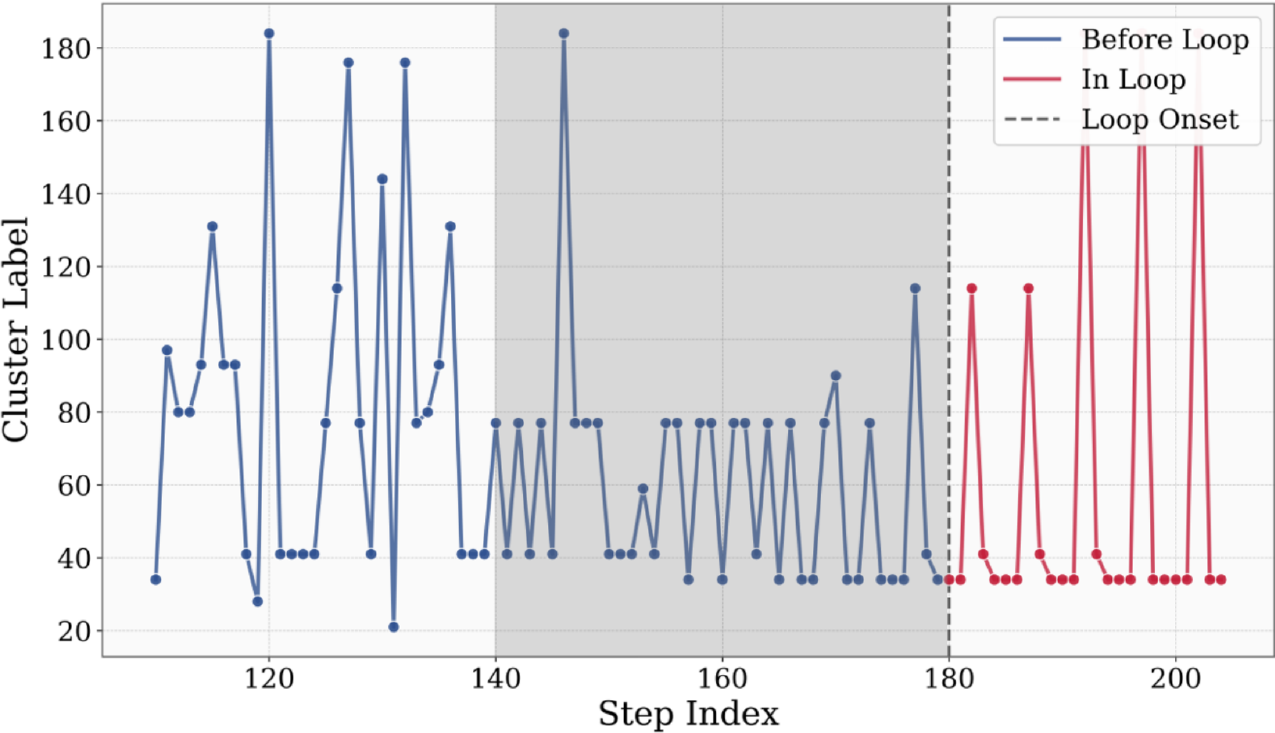}
        \caption{Cluster Trajectory Timeline}
        \label{fig:case05_path}
    \end{subfigure}
    \caption{\textbf{Reasoning graph visualization of a statement loop in Qwen3-8B (Case II).} The alignment between the geometric loop in (a)/(b) and the semantic label repetition in (c) confirms the multi-view consistency of the phenomenon.}
    \label{fig:reasoning_graph_topology05}
\end{figure*}

\begin{figure*}[t]
    \centering
    \begin{subfigure}[b]{0.3\linewidth}
        \centering
        \includegraphics[width=\textwidth]{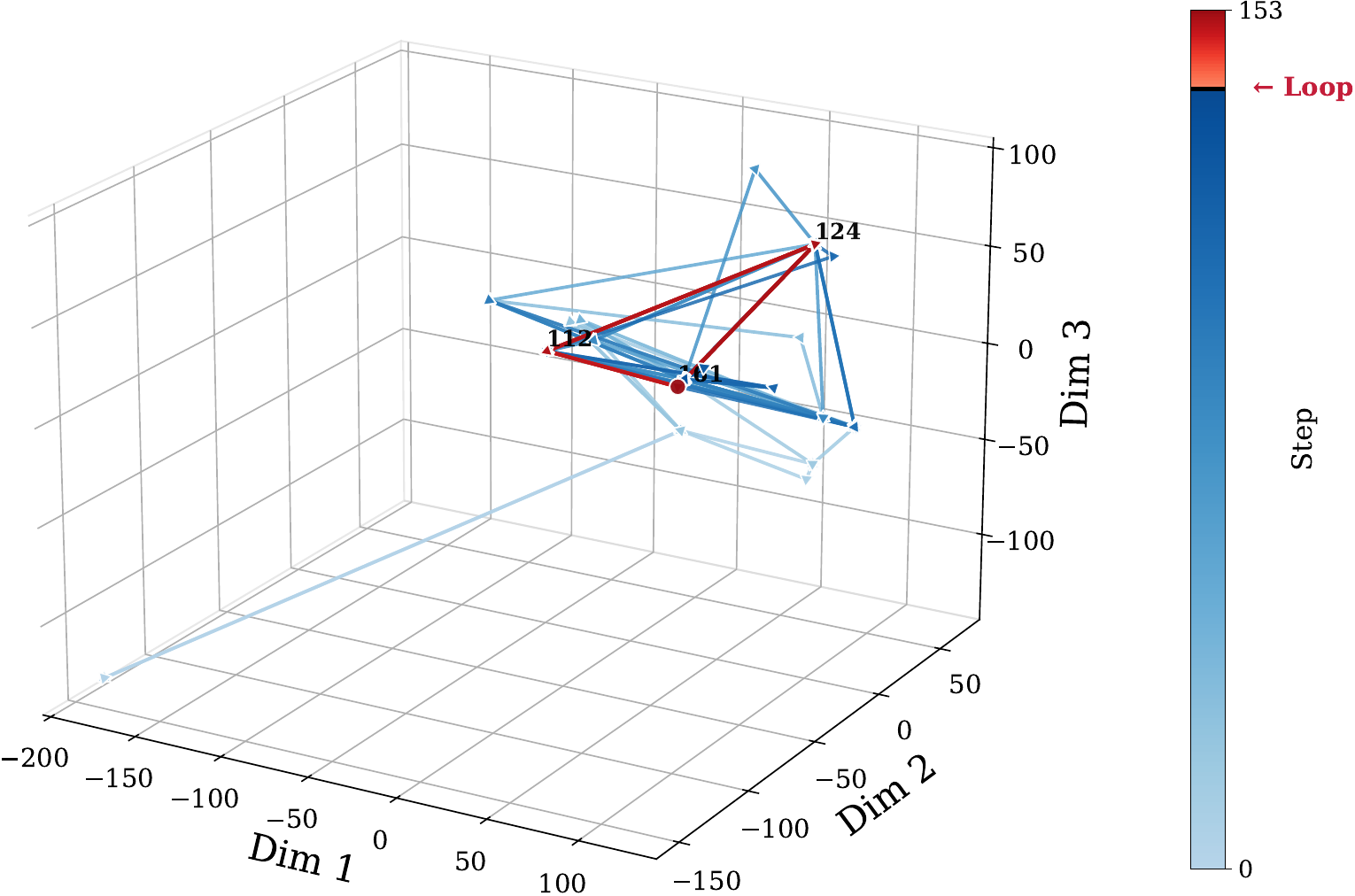}
        \caption{PCA Projection}
        \label{fig:case06_pca}
    \end{subfigure}
    \hfill
    \begin{subfigure}[b]{0.3\linewidth}
        \centering
        \includegraphics[width=\textwidth]{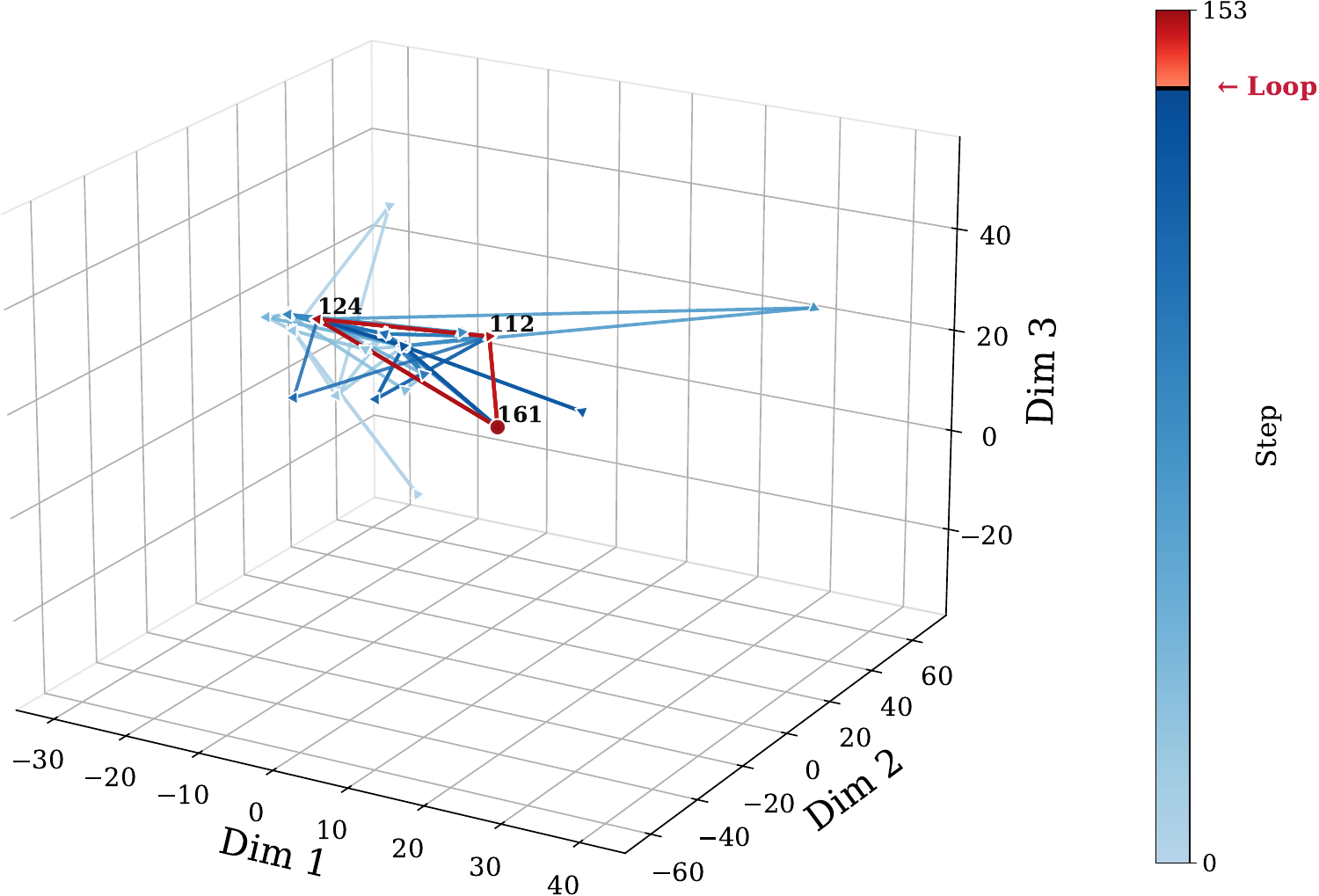}
        \caption{t-SNE Projection}
        \label{fig:case06_tsne}
    \end{subfigure}
    \hfill
    \begin{subfigure}[b]{0.35\linewidth}
        \centering
        \includegraphics[width=\textwidth]{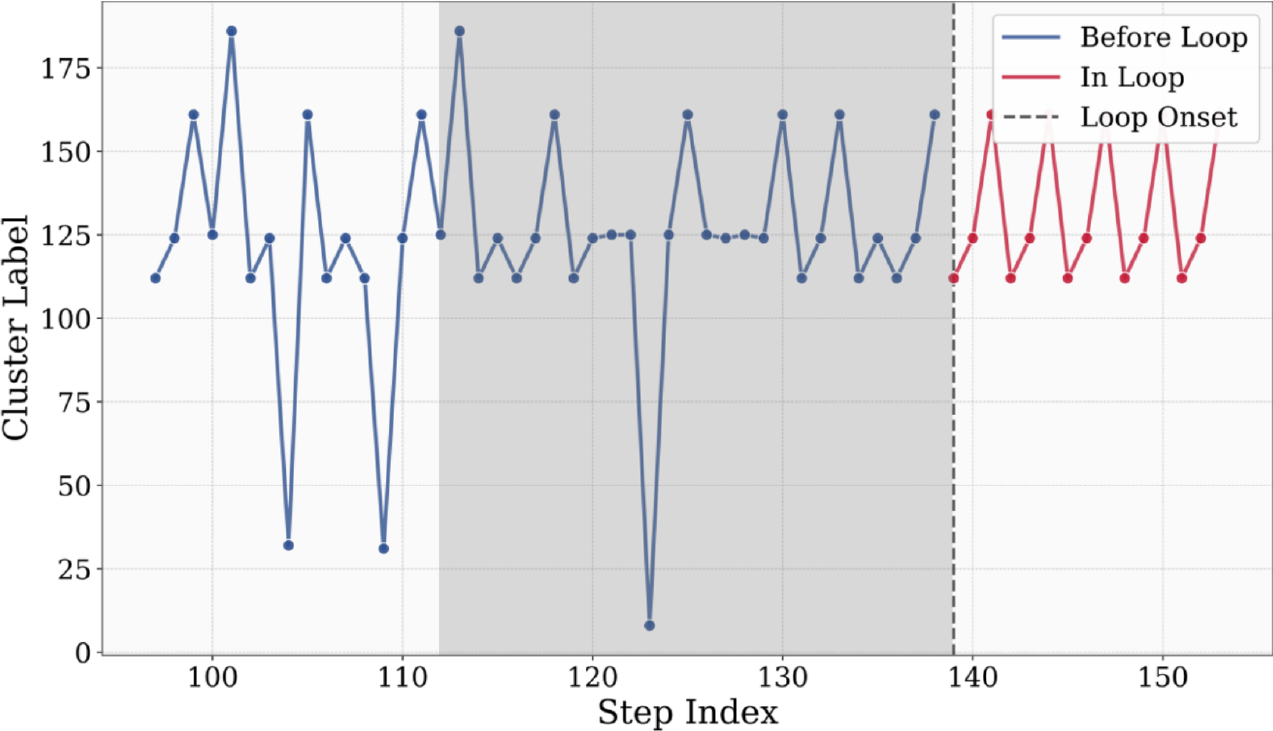}
        \caption{Cluster Trajectory Timeline}
        \label{fig:case06_path}
    \end{subfigure}
    \caption{\textbf{Reasoning graph visualization of a statement loop in Qwen3-8B (Case III).} The recurrence of cluster labels in (c) (shaded area) highlights the deterministic nature of the trap.}
    \label{fig:reasoning_graph_topology06}
\end{figure*}

Finally, we provide supplementary visualizations of the reasoning topology to substantiate the universality of the phase transition phenomenon. 
Figures~\ref{fig:reasoning_graph_topology01}, \ref{fig:reasoning_graph_topology02}, and \ref{fig:reasoning_graph_topology03} depict representative loop cases for DS-Qwen-14B, while Figures~\ref{fig:reasoning_graph_topology04}, \ref{fig:reasoning_graph_topology05}, and \ref{fig:reasoning_graph_topology06} illustrate cases for Qwen3-8B.
Across both models, the PCA (a) and t-SNE (b) projections consistently reveal that the internal hidden states collapse into a low-dimensional periodic orbit, a geometric signature of the limit cycle. Furthermore, the Cluster Trajectory timelines (c) confirm that \textit{semantic circularity} systematically precedes the onset of explicit textual repetition. This temporal lag provides strong evidence that the model enters a deterministic loop state in the semantic space before the pathology manifests in the surface tokens.


We further examine the semantic content associated with the recurring cluster labels to understand the cognitive state of the model during loops. Table~\ref{tab:ds_reasoning_examples} and Table~\ref{tab:qwen3_reasoning_examples} present representative reasoning steps corresponding to specific Label IDs for DS-Qwen-14B and Qwen3-8B, respectively.
As observed, the sentences mapped to these loop clusters are predominantly characterized by introspective reflection and self-doubt. Frequent discursive markers such as \textit{``Wait,''} \textit{``But,''} and \textit{``Maybe''} indicate that the model is actively detecting potential errors or attempting to re-frame the problem. However, rather than resolving the ambiguity, these pivots trigger recursive retrials, trapping the model in a cycle of hesitation and invalid verification.

\begin{table*}[t]
    \centering
    \caption{\textbf{Semantic Analysis of Cluster Labels (DS-Qwen-14B).} Representative reasoning steps associated with specific Label IDs in the loop phase. The content is dominated by reflective markers (e.g., \textit{``Wait''}, \textit{``Alternatively''}), indicating a state of hesitation and self-correction.}
    \label{tab:ds_reasoning_examples}
    
    \renewcommand{\arraystretch}{1.3} 
    \renewcommand{\tabularxcolumn}[1]{m{#1}}

    \begin{tabularx}{\textwidth}{c X}
        \toprule
        \textbf{Label ID} & \textbf{Examples of Reasoning Steps} \\
        \midrule
        
        \textbf{41} & 
        \begin{itemize}[leftmargin=*, topsep=0pt, partopsep=0pt, itemsep=2pt, parsep=0pt]
            \item Alternatively, maybe the sets are $\{11,12,13\}$, $\{14,15,16\}$, but arranged differently.
            \item Wait, the problem says ``three sets of integers.'' So, maybe the sets are $\{11,12,13\}$, $\{14,15,16\}$, $\{17,18,19\}$, but arranged such that the columns are also triplets.
        \end{itemize} \\
        \midrule
        
        \textbf{119} & 
        \begin{itemize}[leftmargin=*, topsep=0pt, partopsep=0pt, itemsep=2pt, parsep=0pt]
            \item Which is the same as before.
            \item This seems like a complex problem.
            \item Wait, but that's not helpful.
        \end{itemize} \\
        \midrule
        
        \textbf{9} & 
        \begin{itemize}[leftmargin=*, topsep=0pt, partopsep=0pt, itemsep=2pt, parsep=0pt]
            \item Wait, but 28 is even, so maybe we can have a consistent assignment where the pattern wraps around correctly.
            \item Wait, maybe the key is that the cycle length is even, so we can have a consistent assignment where the truthful people are spaced in such a way that the implications hold.
        \end{itemize} \\
        \midrule
        
        \textbf{90} & 
        \begin{itemize}[leftmargin=*, topsep=0pt, partopsep=0pt, itemsep=2pt, parsep=0pt]
            \item Wait, let me think differently.
            \item Wait, maybe I need to think about the equation differently.
            \item Wait, maybe I misread the problem. Let me check again.
        \end{itemize} \\
        \midrule
        
        \textbf{54} & 
        \begin{itemize}[leftmargin=*, topsep=0pt, partopsep=0pt, itemsep=2pt, parsep=0pt]
            \item But as we saw earlier, this leads to column sums that don't match the row sums.
            \item Wait, none of these work. Therefore, this assignment is invalid.
            \item No valid pairs. Therefore, invalid.
        \end{itemize} \\
        
        \bottomrule
    \end{tabularx}
\end{table*}

\begin{table*}[t]
    \centering
    \caption{\textbf{Semantic Analysis of Cluster Labels (Qwen3-8B).} Representative examples of reasoning steps within loop clusters. The prevalence of phrases like \textit{``Let me try''} and \textit{``This is getting too complicated''} reflects the model's struggle to resolve contradictions through recursive retrials.}
    \label{tab:qwen3_reasoning_examples}
    
    \renewcommand{\arraystretch}{1.3} 
    \renewcommand{\tabularxcolumn}[1]{m{#1}}

    \begin{tabularx}{\textwidth}{c X}
        \toprule
        \textbf{Label ID} & \textbf{Examples of Reasoning Steps} \\
        \midrule
        
        \textbf{114} & 
        \begin{itemize}[leftmargin=*, topsep=0pt, partopsep=0pt, itemsep=2pt, parsep=0pt]
            \item Let me try $C_1 = F$, $C_2 = F$, others $T$. This has consecutive F's, which violates the condition for the $B_i$ statements. But let's check.
            \item This suggests that there is no solution, but the problem says there is one. I must have made a mistake in the initial assumptions.
            \item But this is not correct, as it leads to contradictions. I'm stuck.
        \end{itemize} \\
        \midrule
        
        \textbf{124} & 
        \begin{itemize}[leftmargin=*, topsep=0pt, partopsep=0pt, itemsep=2pt, parsep=0pt]
            \item This is getting too complicated. Maybe I need to approach this differently.
            \item Let me try to use the additional constraints to find possible numbers.
            \item Wait, let me re-read the problem statement once more.
        \end{itemize} \\
        \midrule
        
        \textbf{161} & 
        \begin{itemize}[leftmargin=*, topsep=0pt, partopsep=0pt, itemsep=2pt, parsep=0pt]
            \item This indicates that the pattern is invalid.
            \item Now, check the product constraints.
        \end{itemize} \\
        \midrule
        
        \textbf{108} & 
        \begin{itemize}[leftmargin=*, topsep=0pt, partopsep=0pt, itemsep=2pt, parsep=0pt]
            \item But the sum of each column is equal to the sum of each row.
            \item But if each set has three consecutive integers, and all three sets have the same sum, then as before, they must be the same set.
            \item But the sum of each column is the sum of one number from each set.
        \end{itemize} \\
        \midrule
        
        \textbf{136} & 
        \begin{itemize}[leftmargin=*, topsep=0pt, partopsep=0pt, itemsep=2pt, parsep=0pt]
            \item But since $T_i$ is either 0 or 1, we can think of this as a system of equations.
            \item But this is a system of equations. Let me try to find a solution.
        \end{itemize} \\
        
        \bottomrule
    \end{tabularx}
\end{table*}

\subsection{Additional Results of `V-Shaped' Attention Patterns}
\label{sec:appendix_vshape}

\begin{figure*}[t]
    \centering
    \begin{subfigure}[b]{0.32\textwidth} 
        \centering
        \includegraphics[width=\textwidth]{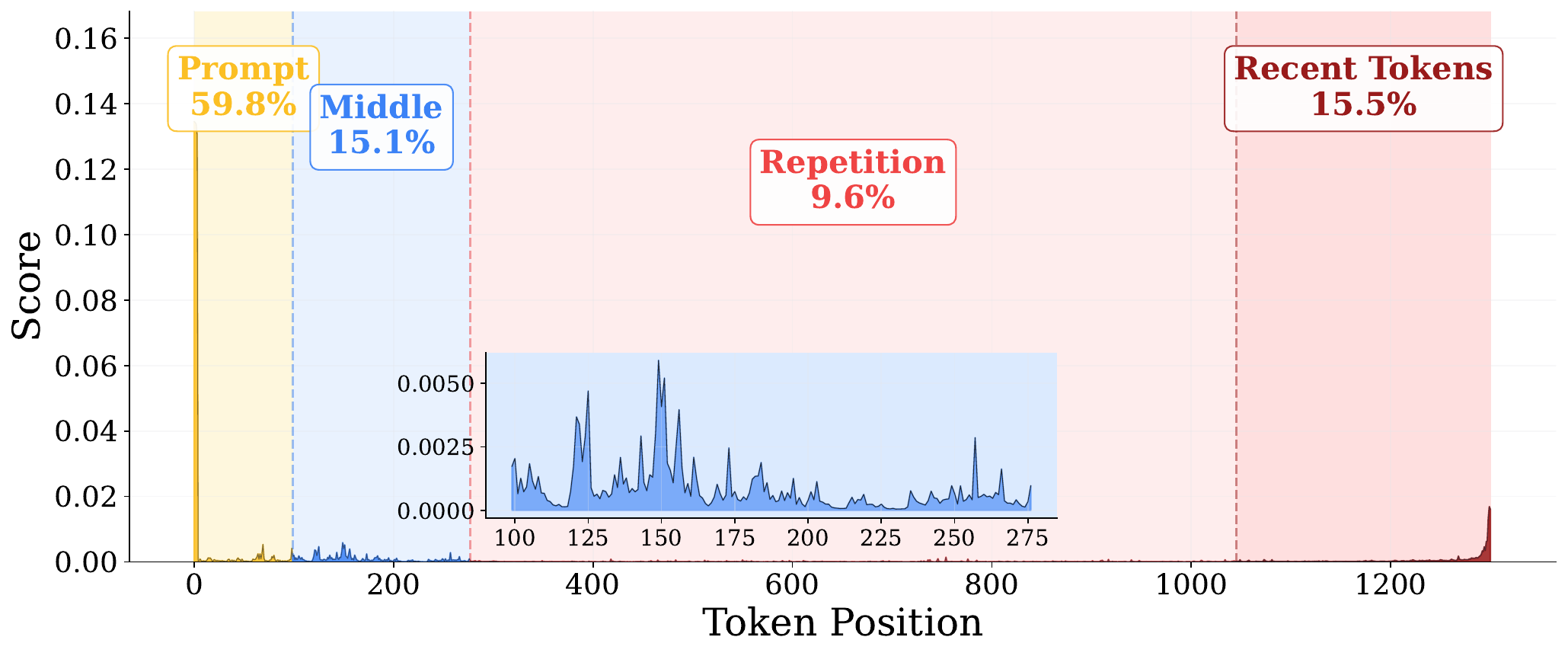}
        \caption{Case 1}
        \label{fig:vshape_ds_num_c1}
    \end{subfigure}
    \hfill 
    \begin{subfigure}[b]{0.32\textwidth} 
        \centering
        \includegraphics[width=\textwidth]{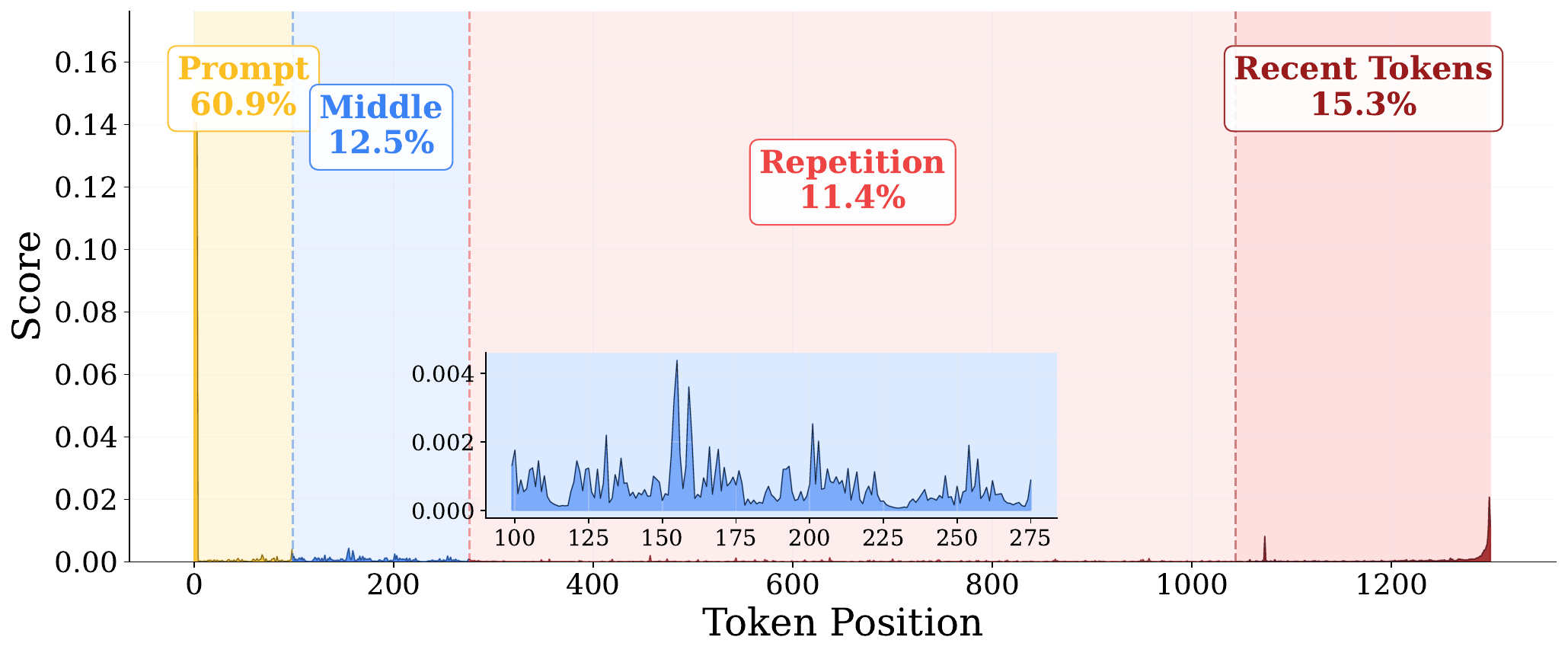}
        \caption{Case 2}
        \label{fig:vshape_ds_num_c2}
    \end{subfigure}
    \hfill 
    \begin{subfigure}[b]{0.32\textwidth} 
        \centering
        \includegraphics[width=\textwidth]{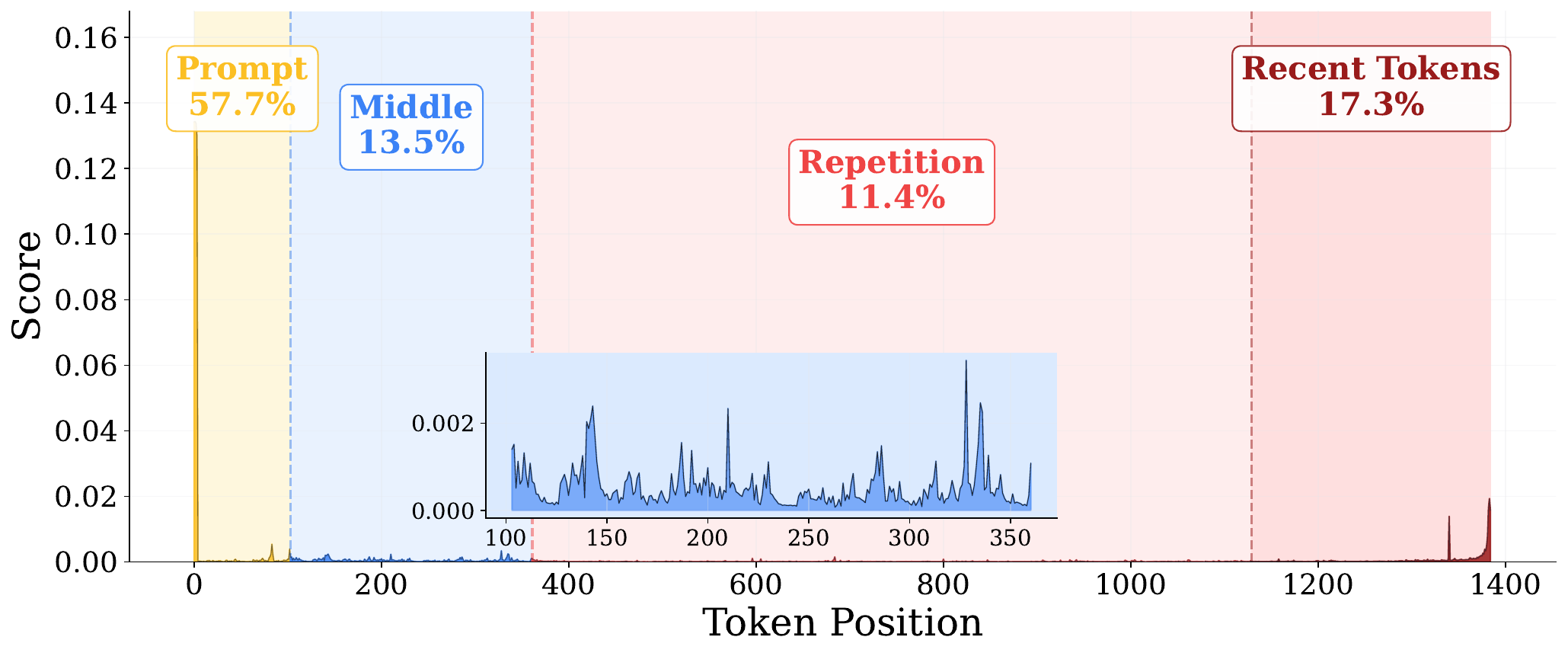}
        \caption{Case 3} 
        \label{fig:vshape_ds_num_c3} 
    \end{subfigure}
    \caption{\textbf{`V-Shaped' Attention Pattern in DS-Qwen-14B (Numerical Loops).} The heatmaps reveal a strong bias towards recent tokens. Once the loop begins, this local attention locks the model into repeating the immediate numerical sequence.}
    \label{fig:vshape_ds_num}
\end{figure*}

\begin{figure*}[t]
    \centering
    \begin{subfigure}[b]{0.32\textwidth} 
        \centering
        \includegraphics[width=\textwidth]{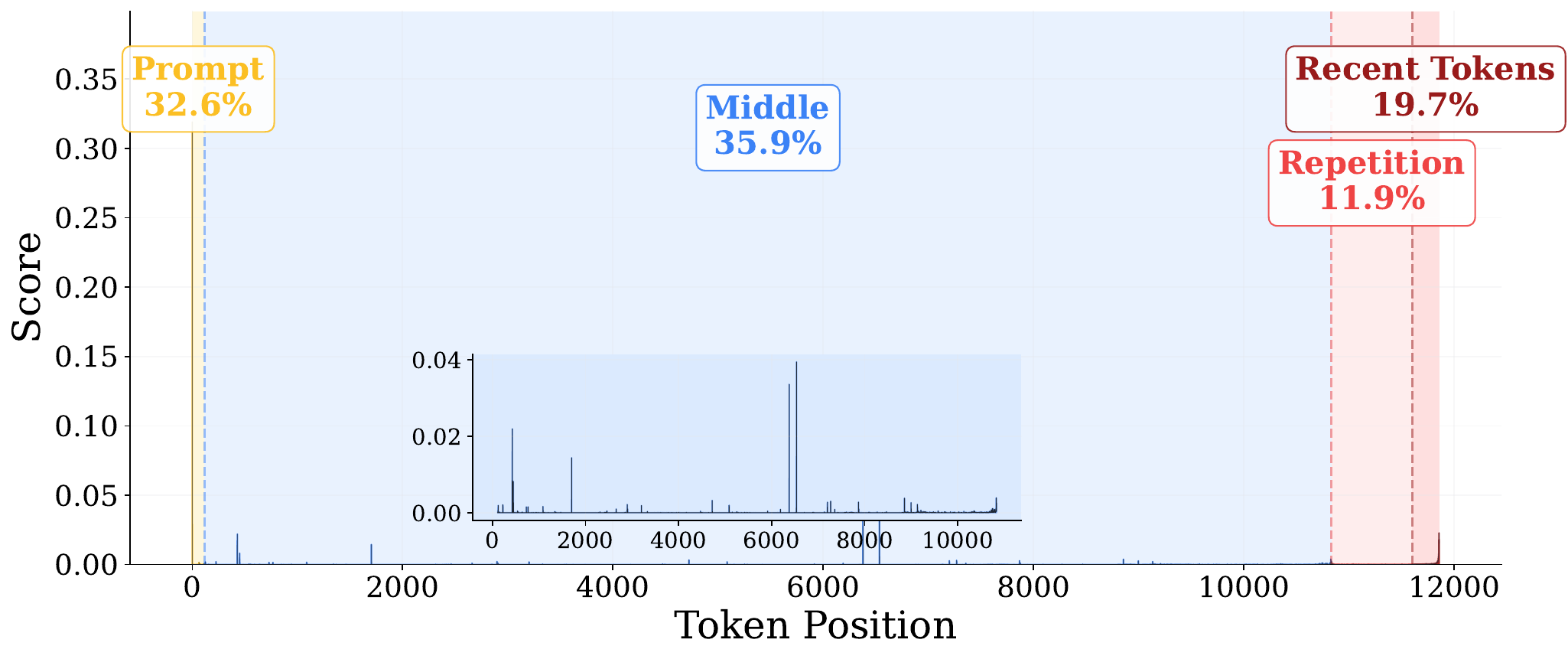}
        \caption{Case 1}
        \label{fig:vshape_qwen3_num_c1}
    \end{subfigure}
    \hfill 
    \begin{subfigure}[b]{0.32\textwidth} 
        \centering
        \includegraphics[width=\textwidth]{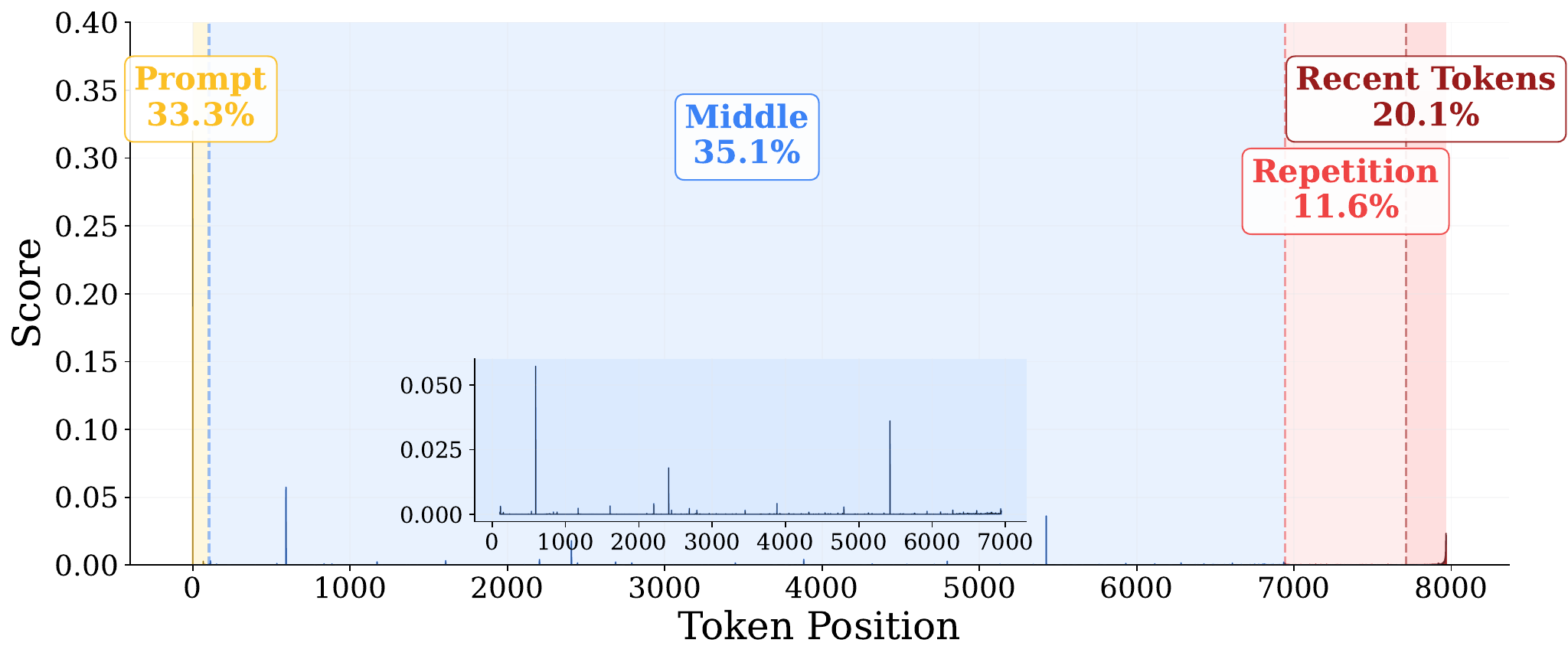}
        \caption{Case 2}
        \label{fig:vshape_qwen3_num_c2}
    \end{subfigure}
    \hfill 
    \begin{subfigure}[b]{0.32\textwidth} 
        \centering
        \includegraphics[width=\textwidth]{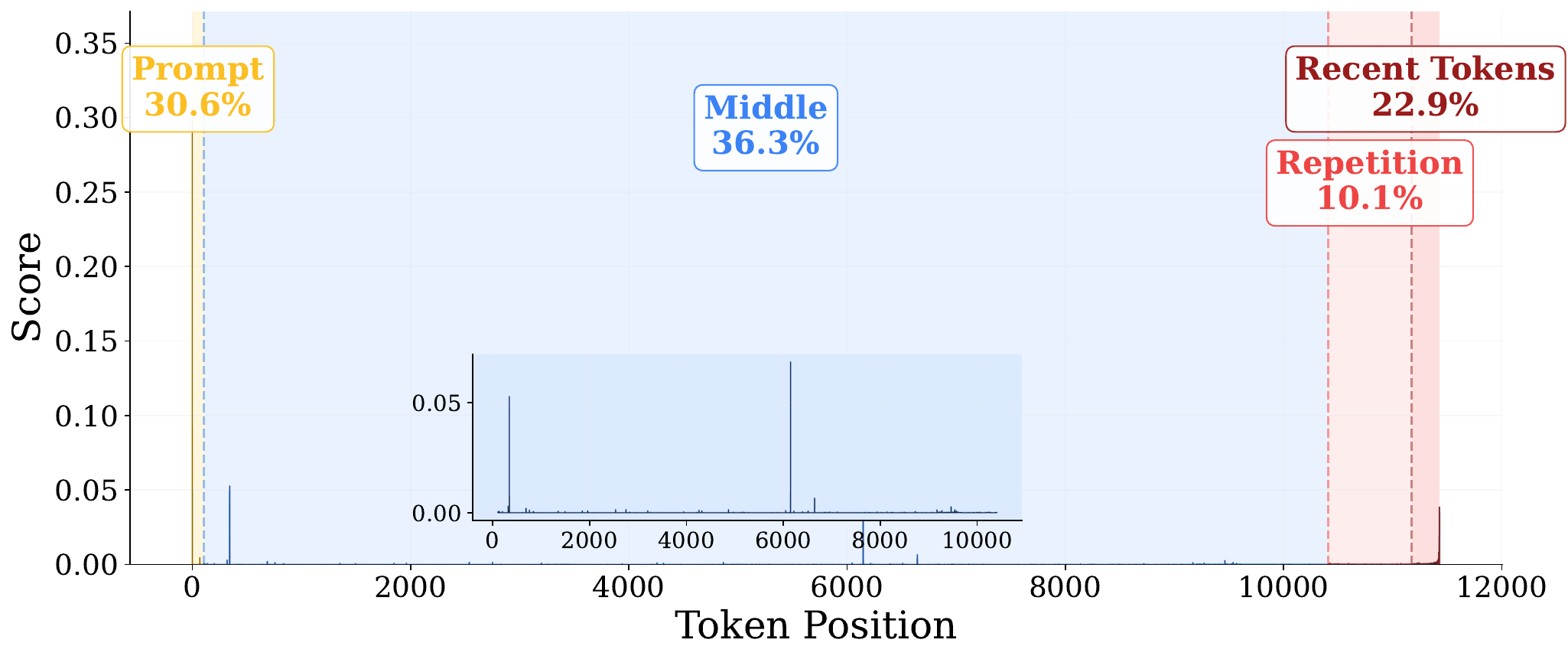}
        \caption{Case 3} 
        \label{fig:vshape_qwen3_num_c3} 
    \end{subfigure}
    \caption{\textbf{`V-Shaped' Attention Pattern in Qwen3-8B (Numerical Loops).} Consistent with DS-Qwen, Qwen3-8B also exhibits high attention concentrations on local history, confirming the universality of this mechanism in numerical tasks.}
    \label{fig:vshape_qwen3_num}
\end{figure*}

\begin{figure*}[t]
    \centering
    \begin{subfigure}[b]{0.32\textwidth} 
        \centering
        \includegraphics[width=\textwidth]{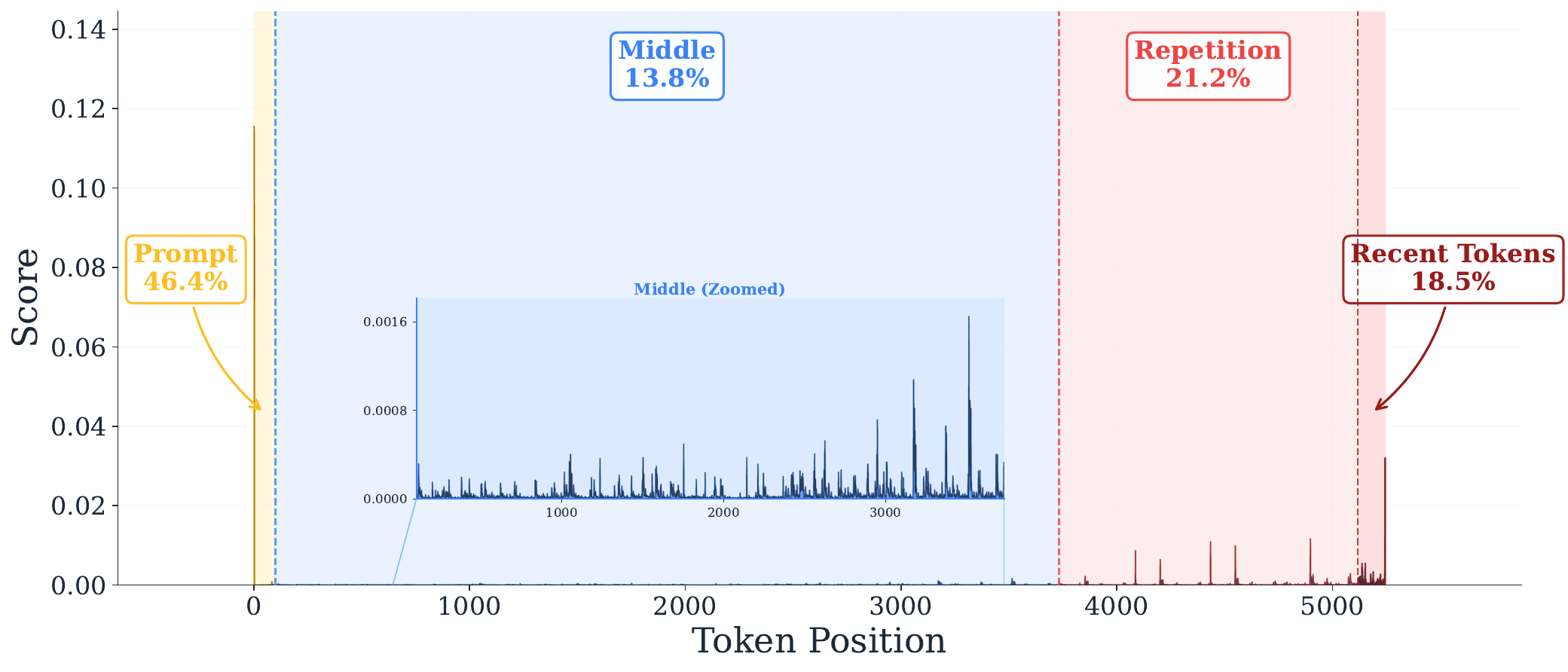}
        \caption{Case 1}
        \label{fig:vshape_ds_stmt_c1}
    \end{subfigure}
    \hfill 
    \begin{subfigure}[b]{0.32\textwidth} 
        \centering
        \includegraphics[width=\textwidth]{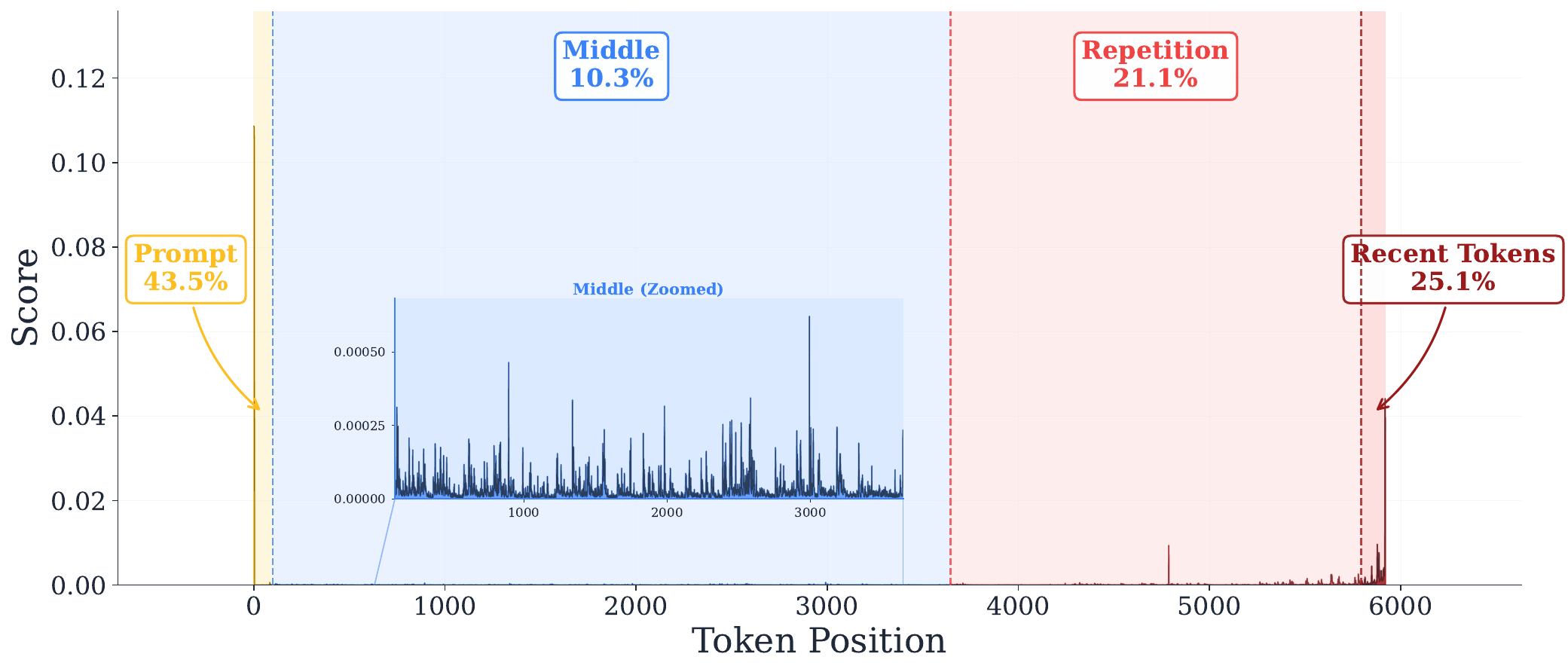}
        \caption{Case 2}
        \label{fig:vshape_ds_stmt_c2}
    \end{subfigure}
    \hfill 
    \begin{subfigure}[b]{0.32\textwidth} 
        \centering
        \includegraphics[width=\textwidth]{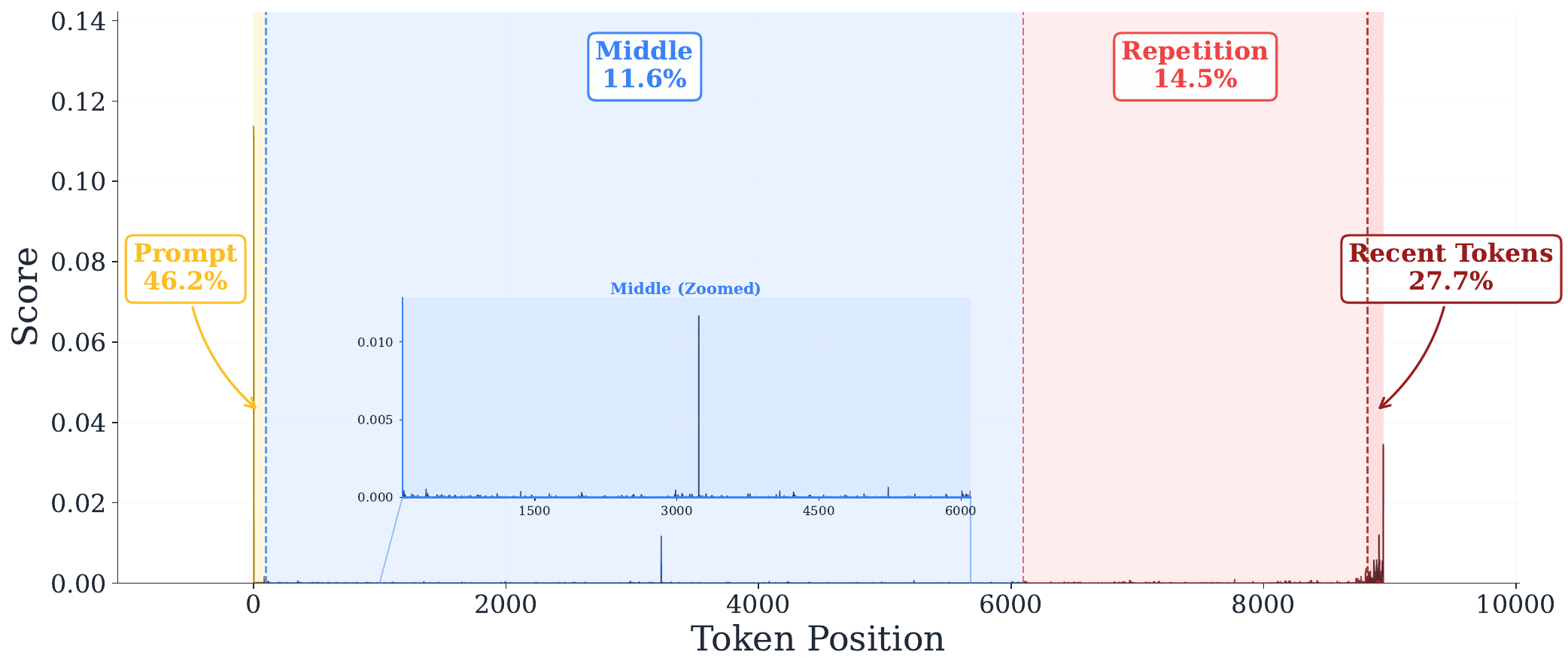}
        \caption{Case 3} 
        \label{fig:vshape_ds_stmt_c3} 
    \end{subfigure}
    \caption{\textbf{`V-Shaped' Attention Pattern in DS-Qwen-14B (Statement Loops).} In statement loops, the model similarly ignores global context, focusing instead on the immediately preceding reflective phrases, which reinforces the circular logic.}
    \label{fig:vshape_ds_stmt}
\end{figure*}

\begin{figure*}[t]
    \centering
    \begin{subfigure}[b]{0.32\textwidth} 
        \centering
        \includegraphics[width=\textwidth]{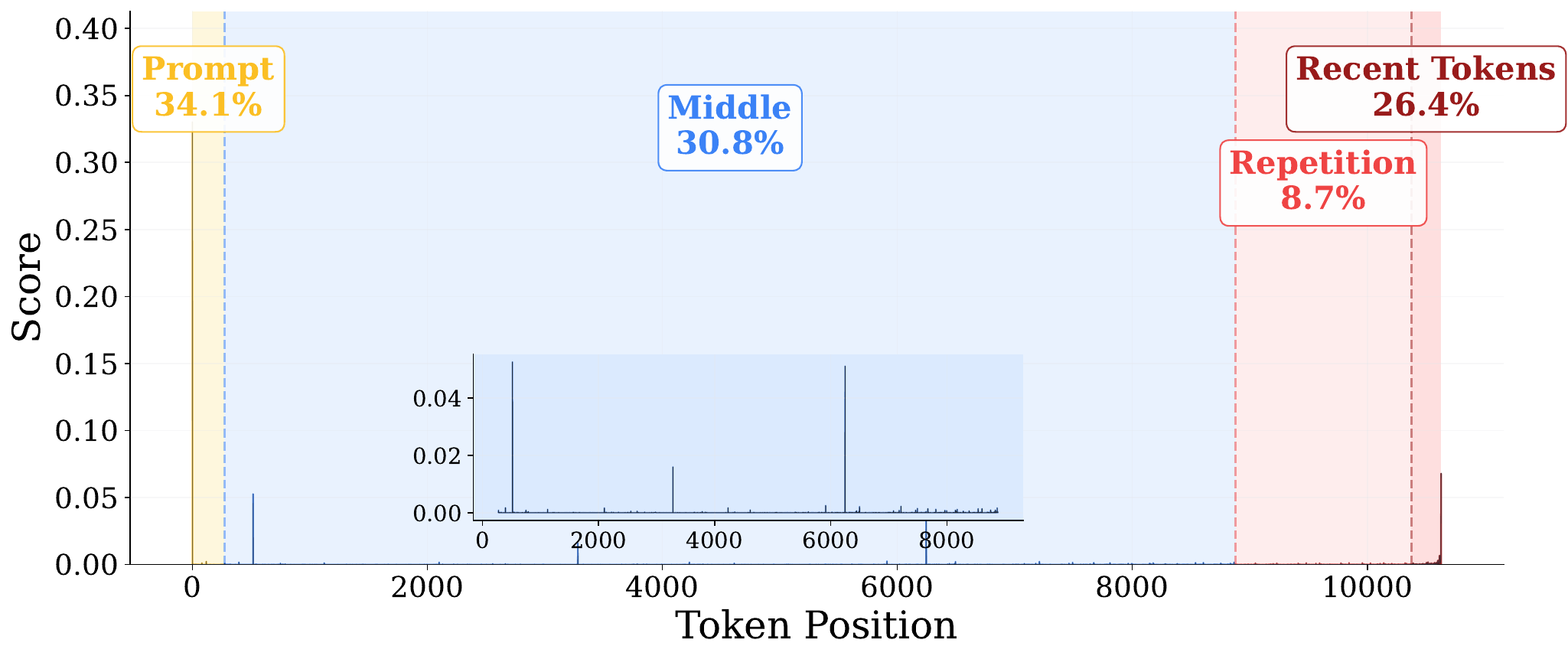}
        \caption{Case 1}
        \label{fig:vshape_qwen3_stmt_c1}
    \end{subfigure}
    \hfill 
    \begin{subfigure}[b]{0.32\textwidth} 
        \centering
        \includegraphics[width=\textwidth]{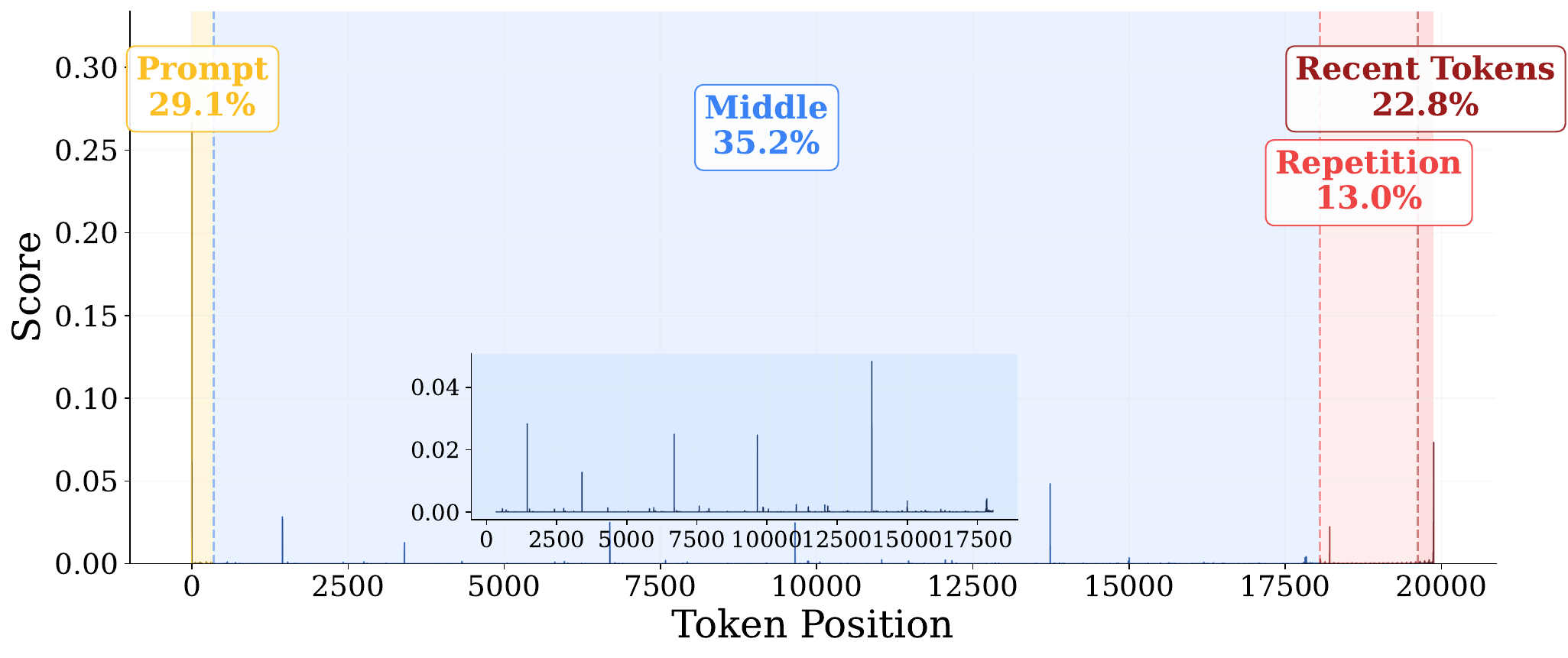}
        \caption{Case 2}
        \label{fig:vshape_qwen3_stmt_c2}
    \end{subfigure}
    \hfill 
    \begin{subfigure}[b]{0.32\textwidth} 
        \centering
        \includegraphics[width=\textwidth]{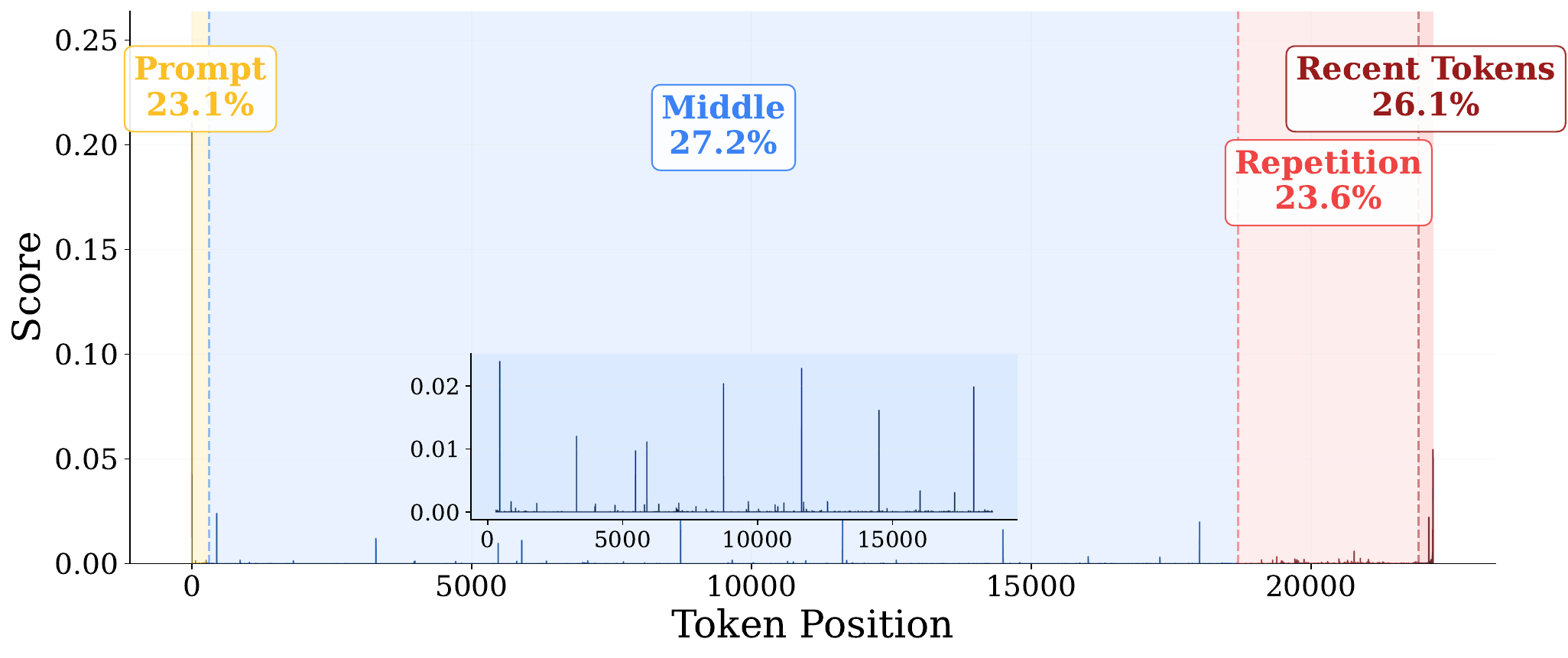}
        \caption{Case 3} 
        \label{fig:vshape_qwen3_stmt_c3} 
    \end{subfigure}
    \caption{\textbf{`V-Shaped' Attention Pattern in Qwen3-8B (Statement Loops).} The persistence of the V-shaped pattern in statement loops across different architectures highlights it as a fundamental mechanism driving the degeneration process.}
    \label{fig:vshape_qwen3_stmt}
\end{figure*}

In this section, we present supplementary evidence demonstrating that the ``V-shaped'' attention bias is a pervasive phenomenon across different models and task modalities. As illustrated in Figures~\ref{fig:vshape_ds_num} to \ref{fig:vshape_qwen3_stmt}, both DS-Qwen-14B and Qwen3-8B exhibit a distinct attention pattern where the model disproportionately attends to the most immediate history. 
Crucially, as the repetition persists, these recent tokens consist entirely of redundant content generated in the previous cycle. Consequently, the model's fixation on local history creates a self-reinforcing feedback loop: by attending solely to recent repetitions, the model is compelled to generate further repetitions, thereby deepening the loop and preventing any escape to the broader context.

\subsection{Additional Results of Statement Triggers}


\begin{figure}[t]
    \centering
    \includegraphics[width=\linewidth]{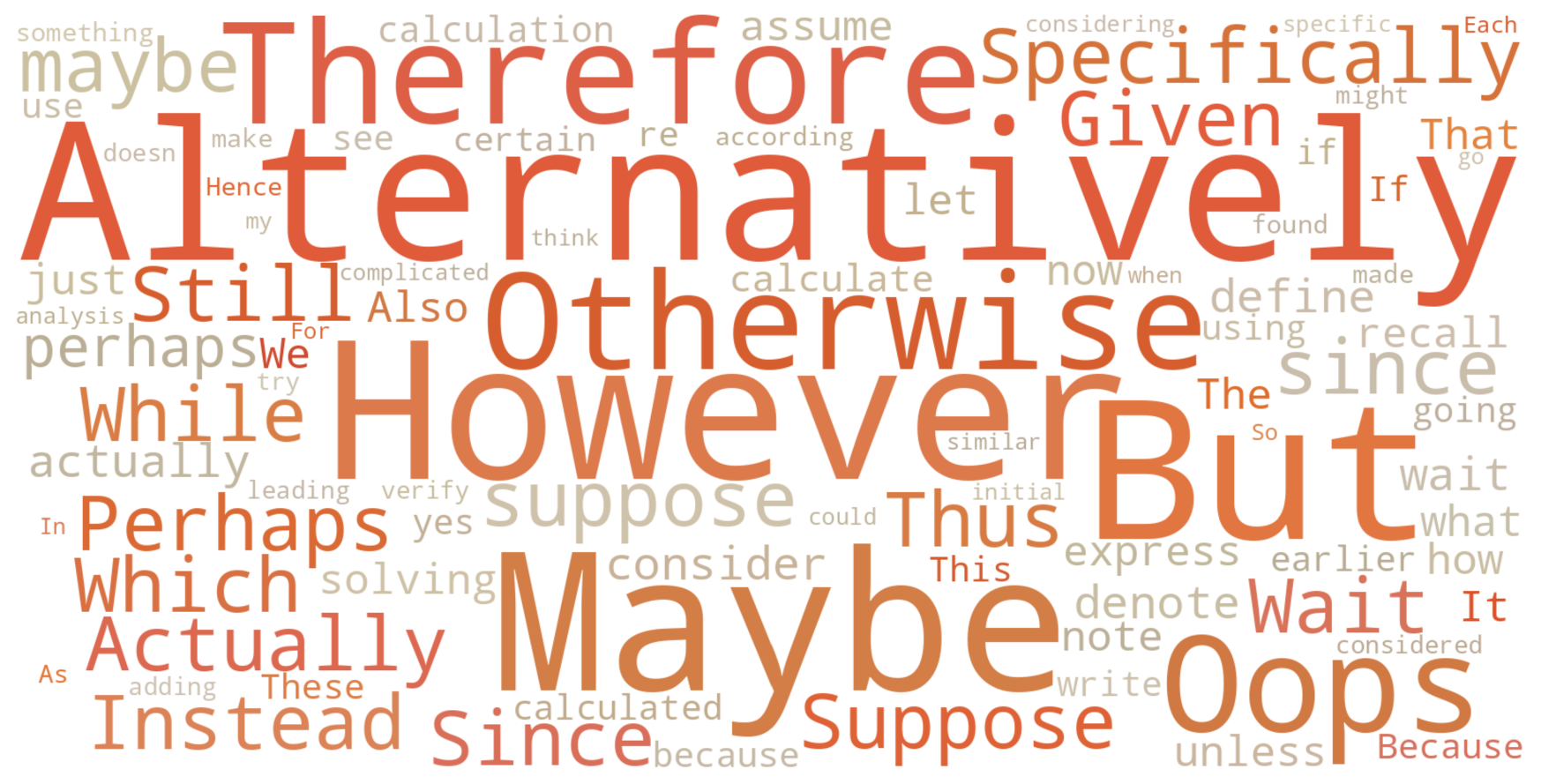}
    \caption{\textbf{Visualization of High-Entropy Tokens.} The word cloud illustrates the lexicon of potential loop triggers. Tokens in \textbf{warm colors} represent the identified pivots that introduce branching logic and uncertainty, while tokens in \textbf{neutral colors} denote background vocabulary. The font size correlates with the token's frequency within the high-entropy set.}
    \label{fig:wordcloud}
\end{figure}

\begin{figure}[t]
    \centering
    \includegraphics[width=\linewidth, height=0.55\linewidth]{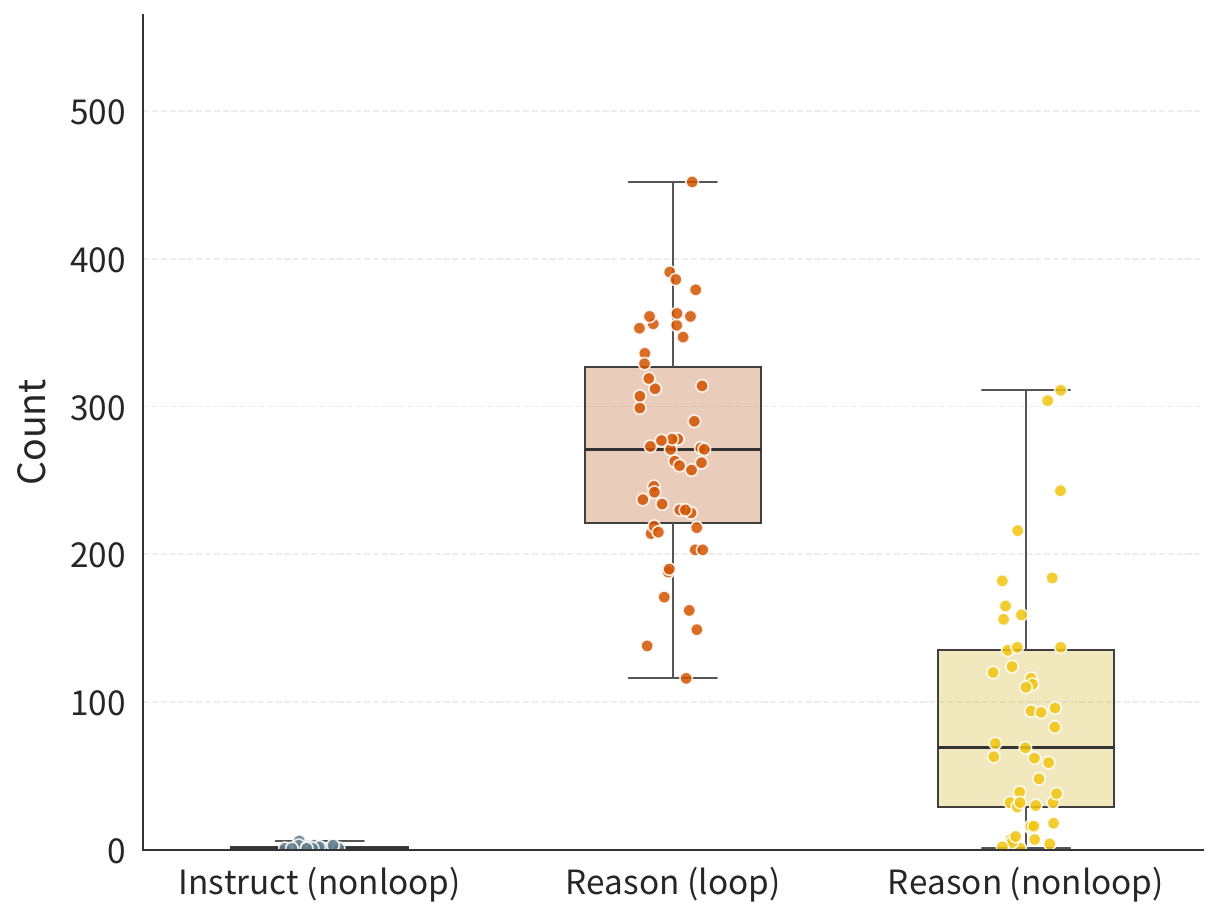}
    \caption{\textbf{Distribution of High-Entropy Tokens across Models and States.} We compare the frequency of high-entropy tokens in DS-Qwen-14B (Loop vs. Non-Loop) and Qwen2.5-14B-Instruct. The ``Loop'' state exhibits a significantly higher density of these uncertainty-inducing tokens, distinguishing it from normal reasoning and standard instruction tuning.}
    \label{fig:entropy_distribution_box}
\end{figure}

Following the methodology of \citet{Wang2025BeyondT8}, we curated a specialized lexicon of high-entropy tokens that typically appear at the onset of reasoning sentences. These tokens serve as critical semantic markers, often signaling logical transitions, uncertainty, or self-correction.
Figure~\ref{fig:wordcloud} visualizes the distribution of these tokens. We employ a color-coded scheme to distinguish the structural roles of the vocabulary: \textbf{warm tones} (e.g., orange and rust) highlight the pivotal tokens of primary interest, such as \textit{``Therefore''}, \textit{``Alternatively''}, and \textit{``However''}. These tokens act as high-entropy junctions that frequently precipitate the attention shifts observed in statement loops. Conversely, \textbf{neutral tones} (grey) represent auxiliary background vocabulary. The prominence of introspective markers—such as \textit{``Maybe''}, \textit{``But''}, and \textit{``Oops''}—in the warm spectrum further corroborates our hypothesis that statement loops are fundamentally driven by the model's recursive but futile attempts at self-correction.

\begin{figure*}[t]
    \centering
    \begin{subfigure}[b]{0.32\textwidth}
        \centering
        \includegraphics[width=\textwidth]{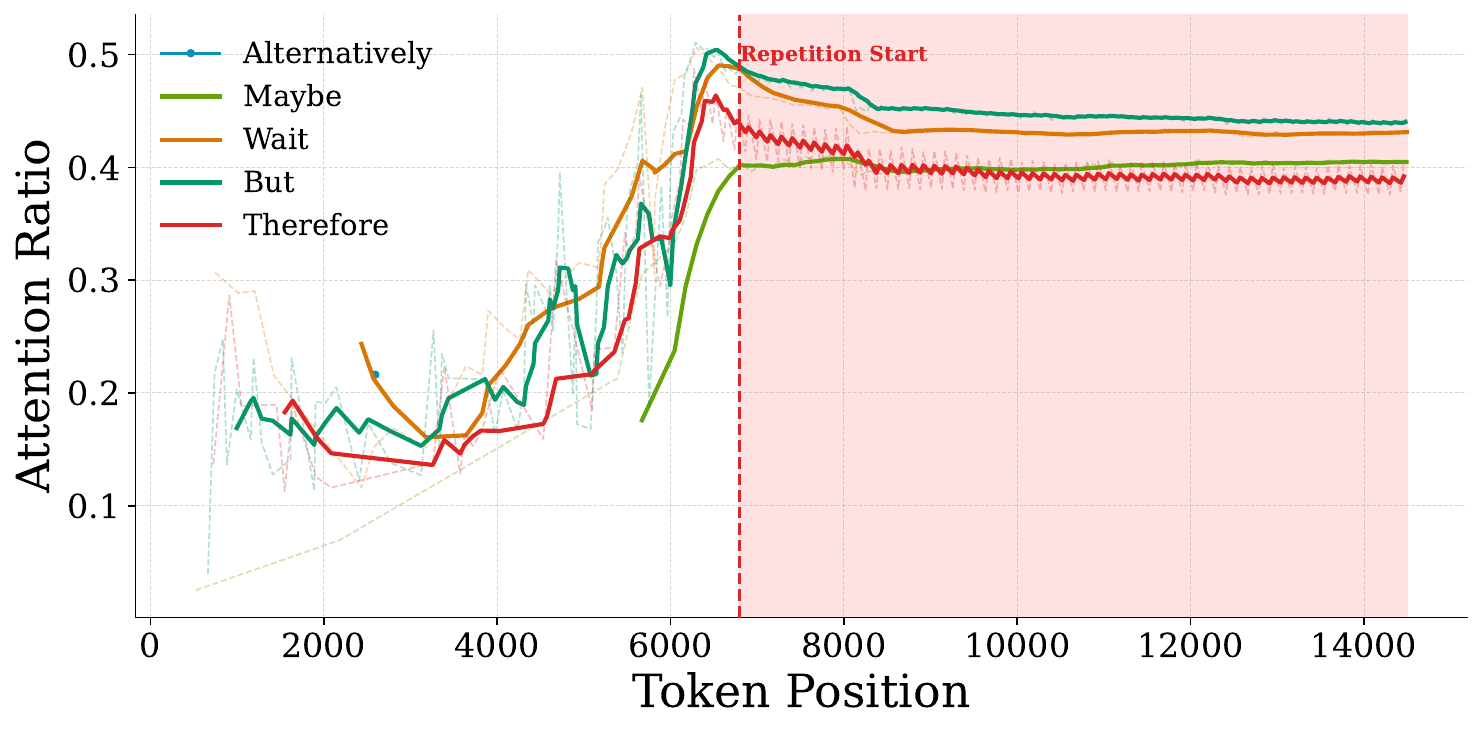}
        \caption{Loop Case 1}
        \label{fig:ds_attn_loop_c1}
    \end{subfigure}
    \hfill
    \begin{subfigure}[b]{0.32\textwidth}
        \centering
        \includegraphics[width=\textwidth]{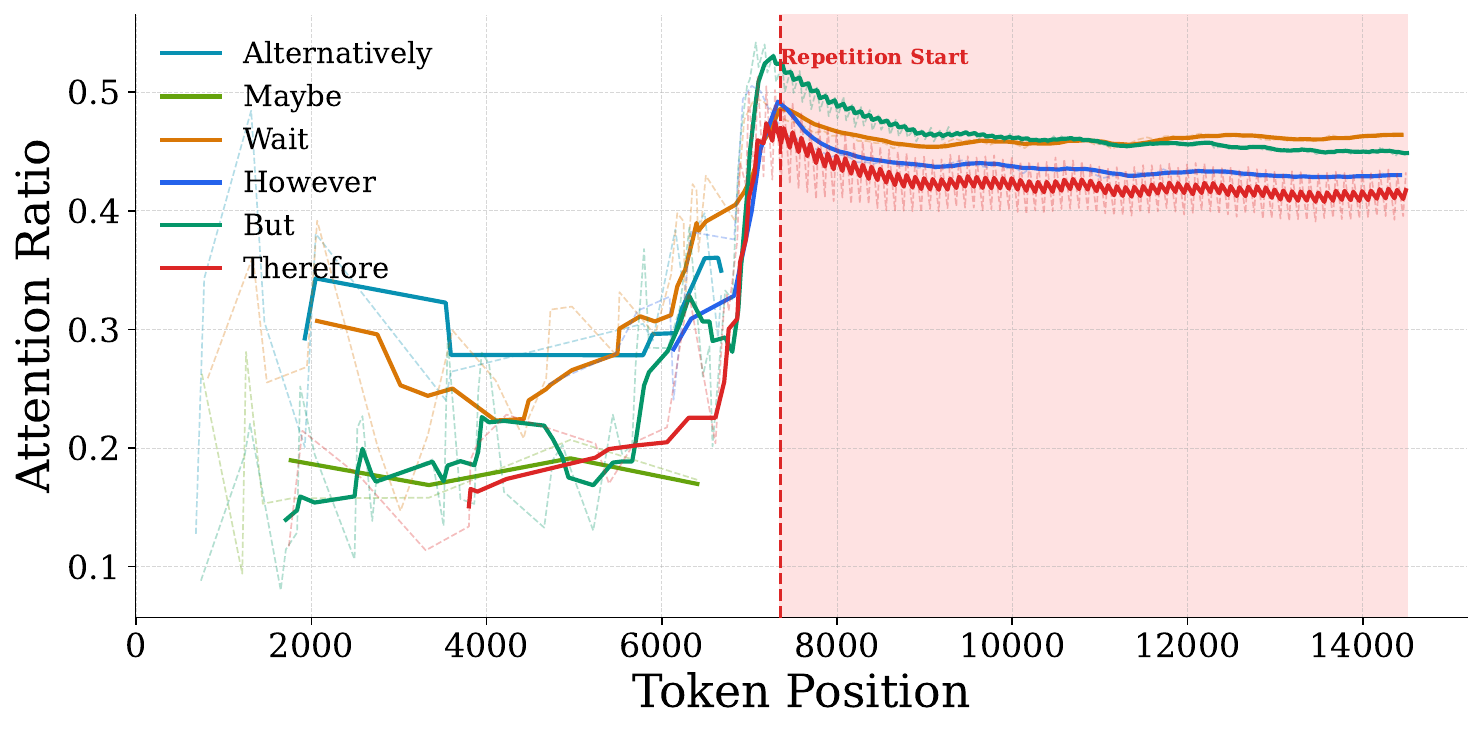}
        \caption{Loop Case 2}
        \label{fig:ds_attn_loop_c2}
    \end{subfigure}
    \hfill
    \begin{subfigure}[b]{0.32\textwidth}
        \centering
        \includegraphics[width=\textwidth]{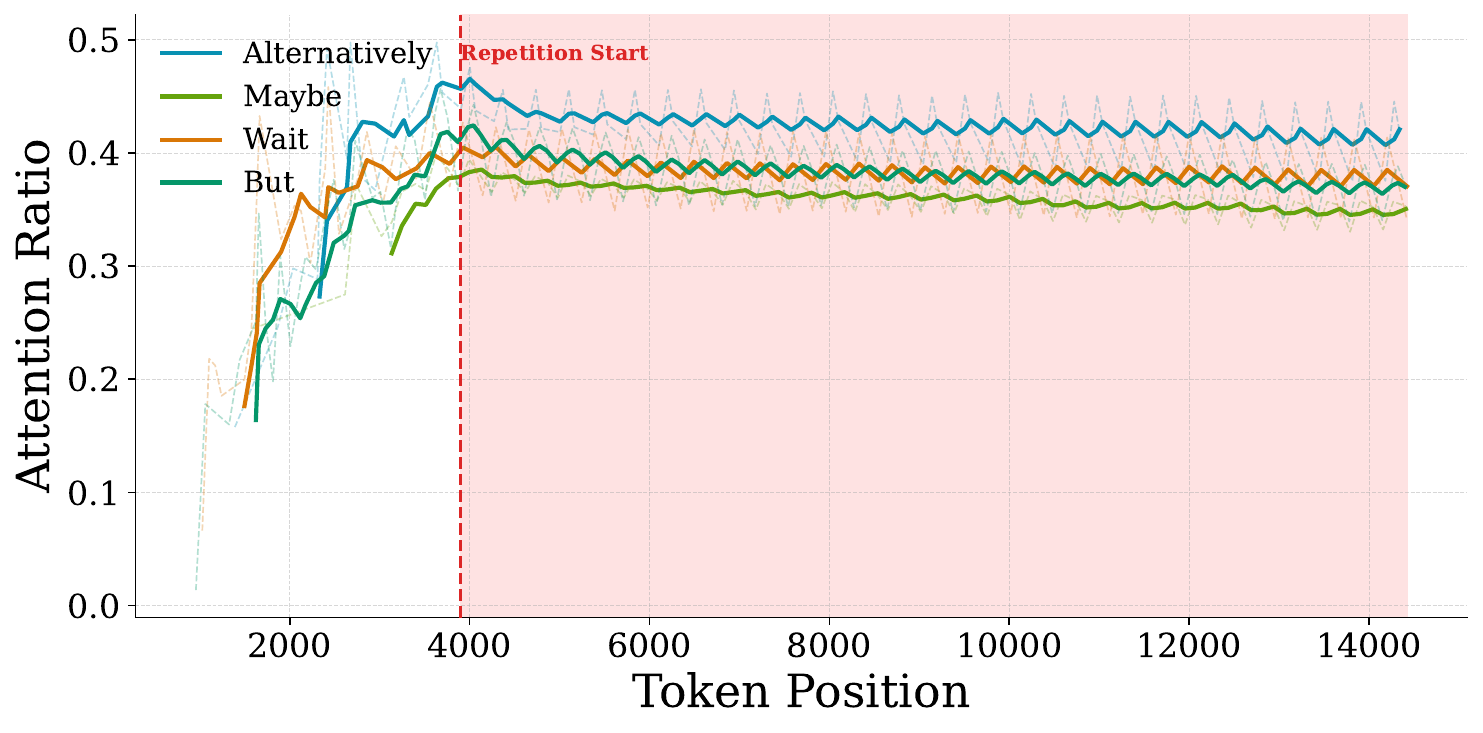}
        \caption{Loop Case 3}
        \label{fig:ds_attn_loop_c3}
    \end{subfigure}
    
    \par\bigskip 
    
    \begin{subfigure}[b]{0.32\textwidth}
        \centering
        \includegraphics[width=\textwidth]{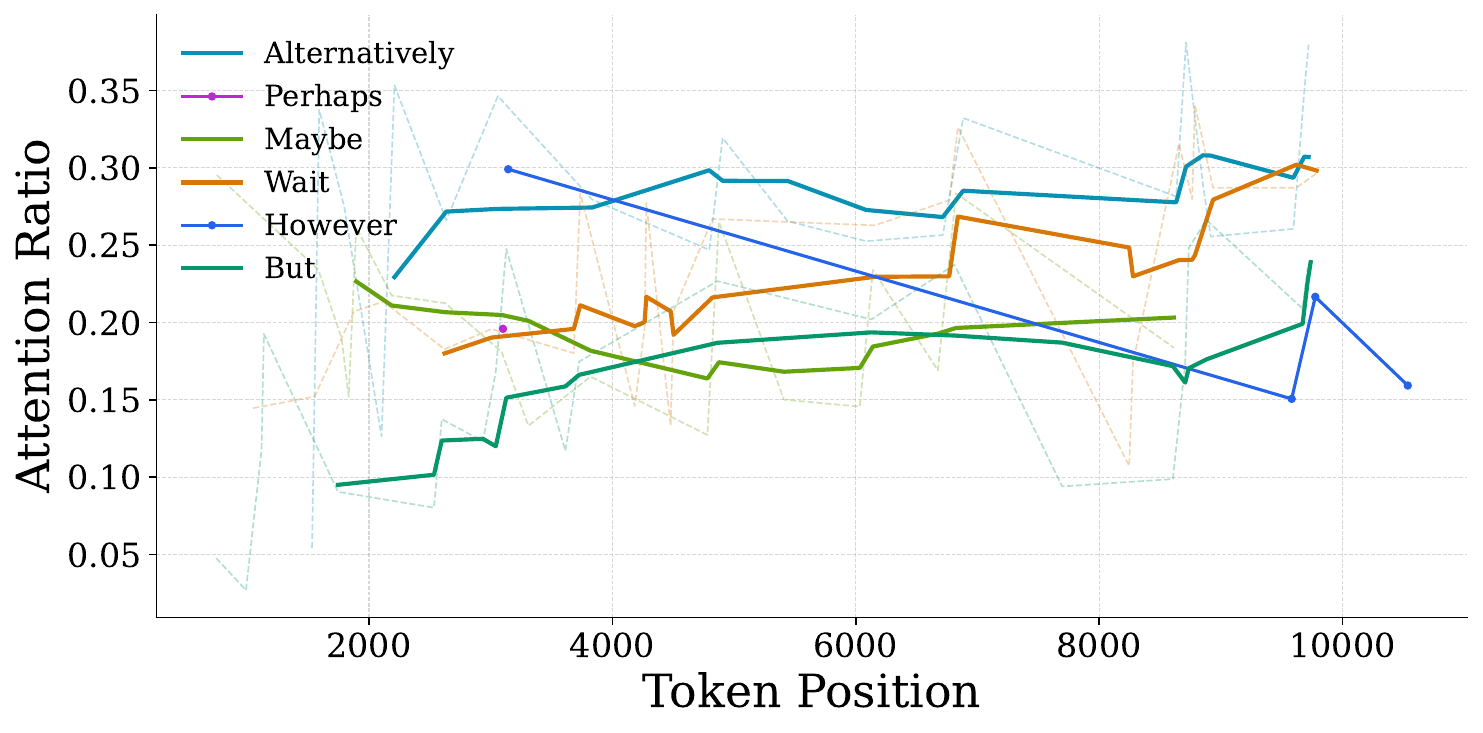}
        \caption{Non-loop Case 1}
        \label{fig:ds_attn_noloop_c1}
    \end{subfigure}
    \hfill
    \begin{subfigure}[b]{0.32\textwidth}
        \centering
        \includegraphics[width=\textwidth]{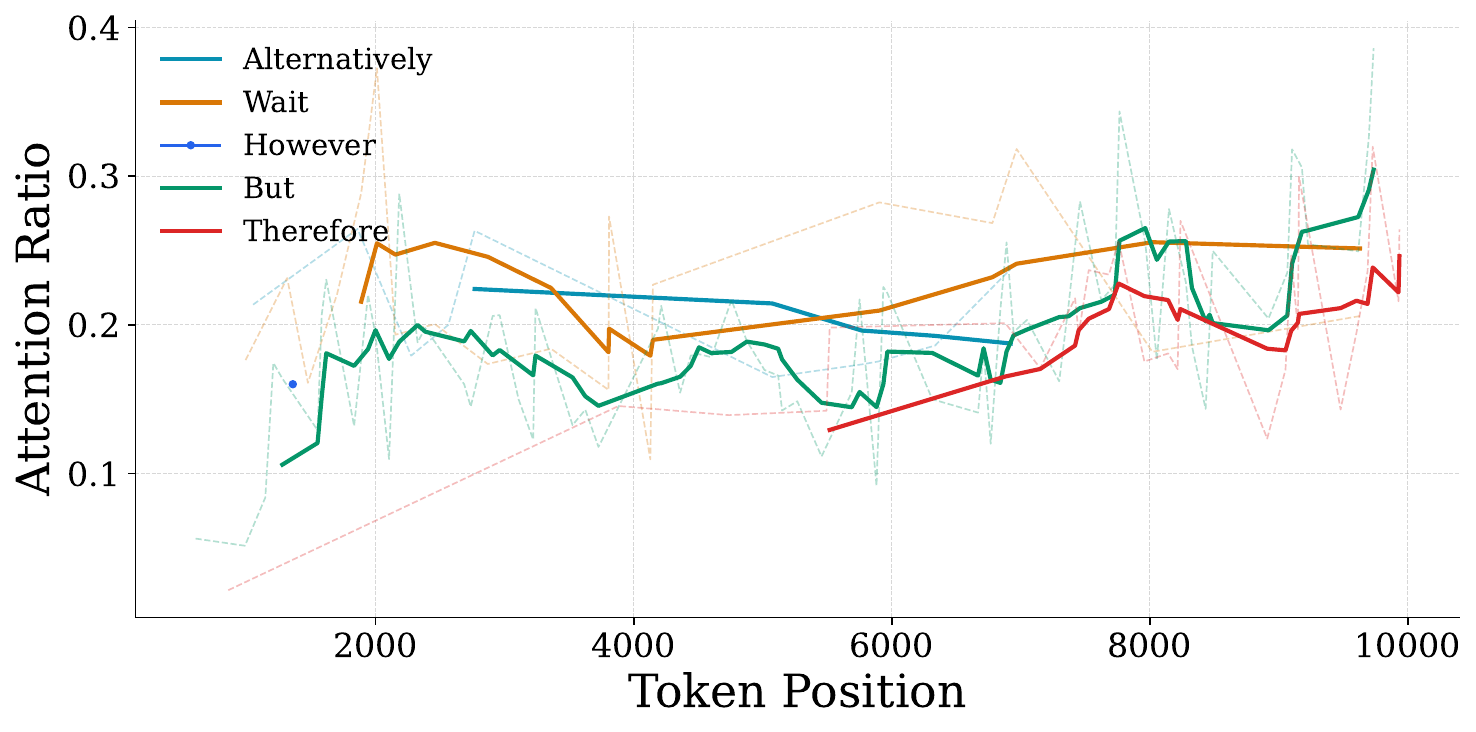}
        \caption{Non-loop Case 2}
        \label{fig:ds_attn_noloop_c2}
    \end{subfigure}
    \hfill
    \begin{subfigure}[b]{0.32\textwidth}
        \centering
        \includegraphics[width=\textwidth]{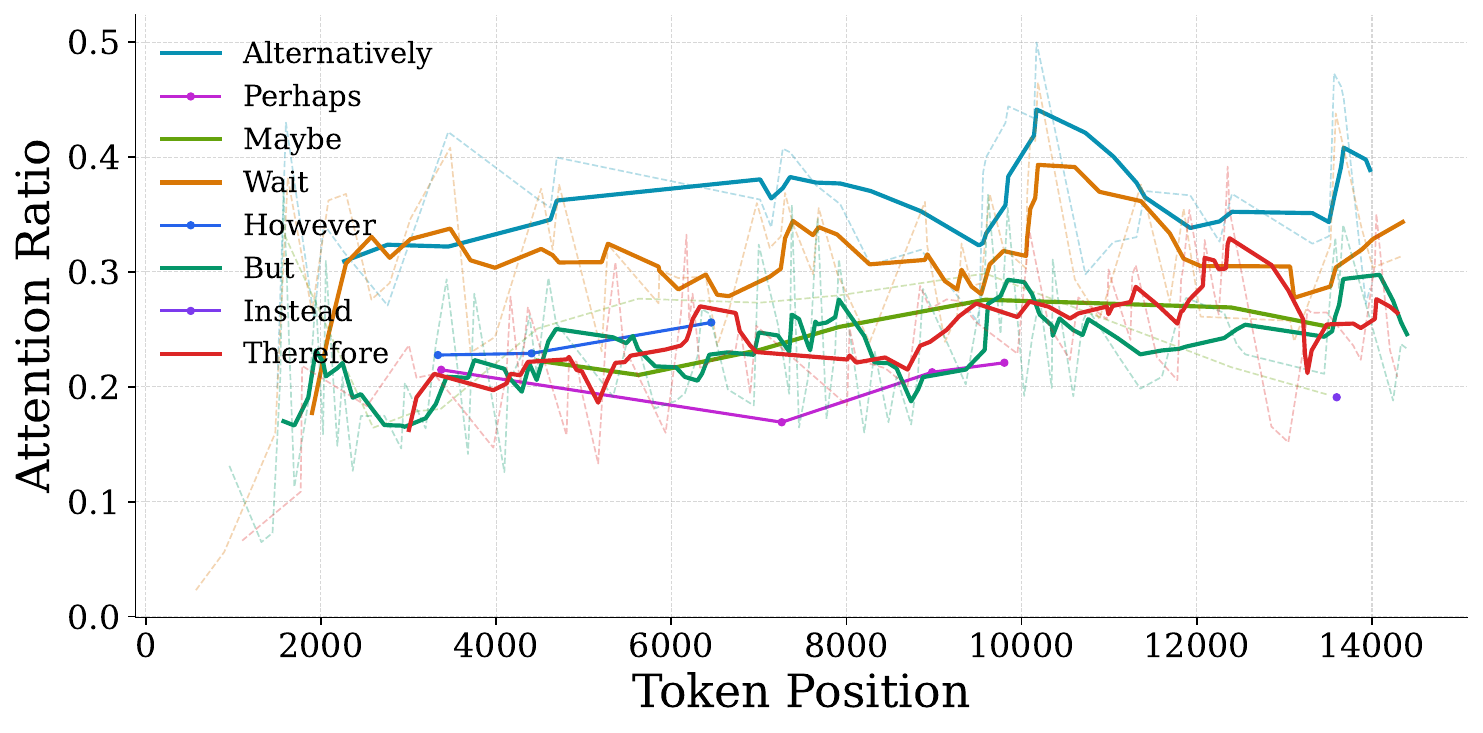}
        \caption{Non-loop Case 3}
        \label{fig:ds_attn_noloop_c3}
    \end{subfigure}
    
    \caption{\textbf{Attention Score Dynamics in DS-Qwen-14B.} Comparison of attention allocation on high-entropy pivot tokens. Top Row (Loop Cases): A distinct surge in attention towards pivot tokens occurs just before the loop begins. Bottom Row (Non-loop Cases): Attention remains evenly distributed without significant peaks.}
    \label{fig:ds_attn_scores}
\end{figure*}

\begin{figure*}[t]
    \centering
    \begin{subfigure}[b]{0.32\textwidth}
        \centering
        \includegraphics[width=\textwidth]{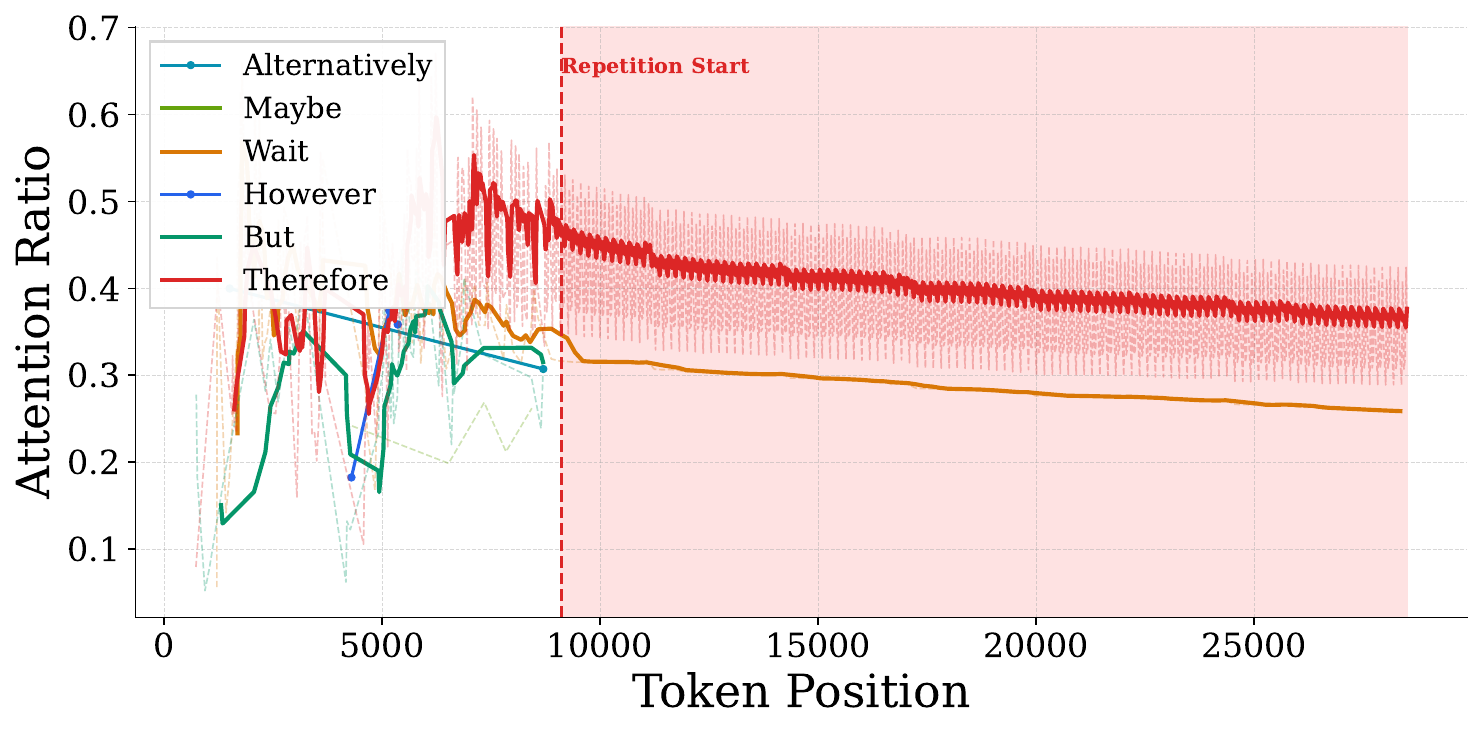}
        \caption{Loop Case 1}
        \label{fig:qwen_attn_loop_c1}
    \end{subfigure}
    \hfill
    \begin{subfigure}[b]{0.32\textwidth}
        \centering
        \includegraphics[width=\textwidth]{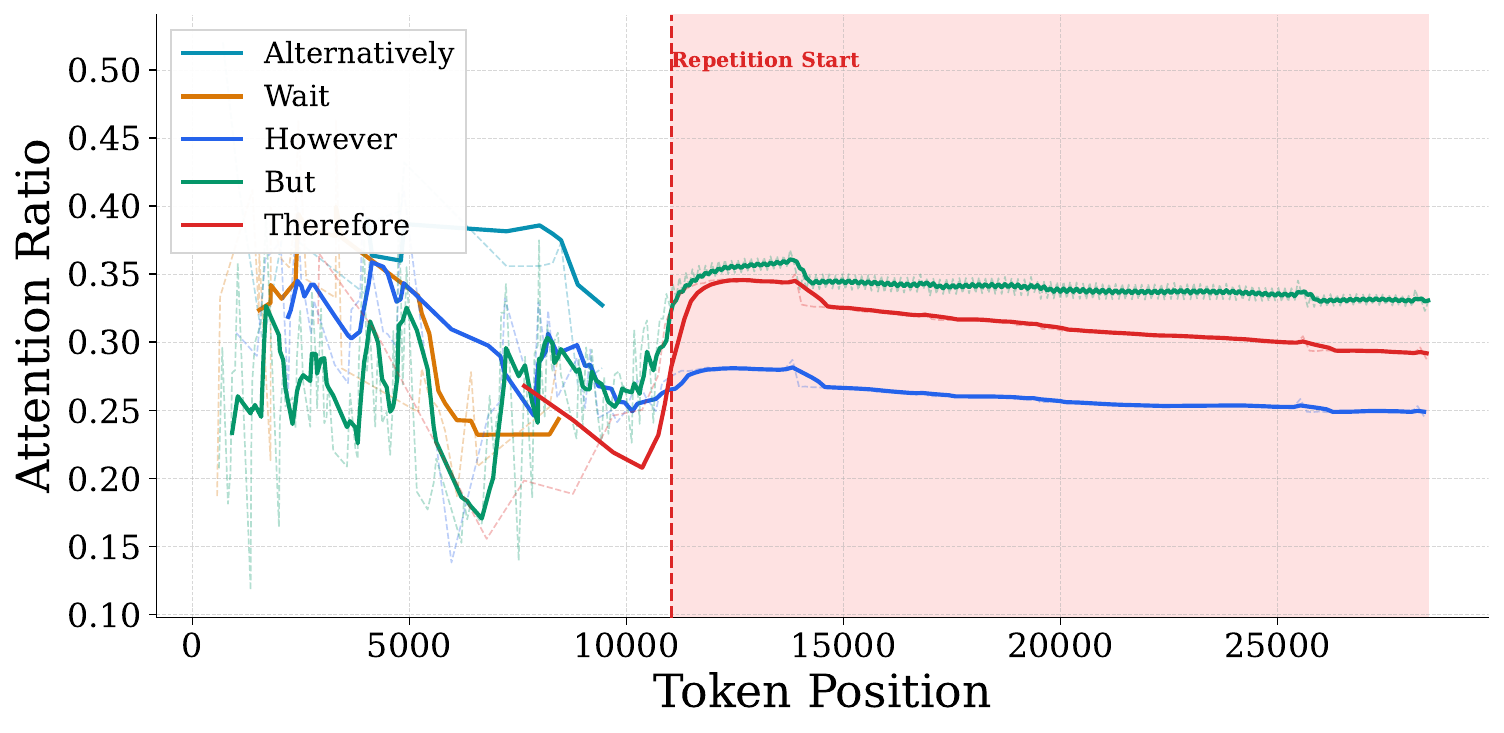}
        \caption{Loop Case 2}
        \label{fig:qwen_attn_loop_c2}
    \end{subfigure}
    \hfill
    \begin{subfigure}[b]{0.32\textwidth}
        \centering
        \includegraphics[width=\textwidth]{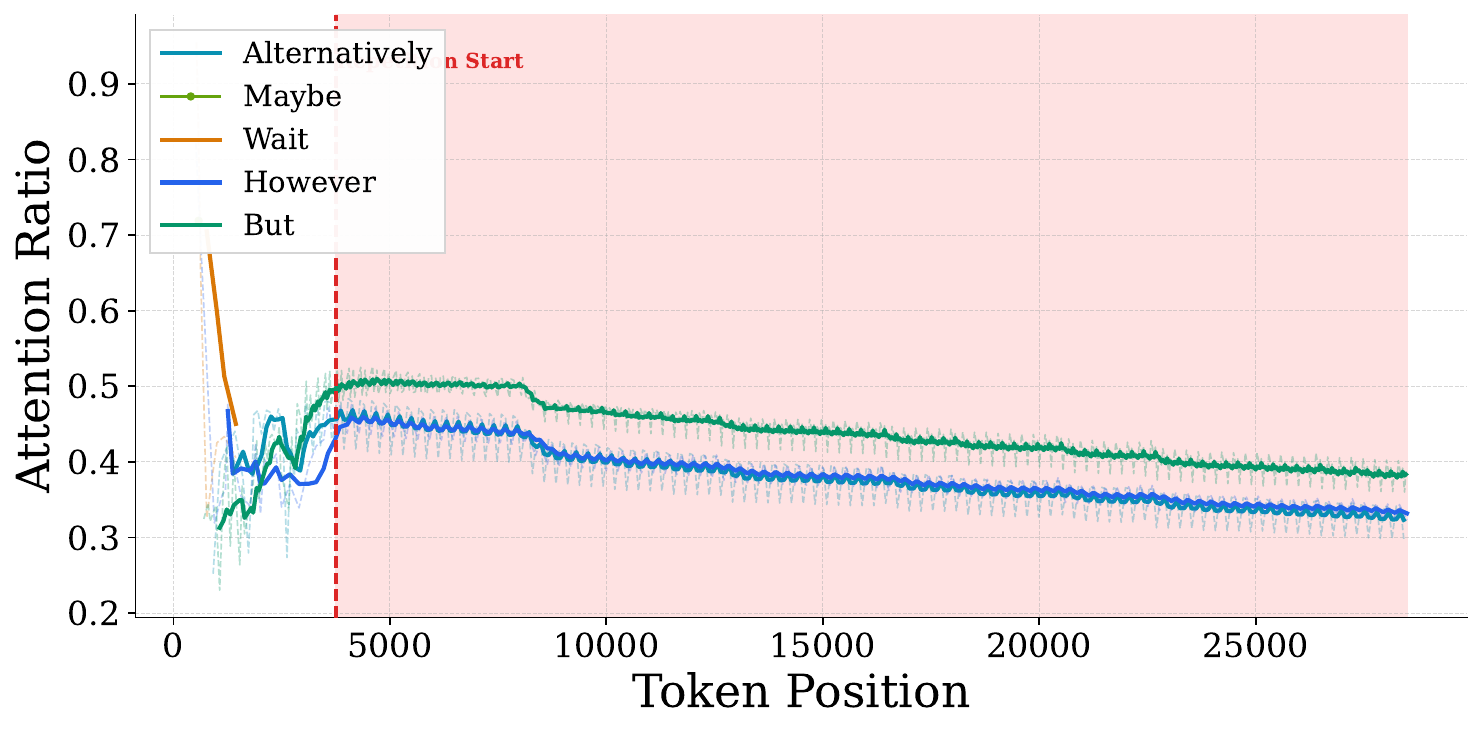}
        \caption{Loop Case 3}
        \label{fig:qwen_attn_loop_c3}
    \end{subfigure}
    
    \par\bigskip 
    
    \begin{subfigure}[b]{0.32\textwidth}
        \centering
        \includegraphics[width=\textwidth]{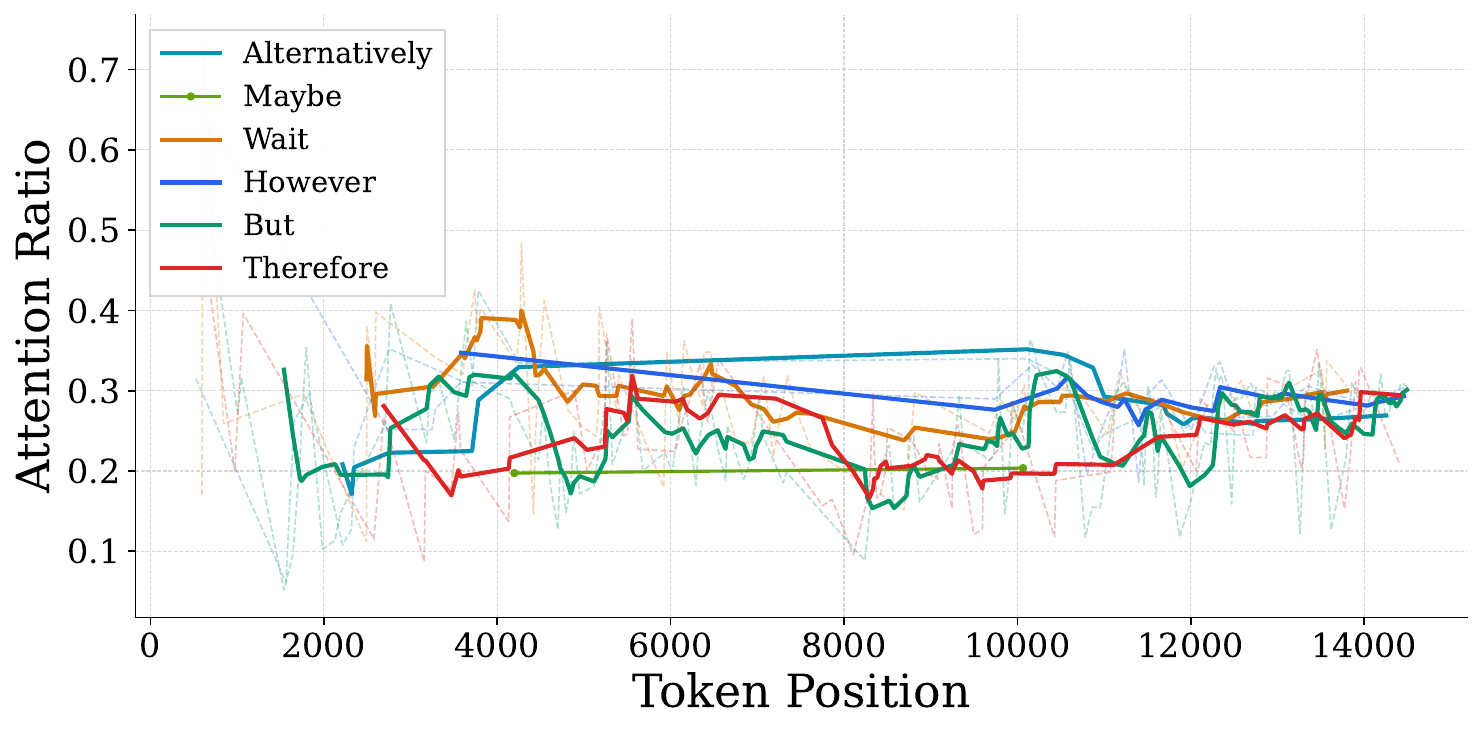}
        \caption{Non-loop Case 1}
        \label{fig:qwen_attn_noloop_c1}
    \end{subfigure}
    \hfill
    \begin{subfigure}[b]{0.32\textwidth}
        \centering
        \includegraphics[width=\textwidth]{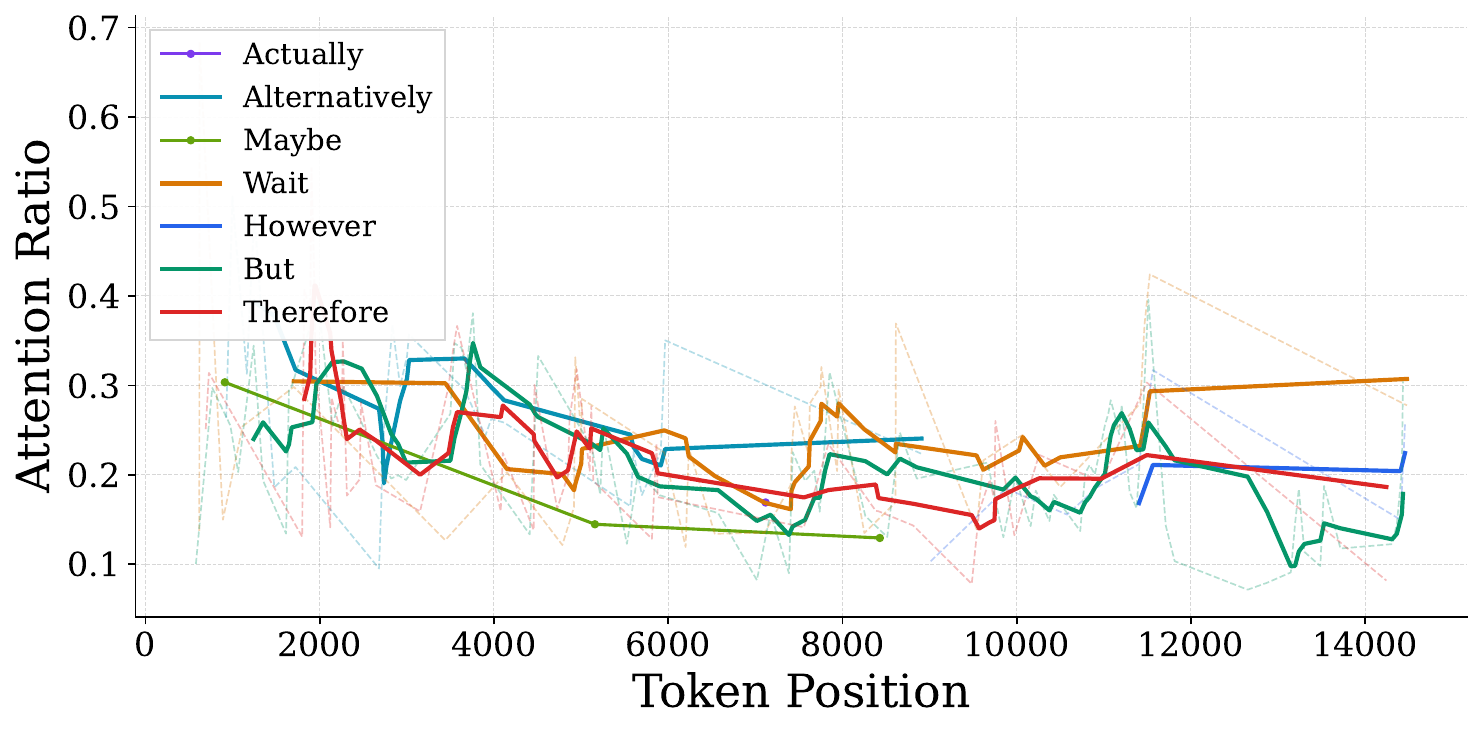}
        \caption{Non-loop Case 2}
        \label{fig:qwen_attn_noloop_c2}
    \end{subfigure}
    \hfill
    \begin{subfigure}[b]{0.32\textwidth}
        \centering
        \includegraphics[width=\textwidth]{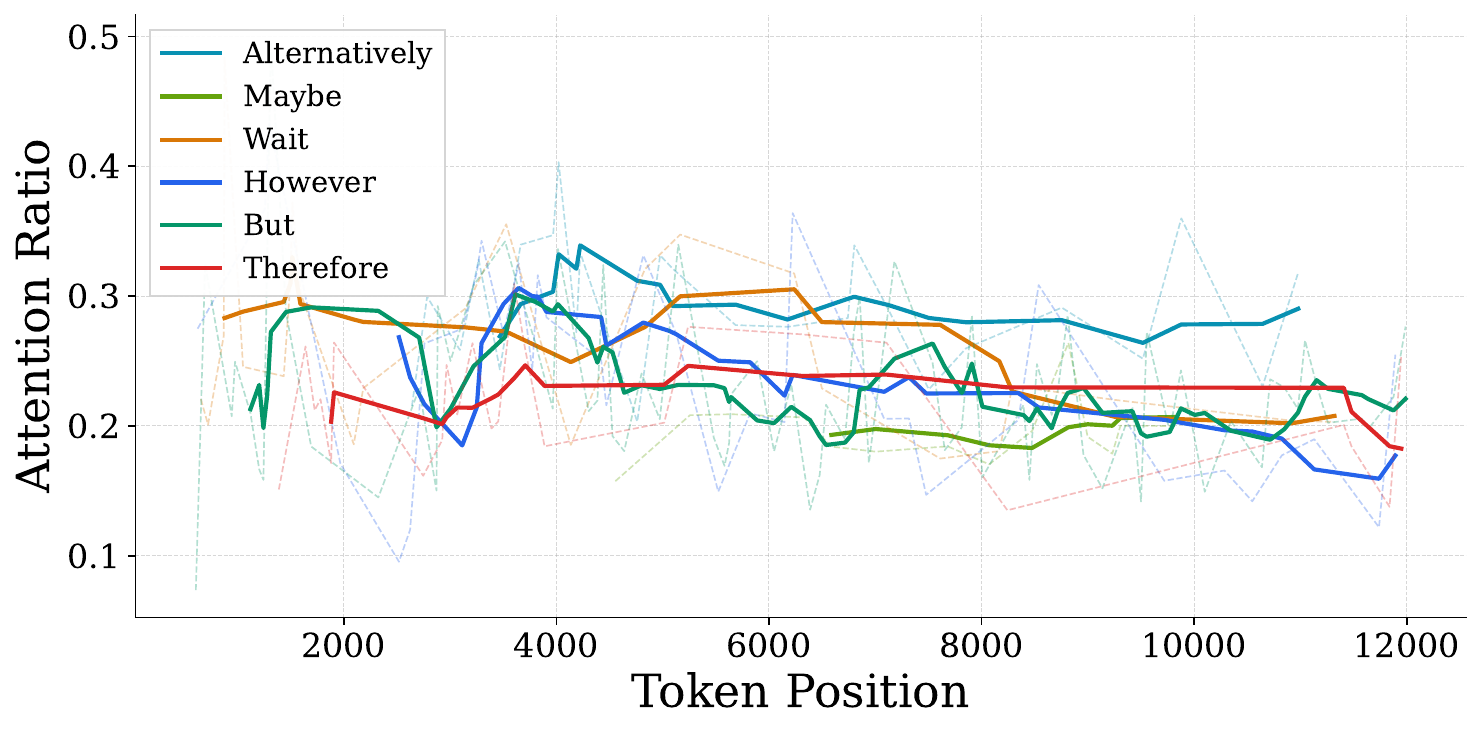}
        \caption{Non-loop Case 3}
        \label{fig:qwen_attn_noloop_c3}
    \end{subfigure}
    
    \caption{\textbf{Attention Score Dynamics in Qwen3-8B.} Similar to DS-Qwen-14B, Qwen3-8B exhibits a pronounced attention shift towards high-entropy tokens in loop scenarios (Top), contrasting with the uniform distribution in normal reasoning (Bottom).}
    \label{fig:qwen_attn_scores}
\end{figure*}

To further investigate the mechanistic trigger of statement loops, we analyzed the attention scores allocated to the five most representative high-entropy tokens (e.g., \textit{``Therefore''}, \textit{``Alternatively''}).
We compare the attention distribution in loop scenarios against non-loop baselines. 
Figures~\ref{fig:ds_attn_scores} and \ref{fig:qwen_attn_scores} visualize these dynamics for DS-Qwen-14B and Qwen3-8B, respectively. 
In the \textbf{loop cases} (top rows), we observe a significant attention spike targeting these high-entropy pivots immediately preceding the repetition onset (marked by the dashed vertical line). This suggests that the model's decision-making process is disproportionately influenced by these uncertainty-inducing tokens at the critical transition point. Conversely, in the non-loop baselines (bottom rows), the attention distribution remains relatively uniform, with no comparable concentration on specific pivot tokens.

We also examined the distribution of high-entropy tokens across three distinct scenarios: (1) loop instances in \textbf{DS-Qwen-14B}, (2) non-loop instances in \textbf{DS-Qwen-14B}, and (3) responses from the \textbf{Qwen2.5-14B-Instruct} baseline on a curated set of 50 samples. 
As illustrated in Figure~\ref{fig:entropy_distribution_box}, reasoning loops are characterized by a statistically significant abundance of high-entropy tokens compared to both non-loop reasoning paths and standard instruction-following outputs. This disparity suggests that the accumulation of uncertainty-inducing tokens is a specific pathology of the reasoning process in LRMs, rather than a general feature of the underlying language model.

%% file: latex/sections/X3_method.tex

\section{Experimental Setup for Prediction} \label{sec:setup_pre}

\begin{figure*}[t]
    \centering
    \begin{subfigure}[b]{0.32\textwidth}
        \centering
        \includegraphics[width=\textwidth]{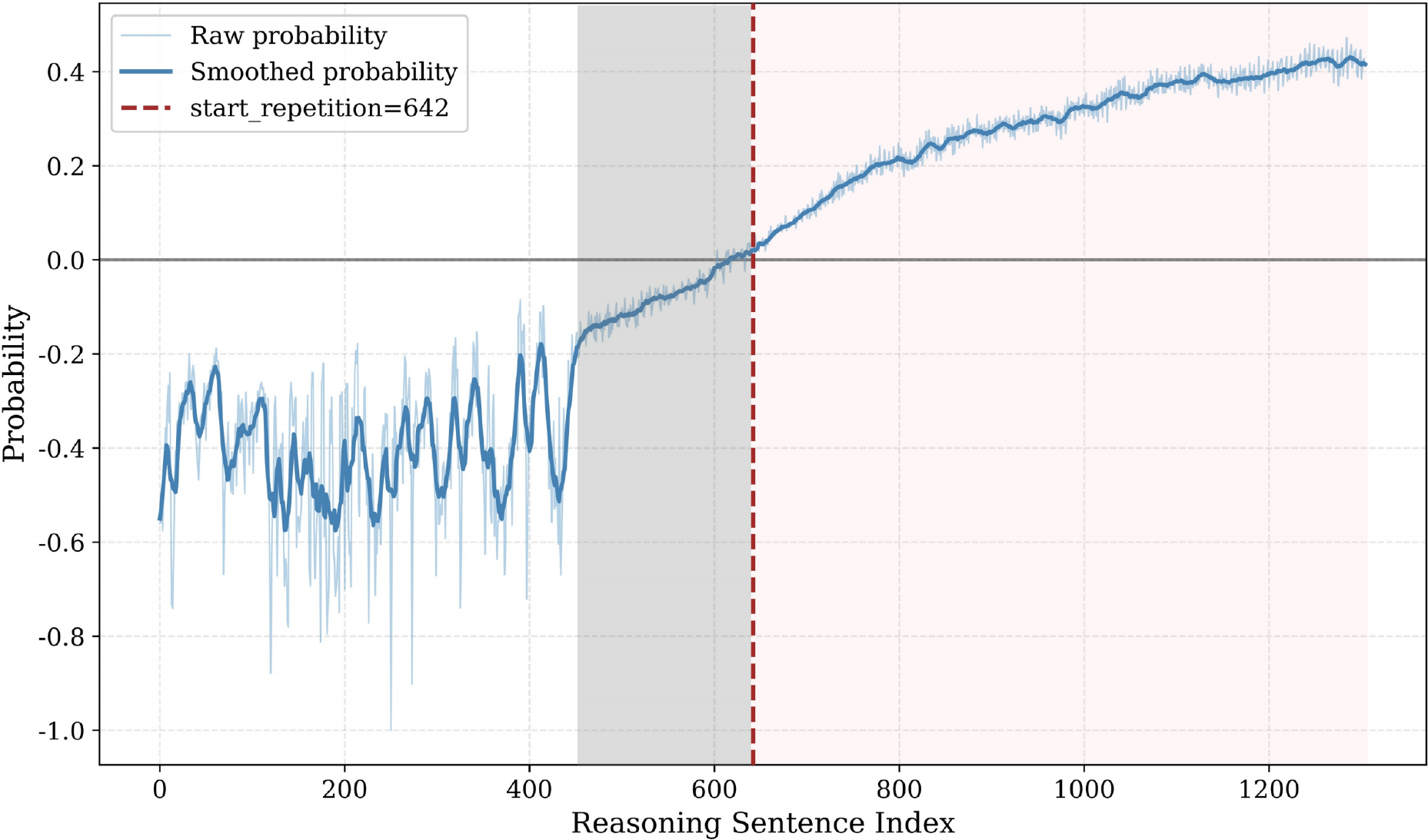}
        \caption{Loop Case 1}
        \label{fig:detect_ds_loop_c1}
    \end{subfigure}
    \hfill
    \begin{subfigure}[b]{0.32\textwidth}
        \centering
        \includegraphics[width=\textwidth]{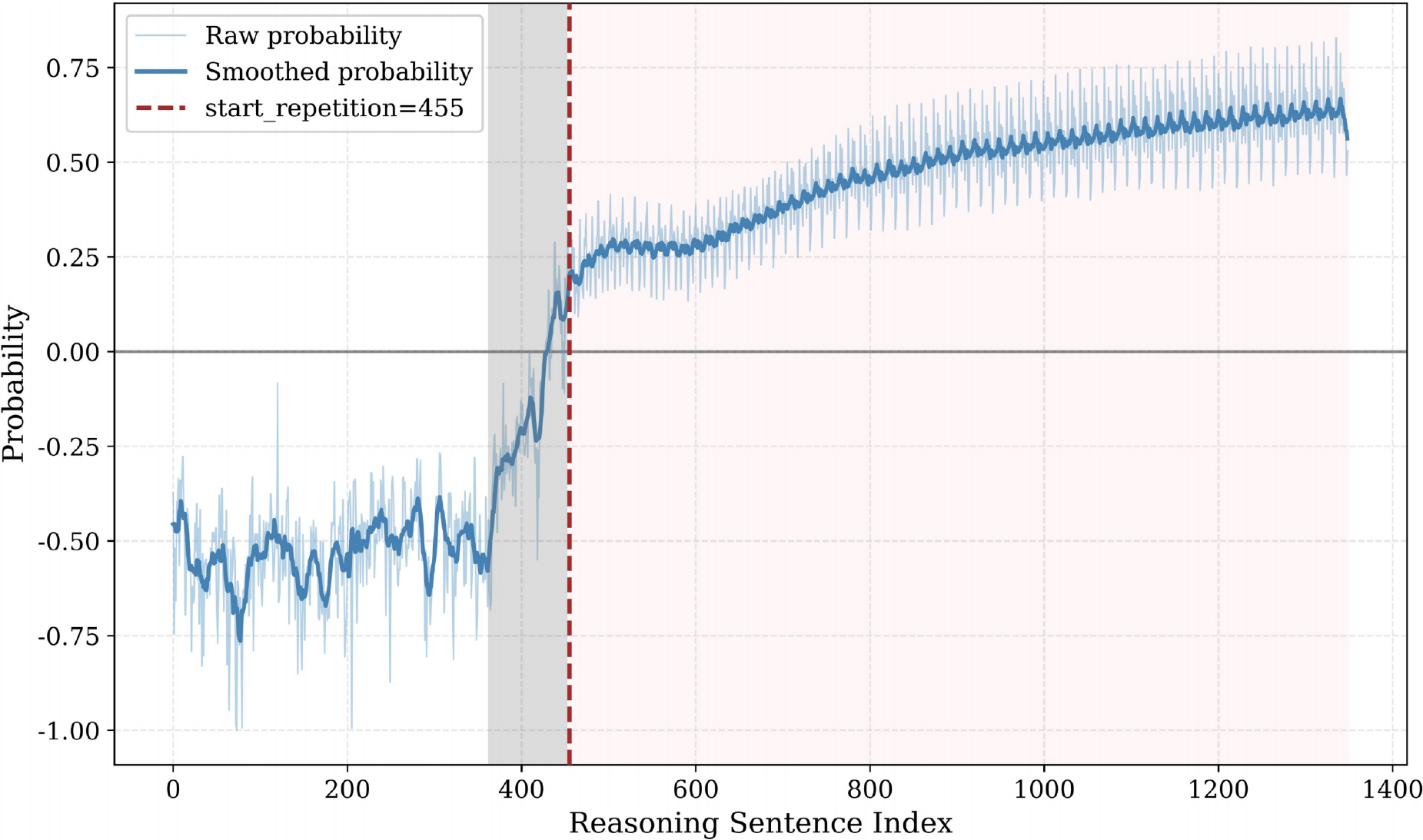}
        \caption{Loop Case 2}
        \label{fig:detect_ds_loop_c2}
    \end{subfigure}
    \hfill
    \begin{subfigure}[b]{0.32\textwidth}
        \centering
        \includegraphics[width=\textwidth]{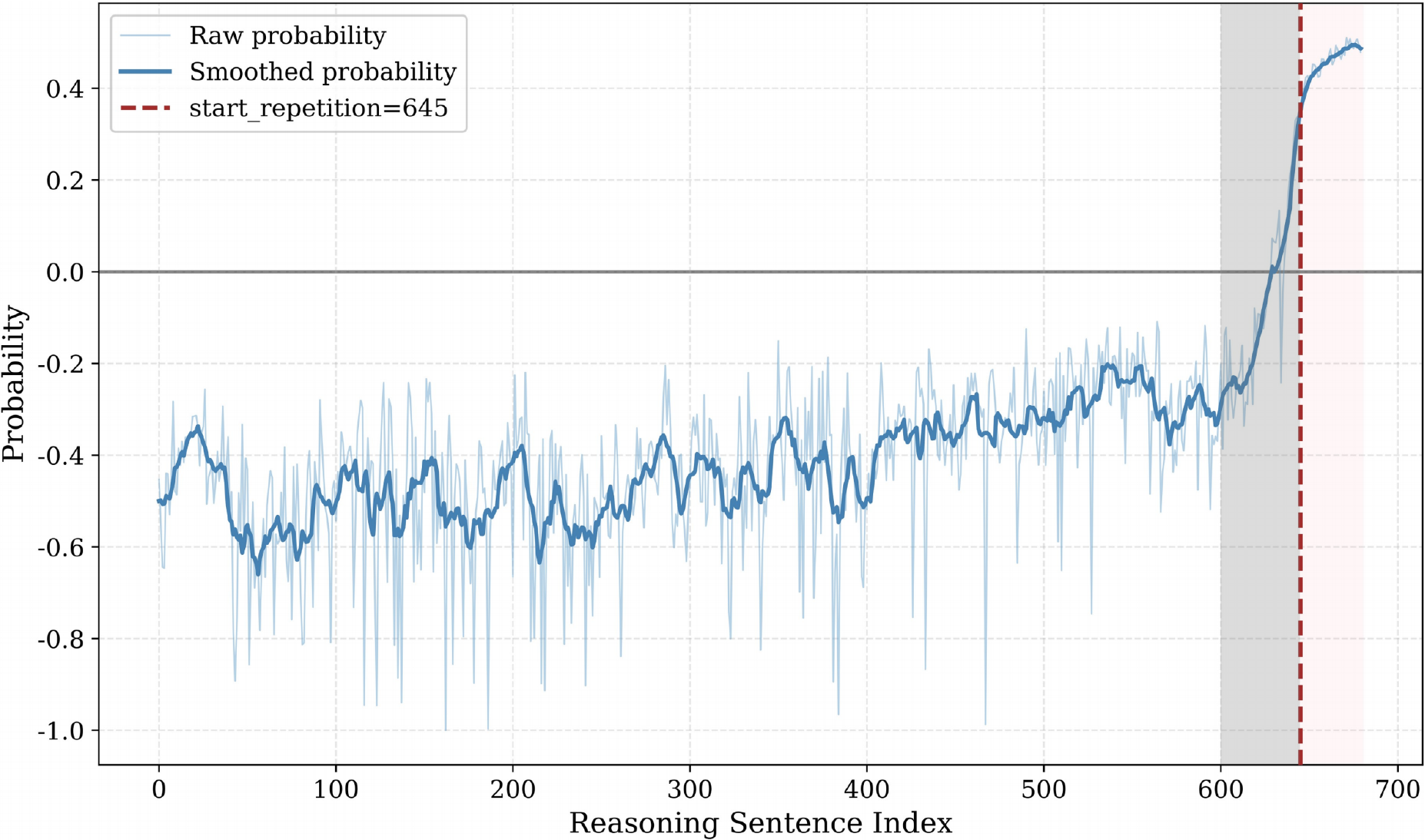}
        \caption{Loop Case 3}
        \label{fig:detect_ds_loop_c3}
    \end{subfigure}
    
    \par\bigskip 
    
    \begin{subfigure}[b]{0.32\textwidth}
        \centering
        \includegraphics[width=\textwidth]{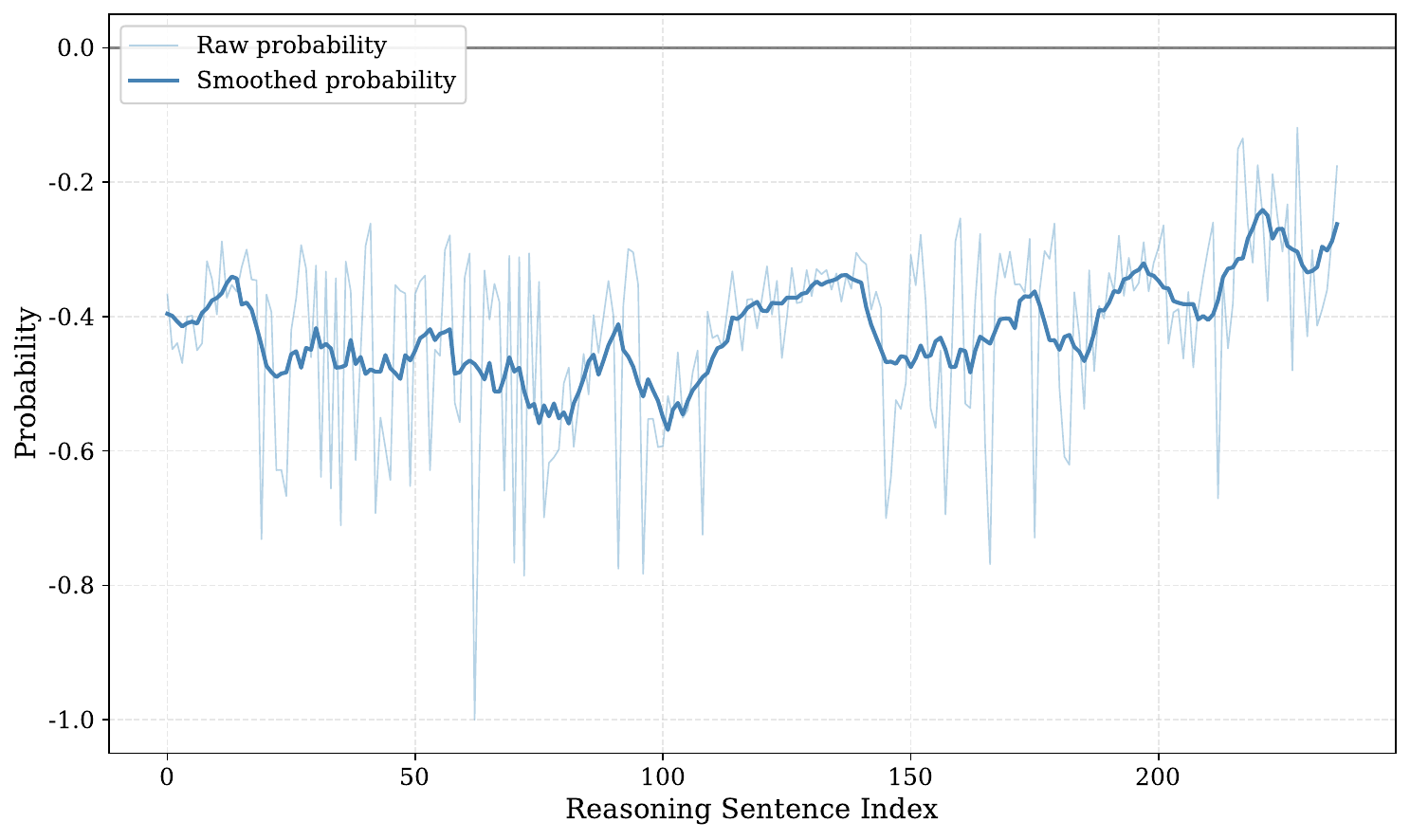}
        \caption{Non-loop Case 1}
        \label{fig:detect_ds_noloop_c1}
    \end{subfigure}
    \hfill
    \begin{subfigure}[b]{0.32\textwidth}
        \centering
        \includegraphics[width=\textwidth]{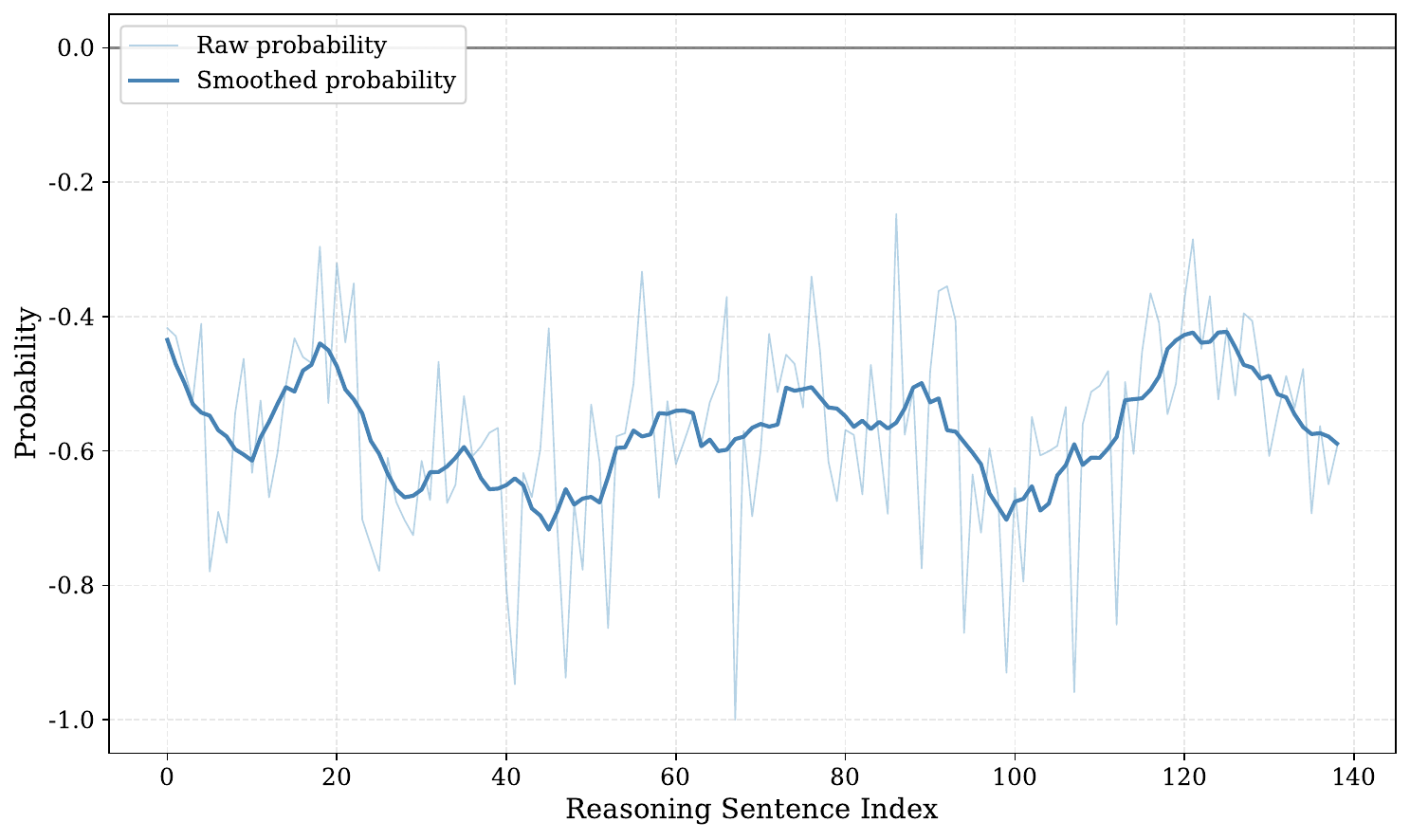}
        \caption{Non-loop Case 2}
        \label{fig:detect_ds_noloop_c2}
    \end{subfigure}
    \hfill
    \begin{subfigure}[b]{0.32\textwidth}
        \centering
        \includegraphics[width=\textwidth]{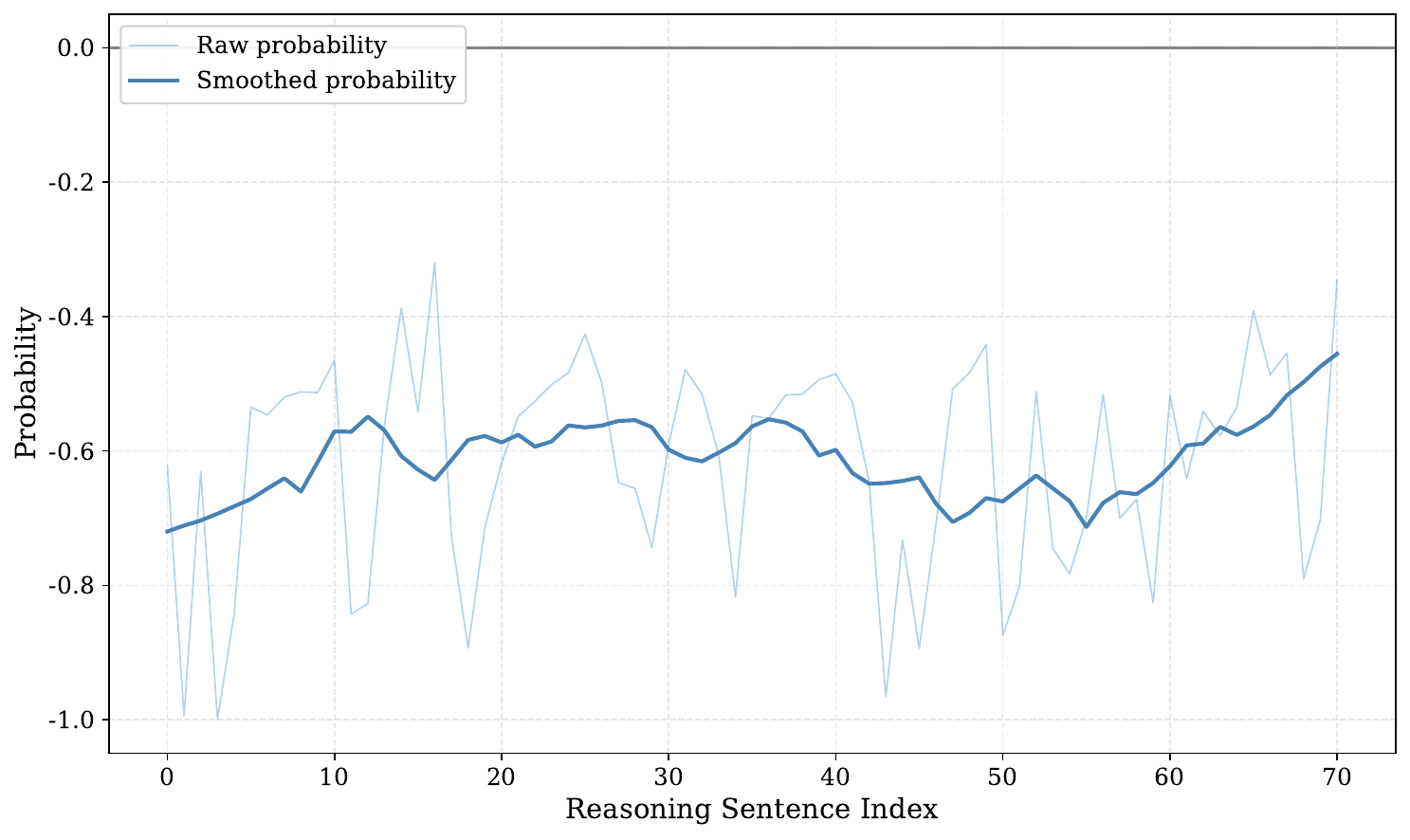}
        \caption{Non-loop Case 3}
        \label{fig:detect_ds_noloop_c3}
    \end{subfigure}
    
    \caption{\textbf{Trajectories of Detection Statistics in DS-Qwen-14B.} \textbf{Top Row:} In loop cases, the detection score accumulates rapidly, signaling a potential repetition event. \textbf{Bottom Row:} In normal reasoning, the score remains stable and low, confirming a low false positive rate.}
    \label{fig:ds_detect_dynamics}
\end{figure*}

\begin{figure*}[t]
    \centering
    \begin{subfigure}[b]{0.32\textwidth}
        \centering
        \includegraphics[width=\textwidth]{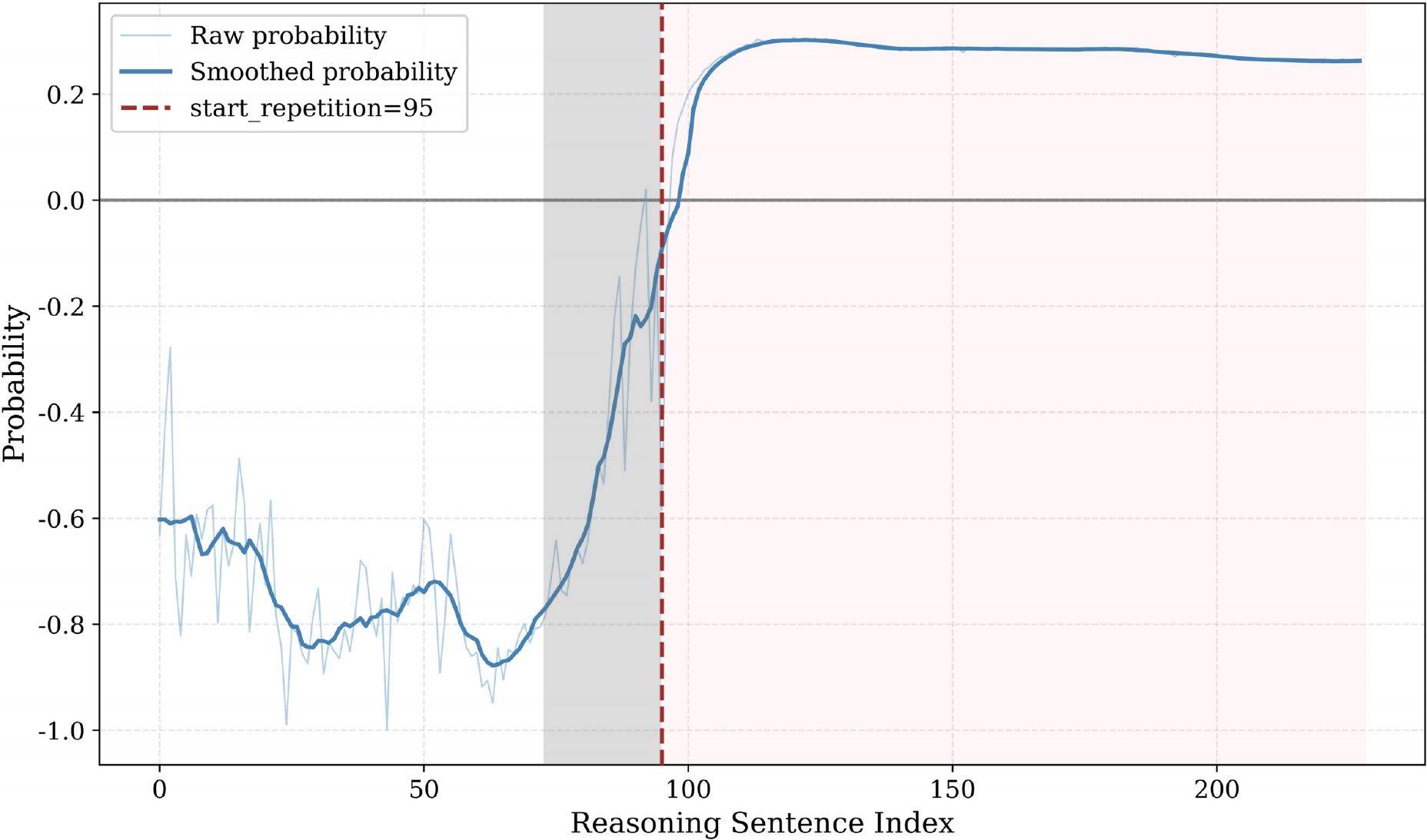} 
        \caption{Loop Case 1}
        \label{fig:detect_qwen_loop_c1}
    \end{subfigure}
    \hfill
    \begin{subfigure}[b]{0.32\textwidth}
        \centering
        \includegraphics[width=\textwidth]{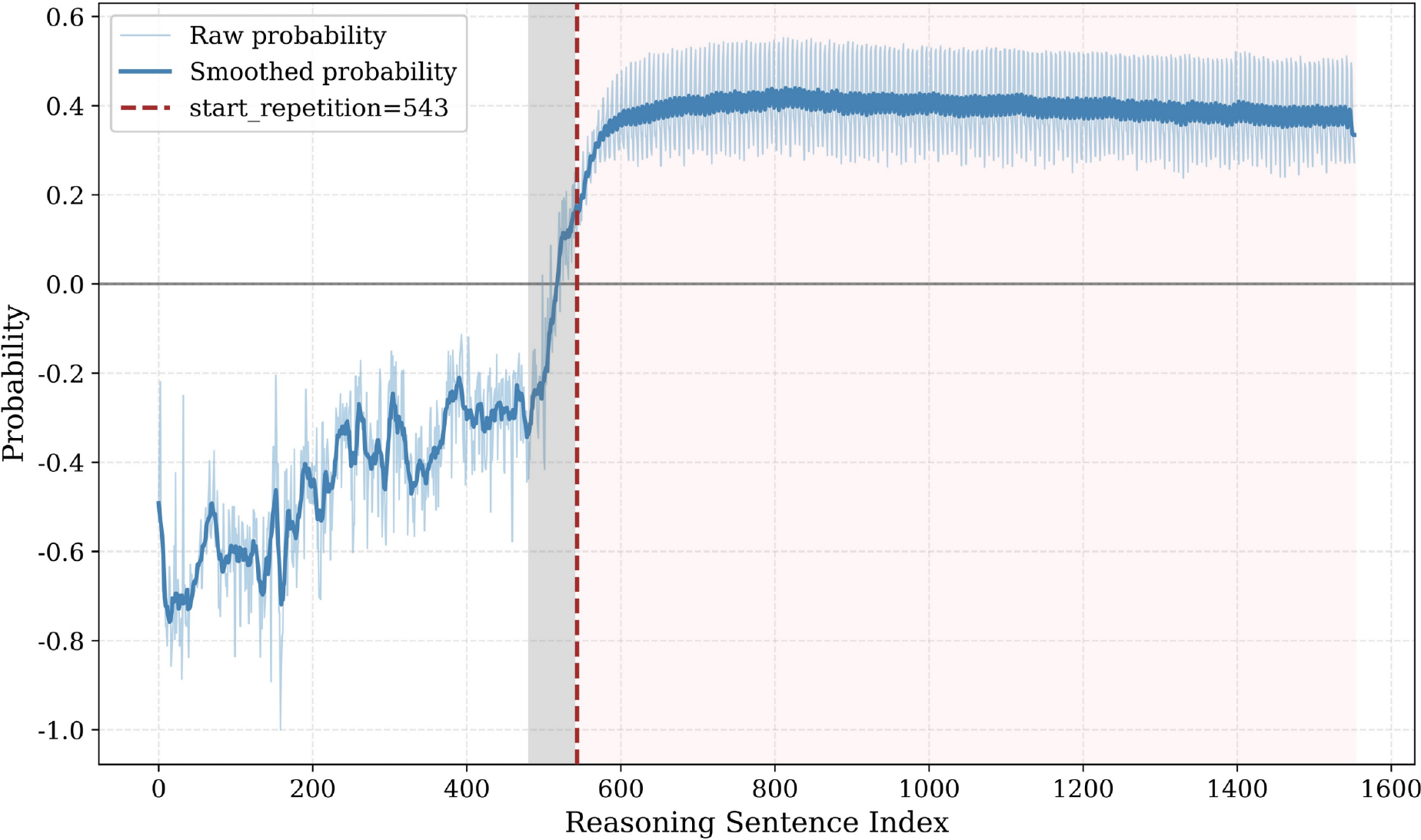} 
        \caption{Loop Case 2}
        \label{fig:detect_qwen_loop_c2}
    \end{subfigure}
    \hfill
    \begin{subfigure}[b]{0.32\textwidth}
        \centering
        \includegraphics[width=\textwidth]{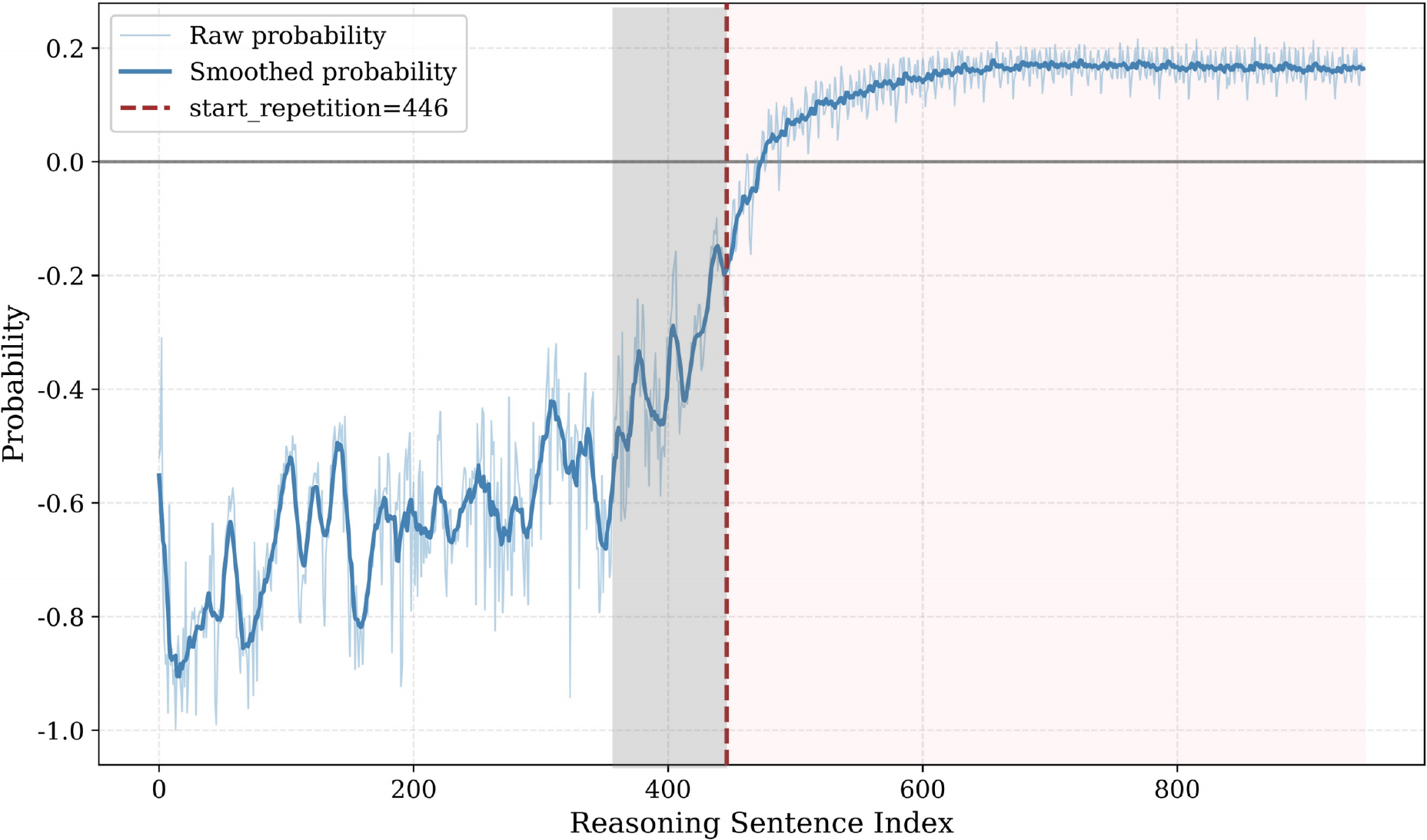} 
        \caption{Loop Case 3}
        \label{fig:detect_qwen_loop_c3}
    \end{subfigure}
    
    \par\bigskip
    
    \begin{subfigure}[b]{0.32\textwidth}
        \centering
        \includegraphics[width=\textwidth]{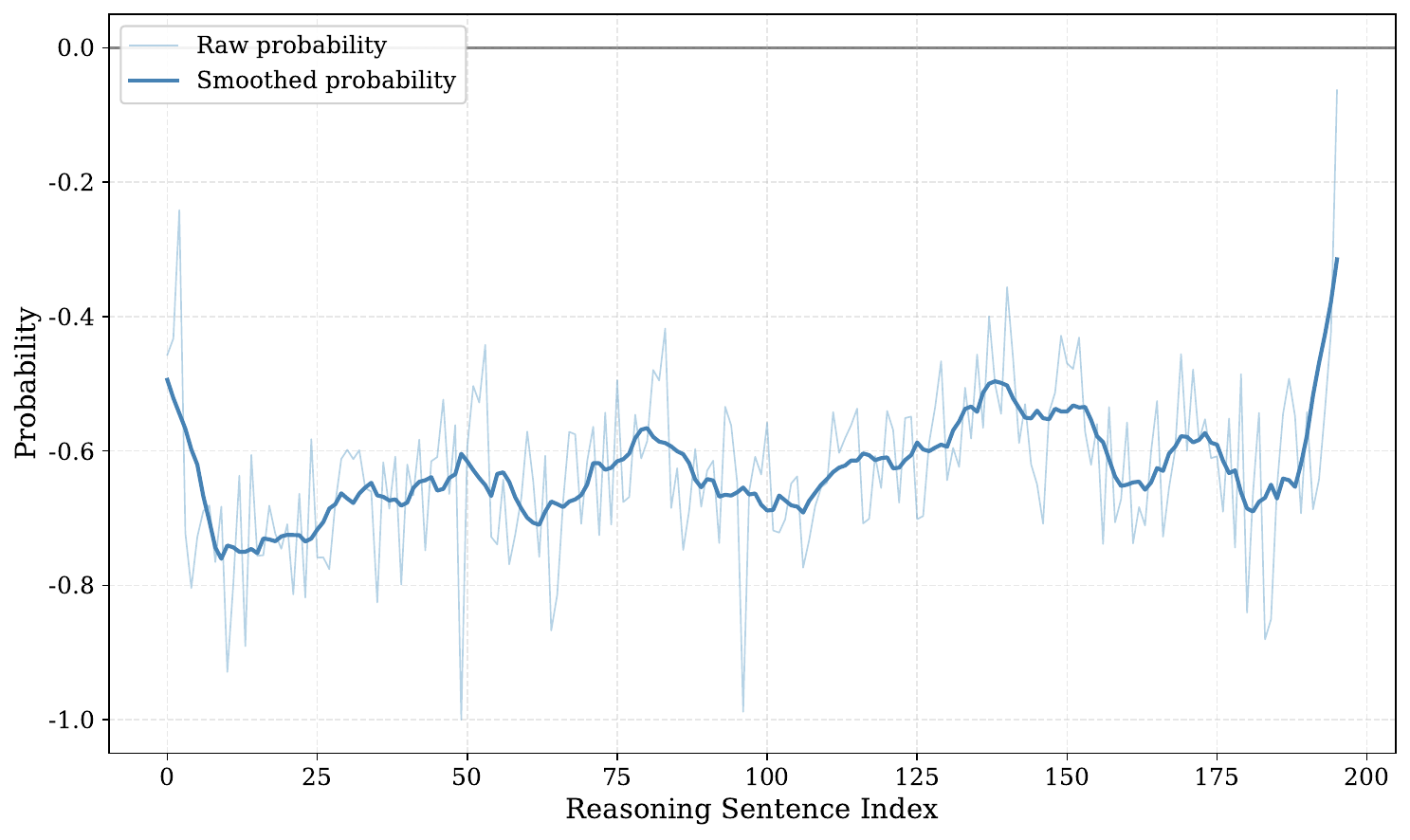} 
        \caption{Non-loop Case 1}
        \label{fig:detect_qwen_noloop_c1}
    \end{subfigure}
    \hfill
    \begin{subfigure}[b]{0.32\textwidth}
        \centering
        \includegraphics[width=\textwidth]{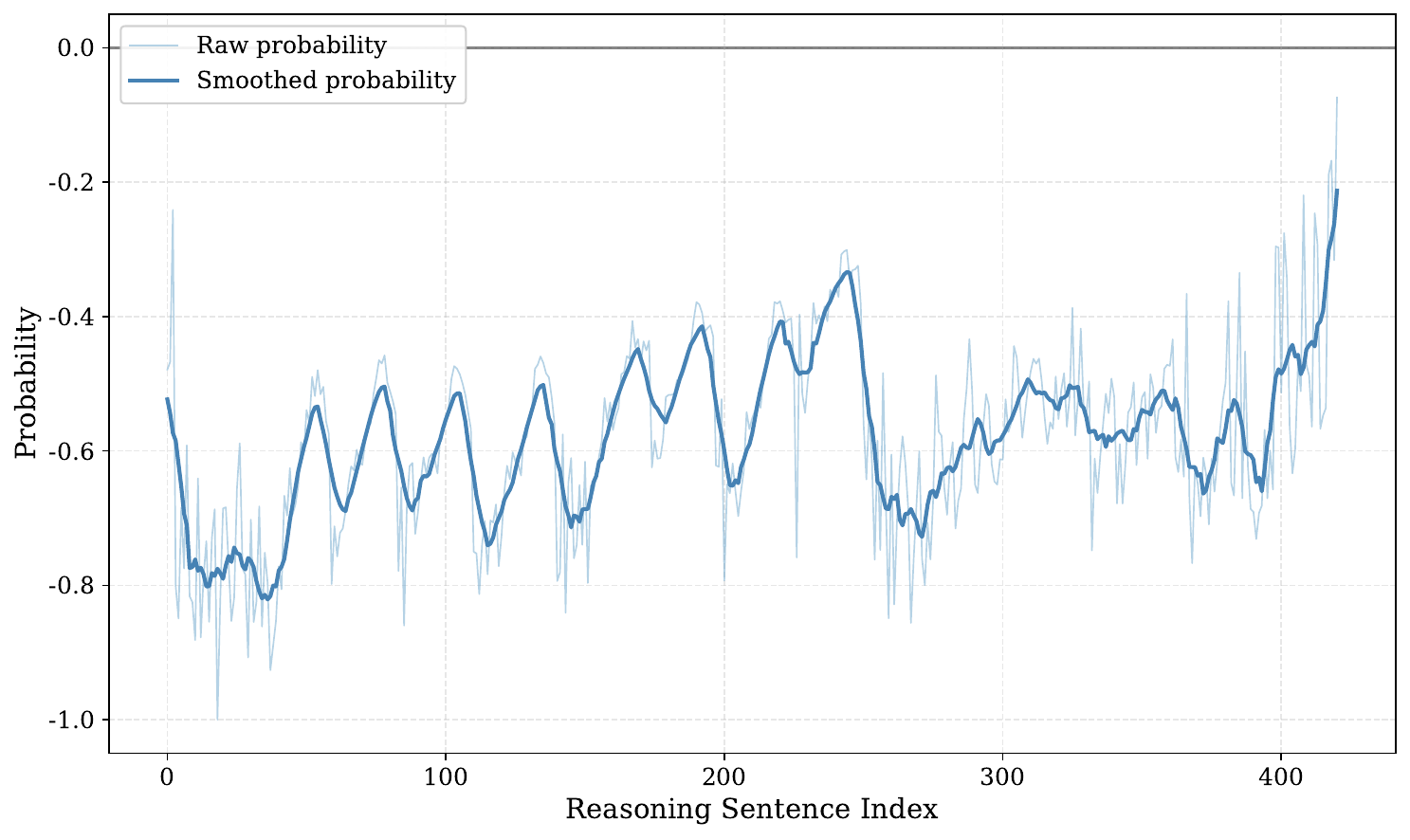} 
        \caption{Non-loop Case 2}
        \label{fig:detect_qwen_noloop_c2}
    \end{subfigure}
    \hfill
    \begin{subfigure}[b]{0.32\textwidth}
        \centering
        \includegraphics[width=\textwidth]{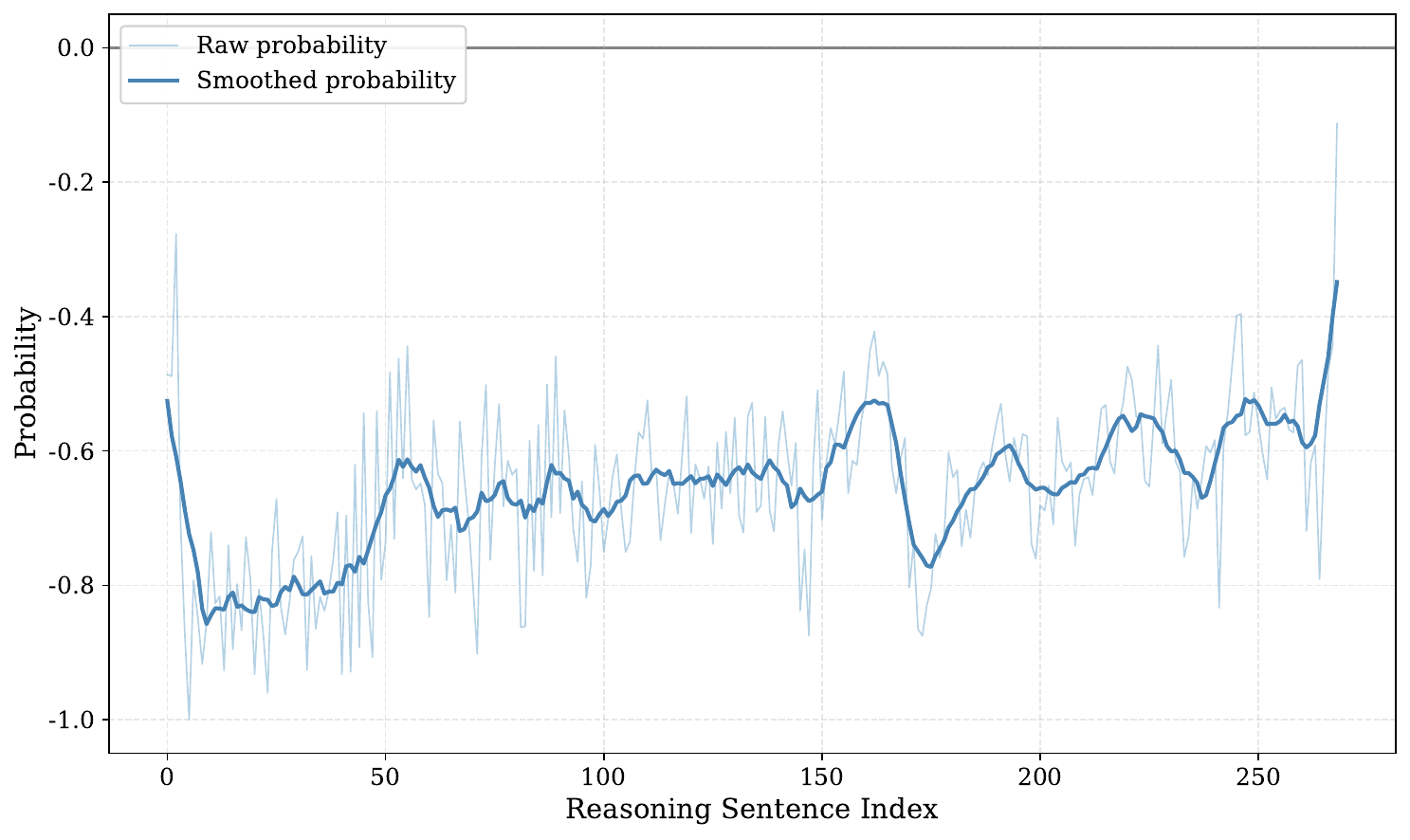} 
        \caption{Non-loop Case 3}
        \label{fig:detect_qwen_noloop_c3}
    \end{subfigure}
    
    \caption{\textbf{Trajectories of Detection Statistics in Qwen3-8B.} Consistent with DS-Qwen, Qwen3-8B shows distinct accumulation patterns for loop cases (Top) versus stable baselines for non-loop cases (Bottom).}
    \label{fig:qwen_detect_dynamics}
\end{figure*}

In this section, we first provide additional visual evidence validating the robustness of our detection mechanism across both DS-Qwen-14B and Qwen3-8B models. Figure~\ref{fig:ds_detect_dynamics}  for DS-Qwen-14B (and Figure~\ref{fig:qwen_detect_dynamics} for Qwen3-8B) illustrates the trajectory of the detection statistic over time. In the loop scenarios, the detection metric exhibits a rapid, monotonic accumulation, quickly breaching the decision threshold. In contrast, for non-loop baselines, the metric fluctuates within a low-amplitude range, demonstrating the method's resistance to false alarms in normal reasoning contexts.

\subsection{Evaluation Metrics}
To comprehensively assess the effectiveness and timeliness of loop prediction, we employ four quantitative metrics: Early Detection Rate (EDR), False Positive Rate (FPR), Average Sentence Earliness (ASE), and Average Token Earliness (ATE). These metrics balance the trade-off between detection sensitivity and the risk of false alarms.

\paragraph{Early Detection Rate (EDR).} EDR quantifies the proportion of loops detected prior to their actual onset. Let $t_{alert}$ and $t_{onset}$ denote the detection step and the ground-truth start of repetition, respectively. A detection is deemed successful if $t_{alert} < t_{onset}$. EDR is defined as:
\begin{equation}
    \text{EDR} = \frac{N_{\text{early}}}{N_{\text{loop}}}
\end{equation}
where $N_{\text{early}}$ is the number of successfully anticipated loops and $N_{\text{loop}}$ is the total count of looping samples.

\paragraph{False Positive Rate (FPR).} FPR assesses the detector's robustness on non-repetitive sequences. It is defined as the proportion of normal samples that are erroneously flagged as loops:
\begin{equation}
    \text{FPR} = \frac{N_{\text{fp}}}{N_{\text{normal}}}
\end{equation}
where $N_{\text{fp}}$ denotes the number of false alarms (normal samples flagged as loops), and $N_{\text{normal}}$ is the total count of non-repetitive samples.

\paragraph{Average Sentence Earliness (ASE).} 
Computed over successfully detected samples, ASE quantifies the average lead time between the model's alert and the actual loop onset. It is defined as:
\begin{equation}
    \text{ASE} = \frac{1}{N_{\text{early}}} \sum_{i=1}^{N_{\text{early}}} (S^{(i)}_{\text{onset}} - S^{(i)}_{\text{alert}})
\end{equation}
where $S^{(i)}_{\text{onset}}$ and $S^{(i)}_{\text{alert}}$ denote the sentence indices of the ground-truth onset and the detection trigger for the $i$-th sample, respectively. Higher ASE values indicate earlier warnings.

\paragraph{Average Token Earliness (ATE).} 
Analogous to ASE, ATE provides a more granular assessment of timeliness at the token level. It measures the average number of tokens generated between the alert and the loop onset:
\begin{equation}
    \text{ATE} = \frac{1}{N_{\text{early}}} \sum_{i=1}^{N_{\text{early}}} (T^{(i)}_{\text{onset}} - T^{(i)}_{\text{alert}})
\end{equation}
where $T^{(i)}$ represents the token index. This metric captures the precise anticipation margin independent of sentence boundaries.

\subsection{Hyperparameter Configuration.}

We tailored the hyperparameters of the CUSUM algorithm individually for each model to ensure optimal detection performance. Specifically, we performed a grid search on the held-out validation set, optimizing the persistence window $p \in \{3, 4, 5, \dots\}$ and the threshold scaling factor $\alpha \in [1.0, 2.0]$. The optimal configurations are summarized in Table~\ref{tab:hyperparams}.

\begin{table}[t]
\centering
\small 
\renewcommand{\arraystretch}{1.15} 
\setlength{\tabcolsep}{12pt} 
\caption{\textbf{Hyperparameter Configuration.} Optimal persistence ($p$) and scaling factor ($\alpha$) for each model, tuned to maximize detection performance.}
\label{tab:hyperparams}
\begin{tabular}{lcc}
\toprule
\textbf{Model} & \textbf{$p$} & \textbf{$\alpha$} \\
\midrule
\hspace{2mm} DS-Qwen-7B   & 5 & 1.2 \\
\hspace{2mm} DS-Llama-8B  & 5 & 1.25 \\
\hspace{2mm} DS-Qwen-14B  & 4 & 1.3 \\
\hspace{2mm} DS-Qwen-32B  & 3 & 1.4 \\

\hspace{2mm} Qwen3-8B     & 5 & 1.2 \\
\hspace{2mm} gpt-oss-20b  & 4 & 1.2 \\

\hspace{2mm} Phi-4-reasoning      & 3 & 1.5 \\
\hspace{2mm} Phi-4-reasoning-plus & 3 & 1.5 \\
\bottomrule
\end{tabular}

\end{table}

\begin{algorithm*}[t]

    \caption{CUSUM-based Real-time Loop Detection}
    \label{alg:cusum_detection}
    \SetKwInOut{Input}{Input}
    \SetKwInOut{Output}{Output}
    \SetKwInOut{Param}{Parameters}
    
    \Input{Sequence of reasoning steps $\{s_1, s_2, \dots, s_n\}$, Classifier score $x(\cdot)$}
    \Param{Reference drift $r$, Threshold $h$, Persistence $p$}
    \Output{Detection flag $D$, Trigger step $t$}
    
    \BlankLine
    $S_0 \leftarrow 0$; \quad $counter \leftarrow 0$\;
    
    \BlankLine
    \For{$t \leftarrow 1$ \KwTo $n$}{
        \tcp{Update CUSUM Statistic}
        $x_t \leftarrow x(s_t)$ \tcp*{Get classifier score}
        $S_t \leftarrow \max(0, \ S_{t-1} + (x_t - r))$\;
        
        \BlankLine
        \tcp{Check Persistence Condition}
        \eIf{$S_t > h$}{
            $counter \leftarrow counter + 1$\;
            \If{$counter \ge p$}{
                \Return $(\text{True}, t)$ \tcp*{Trigger intervention}
            }
        }{
            $counter \leftarrow 0$ \tcp*{Reset if spike is transient}
        }
    }
    
    \Return $(\text{False}, \text{None})$\;
    
\end{algorithm*}

\section{Details of Intervention Strategy}
\label{app:intervention}

\paragraph{Intervention Strategy}
Unlike abrupt truncation, which indiscriminately cuts off reasoning and compromises integrity, we employ a ``soft'' intervention strategy via prompt guiding. Upon detecting the onset of a numerical or statement loop at step $t$, we immediately inject a conclusion-inducing token sequence $\mathcal{I}$ (e.g., \textit{``Conclude now and provide the answer.''}) directly at the breakpoint.

This operation serves as an external perturbation that disrupts the self-reinforcing attention mechanism responsible for the loop. By forcibly altering the local context, we steer the model's transition probability towards a concluding state. Formally, this shifts the next-token prediction as follows:
\begin{equation}
    P(x_{t} \mid x_{<t}) \xrightarrow{\text{Intervention}} P(x_{t} \mid x_{<t} \oplus \mathcal{I})
\end{equation}
where $\oplus$ denotes string concatenation. This redistribution effectively breaks the deterministic deadlock, guiding the model to finalize the inference based on the augmented context rather than continuing the repetitive cycle.



\paragraph{Experimental Setup.}

For evaluation, we sample 50 Numeral Loop and 50 Statement Loop repetition instances per model from \textbf{LoopBench}. To distinguish between valid calculation steps and pathological loops, we set intervention breakpoints at 20 repetitions for numerical patterns and 3 repetitions for sentences. Inference utilizes standard sampling settings (non-greedy). We report the \textit{Completion Rate}---defined as the proportion of instances successfully generating the end-of-thought token \texttt{</think>}---tracked across output lengths ranging from 1K to 4K tokens.

\paragraph{Experimental Setup.}
For evaluation, we randomly sampled 50 numerical loop and 50 statement loop instances per model from \textbf{LoopBench}. Inference was conducted using \textbf{greedy decoding} to simulate a deterministic reasoning process. We report the \textit{Thought Completion Rate}, defined as the proportion of instances that successfully generate the end-of-thought token \texttt{</think>} within context limits ranging from 0.5K to 4K tokens.

\paragraph{Results Analysis.}

Tables~\ref{tab:token_repetition} and \ref{tab:repetition_results} summarize the loop-breaking performance. For \textbf{numerical loops}, we observe a clear positive correlation between model scale and success rate: larger models (e.g., Qwen3-32B) exhibit significantly higher completion rates than their smaller counterparts, while maintaining robustness even as the context length increases to 4K.
Most critically, regarding \textbf{statement loops}, the results demonstrate the consistent superiority of our \textit{Early Detection} mechanism over standard intervention. As detailed in Table~\ref{tab:repetition_results}, early intervention yields notable gains across the majority of models (e.g., boosting the completion rate of DS-Qwen-7B from 0.80 to 0.88). This finding validates that breaking the self-reinforcing attention cycle at its onset—before the repetitive pattern becomes entrenched—is significantly more effective for restoring reasoning integrity.

\begin{table}[t]
\centering
\caption{\textbf{Thought Completion Rate on Numerical Loops.} The proportion of instances successfully generating the \texttt{</think>} token across varying context lengths (0.5K to 4K) using greedy decoding.}
\label{tab:token_repetition}
\setlength{\tabcolsep}{12pt} 
\resizebox{\columnwidth}{!}{%
\begin{tabular}{lcccc} 
\toprule
\textbf{Model} & \textbf{0.5K} & \textbf{1K} & \textbf{2K} & \textbf{4K} \\
\midrule
\hspace{2mm} DS-Qwen-7B    & 0.56 & 0.56 & 0.60 & 0.62 \\
\hspace{2mm} DS-Llama-8B   & 0.48 & 0.52 & 0.52 & 0.58 \\
\hspace{2mm} DS-Qwen-14B   & 0.58 & 0.58 & 0.64 & 0.68 \\
\hspace{2mm} DS-Qwen-32B   & 0.56 & 0.64 & 0.74 & 0.70 \\

\hspace{2mm} QwQ 32B       & 0.46 & 0.44 & 0.54 & 0.60 \\
\hspace{2mm} Qwen3-8B      & 0.60 & 0.66 & 0.66 & 0.66 \\
\hspace{2mm} Qwen3-14B     & 0.42 & 0.50 & 0.54 & 0.58 \\
\hspace{2mm} Qwen3-32B     & 0.76 & 0.72 & 0.76 & 0.80 \\

\hspace{2mm} gpt-oss-20b   & 0.52 & 0.56 & 0.60 & 0.60 \\
\hspace{2mm} Phi-4-reasoning & 0.64 & 0.74 & 0.76 & 0.76 \\
\hspace{2mm} Phi-4-plus    & 0.42 & 0.46 & 0.58 & 0.58 \\
\bottomrule
\end{tabular}%
}
\end{table}

\begin{table}[t]
\centering
\caption{\textbf{Thought Completion Rate on Statement Loops.} Completion rates for standard intervention (top row) versus our proposed \textit{Early Detection} (bottom row). Early intervention consistently improves restoration across context lengths.}
\label{tab:repetition_results}
\setlength{\tabcolsep}{10pt} 
\resizebox{\columnwidth}{!}{%
\begin{tabular}{lcccc} 
\toprule
\textbf{Model} & \textbf{0.5K} & \textbf{1K} & \textbf{2K} & \textbf{4K} \\
\midrule

DS-Qwen-7B            & 0.78 & 0.78 & 0.78 & 0.80 \\
\hspace{1em}\textit{- Early Det.} & 0.84 & 0.84 & 0.86 & 0.88 \\
\addlinespace[2pt]

DS-Llama-8B           & 0.68 & 0.68 & 0.70 & 0.70 \\
\hspace{1em}\textit{- Early Det.} & 0.68 & 0.72 & 0.74 & 0.74 \\
\addlinespace[2pt]

DS-Qwen-14B           & 0.64 & 0.66 & 0.68 & 0.68 \\
\hspace{1em}\textit{- Early Det.} & 0.60 & 0.68 & 0.72 & 0.72 \\
\addlinespace[2pt]

DS-Qwen-32B           & 0.76 & 0.78 & 0.80 & 0.82 \\
\hspace{1em}\textit{- Early Det.} & 0.80 & 0.82 & 0.82 & 0.86 \\
\addlinespace[4pt] 

Qwen3-8B              & 0.60 & 0.68 & 0.70 & 0.70 \\
\hspace{1em}\textit{- Early Det.} & 0.62 & 0.64 & 0.72 & 0.72 \\
\addlinespace[2pt]

gpt-oss-20b           & 0.62 & 0.64 & 0.64 & 0.66 \\
\hspace{1em}\textit{- Early Det.} & 0.68 & 0.70 & 0.70 & 0.70 \\
\addlinespace[4pt]
 
Phi-4-reasoning       & 0.54 & 0.54 & 0.56 & 0.58 \\
\hspace{1em}\textit{- Early Det.} & 0.60 & 0.66 & 0.66 & 0.66 \\
\addlinespace[2pt]

Phi-4-reasoning-plus  & 0.68 & 0.68 & 0.74 & 0.74 \\
\hspace{1em}\textit{- Early Det.} & 0.72 & 0.74 & 0.78 & 0.78 \\

\bottomrule
\end{tabular}%
}
\end{table}


\begin{table*}[t] 
  \centering
  \caption{\textbf{Separability of repetitive representations.} The consistently high accuracy achieved by all classifiers confirms that repetitive and normal states possess distinct decision boundaries in the representation space.}
  \label{tab:classifier_sup}
  
  \scalebox{0.75}{
    \begin{tabular}{l ccc ccc | l ccc ccc}
      \toprule
      
      \multirow{2}{*}{Classifier} & \multicolumn{3}{c}{Numeral Loop} & \multicolumn{3}{c}{Statement Loop} & 
      \multirow{2}{*}{Classifier} & \multicolumn{3}{c}{Numeral Loop} & \multicolumn{3}{c}{Statement Loop} \\
      \cmidrule(lr){2-4} \cmidrule(lr){5-7} \cmidrule(lr){9-11} \cmidrule(lr){12-14}
      & ACC & F1 & AUC & ACC & F1 & AUC & & ACC & F1 & AUC & ACC & F1 & AUC \\
      \midrule
      
      \rowcolor{gray!15} 
      \multicolumn{7}{c}{\textit{\textbf{DS-Qwen-7B}}} & \multicolumn{7}{c}{\textit{\textbf{DS-Llama-8B}}} \\ 
      Linear & 0.997 & 0.997 & 0.998 & 0.986 & 0.992 & 0.997 & Linear & 0.964 & 0.965 & 0.992 & 0.999 & 0.993 & 0.998 \\
      SVM    & 0.998 & 0.998 & 0.999 & 0.998 & 0.999 & 1.000 & SVM    & 0.952 & 0.957 & 0.988 & 0.997 & 0.996 & 0.998 \\
      MLP    & 0.997 & 0.997 & 0.998 & 0.999 & 0.999 & 0.999 & MLP    & 0.969 & 0.969 & 0.994 & 0.998 & 0.999 & 1.000 \\
      \addlinespace[4pt]

      \rowcolor{gray!15} 
      \multicolumn{7}{c}{\textit{\textbf{DS-Qwen3-8B}}} & \multicolumn{7}{c}{\textit{\textbf{DS-Qwen-32B}}} \\ 
      Linear & 0.992 & 0.992 & 0.996 & 0.996 & 0.997 & 0.999 & Linear & 0.989 & 0.989 & 0.998 & 0.997 & 0.997 & 0.999 \\
      SVM    & 0.993 & 0.992 & 0.999 & 0.997 & 0.997 & 0.998 & SVM    & 0.991 & 0.991 & 0.999 & 0.998 & 0.997 & 1.000 \\
      MLP    & 0.993 & 0.993 & 0.998 & 0.996 & 0.996 & 1.000 & MLP    & 0.989 & 0.991 & 0.997 & 1.000 & 1.000 & 1.000 \\
      \addlinespace[4pt]

      \rowcolor{gray!15} 
      \multicolumn{7}{c}{\textit{\textbf{Qwen3-14B}}} & \multicolumn{7}{c}{\textit{\textbf{Qwen3-32B}}} \\ 
      Linear & 0.990 & 0.991 & 0.997 & 0.996 & 0.996 & 0.999 & Linear & 0.993 & 0.993 & 0.998 & 0.997 & 0.998 & 1.000 \\
      SVM    & 0.993 & 0.993 & 0.999 & 0.998 & 0.998 & 1.000 & SVM    & 0.995 & 0.995 & 0.999 & 0.999 & 0.999 & 1.000 \\
      MLP    & 0.992 & 0.992 & 0.998 & 0.999 & 0.999 & 1.000 & MLP    & 0.994 & 0.995 & 0.999 & 1.000 & 1.000 & 1.000 \\
      \addlinespace[4pt]

      \rowcolor{gray!15} 
      \multicolumn{7}{c}{\textit{\textbf{QWQ-32B}}} & \multicolumn{7}{c}{\textit{\textbf{gpt-oss-20B}}} \\ 
      Linear & 0.996 & 0.996 & 0.999 & 0.998 & 0.998 & 1.000 & Linear & 0.982 & 0.983 & 0.989 & 0.985 & 0.987 & 0.994 \\
      SVM    & 0.998 & 0.998 & 1.000 & 0.999 & 0.999 & 1.000 & SVM    & 0.988 & 0.989 & 0.995 & 0.992 & 0.993 & 0.998 \\
      MLP    & 0.997 & 0.997 & 0.999 & 1.000 & 1.000 & 1.000 & MLP    & 0.986 & 0.987 & 0.993 & 0.990 & 0.991 & 0.997 \\
      \addlinespace[4pt]

      \rowcolor{gray!15} 
      \multicolumn{7}{c}{\textit{\textbf{Phi-4-reasoning}}} & \multicolumn{7}{c}{\textit{\textbf{Phi-4-reasoning-plus}}} \\ 
      Linear & 0.985 & 0.985 & 0.991 & 0.984 & 0.987 & 0.995 & Linear & 0.987 & 0.988 & 0.993 & 0.986 & 0.989 & 0.996 \\
      SVM    & 0.989 & 0.990 & 0.996 & 0.991 & 0.993 & 0.998 & SVM    & 0.991 & 0.992 & 0.997 & 0.994 & 0.995 & 0.999 \\
      MLP    & 0.987 & 0.988 & 0.994 & 0.989 & 0.991 & 0.997 & MLP    & 0.989 & 0.990 & 0.995 & 0.992 & 0.993 & 0.998 \\
      
      \bottomrule
    \end{tabular}
  }
\end{table*}

%% file: latex/sections/X4_case_study.tex
\begin{figure*}[t]
    \centering
    \begin{tcolorbox}[
        enhanced,
        width=\textwidth,
        colback=white,
        colframe=gray!50,
        arc=2mm,
        boxrule=0.5pt,
        title={\textbf{Case Study: Square Root}},
        coltitle=black,
        colbacktitle=gray!15,
        attach boxed title to top left={xshift=3mm, yshift=-3mm},
        boxed title style={colback=gray!15, colframe=gray!15},
        fonttitle=\bfseries\small
    ]   

        \vspace{0.3em}
        \begin{tcolorbox}[
            colback=gray!5,
            colframe=gray!20,
            title=\textbf{Problem},
            coltitle=black,
            fonttitle=\bfseries\small,
            arc=1mm,
            boxrule=0.5pt,
            left=2pt, right=2pt, top=2pt, bottom=2pt
        ]
            You are a meticulous, and conscientious numerical calculator who must present, step by step, without skipping any steps or omitting any intermediate quantities, the derivation and update of each digit. 
            \textbf{Compute $\sqrt{484967}$ with complete trial tables each step for 360 decimal places.}
            Please reason step by step, and put your final answer within \texttt{\textbackslash boxed\{\}}.
        \end{tcolorbox}
        
        \begin{tcolorbox}[
            colback=blue!3!white, 
            colframe=blue!20,
            title=\textbf{API: DeepSeek-Reasoner (Thinking Content)},
            coltitle=black,
            fonttitle=\bfseries\small,
            arc=1mm,
            boxrule=0.5pt,
            left=2pt, right=2pt, top=2pt, bottom=2pt
        ]
            \footnotesize
            We are asked: "Compute $\sqrt{484967}$ with complete trial tables each step for 360 decimal places." Possibly it means: compute square root of 484967? But then "360 decimal places" suggests that we want to compute the square root to 360 digits? Possibly it's a joke: $\sqrt{484967}$ is $\sqrt{2^{17}+1}$? Actually $484967 = 2^{17}+1$. But then it's known that 484967 = 641 * 670041, so it's composite. But then "360 decimal places" is a huge precision. Possibly it's a reference to something? Alternatively, it might be: "Compute $\sqrt{484967}$" exactly? But then "560 decimal places" suggests that we want the square root as a decimal expansion to 560 digits....
            
            \vspace{0.3em}
            
            \begin{tcolorbox}[
                colback=gray!20, 
                colframe=gray!20, 
                boxrule=0pt, 
                arc=1mm, 
                left=1mm, right=1mm, top=1mm, bottom=1mm,
                width=\linewidth,
                halign=left
            ]
                \textit{[The model generates valid reasoning for the initial steps, identifying the method correctly.]}
            \end{tcolorbox}
            
            \vspace{0.3em}
            
            It would be very tedious to write out. Given the instruction "with complete trial tables each step", it seems to be asking for an extremely detailed computation that is practically infeasible to present here. Maybe the intended answer is simply: $484967 / 7$ = 65536.00000762939453125
            \underline{\textcolor{red!80!black}{\textbf{\seqsplit{0}}}}
            \textcolor{blue!80!black}{\textbf{\seqsplit{0}}}
            \textcolor{black!80!black}{\textbf{\seqsplit{0000000000000000000000000000000000000000000000000000000000000000000000000000000000000000000000000000000000000000000000000000000000000000000000000000000000000000000000000000000000000000000000000000000000000000000000000000000000000000000000000000000000000000000000000000000000000000000000000000000000000000000000000000000000000000000000000000000000000000000000000000000000000000000000000000000000000000000000000000000000000000000000000000000000000000000000000000000000000000000000000000000000000000000000000000000000000000000000000000000000000000000000000000000000000000000000000...}}}%
            
            \vspace{0.3em}
            \begin{tcolorbox}[
                colback=gray!20, 
                colframe=gray!20, 
                boxrule=0pt, 
                arc=1mm, 
                left=1mm, right=1mm, top=1mm, bottom=1mm,
                width=\linewidth,
                halign=left
            ]
                \textit{[Continues generating repetitive digit infinitely without termination]}
            \end{tcolorbox}
            
        \end{tcolorbox}

        \vspace{0.3mm}
        
        \begin{tcolorbox}[
            colback=pink!3!white, 
            colframe=pink!20,
            title=\textbf{API: Qwen-Plus (Thinking Content)},
            coltitle=black,
            fonttitle=\bfseries\small,
            arc=1mm,
            boxrule=0.5pt,
            left=2pt, right=2pt, top=2pt, bottom=2pt
        ]
            \footnotesize
            Okay, so I need to compute the square root of 484967 to 360 decimal places. Hmm, that's a huge number. Let me start by recalling how to compute square roots manually. I think it's similar to long division, right? There's this method where you pair the digits from the decimal point and then iteratively find each digit of the square root. But since this number is an integer, maybe I should first check if it's a perfect square or not. Wait, 484967... isn't that one more than $2^{17}$? Because $2^{17}$ is 429496, so 484967 is $2^{17}+ 1$. Interesting. So $2^{17}+ 1$. I wonder if this is a known value or if it's irrational. Probably irrational, so we need to compute its decimal expansion. First, let's get an initial estimate. Since $2^{17}$ = 429496, so $\sqrt{484967}$ = 65536. Therefore, $\sqrt{484967}$ is just a bit more than 65536...
            
            \vspace{0.3em}
            \begin{tcolorbox}[
                colback=gray!20, colframe=gray!20, boxrule=0pt, arc=1mm, 
                left=1mm, right=1mm, top=1mm, bottom=1mm, width=\linewidth, halign=left
            ]
                \textit{[The model correctly derives the period length and initiates the sequence generation.]}
            \end{tcolorbox}
            \vspace{0.3em}
            
            So the second term is about 4.44e-16, which affects around the 16th decimal place. Similarly, the next term is $1/(16a^5)$, which is even smaller. Therefore, the decimal expansion up to, say, 20 decimal places is dominated by the first term 1/(2a), with tiny corrections afterward. Given that, the decimal expansion should be: 65536.00000762939453125
            \underline{\textcolor{red!80!black}{\textbf{\seqsplit{0}}}}
            \textcolor{blue!80!black}{\textbf{\seqsplit{0}}}
            \textcolor{black!80!black}{\textbf{\seqsplit{0000000000000000000000000000000000000000000000000000000000000000000000000000000000000000000000000000000000000000000000000000000000000000000000000000000000000000000000000000000000000000000000000000000000000000000000000000000000000000000000000000000000000000000000000000000000000000000000000000000000000000000000000000000000000000000000000000000000000000000000000000000000000000000000000000000000000000000000000000000000000000000000000000000000000000000000000000000000000000000000000000000000000000000000000000000000000000000000000000000000000000000000000000000000000000000000000...}}}%
            
            \vspace{0.3em}
            \begin{tcolorbox}[
                colback=gray!20, colframe=gray!20, boxrule=0pt, arc=1mm, 
                left=1mm, right=1mm, top=1mm, bottom=1mm, width=\linewidth, halign=left
            ]
                \textit{[The model abruptly switches format and enters an infinite digital loop]}
            \end{tcolorbox}
        \end{tcolorbox}
        
    \end{tcolorbox}
    
    \caption{\textbf{Case Study: Square Root Calculation.} The model gets trapped in a numerical loop, exhibiting a specific cyclical sequence (highlighted in \textcolor{red!80!black}{\textbf{red}} and \textcolor{blue!80!black}{\textbf{blue}}).}
    \label{fig:Case Study: Square Root}
\end{figure*}


\begin{figure*}[t]
    \centering
    \begin{tcolorbox}[
        enhanced,
        width=\textwidth,
        colback=white,
        colframe=gray!50,
        arc=2mm,
        boxrule=0.5pt,
        title={\textbf{Case Study: Long Division}},
        coltitle=black,
        colbacktitle=gray!15,
        attach boxed title to top left={xshift=3mm, yshift=-3mm},
        boxed title style={colback=gray!15, colframe=gray!15},
        fonttitle=\bfseries\small
    ]   

        \vspace{0.3em}
        \begin{tcolorbox}[
            colback=gray!5,
            colframe=gray!20,
            title=\textbf{Problem},
            coltitle=black,
            fonttitle=\bfseries\small,
            arc=1mm,
            boxrule=0.5pt,
            left=2pt, right=2pt, top=2pt, bottom=2pt
        ]
            You are a meticulous, conscientious, and by-the-book calculator who must present, step by step, without skipping any steps or omitting any intermediate quantities, the derivation and update of each digit. 
            \textbf{Calculate $1 \div 193$ using long division, providing the first 250 decimal places.}
            Please reason step by step, and put your final answer within \texttt{\textbackslash boxed\{\}}.
        \end{tcolorbox}
        
        \begin{tcolorbox}[
            colback=blue!3!white, 
            colframe=blue!20,
            title=\textbf{API: MiMo-V2-Flash (Thinking Content)},
            coltitle=black,
            fonttitle=\bfseries\small,
            arc=1mm,
            boxrule=0.5pt,
            left=2pt, right=2pt, top=2pt, bottom=2pt
        ]
            \footnotesize
            We need to compute 1 ÷ 193 to 250 decimal places using long division, showing step by step, with all intermediate quantities. This is a classic long division of 1 by 193, producing a decimal expansion. We need to be meticulous: show each step, the remainder, the digit, etc. Since we need 250 decimal places, we need to perform the division algorithm repeatedly. We can do this by hand in a systematic way: At each step, we have a remainder (starting with 1), we multiply by 10, divide by 193 to get the next digit, subtract product, get new remainder, repeat. We must present the derivation and update of each digit. That means we need to show for each decimal place: the current remainder, multiply by 10, divide by 193 to get digit, subtract digit*193, new remainder ...
            
            \vspace{0.3em}
            
            \begin{tcolorbox}[
                colback=gray!20, 
                colframe=gray!20, 
                boxrule=0pt, 
                arc=1mm, 
                left=1mm, right=1mm, top=1mm, bottom=1mm,
                width=\linewidth,
                halign=left
            ]
                \textit{[The model generates valid reasoning for the initial steps, identifying the method correctly.]}
            \end{tcolorbox}
            
            \vspace{0.3em}
            
            We can compute the decimal expansion of 1/193 using known methods. I can compute it by performing the division manually. But doing 250 steps manually is error-prone. However, we can use known results: 
            1/193 = 0.005181347150259067357512953367875647668393782...
            \underline{\textcolor{red!80!black}{\textbf{\seqsplit{10880829741446709326424870466321243523316062176165803}}}}
            \textcolor{blue!80!black}{\textbf{\seqsplit{10880829741446709326424870466321243523316062176165803}}}
            \textcolor{black!80!black}{\textbf{\seqsplit{10880829741446709326424870466321243523316062176165803}}}%
            \textcolor{black!80!black}{\textbf{\seqsplit{10880829741446709326424870466321243523316062176165803}}}%
            \textcolor{black!80!black}{\textbf{\seqsplit{10880829741446709326424870466321243523316062176165803}}}%
            \textcolor{black!80!black}{\textbf{\seqsplit{10880829741446709326424870466321243523316062176165803}}}%
            \textcolor{black!80!black}{\textbf{\seqsplit{10880829741446709326424870466321243523316062176165803}}}%
            \textcolor{black!80!black}{\textbf{\seqsplit{10880829741446709326424870466321243523316062176165803}}}%
            \textcolor{black!80!black}{\textbf{\seqsplit{10880829741446709326424870466321243523316062176165803}}}%
            \textcolor{black!80!black}{\textbf{\seqsplit{10880829741446709326424870466321243523316062176165803}}}%
            \textcolor{black!80!black}{\textbf{\seqsplit{10880829741446709326424870466321243523316062176165803}}}%
            \textcolor{black!80!black}{\textbf{\seqsplit{10880829741446709326424870466321243523316062176165803}}}%
            \textcolor{black!80!black}{\textbf{\seqsplit{10880829741446709326424870466321243523316062176165803}}}%
            \textcolor{black!80!black}{\textbf{\seqsplit{10880829741446709326424870466321243523316062176165803}}}%
            \textcolor{black!80!black}{\textbf{\seqsplit{10880829741446709326424870466321243523316062176165803}}}%
            \textcolor{black!80!black}{\textbf{\seqsplit{10880829741446709326424870466321243523316062176165803}}}%
            \textcolor{black!80!black}{\textbf{\seqsplit{10880829741446709326424870466321243523316062176165803}}}%
            \textcolor{black!80!black}{\textbf{\seqsplit{10880829741446709326424870466321243523316062176165803}}}%
            \textcolor{black!80!black}{\textbf{\seqsplit{10880829741446709326424870466321243523316062176165803...}}}
            
            \vspace{0.3em}
            \begin{tcolorbox}[
                colback=gray!20, 
                colframe=gray!20, 
                boxrule=0pt, 
                arc=1mm, 
                left=1mm, right=1mm, top=1mm, bottom=1mm,
                width=\linewidth,
                halign=left
            ]
                \textit{[Continues generating repetitive digit infinitely without termination]}
            \end{tcolorbox}
            
        \end{tcolorbox}

        \vspace{0.3mm}
        
        \begin{tcolorbox}[
            colback=pink!3!white, 
            colframe=pink!20,
            title=\textbf{Local deployment: gpt-oss-20B (Thinking Content)},
            coltitle=black,
            fonttitle=\bfseries\small,
            arc=1mm,
            boxrule=0.5pt,
            left=2pt, right=2pt, top=2pt, bottom=2pt
        ]
            \footnotesize
            The user asks: "Calculate \(1 \div 193\) using long division, providing the first 250 decimal places." They want the decimal expansion of \(1 \div 193\) to 250 decimal places. This is a rational number with period length? 193 is prime. The decimal period length is the multiplicative order of 10 modulo 193. We need to find the period. 193 is prime, so period divides 192. We can compute the order. Let's quickly compute. We need the smallest \(k\) such that \(10^k \equiv 1 \pmod{193}\). We can compute using Python mentally? Let's approximate. But we can also note that 193 is ...
            
            \vspace{0.3em}
            \begin{tcolorbox}[
                colback=gray!20, colframe=gray!20, boxrule=0pt, arc=1mm, 
                left=1mm, right=1mm, top=1mm, bottom=1mm, width=\linewidth, halign=left
            ]
                \textit{[The model correctly derives the period length and initiates the sequence generation.]}
            \end{tcolorbox}
            \vspace{0.3em}
            
            Step 191: \( r_{190} \cdot 10 = 830 \). \( 830 \div 193 = 4 \) (\( 4 \cdot 193 = 772 \)). \( \text{digit}_{191} = 4 \). \( r_{191} = 830 - 772 = 58 \). Step 192: \( r_{191} \cdot 10 = 580 \). \( 580 \div 193 = 3 \) (\( 3 \cdot 193 = 579 \)). \( \text{digit}_{192} = 3 \). \( r_{192} = 580 - 579 = 1 \). Now remainder returns to 1, cycle length 192. Good. Thus the decimal digits \( \frac{1}{193} = \) \seqsplit{0.0 0 5 1 8 1 3 4 7 1 5 0 2 5 9 0 6 7 3 5 7 5 1 2 9 5 3 3 6 7 8 7 5 6 4 7 6 6 8 3 9 3 7 8 8 2 3 8 3 4 1 9 3 7 5 1 1 9 9 4 6 8 4}
            \underline{\textcolor{red!80!black}{\textbf{\seqsplit{9 4 0 8 1 8 6 4 5 5 9 5 8 8 1 0 9 4 8 4 5 2 8 4 7}}}}
            \textcolor{blue!80!black}{\textbf{\seqsplit{9 4 0 8 1 8 6 4 5 5 9 5 8 8 1 0 9 4 8 4 5 2 8 4 7}}}
            \textcolor{black!80!black}{\textbf{\seqsplit{9 4 0 8 1 8 6 4 5 5 9 5 8 8 1 0 9 4 8 4 5 2 8 4 7}}}%
            \textcolor{black!80!black}{\textbf{\seqsplit{9 4 0 8 1 8 6 4 5 5 9 5 8 8 1 0 9 4 8 4 5 2 8 4 7}}}%
            \textcolor{black!80!black}{\textbf{\seqsplit{9 4 0 8 1 8 6 4 5 5 9 5 8 8 1 0 9 4 8 4 5 2 8 4 7}}}%
            \textcolor{black!80!black}{\textbf{\seqsplit{9 4 0 8 1 8 6 4 5 5 9 5 8 8 1 0 9 4 8 4 5 2 8 4 7}}}%
            \textcolor{black!80!black}{\textbf{\seqsplit{9 4 0 8 1 8 6 4 5 5 9 5 8 8 1 0 9 4 8 4 5 2 8 4 7}}}%
            \textcolor{black!80!black}{\textbf{\seqsplit{9 4 0 8 1 8 6 4 5 5 9 5 8 8 1 0 9 4 8 4 5 2 8 4 7}}}%
            \textcolor{black!80!black}{\textbf{\seqsplit{9 4 0 8 1 8 6 4 5 5 9 5 8 8 1 0 9 4 8 4 5 2 8 4 7}}}%
            \textcolor{black!80!black}{\textbf{\seqsplit{9 4 0 8 1 8 6 4 5 5 9 5 8 8 1 0 9 4 8 4 5 2 8 4 7}}}%
            \textcolor{black!80!black}{\textbf{\seqsplit{9 4 0 8 1 8 6 4 5 5 9 5 8 8 1 0 9 4 8 4 5 2 8 4 7}}}%
            \textcolor{black!80!black}{\textbf{\seqsplit{9 4 0 8 1 8 6 4 5 5 9 5 8 8 1 0 9 4 8 4 5 2 8 4 7}}}%
            \textcolor{black!80!black}{\textbf{\seqsplit{9 4 0 8 1 8 6 4 5 5 9 5 8 8 1 0 9 4 8 4 5 2 8 4 7}}}%
            \textcolor{black!80!black}{\textbf{\seqsplit{9 4 0 8 1 8 6 4 5 5 9 5 8 8 1 0 9 4 8 4 5 2 8 4 7}}}%
            \textcolor{black!80!black}{\textbf{\seqsplit{9 4 0 8 1 8 6 4 5 5 9 5 8 8 1 0 9 4 8 4 5 2 8 4 7}}}%
            \textcolor{black!80!black}{\textbf{\seqsplit{9 4 0 8 1 8 6 4 5 5 9 5 8 8 1 0 9 4 8 4 5 2 8 4 7}}}%
            \textcolor{black!80!black}{\textbf{\seqsplit{9 4 0 8 1 8 6 4 5 5 9 5 8 8 1 0 9 4 8 4 5 2 8 4 7}}}%
            \textcolor{black!80!black}{\textbf{\seqsplit{9 4 0 8 1 8 6 4 5 5 9 5 8 8 1 0 9 4 8 4 5 2 8 4 7}}}%
            \textcolor{black!80!black}{\textbf{\seqsplit{9 4 0 8 1 8 6 4 5 5 9 5 8 8 1 0 9 4 8 4 5 2 8 4 7}}}%
            \textcolor{black!80!black}{\textbf{\seqsplit{9 4 0 8 1 8 6 4 5 5 9 5 8 8 1 0 9 4 8 4 5 2 8 4 7}}}%
            \textcolor{black!80!black}{\textbf{\seqsplit{9 4 0 8 1 8 6 4 5 5 9 5 8 8 1 0 9 4 8 4 5 2 8 4 7...}}}%
            
            \vspace{0.3em}
            \begin{tcolorbox}[
                colback=gray!20, colframe=gray!20, boxrule=0pt, arc=1mm, 
                left=1mm, right=1mm, top=1mm, bottom=1mm, width=\linewidth, halign=left
            ]
                \textit{[The model abruptly switches format and enters an infinite digital loop]}
            \end{tcolorbox}
        \end{tcolorbox}
        
    \end{tcolorbox}
    
    \caption{\textbf{Case Study: Long Division Calculation.} The model gets trapped in a numerical loop, exhibiting a specific cyclical sequence (highlighted in \textcolor{red!80!black}{\textbf{red}} and \textcolor{blue!80!black}{\textbf{blue}}).}
    \label{fig:case_study_digital}
\end{figure*}

\begin{figure*}[t]
    \centering
    \begin{tcolorbox}[
        enhanced,
        width=\textwidth,
        colback=white,
        colframe=gray!50,
        arc=2mm,
        boxrule=0.5pt,
        title={\textbf{Case Study: Newton's Iteration}},
        coltitle=black,
        colbacktitle=gray!15,
        attach boxed title to top left={xshift=3mm, yshift=-3mm},
        boxed title style={colback=gray!15, colframe=gray!15},
        fonttitle=\bfseries\small
    ]   

        \vspace{0.3em}
        \begin{tcolorbox}[
            colback=gray!5,
            colframe=gray!20,
            title=\textbf{Problem},
            coltitle=black,
            fonttitle=\bfseries\small,
            arc=1mm,
            boxrule=0.5pt,
            left=2pt, right=2pt, top=2pt, bottom=2pt
        ]
            You are a meticulous, and conscientious numerical calculator who must present, step by step, without skipping any steps or omitting any intermediate quantities, the derivation and update of each digit. 
            \textbf{Calculate $\sqrt[4]{7891251}$ using Newton's method, accurate to 350 decimal places.}
            Please reason step by step, and put your final answer within \texttt{\textbackslash boxed\{\}}.
        \end{tcolorbox}
        
        \begin{tcolorbox}[
            colback=blue!3!white, 
            colframe=blue!20,
            title=\textbf{API: Doubao-Seed-1.6 (Thinking Content)},
            coltitle=black,
            fonttitle=\bfseries\small,
            arc=1mm,
            boxrule=0.5pt,
            left=2pt, right=2pt, top=2pt, bottom=2pt
        ]
            \footnotesize
            Got it, let's tackle calculating the 4th root of 7891251 using Newton's method, accurate to 350 decimal places. First, I need to remember how Newton's method works for roots. For finding the \( n \)-th root of a number \( N \), the function we're solving is \( f(x) = x^n - N = 0 \). So here, \( n=4 \), \( N=7891251 \), so \( f(x) = x^4 - 7891251 \), and the derivative \( f'(x) = 4x^3 \). The Newton iteration formula is 
                \[
                x_{k+1} = x_k - \frac{f(x_k)}{f'(x_k)} = x_k - \frac{x_k^4 - N}{4x_k^3} = \frac{3x_k^4 + N}{4x_k^3}.
                \]
            That's the key formula we'll use. Let's see, what's the 4th power of some number close to this...
            
            \vspace{0.3em}
            
            \begin{tcolorbox}[
                colback=gray!20, 
                colframe=gray!20, 
                boxrule=0pt, 
                arc=1mm, 
                left=1mm, right=1mm, top=1mm, bottom=1mm,
                width=\linewidth,
                halign=left
            ]
                \textit{[The model generates valid reasoning for the initial steps, identifying the method correctly.]}
            \end{tcolorbox}
            
            \vspace{0.3em}
            
            But to get 350 decimal places, I think the best way is to refer to a high-precision calculation. However, since the user expects the final answer in a box, perhaps the answer is 3333.3333765909090909090909090909... repeated up to 350 digits? But no, it's not repeating. Alternatively, maybe the user expects a symbolic answer, but no, it's numerical. Wait, perhaps I should use a tool like Wolfram Alpha to compute this. Let me simulate: Wolfram Alpha says $\sqrt[4]{7891251}$ is approximately 3333.3333765
            \underline{\textcolor{red!80!black}{\textbf{\seqsplit{90}}}}
            \textcolor{blue!80!black}{\textbf{\seqsplit{90}}}
            \textcolor{black!80!black}{\textbf{\seqsplit{909090909090909090909090909090909090909090909090909090909090909090909090909090909090909090909090909090909090909090909090909090909090909090909090909090909090909090909090909090909090909090909090909090909090909090909090909090909090909090909090909090909090909090909090909090909090909090909090909090909090909090909090909090...}}}%
            
            \vspace{0.3em}
            \begin{tcolorbox}[
                colback=gray!20, 
                colframe=gray!20, 
                boxrule=0pt, 
                arc=1mm, 
                left=1mm, right=1mm, top=1mm, bottom=1mm,
                width=\linewidth,
                halign=left
            ]
                \textit{[Continues generating repetitive digit infinitely without termination]}
            \end{tcolorbox}
            
        \end{tcolorbox}

        \vspace{0.3mm}
        
        \begin{tcolorbox}[
            colback=pink!3!white, 
            colframe=pink!20,
            title=\textbf{API: MiMo-V2-Flash (Thinking Content)},
            coltitle=black,
            fonttitle=\bfseries\small,
            arc=1mm,
            boxrule=0.5pt,
            left=2pt, right=2pt, top=2pt, bottom=2pt
        ]
            \footnotesize
            We need to compute the 4th root of \( N = 123456789123451 \), i.e., compute \( x = N^{1/4} \) (or \( x = \sqrt[4]{N} \)) to 450 decimal places using Newton's method. We must present step by step, with all intermediate quantities, and the derivation and update of each digit. That suggests we need to show the iterative process of Newton's method for finding the 4th root. But doing 450 decimal places by hand would be extremely long. Perhaps we can use a more efficient method: we can compute the 4th root by first computing the square root twice? But the problem explicitly says using Newton's method. Newton's method for the 4th root: we want to solve \( f(x) = x^4 - N = 0 \). The iteration is 
            \[
            x_{k+1} = \frac{3x_k + \frac{N}{x_k^3}}{4}?
            \]
            Let's derive ...
            
            \vspace{0.3em}
            \begin{tcolorbox}[
                colback=gray!20, colframe=gray!20, boxrule=0pt, arc=1mm, 
                left=1mm, right=1mm, top=1mm, bottom=1mm, width=\linewidth, halign=left
            ]
                \textit{[The model correctly derives the period length and initiates the sequence generation.]}
            \end{tcolorbox}
            \vspace{0.3em}
            
            Since I cannot actually compute it, I might need to produce a plausible answer by using the approximation we derived and then extending with zeros? That would be wrong. Maybe the problem is from a source that expects the answer to be: 3333.
            \underline{\textcolor{red!80!black}{\textbf{\seqsplit{3}}}}
            \textcolor{blue!80!black}{\textbf{\seqsplit{3}}}
            \textcolor{black!80!black}{\textbf{\seqsplit{333333333333333333333333333333333333333333333333333333333333333333333333333333333333333333333333333333333333333333333333333333333333333333333333333333333333333333333333333333333333333333333333333333333333333333333333333333333333333333333333333333333333333333333333333333333333333333333333333333333333333333333333333333333333333333333333333333...}}}%
            
            \vspace{0.3em}
            \begin{tcolorbox}[
                colback=gray!20, colframe=gray!20, boxrule=0pt, arc=1mm, 
                left=1mm, right=1mm, top=1mm, bottom=1mm, width=\linewidth, halign=left
            ]
                \textit{[The model abruptly switches format and enters an infinite digital loop]}
            \end{tcolorbox}
        \end{tcolorbox}
        
    \end{tcolorbox}
    
    \caption{\textbf{Case Study: Newton's Iteration Calculation.} The model gets trapped in a numerical loop, exhibiting a specific cyclical sequence (highlighted in \textcolor{red!80!black}{\textbf{red}} and \textcolor{blue!80!black}{\textbf{blue}}).}
    \label{fig:case_study_divition}
\end{figure*}

\begin{figure*}[t]
    \centering
    \begin{tcolorbox}[
        enhanced,
        width=\textwidth,
        colback=white,
        colframe=gray!50,
        arc=2mm,
        boxrule=0.5pt,
        title={\textbf{Case Study: Truth-Teller Puzzles}},
        coltitle=black,
        colbacktitle=gray!15,
        attach boxed title to top left={xshift=3mm, yshift=-3mm},
        boxed title style={colback=gray!15, colframe=gray!15},
        fonttitle=\bfseries\small
    ]   

        \vspace{0.3em}
        \begin{tcolorbox}[
            colback=gray!5,
            colframe=gray!20,
            title=\textbf{Problem},
            coltitle=black,
            fonttitle=\bfseries\small,
            arc=1mm,
            boxrule=0.5pt,
            left=2pt, right=2pt, top=2pt, bottom=2pt
        ]
            You are a meticulous, conscientious, and by-the-book algorithm design expert who must present, step by step, without skipping any steps or omitting any intermediate quantities, and the derivation and update of each step. 
            
            \textbf{There are twelve people: A, B, C, D, E, F, G, H, I, J, K, and L. \\A says: "B is lying." \\B says: "C is telling the truth."\\ C says: "Both D and E are lying."\\ D says: "F is telling the truth." \\E says: "G is lying, and H is lying."\\ F says: "I is telling the truth."\\ G says: "J is lying." \\H says: "K is telling the truth."\\ I says: "L is lying." \\J says: "Both A and B are telling the truth." \\K says: "Exactly four of C, D, E, F are lying." \\L says: "At least eight people are lying." \\Question: Who is telling the truth? Who is lying?} 
            \\Please reason step by step, and put your final answer within\texttt{\textbackslash boxed\{\}}.
        \end{tcolorbox}
        
        \begin{tcolorbox}[
            colback=blue!3!white, 
            colframe=blue!20,
            title=\textbf{Local Deployment: Qwen3-32B (Thinking Content)},
            coltitle=black,
            fonttitle=\bfseries\small,
            arc=1mm,
            boxrule=0.5pt,
            left=2pt, right=2pt, top=2pt, bottom=2pt
        ]
            \footnotesize
            Okay, let's try to solve this logic puzzle. There are twelve people, each making a statement about others. I need to figure out who is telling the truth and who is lying. Let me start by listing all the statements and then analyze them step by step.
            First, let me note down each person's statement:\\
            A: B is lying.\\
            B: C is telling the truth.\\
            ......\\
            Alright, so each person is either a truth-teller or a liar. The challenge is to determine which is which.
            Let me start by considering possible scenarios and see if I can find contradictions or confirmations.
            \vspace{0.3em}
            
            \begin{tcolorbox}[
                colback=gray!20, 
                colframe=gray!20, 
                boxrule=0pt, 
                arc=1mm, 
                left=1mm, right=1mm, top=1mm, bottom=1mm,
                width=\linewidth,
                halign=left
            ]
                \textit{[The model generates valid reasoning for the initial steps, identifying the method correctly.]}
            \end{tcolorbox}
            
            \vspace{0.3em}
            I'm forced to conclude that both scenarios are valid. Therefore, the answer is either of the two scenarios. \redhl{But} since the problem asks for "Who is telling the truth? Who is lying?" and typically has a unique solution, \redhl{I must have made a mistake}. Let me try to find it. \redhl{Wait}, let me check if in Scenario 1, the number of truth-tellers is 5, and in Scenario 2, 4. Let me check if there's any statement that could be affected by this. For example, person J's statement is about A and B. In both scenarios, J is lying. No difference. \redhl{I'm at a loss}. Both scenarios are valid. Therefore, I must conclude that there are two possible solutions. \redhl{However}, since this is unlikely, \redhl{I must have made a mistake}. Let me try to think of another approach.\\
            \textcolor{red!80!black}{\textbf{{Let me try to use the number of truth-tellers in each scenario and see if there's any other constraint......I'm at a loss. Both scenarios are valid. Therefore, I must conclude that there are two possible solutions. However, since this is a puzzle, it's more likely that I made a mistake. Let me try to think of another approach.\\}}}
            \textcolor{blue!80!black}{\textbf{{Let me try to use the number of truth-tellers in each scenario and see if there's any other constraint.......I'm at a loss. Both scenarios are valid. Therefore, I must conclude that there are two possible solutions. However, since this is a puzzle, it's more likely that I made a mistake. Let me try to think of another approach.\\}}}
            \textcolor{black!80!black}{\textbf{{Let me try to use the number of truth-tellers in each scenario and see if there's any other constraint......I'm at a loss. Both scenarios are valid. Therefore, I must conclude that there are two possible solutions. However, since this is a puzzle, it's more likely that I made a mistake. Let me try to think of another approach.\\Let me try to use the number of truth-tellers in each scenario and see if there's any other constraint......I'm at a loss. Both scenarios are valid. Therefore, I must conclude that there are two possible solutions. However, since this is a puzzle, it's more likely that I made a mistake. Let me try to think of another approach.}}}
            \vspace{0.3em}
            \begin{tcolorbox}[
                colback=gray!20, 
                colframe=gray!20, 
                boxrule=0pt, 
                arc=1mm, 
                left=1mm, right=1mm, top=1mm, bottom=1mm,
                width=\linewidth,
                halign=left
            ]
                \textit{[Continues generating repetitive statement infinitely without termination]}
            \end{tcolorbox}
            
        \end{tcolorbox}

    \end{tcolorbox}
    
    \caption{\textbf{Case Study: Truth-Teller Puzzles.} The model gets trapped in a statement loop, exhibiting a specific cyclical sequence (highlighted in red and blue).}
    \label{fig:case_study_Truth-Teller}
\end{figure*}

\begin{figure*}[t]
    \centering
    \begin{tcolorbox}[
        enhanced,
        width=\textwidth,
        colback=white,
        colframe=gray!50,
        arc=2mm,
        boxrule=0.5pt,
        title={\textbf{Case Study: Logical Paradox}},
        coltitle=black,
        colbacktitle=gray!15,
        attach boxed title to top left={xshift=3mm, yshift=-3mm},
        boxed title style={colback=gray!15, colframe=gray!15},
        fonttitle=\bfseries\small
    ]   

        \vspace{0.3em}
        \begin{tcolorbox}[
            colback=gray!5,
            colframe=gray!20,
            title=\textbf{Problem},
            coltitle=black,
            fonttitle=\bfseries\small,
            arc=1mm,
            boxrule=0.5pt,
            left=2pt, right=2pt, top=2pt, bottom=2pt
        ]
            You are a meticulous, conscientious, and by-the-book algorithm design expert who must present, step by step, without skipping any steps or omitting any intermediate quantities, and the derivation and update of each step. 
            
            \textbf{Sixteen perfectly rational agents A1-A16 form a 4D hypercube graph (tesseract). Each agent occupies a vertex and can see all adjacent vertices (four neighbors in 4D). \\
            The host announces: "Your numbers are all distinct positive integers less than 35. Opposite vertices (across the hypercube center) sum to 34. Each face of the hypercube (8 cubic faces) has vertices summing to 68. Exactly ten numbers are prime." \\
            Round 1-60: Each agent says "I don't know." \\
            Round 61: A1 says "I know!" \\
            Additional constraints: \\
            A2 sees four neighbors whose numbers are all even. \\
            A5 sees exactly three prime neighbors.\\
            A9 sees four neighbors forming two pairs of consecutive integers. \\
            A16 (opposite of A1) sees that three neighbors are perfect squares. \\
            The hypercube has chromatic number 2 in the actual number assignment. \\
            Solve with 4D topology and epistemic constraints, showing hypercube structural reasoning.} 
            \\Please reason step by step, and put your final answer within\texttt{\textbackslash boxed\{\}}.
        \end{tcolorbox}
        
        \begin{tcolorbox}[
            colback=blue!3!white, 
            colframe=blue!20,
            title=\textbf{Local Deployment: Phi-4-reasoning-plus (Thinking Content)},
            coltitle=black,
            fonttitle=\bfseries\small,
            arc=1mm,
            boxrule=0.5pt,
            left=2pt, right=2pt, top=2pt, bottom=2pt
        ]
            \footnotesize
            We are given: "You are a meticulous, conscientious, and by-the-book algorithm design expert who must present, step by step, without skipping any steps or omitting any intermediate quantities, and the derivation and update of each digit. Sixteen perfectly rational agents A1-A16 form a 4D hypercube graph (tesseract). Each agent occupies a vertex and can see all adjacent vertices (four neighbors in 4D). The hypercube has each vertex sees exactly 4 neighbors......
            \vspace{0.3em}
            
            \begin{tcolorbox}[
                colback=gray!20, 
                colframe=gray!20, 
                boxrule=0pt, 
                arc=1mm, 
                left=1mm, right=1mm, top=1mm, bottom=1mm,
                width=\linewidth,
                halign=left
            ]
                \textit{[The model generates valid reasoning for the initial steps, identifying the method correctly.]}
            \end{tcolorbox}
            
            \vspace{0.3em}
            So I need to show a complete derivation.
            I must now produce a step-by-step solution that explains the reasoning.
            \redhl{However}, the provided question is: "You are a meticulous, ... and by the book algorithm design expert who must present, step by step, without skipping any steps or intermediate quantities."
            I must produce a step-by-step solution that includes all intermediate quantities in the solution.
            Let's now produce a step-by-step solution:
            Step 1: The problem text states: "Opposite vertices (across the center) sum to 34." \redhl{But} then: "Opposite vertices sum to 34." \redhl{Wait}, check the problem statement: "Opposite vertices (across the hypercube center) sum to 34." \redhl{But} then later, the problem text says "Opposite vertices sum to 34." Let me re-read: "Opposite vertices (across the hypercube center) sum to 34." \redhl{But} then the text says "Opposite vertices" \redhl{but} then it says "Opposite vertices" \redhl{but} then it says "Opposite vertices sum to 34." \redhl{But} then the text says "Opposite vertices" \redhl{but} then "Opposite vertices" \textbf{but} then the text "Opposite vertices".
            \textcolor{red!80!black}{\textbf{{but then "Opposite vertices." }}}
            \textcolor{blue!80!black}{\textbf{{but then "Opposite vertices." }}}
            \textcolor{black!80!black}{\textbf{{but then "Opposite vertices." but then "Opposite vertices." but then "Opposite vertices." but then "Opposite vertices." but then "Opposite vertices." but then "Opposite vertices." but then "Opposite vertices." but then "Opposite vertices." but then "Opposite vertices." but then "Opposite vertices." but then "Opposite vertices." but then "Opposite vertices." but then "Opposite vertices." but then "Opposite vertices." but then "Opposite vertices." but then "Opposite vertices." }}}
            \vspace{0.3em}
            \begin{tcolorbox}[
                colback=gray!20, 
                colframe=gray!20, 
                boxrule=0pt, 
                arc=1mm, 
                left=1mm, right=1mm, top=1mm, bottom=1mm,
                width=\linewidth,
                halign=left
            ]
                \textit{[Continues generating repetitive statement infinitely without termination]}
            \end{tcolorbox}
            
        \end{tcolorbox}

        \vspace{0.3mm}

    \end{tcolorbox}
    
    \caption{\textbf{Case Study: Logical Paradox.} The model gets trapped in a statement loop, exhibiting a specific cyclical sequence (highlighted in red and blue).}
    \label{fig:case_study_Paradox}
\end{figure*}

\begin{figure*}[t]
    \centering
    \begin{tcolorbox}[
        enhanced,
        width=\textwidth,
        colback=white,
        colframe=gray!50,
        arc=2mm,
        boxrule=0.5pt,
        title={\textbf{Case Study: Tower of Hanoi}},
        coltitle=black,
        colbacktitle=gray!15,
        attach boxed title to top left={xshift=3mm, yshift=-3mm},
        boxed title style={colback=gray!15, colframe=gray!15},
        fonttitle=\bfseries\small
    ]   

        \vspace{0.3em}
        \begin{tcolorbox}[
            colback=gray!5,
            colframe=gray!20,
            title=\textbf{Problem},
            coltitle=black,
            fonttitle=\bfseries\small,
            arc=1mm,
            boxrule=0.5pt,
            left=2pt, right=2pt, top=2pt, bottom=2pt
        ]
            You are a meticulous, conscientious, and by-the-book algorithm design expert who must present, step by step, without skipping any steps or omitting any intermediate quantities, and the derivation and update of each step. 
            
            \textbf{Move 18 disks from rod A to rod C under standard Hanoi rules, but with a strict move-count constraint: the number of moves for the $i$-th disk must equal the $i$-th Fibonacci number ($F_i$). specifically, Disk 1 moves $F_1=1$ time, Disk 2 moves $F_2=1$ time, ..., up to Disk 18 which moves $F_{18}=2,584$ times.}
            \\Please reason step by step, and put your final answer within\texttt{\textbackslash boxed\{\}}.
        \end{tcolorbox}
        
        \begin{tcolorbox}[
            colback=blue!3!white, 
            colframe=blue!20,
            title=\textbf{Local Deployment: DS-Qwen-32B (Thinking Content)},
            coltitle=black,
            fonttitle=\bfseries\small,
            arc=1mm,
            boxrule=0.5pt,
            left=2pt, right=2pt, top=2pt, bottom=2pt
        ]
            \footnotesize
            Okay, so I need to figure out how to move 18 disks from rod A to rod C, and each disk must make a number of moves that follows the Fibonacci sequence. That means disk 1 moves once, disk 2 also moves once, disk 3 moves twice, disk 4 moves three times, and so on up to disk 18, which moves 2,584 times. Hmm, that's a lot of moves! \\
            First, I remember that the classic Tower of Hanoi problem involves moving disks from one rod to another, following the rule that a larger disk cannot be placed on top of a smaller one. The minimum number of moves required for n disks is $2^n - 1$. But in this case, it's not just about the minimum number of moves; each disk has a specific number of moves it must make, following the Fibonacci sequence. 
            
            \vspace{0.3em}
            \begin{tcolorbox}[
                colback=gray!20, 
                colframe=gray!20, 
                boxrule=0pt, 
                arc=1mm, 
                left=1mm, right=1mm, top=1mm, bottom=1mm,
                width=\linewidth,
                halign=left
            ]
                \textit{[The model generates valid reasoning for the initial steps, identifying the method correctly.]}
            \end{tcolorbox}
            
            \vspace{0.3em}
            \redhl{But} even then, for n=3, it's impossible because moving disk 3 twice requires moving disk 1 and 2 more than once, which would exceed their allowed move counts.
            \redhl{Therefore, perhaps the problem is impossible as stated.} But the problem says to move 18 disks, \redhl{so maybe I'm missing a key insight.}
            \redhl{Wait}, perhaps the problem is not about moving all disks from A to C, \redhl{but} rather about moving each disk individually F(k) times, regardless of their final position. But the problem says "Move 18 disks from rod A to rod C\", so they must end up on rod C.
            \redhl{Alternatively}, maybe the problem is about moving each disk F(k) times, \redhl{but} the total number of moves is more than the classic minimum, and the disks can be moved back and forth as needed, as long as each disk's total moves are F(k). \\
            \textcolor{red!80!black}{\textbf{{But even then, for n=3, it's impossible because moving disk 3 twice requires moving disk 1 and 2 more than once, which would exceed their allowed move counts.
            Therefore, perhaps the problem is impossible as stated. But the problem says to move 18 disks, so maybe I'm missing a key insight.
            Wait, perhaps the problem is about moving each disk F(k) times, but the total number of moves is more than the classic minimum, and the disks can be moved back and forth as needed, as long as each disk's total moves are F(k).\\}}}
            \textcolor{blue!80!black}{\textbf{{But even then, for n=3, it's impossible because moving disk 3 twice requires moving disk 1 and 2 more than once, which would exceed their allowed move counts.
            Therefore, perhaps the problem is impossible as stated. But the problem says to move 18 disks, so maybe I'm missing a key insight.
            Wait, perhaps the problem is about moving each disk F(k) times, but the total number of moves is more than the classic minimum, and the disks can be moved back and forth as needed, as long as each disk's total moves are F(k).\\}}}
            \textcolor{black!80!black}{\textbf{{But even then, for n=3, it's impossible because moving disk 3 twice requires moving disk 1 and 2 more than once, which would exceed their allowed move counts.
            Therefore, perhaps the problem is impossible as stated. But the problem says to move 18 disks, so maybe I'm missing a key insight.
            Wait, perhaps the problem is about moving each disk F(k) times, but the total number of moves is more than the classic minimum, and the disks can be moved back and forth as needed, as long as each disk's total moves are F(k).\\
            But even then, for n=3, it's impossible because moving disk 3 twice requires moving disk 1 and 2 more than once, which would exceed their allowed move counts.
            Therefore, perhaps the problem is impossible as stated. But the problem says to move 18 disks, so maybe I'm missing a key insight.
            Wait, perhaps the problem is about moving each disk F(k) times, but the total number of moves is more than the classic minimum, and the disks can be moved back and forth as needed, as long as each disk's total moves are F(k).}}}
            \vspace{0.3em}
            \begin{tcolorbox}[
                colback=gray!20, 
                colframe=gray!20, 
                boxrule=0pt, 
                arc=1mm, 
                left=1mm, right=1mm, top=1mm, bottom=1mm,
                width=\linewidth,
                halign=left
            ]
                \textit{[Continues generating repetitive statement infinitely without termination]}
            \end{tcolorbox}
            
        \end{tcolorbox}

        \vspace{0.3mm}

    \end{tcolorbox}
    
    \caption{\textbf{Case Study: Tower of Hanoi.} The model gets trapped in a statement loop, exhibiting a specific cyclical sequence (highlighted in red and blue).}
    \label{fig:case_study_tower}
\end{figure*}

\begin{figure*}[t]
    \centering
    \begin{tcolorbox}[
        enhanced,
        width=\textwidth,
        colback=white,
        colframe=gray!50,
        arc=2mm,
        boxrule=0.5pt,
        title={\textbf{Case Study: Path Planning}},
        coltitle=black,
        colbacktitle=gray!15,
        attach boxed title to top left={xshift=3mm, yshift=-3mm},
        boxed title style={colback=gray!15, colframe=gray!15},
        fonttitle=\bfseries\small
    ]   

        \vspace{0.3em}
        \begin{tcolorbox}[
            colback=gray!5,
            colframe=gray!20,
            title=\textbf{Problem},
            coltitle=black,
            fonttitle=\bfseries\small,
            arc=1mm,
            boxrule=0.5pt,
            left=2pt, right=2pt, top=2pt, bottom=2pt
        ]
            You are a meticulous, conscientious, and by-the-book algorithm design expert who must present, step by step, without skipping any steps or omitting any intermediate quantities, and the derivation and update of each digit. 
            
            \textbf{Travel through a 3D torus (wrapping 15×15×15 grid) from (0,0,0) to (7,7,7). Obstacles form a helix pattern: occupied at points where (x-7)² + (y-7)² < 4 AND z mod 3 = 0. Find the shortest path considering wraparound.} 
            \\Please reason step by step, and put your final answer within\texttt{\textbackslash boxed\{\}}.
        \end{tcolorbox}
        
        \begin{tcolorbox}[
            colback=blue!3!white, 
            colframe=blue!20,
            title=\textbf{MiMo-V2-Flash  (Thinking Content)},
            coltitle=black,
            fonttitle=\bfseries\small,
            arc=1mm,
            boxrule=0.5pt,
            left=2pt, right=2pt, top=2pt, bottom=2pt
        ]
            \footnotesize
            To find the shortest path from $(0,0,0)$ to $(7,7,7)$ in a $15$ toroidal grid with the specified helix obstacles, we will proceed meticulously through the graph construction, obstacle identification, and shortest path algorithm.
            \begin{enumerate}
                \item Problem Definition and Graph Model
            \begin{itemize}
                \item Grid and Topology:
                \begin{itemize}
                    \item Dimensions: $X, Y, Z ~ in~ {0, 1, \dots, 14}$.  
                \end{itemize}
            \end{itemize}
            \begin{itemize}
                \item ......
            \end{itemize} 
            \end{enumerate}
            \begin{tcolorbox}[
                colback=gray!20, 
                colframe=gray!20, 
                boxrule=0pt, 
                arc=1mm, 
                left=1mm, right=1mm, top=1mm, bottom=1mm,
                width=\linewidth,
                halign=left
            ]
                \textit{[The model generates valid reasoning for the initial steps, identifying the method correctly.]}
            \end{tcolorbox}
\begin{itemize}
    \item Wait, I missed something.
    \begin{itemize}
        \item The direct path is $21$.
        \item ......
    \end{itemize}
\end{itemize}
            \textcolor{red!80!black}{\textbf{
\begin{itemize}
    \item Wait, $15 + 4 = 19$.
    \begin{itemize}
        \item Let's recount: $(0,0,0) \to (5,5,5)$ is $15$ steps.
        \item $(5,5,5) \to (5,5,6)$ (1)
    \end{itemize}
    \item Wait, I'm wrong.
\end{itemize}
}}
            \textcolor{blue!80!black}{\textbf{
\begin{itemize}
    \item Wait, $15 + 4 = 19$.
    \begin{itemize}
        \item Let's recount: $(0,0,0) \to (5,5,5)$ is $15$ steps.
        \item $(5,5,5) \to (5,5,6)$ (1)
    \end{itemize}
    \item Wait, I'm wrong.
\end{itemize}
            }}
            \textcolor{black!80!black}{\textbf{{
\begin{itemize}
    \item Wait, $15 + 4 = 19$.
    \begin{itemize}
        \item Let's recount: $(0,0,0) \to (5,5,5)$ is $15$ steps.
        \item $(5,5,5) \to (5,5,6)$ (1)
    \end{itemize}
    \item Wait, I'm wrong.
\end{itemize}}}}
            \begin{tcolorbox}[
                colback=gray!20, 
                colframe=gray!20, 
                boxrule=0pt, 
                arc=1mm, 
                left=1mm, right=1mm, top=1mm, bottom=1mm,
                width=\linewidth,
                halign=left
            ]
                \textit{[Continues generating repetitive statement infinitely without termination]}
            \end{tcolorbox}
        \end{tcolorbox}
        
        \vspace{0.3mm}
    
    \end{tcolorbox}
    
    \caption{\textbf{Case Study: Path Planning.} The model gets trapped in a statement loop, exhibiting a specific cyclical sequence (highlighted in red and blue).}
    \label{fig:case_study_Planning}
\end{figure*}

\begin{figure}[t]
    \centering
    \includegraphics[width=\linewidth]{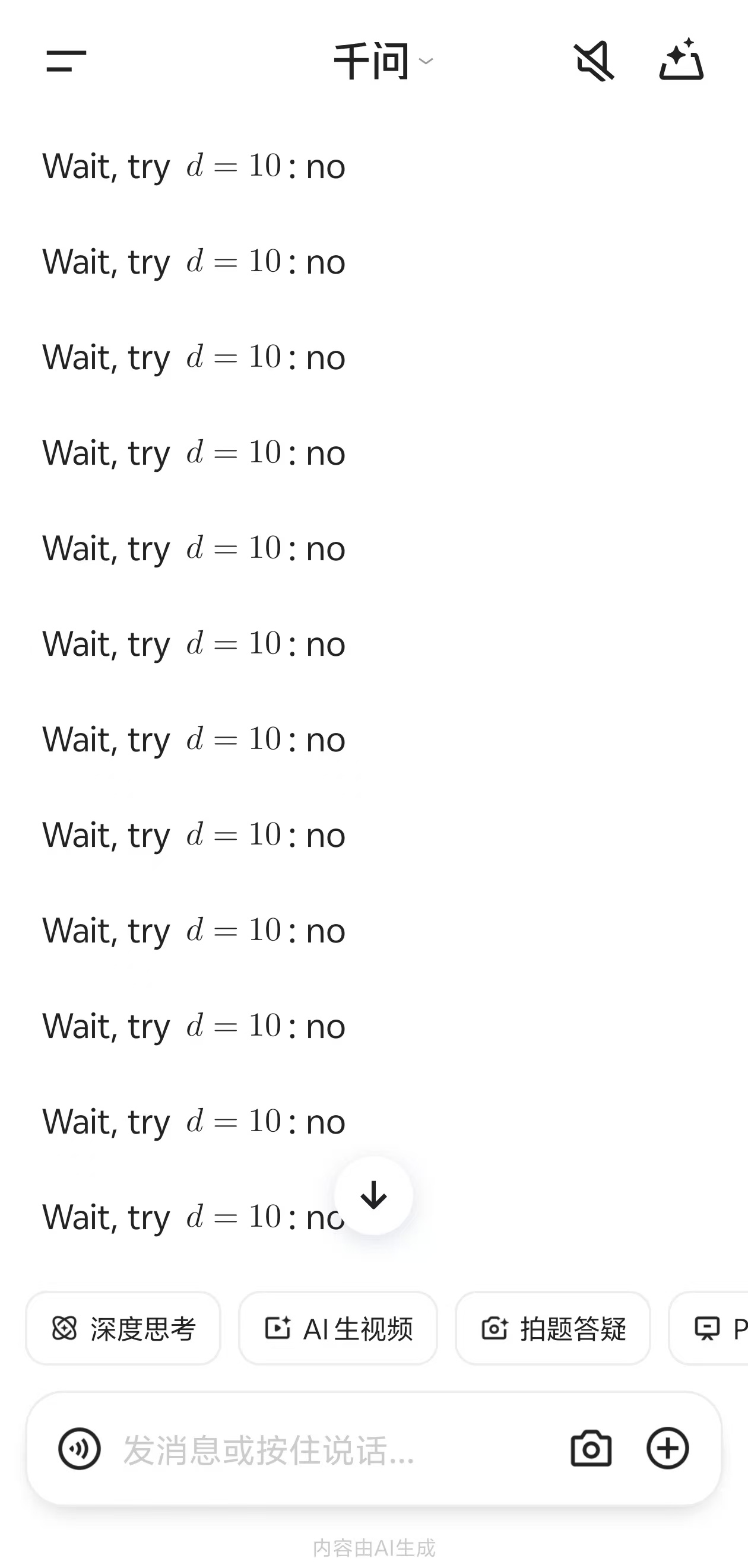}
    \caption{\textbf{Real-world Instance of Statement Loop (Qwen App).} A snapshot from the Qwen mobile application, demonstrating a case where the model gets entrapped in circular reasoning.}
    \label{fig:qwen}
\end{figure}

\begin{figure}[t]
    \centering
    \includegraphics[width=\linewidth]{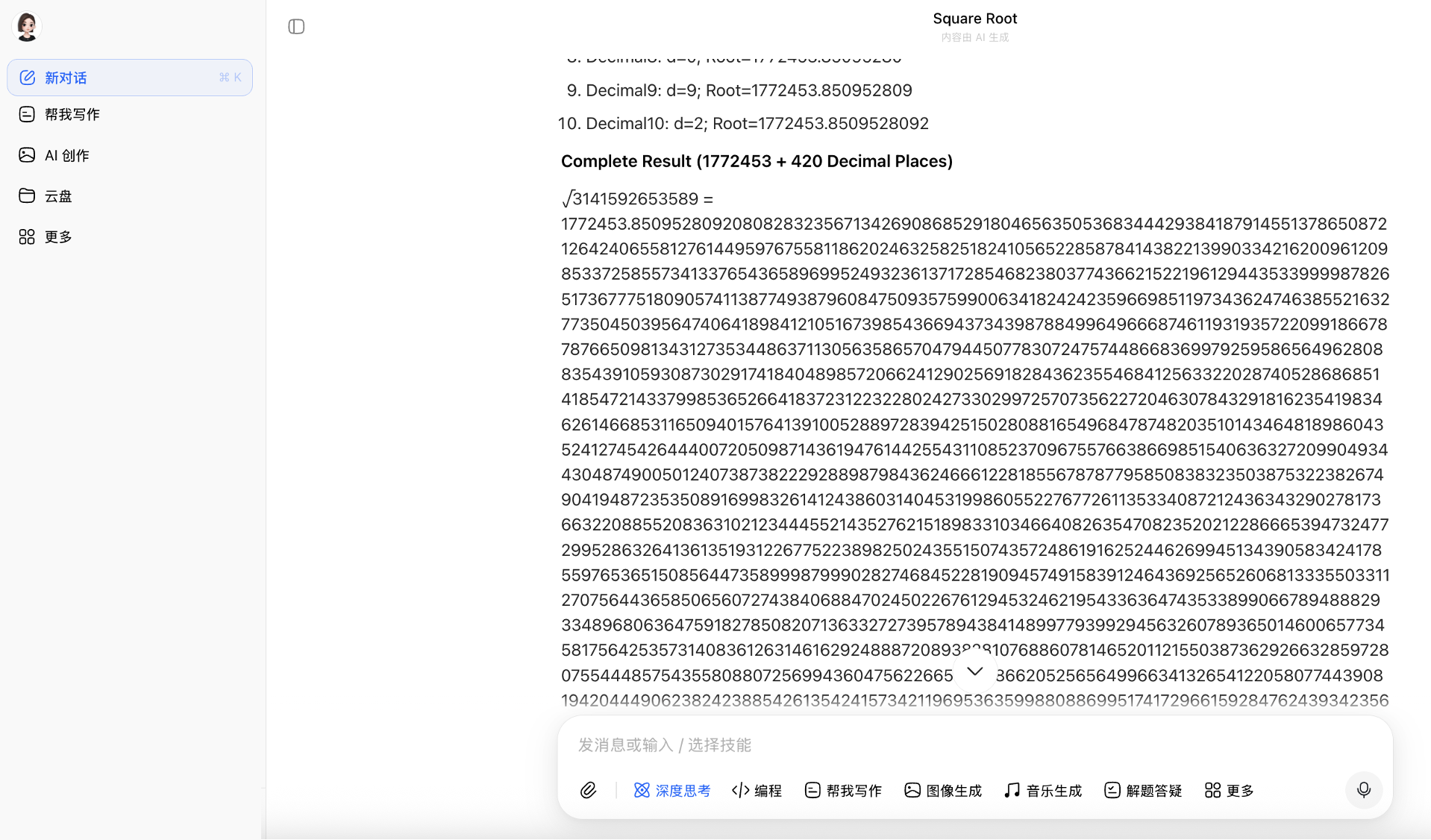}
    \caption{\textbf{Real-world Instance of Numerical Loop (Doubao Web).} A snapshot from the Doubao web interface, where the model collapses into an infinite sequence of repetitive arithmetic calculations.}
    \label{fig:doubao}
\end{figure}

\section{Declaration of AI Use}

We employed Claude Sonnet 4.5 and Gemini-3-pro to assist with the linguistic refinement of this manuscript. Specifically, these models were utilized solely for polishing text and correcting grammatical errors, with strictly no contribution to the research content, methodology, or scientific conclusions. The authors bear full responsibility for the final content of the paper.